\pdfoutput=1
\documentclass[11pt]{article}
\usepackage[margin=1in]{geometry}
\usepackage[T1]{fontenc}
\usepackage[utf8]{inputenc}
\usepackage{lmodern}
\usepackage{textcomp}
\usepackage{amsmath,amssymb,bm}
\usepackage{newunicodechar}
\newunicodechar{τ}{\ensuremath{\tau}}
\newunicodechar{Δ}{\ensuremath{\Delta}}
\newunicodechar{·}{\textperiodcentered}
\newunicodechar{↵}{\ensuremath{\hookleftarrow}}
\usepackage{graphicx}
\usepackage{longtable,booktabs,array,calc}
\usepackage[numbers,sort&compress]{natbib}
\usepackage{xcolor}
\usepackage[colorlinks=true,linkcolor=blue!50!black,citecolor=blue!50!black,urlcolor=blue!50!black]{hyperref}
\usepackage{fancyvrb}
\usepackage{caption}
\providecommand{\tightlist}{\setlength{\itemsep}{0pt}\setlength{\parskip}{0pt}}

\let\oldunderscore\_
\renewcommand{\_}{\oldunderscore\allowbreak}
\title{Interpreting Language Model Parameters}
\author{\parbox{0.95\textwidth}{\centering Lucius Bushnaq$^{1,*}$, Dan Braun$^{1,*,\dagger}$, Oliver Clive-Griffin$^{1,*,\dagger}$, Bart Bussmann$^{2}$, Nathan Hu$^{2}$, Michael Ivanitskiy$^{2}$, Linda Linsefors$^{3}$, Lee Sharkey$^{1,*}$\\[0.5em]
\small $^1$Goodfire\quad $^2$MATS\quad $^3$Independent\\
\small Correspondence: \href{mailto:lee@goodfire.com}{lee@goodfire.com}\\
\small $^*$Core contributor.\quad $^\dagger$Equal contribution.\\
\small Interactive version: \url{https://www.goodfire.com/research/interpreting-lm-parameters}}}
\date{April 2026}

\begin{document}
\maketitle
\begin{abstract}
Mechanistic interpretability aims to uncover how neural networks use their parameters to implement their impressive neural algorithms. Although previous work has uncovered substantial structure in the intermediate representations that networks use, little progress has been made to understand how the parameters and nonlinearities of networks perform computations on those representations. In this work, we present a method that brings us closer to this understanding by decomposing a language model's parameters into subcomponents that each implement only a small part of the model's learned algorithm, while simultaneously requiring only a small fraction of those subcomponents to account for the network's behavior on any input. The method, adVersarial Parameter Decomposition (VPD), optimizes for decompositions of neural network parameters into simple subcomponents that preserve the network's input-output behavior even when many subcomponents are ablated, including under ablations that are adversarially selected to destroy behavior. This encourages learning subcomponents that provide short, mechanistically faithful descriptions of the network's behavior that should aggregate appropriately into more global descriptions of the network's learned algorithm. We study how sequences of interactions between these parameter subcomponents produce the network's output on particular inputs, enabling a new kind of 'circuit' analysis. While more work remains to be done to deepen our understanding of how neural networks use their parameters to compute their behavior, our work suggests an approach to identify a small set of simple, mechanistically faithful subcomponents on which further mechanistic analysis can be based.

\medskip\noindent\emph{This paper was written to be read online, with interactive figures, at
\url{https://www.goodfire.com/research/interpreting-lm-parameters}.}
\end{abstract}
\tableofcontents
\newpage
\section{Introduction}\label{introduction}

Mechanistic interpretability aims to reverse engineer neural networks,
such as language models, so that we can understand the neural algorithms
they have learned. Reverse engineering requires decomposing a system
into simpler parts that we can study in relative isolation.
Unfortunately, it is not obvious how best to decompose neural networks
into such parts
\citep{mueller2024questrightmediatorhistory, sharkey2025openproblemsmechanisticinterpretability}.
The most straightforward candidates for these parts, such as neurons,
attention heads, or whole layers, don't always map to individual,
interpretable computations
\citep{hinton1981parallel, wei2015understandingintraclassknowledgeinside, nguyen2016multifacetedfeaturevisualizationuncovering, olah2017feature, janiak2023polysemantic, jermyn2023attention, yun2021sparse, lindsey2024crosscoders}.

Alternative approaches to decomposition, such as transcoders
\citep{dunefsky2024transcodersinterpretablellmfeature, ameisen2025circuit}
or mixtures of linear transforms
\citep{oldfield2025towards, lindsey2025molts}, typically involve fitting
a set of simple functions to the transitions between activations at
different layers in the network, and linearly combining the outputs of
these simple functions. The idea here is to approximate the complex,
nonlinear function implemented by the network's layers using a simpler,
easier-to-understand function. These methods, sometimes called
\emph{activation-based decomposition} methods, have led to significant
advances in our understanding of the intermediate representations inside
neural networks when computing their outputs
\citep{dunefsky2024transcodersinterpretablellmfeature, ameisen2025circuit}.

Unfortunately, because the simpler functions that these methods use are
of a different functional form to the original network, it is hard to
relate their accounts of network function to the actual objects that are
doing the computations, namely the network's parameters and
nonlinearities.

This is not just a theoretical issue. It prevents us from achieving
practical engineering goals. For example, it makes it challenging to
know how to make precise, predictable modifications to a model's neural
algorithm by making edits to its parameters. It also makes it hard to
predict how the model's neural algorithm will perform on a different
distribution to the one it was studied on.

The mismatch of functional form between models and their
activation-based decompositions is an important issue, but it is not the
only one: Activation-based methods have not yet yielded decompositions
that exhibit a fully satisfactory level of mechanistic faithfulness
\citep{ameisen2025circuit}, and suffer from a number of other issues
(See \citep{sharkey2025openproblemsmechanisticinterpretability} for
review).

These issues motivate alternative approaches to mechanistic
decomposition, including parameter decomposition methods
\citep{braun2025interpretabilityparameterspaceminimizing, bushnaq2025spd, chrisman2025identifyingsparselyactivecircuits},
which give accounts of network function in terms of the parameters that
the network uses on each datapoint. \emph{Ablation-based parameter
decomposition methods}
\citep{braun2025interpretabilityparameterspaceminimizing, bushnaq2025spd}
aim to identify a set of parameter components where as few components as
possible are necessary to perform the same computations as the original
network on any datapoint, and ``unnecessary'' components can be ablated
on a given datapoint in any combination without adversely affecting
output reconstruction error. Simultaneously, the parameter components
are selected to implement as simple computations as possible and to sum
collectively to the target network's parameters. If parameter components
exhibit all these properties, then they are strong candidates for the
network's `ground truth' mechanisms\footnote{Though one would first need
  to accept philosophically that such mechanisms can be said to exist in
  non-toy networks!}.

Parameter decomposition methods can identify known ground truth
mechanisms in toy models that: Are not necessarily aligned to
architectural components such as neurons, individual attention heads, or
layers; operate on representations in superposition; or are
multidimensional. And, due to the requirement that unnecessary
components can be ablated in any combination rather than just all
simultaneously, parameter decomposition methods should not exhibit
feature splitting. Notably, parameter decomposition methods can readily
be applied to any architecture, unlike activation-based methods, where
it has been challenging to use the same decomposition methods to
decompose both attention layers and MLPs
\citep{kamath2025tracing, ameisen2025circuit, wynroe2024decomposing, ge2024localglobal}.
In demonstration of this ability, previous work has used ablation-based
parameter decomposition to identify induction heads in a transformer
trained on a toy model of induction
\citep{christensen2025decomposition}.

Ablation-based parameter decomposition methods thus promise solutions to
many of the issues of activation-based decomposition methods. However,
prior parameter decomposition proposals have several important
shortcomings, some of which we address in this work with a new method
that we introduce, called \emph{ad\textbf{V}ersarial \textbf{P}arameter
\textbf{D}ecomposition} (\textbf{VPD})\footnote{Regrettably, the acronym
  APD was taken by our previous work, Attribution-based Parameter
  Decomposition!
  \citep{braun2025interpretabilityparameterspaceminimizing}}. Our main
contributions are:

\begin{itemize}
\item
  \textbf{We scale parameter decomposition to full language models}:
  While the most recent parameter decomposition method, Stochastic
  Parameter Decomposition (SPD) \citep{bushnaq2025spd} is more scalable
  than its predecessor, Attribution-based Parameter Decomposition
  \citep{braun2025interpretabilityparameterspaceminimizing}, it has not
  yet been applied to full language models. We use VPD to decompose a
  small language model (\(67\)M parameters, four layers) trained on the
  Pile \citep{gao2020pile}. We find parameter subcomponents that are
  highly interpretable (\autoref{sec:param-comps-interpretable}), both
  in terms of the dataset examples that they activate on and how they
  interact with other subcomponents to produce specific behaviors
  (\autoref{sec:circuits}).
\item
  \textbf{We introduce a stronger notion of ablatability to achieve more
  mechanistic faithfulness}: While some work has applied SPD to a single
  layer of GPT2-small \citep{christensen2025decomposition}, no
  application of SPD so far has measured key metrics that would be
  necessary to ensure mechanistic faithfulness, such as having good
  output reconstruction loss even under adversarially chosen ablations
  (rather than under only stochastically chosen ablations). We resolve
  this issue with VPD, which builds heavily on the SPD method but has
  several important modifications, which together make it more
  mechanistically faithful and scalable to larger models than those
  decomposed in previous work. The primary difference between VPD and
  SPD is in the ablations. On each datapoint, both SPD and VPD sample
  from the space of possible partial ablations of parameter
  subcomponents in order to check whether those parameter subcomponents
  can be partially ablated in any combination, thus identifying whether
  they are ``necessary'' for that datapoint. However, where SPD samples
  from the space of partial ablations using \emph{stochastic} samples
  from the space, VPD uses \emph{adversarially chosen samples}
  (\autoref{sec:opt-mech-faithfulness}) \footnote{Both approaches are
    nonetheless designed to approximate what would happen if we could
    check \emph{all} potential partial ablations.}. The core details of
  the method are discussed in \autoref{sec:method}.
\item
  \textbf{We compare VPD to other decomposition methods}: We compare the
  parameter subcomponents that we find to the objects found by other
  decomposition methods, such as per-layer and cross-layer transcoder
  (CLT) latents. We find that VPD achieves a better tradeoff between
  sparsity and reconstruction under standard training objectives and is
  more robust to mismatches between training and evaluation protocols
  compared to end-to-end trained methods
  (\autoref{sec:decomp-model-behav-sim},
  \autoref{app:vpd-sparsity-acc-tradeoff}). VPD also has comparable
  interpretability (\autoref{sec:param-comps-interpretable}) and
  exhibits less feature splitting (\autoref{sec:splitting}) than
  activation-based comparisons.
\item
  \textbf{We decompose attention layers into computations that are
  distributed across multiple heads}: Our approach decomposes parameters
  in attention layers into functionally specialized subcomponents that
  span multiple heads. These subcomponents interact to perform
  interpretable computations. Perhaps for the first time, our approach
  yields a satisfying decomposition of computations in attention layers
  even though those computations may involve multiple heads
  (\autoref{sec:attn-analysis}).
\item
  \textbf{We develop attribution graphs to study information flow
  between parameter subcomponents}: We demonstrate that the parameter
  subcomponents found by VPD can be used to construct interpretable
  attribution graphs that let us study the circuits that underlie some
  language model behaviors (\autoref{sec:circuits}).
\item
  \textbf{We use parameter subcomponents to manually edit a model}:
  Finally, we provide a proof of concept showing that we can use our
  understanding of the network's parameters to manually edit a model in
  a predictable, interpretable way. In particular, we rewrite the part
  of its neural algorithm involved in emoticon predictions
  (\autoref{sec:model-editing}).
\end{itemize}

Additionally, we also introduce an approach for clustering parameter
subcomponents into full parameter components. Previous methods left this
clustering step implicit \citep{bushnaq2025spd}
(\autoref{app:clustering}). We introduce an explicit clustering method,
but found that subcomponents were usually interpretable even without
clustering, and therefore used clustering only rarely in our analyses.

We release a library for reproducing our experiments and running VPD at
\url{https://github.com/goodfire-ai/param-decomp}.

\section{The core method: adVersarial Parameter
Decomposition}\label{the-core-method-adversarial-parameter-decomposition}

\label{sec:method}

In this section, we introduce ablation-based parameter decomposition
methods from scratch and highlight key differences between VPD and prior
methods in this class. Although our method, VPD, builds heavily on SPD
\citep{bushnaq2025spd}, the following explanation of VPD does not assume
familiarity with SPD or its predecessor
\citep{braun2025interpretabilityparameterspaceminimizing}.

Our goal is to decompose a neural network into the \emph{mechanisms}
that it uses to compute its behavior. Its mechanisms are what it uses to
take input activations, compute its hidden activations, and finally
compute its output. We don't approach this goal with strong
presuppositions of what a ``mechanism'' is. But we take for granted that
a typical network doesn't use all of its mechanisms on every input (or,
at least, it doesn't use all of its mechanisms by the same amount). If
that were not the case, then networks could not be said to be
\emph{modular}, having distinct parts that do different things on
different inputs. Without modularity, networks simply couldn't be
decomposed into separable functional units.

One candidate for the network's mechanisms is the network's parameters.
Like mechanisms, networks appear not to use all of their parameters
simultaneously on every datapoint
\citep{veit2016residual, zhang2022moefication, dong2023attention}. This
happens, for instance, when a network's parameters ``read from''
activation subspaces that are orthogonal to the activations on that
datapoint, thus projecting the activations to zero, thereafter having no
downstream causal effect. Alternatively, if the activations fail to
``activate'' a given ReLU neuron, the activation of that neuron is zero,
thereafter having no downstream causal effect. However, the network's
parameters are in fact a single vector in the network's parameter space,
and do not have an obvious decomposition into parts. How should they be
decomposed into parts that comprise the network's mechanisms?

On a high level, parameter decomposition methods use the idea that it
should be possible, for a given datapoint, to identify the ``subset'' of
the network's parameters that are necessary and sufficient for computing
its output on that datapoint. That ``subset'' of parameters should
contain all the mechanisms used by the network on that datapoint. If
particular ``subsets'' of the network's parameters are repeatedly used
together by different datapoints, then they may be part of the same
mechanism. Parameter decomposition methods therefore aim to find
particular ``subsets'' of the network's parameters that tend to be used
together, where as few of them as possible are necessary and sufficient
for computing the network's output on any input\footnote{We use the word
  ``subset'' loosely here. In practice, parameters are not divided into
  discrete sets. The network's parameters are a vector in parameter
  space, and we want some way to divide up that vector into `parts' in a
  way that they still `make up' the original parameters.}.\footnote{An
  analogy that is sometimes helpful for understanding VPD is that it is
  similar to Singular Value Decomposition on a weight matrix, except
  where we decompose the matrix into more subcomponents than the rank of
  the matrix, and where the subcomponents we identify are the parts of
  the matrix that have similar downstream causal effects, thereby taking
  downstream nonlinearities into account.}

More concretely: If particular parameters are unused by the network on a
particular datapoint, then we should be able to ablate them (including
partially) on that datapoint without adversely affecting the network's
output. Ablation-based parameter decomposition methods thus aim to
decompose network parameters into a set of vectors in parameter space
called \emph{parameter components}. Parameter components are trained to
exhibit a number of specific properties such that, if they exhibit those
properties, they would be good candidates for the network's
``mechanisms''. They are trained to be:

\begin{itemize}
\tightlist
\item
  \textbf{Parameter-faithful}: They sum to the network's total parameter
  vector;
\item
  \textbf{Minimal}: As few components as possible are causally important
  for computing the network's output on any particular input;
\item
  \textbf{Mechanistically faithful}: Every subset of components that
  includes the causally important components is sufficient to compute
  the network's output on any particular input;
\item
  \textbf{Simple}: Each component should involve as little computational
  machinery as possible.
\end{itemize}

In the following sections, we define parameter components concretely and
explain how they are optimized to exhibit each of these four properties.

\subsection{Parameter components consist of
subcomponents}\label{parameter-components-consist-of-subcomponents}

\label{sec:method-components}

Suppose we have a neural network \(f(x;\theta)\) with parameters
\(\theta\). We would like to decompose this parameter vector into a sum
of \emph{parameter components} with the above properties.

It would be computationally expensive to decompose models into whole
parameter vectors, since each such vector would have a memory cost
equivalent to the whole target model. Therefore, as in
\citep{bushnaq2025spd}, we use a less expensive way to parameterize
parameter components: Although its parameters \(\theta\) can be
expressed as a single large vector, they are more commonly
conceptualized as a set of matrices \(\theta = \{W_1, \dots, W_L\}\). We
further decompose individual matrices into sums of rank-one matrices
called \emph{subcomponents}, each parameterized as an outer product of
two vectors:

\[W_l \approx \sum_{c} \vec{U}^l_c (\vec{V}_c^l)^\top = U^l (V^l)^\top , \]

where there may be more subcomponents than rows and columns in the
matrix. Permitting more subcomponents than rows and columns in the
matrix allows VPD to identify mechanisms that operate on representations
in superposition
\citep{Vaintrob_Mendel_Kaarel_2024, Bushnaq_Mendel_2024, elhage2022toy}.

\begin{figure}[htbp]
\centering
\includegraphics[width=\textwidth,height=0.8\textheight,keepaspectratio]{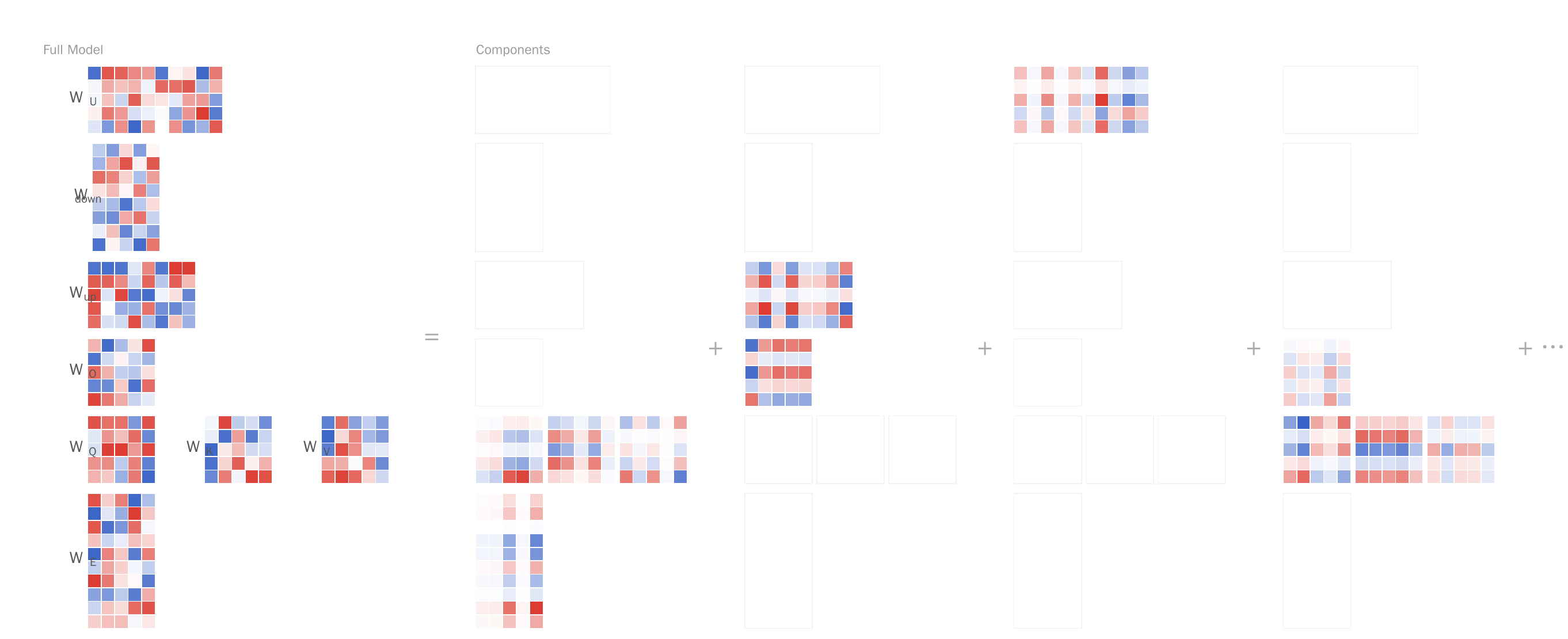}
\caption[]{Parameter decomposition methods decompose target model parameters into vectors in parameter space (parameter components) that are optimized to approximate the model's mechanisms.}\label{fig:sum_components}
\end{figure}

Although a single subcomponent \emph{explicitly} parameterizes only a
single weight matrix, it \emph{implicitly} parametrizes a full parameter
vector if we assume it takes values of \(0\) in all other weight
matrices. It is therefore possible to combine these subcomponents into
full parameter components by adding them together in the right way. We
identify these components using a subcomponent clustering method.
Previous work left this clustering step implicit, but in this paper we
introduce an explicit method (\autoref{app:clustering}).

\subsection{\texorpdfstring{Enforcing parameter faithfulness with
\(\Delta\)-components}{Enforcing parameter faithfulness with \textbackslash Delta-components}}\label{enforcing-parameter-faithfulness-with-delta-components}

\label{sec:method-delta-components}

To ensure the components collectively sum to the parameter vector of the
target model, we define additional \(\Delta\)-components, \(\Delta^l\),
that parametrize the difference between our subcomponents and the
original model's matrices:

\begin{equation}
\Delta^l
  :=
  W^{l}
  -
    \sum_{c}
    \vec{U}^l_c (\vec{V}_c^l)^\top
\label{eq:delta_l2}
\end{equation}

We also encourage the \(\Delta^l\)-components to be small with an
auxiliary MSE loss (\(\mathcal{L}_{\text{Delta-L2}}\))
(\autoref{sec:vpd_delta_l2}).

\subsection{Optimizing for minimality}\label{optimizing-for-minimality}

\label{sec:opt-minimality}

We want as few subcomponents as possible to be causally important for
computing the network's output on any particular input. We therefore
need some way to estimate which parameter subcomponents are
``necessary'' for computing the network's output on a given datapoint.
We also require a notion of how well the ``necessary'' subcomponents
have reconstructed the network's output.

Ablation-based parameter decomposition methods contend that a parameter
subcomponent is ``necessary'' if it cannot be ablated without affecting
the model's output on that datapoint. As in \citep{bushnaq2025spd}, we
train a \emph{causal importance function} to predict how ablatable each
subcomponent is on each batch and sequence position. We also implement
the causal importance function using a neural network, though we use a
different architecture (\autoref{sec:vpd_ci_function}).

We call the output of this function the \emph{causal importance values},
\(g^l_{b,t,c}\in[0,1]\) (for each subcomponent \(c\) of weight matrix
\(l\) at a given batch index \(b\) and sequence position \(t\)):

\begin{itemize}
\item
  If \(g^l_{b,t,c} = 0\), then we should be able to fully or partially
  ablate that subcomponent on the forward pass at position \(b,t\)
  without affecting the final model output.
\item
  If \(g^l_{b,t,c} = 1\), then it should not be possible to ablate that
  subcomponent without affecting the model's output on that
  datapoint\footnote{The Delta-components \(\Delta^l\) should always be
    ablatable, so they are assigned a causal importance of \(0\) by
    definition.}.
\end{itemize}

We want as few subcomponents as possible to be required to compute the
output, so we train the causal importance values \(g^l_{b,t,c}\) to take
minimal values with an \emph{importance minimality loss}:

\label{eq:minimal}

\[\begin{aligned}
\mathcal{L}_{\text{importance-minimality}}
  =
  \frac{1}{BT}
  \sum^{B}_{b=1}
  \sum^{T}_{t=1}
  \sum^{L}_{l=1}
  \sum^C_{c=1}
  \vert g^l_{b,t,c} \vert^p,
\end{aligned}\]

where \(p>0\).\footnote{The \(\Delta^l\)-components are defined always
  to have causal importance values of zero, since they should never be
  ``necessary'' to compute the model output.}

\subsection{Optimizing for mechanistic
faithfulness}\label{optimizing-for-mechanistic-faithfulness}

\label{sec:opt-mech-faithfulness}

Components and their causal importances should be mechanistically
faithful to the original model. One way of operationalizing this is to
insist that, on any given data point, it should ideally be possible to
ablate all causally unimportant components from the model weights, using
any combination of ablations, without changing the model output.
Another, more succinct, way of saying this is that \emph{every} subset
of components that includes the causally important components should be
sufficient to compute the network's output on any particular input.

This is a much stricter requirement than merely demanding that the
output should be invariant to the joint ablation of all causally
unimportant components together. To see why it is stricter, suppose that
two components \(\theta_A\) and \(\theta_B\) can be \emph{jointly}
ablated, but not \emph{individually} ablated, on a data point without
affecting the output\footnote{One way this could happen if \(\theta_A\)
  and \(\theta_B\) cancel each other out by influencing the final model
  output vector in opposite directions.}. Then we would consider both
\(\theta_A\) and \(\theta_B\) to be causally important on that
datapoint, whereas the less strict criterion might consider them both
causally unimportant because they happen to be jointly ablatable. In
other words, the stricter criterion demands an unchanged model output
over a whole set of points in parameter space, whereas the less strict
one demands it only for a single point. For an illustration of why this
stricter condition is necessary, see \autoref{sec:vpd_recon_motivation}.

VPD works on the level of rank-1 subcomponents instead of full
components, but the same principle applies.\footnote{Demanding that
  every combination of causally unimportant subcomponents is ablatable
  is actually stricter than demanding that every combination of causally
  important components is ablatable. See \autoref{sec:limitations} for
  some discussion of this.} To check whether subcomponents are
ablatable, we define ablation masks \(m^l_{b,t,c}\in[g^l_{b,t,c},1]\)
for each subcomponent at each batch index \(b\) and sequence position
\(t\). So, if a subcomponent has causal importance \(g^l_{b,t,c}=1\),
the only permitted value for the mask \(m^l_{b,t,c}\) is also \(1\),
whereas if the causal importance is \(0\), its mask can take any value
between \(0\) and \(1\). These masks define new weight matrices
\(W^{\prime l}_{b,t}\) which we should be able to insert in place of the
original model matrices \(W^l\) without substantially changing the
model's final output.

We operationalize this by demanding that the KL-divergence \(D\) between
the model output on the original forward pass and on forward passes
using the masked weights should be small:

\begin{equation}
\begin{aligned}
  \mathcal{L}_{\text{masked-recon}}
  &=
  \frac{1}{B}
  \sum^{B}_{b=1}
    D
    \Big(
        f(
          \vec{x}_b
          \vert
            W^1,\dots,W^L
        )
      ,
        f(
          \vec{x}_b
          \vert
            {W'}^1_b(
              m^1
            ),\dots,{W'}^L_b(
              m^L
            )
        )
    \Big)
   \\
  \end{aligned}
\label{eq:random_recon}
\end{equation}

Ideally, we would calculate this masked reconstruction loss for every
permitted combination of ablation masks \(m\) for all
subcomponents\footnote{And Delta components.} in all the model's weight
matrices, but this would require performing an intractably large number
of forward passes. So we instead use ablation masks \(m\) drawn using
two types of sampling:

\begin{enumerate}
\def\labelenumi{\arabic{enumi}.}
\tightlist
\item
  \textbf{Stochastic sampling}, with ablation masks \(m^{\text{stoch}}\)
  drawn from uniform distributions. This yields the \emph{stochastic
  reconstruction loss}, \(\mathcal{L}_{\text{stochastic-recon}}\).
\item
  \textbf{Adversarial sampling}, using ablation masks \(m^{\text{adv}}\)
  optimized via gradient ascent to maximise the reconstruction loss.
  This yields the \emph{adversarial reconstruction loss},
  \(\mathcal{L}_{\text{adversarial-recon}}\).
\end{enumerate}

For details on the stochastic and adversarial sampling, see
\autoref{sec:recon}.

\subsection{Optimizing for simplicity}\label{optimizing-for-simplicity}

\label{sec:methods-simplicity}

Each component ought to contain as little computational machinery as
possible. Otherwise, we could say that the target model is one big
parameter component, and proclaim our decomposition complete without
doing any actual decomposition!

We both constrain and train our subcomponents to be simple. Our
subcomponents are rank-one, which constrains them to be simpler objects
than full matrices. Unfortunately, this is not enough of a simplicity
constraint, because some rank-one solutions can be ``simpler'' than
others: In some situations, it is possible to add multiple subcomponents
parametrizing independent mechanisms used on disjoint subsets of the
data together and have the resulting sum also be rank-one.\footnote{We
  observed indications that some VPD decompositions suffered from this
  failure mode. Sometimes, subcomponents seemed to be involved in
  multiple (usually two) unrelated computations, which depended on
  whether the incoming activations had strong positive or negative inner
  products with the subcomponent's right singular vector.} \footnote{A
  theoretically clean motivating example of this phenomenon is the toy
  model of ping pong superposition \citep{gibson2025}. In the ping pong
  superposition construction, \(64\) superposed rank \(1\) circuits can
  be implemented in layers of width \(21\). Only one circuit is ever
  active at a time, and groups of eight circuits each share the exact
  same origin or target neurons. Subcomponents for circuits in the same
  group can then be summed, and the result will again be exactly
  representable as a rank \(1\) matrix, which is causally important for
  computing the output exactly when any of the circuits in the group are
  causally important for computing the output. Hence if we apply VPD to
  this toy model, the importance minimality loss alone will provide no
  incentive to further separate the eight rank \(1\) matrices for the
  eight circuit groups into \(64\) rank \(1\) matrices for the \(64\)
  individual circuits, leaving us with components that activate
  polysemantically and contain more computational machinery than they
  need to.}

We therefore encourage breaking up subcomponents into multiple that are
causally important on as few data points as possible by introducing an
additional, slightly superlinear, penalty on subcomponent activation
frequency:

\begin{equation}
\begin{aligned}
  \mathcal{L}_{\text{frequency-minimality}}
  =
  \frac{1}{B T}
  \sum^{B}_{b=1}\sum^{T}_{t=1}\sum^L_{l=1}\sum^C_{c=1}
  \vert g^l_{b,t,c} \vert^p
    \log_2(
    1 +
    \sum^{B}_{b'=1}\sum^{T}_{t'=1}
    \vert
      g^l_{b',t',c}
    \vert^p
  ),
  \end{aligned}
\label{eq:freq_minimality}
\end{equation}

There are probably multiple ways to optimize for the computational
simplicity of parameter subcomponents, and we are not confident this
choice is optimal (nor our choices for the other losses). Nonetheless,
we found it to work well enough in practice. See
\autoref{sec:vpd_frequency_penalty} for a more detailed motivation of
this loss.

\subsection{Summary of loss terms}\label{summary-of-loss-terms}

\label{sec:methods-summary}

In total, our loss function has five terms:

\[\begin{aligned}
\mathcal{L}_{\text{VPD}} ={}
  & \beta_1 \mathcal{L}_{\text{adversarial-recon}} \\
  + & \beta_2 \mathcal{L}_{\text{stochastic-recon}} \\
  + & \beta_3 \mathcal{L}_{\text{importance-minimality}} \\
  + & \beta_4 \mathcal{L}_{\text{frequency-minimality}} \\
  + & \beta_5 \mathcal{L}_{\text{Delta-L2}}
\end{aligned}\]

They each optimize the parameter subcomponents to exhibit particular
properties:

\begin{itemize}
\tightlist
\item
  The \(\mathcal{L}_{\text{adversarial-recon}}\) and
  \(\mathcal{L}_{\text{stochastic-recon}}\) losses optimize for
  \textbf{mechanistic faithfulness} (\autoref{eq:random_recon}).
\item
  The \(\mathcal{L}_{\text{importance-minimality}}\) loss optimizes for
  \textbf{minimality} (\autoref{eq:minimal}).
\item
  The \(\mathcal{L}_{\text{frequency-minimality}}\) loss optimizes
  subcomponents for \textbf{simplicity}. They are also constrained to be
  rank-1 matrices, which imposes one aspect of simplicity
  (\autoref{eq:freq_minimality}).
\item
  The \(\mathcal{L}_{\text{Delta-L2}}\) auxiliary loss optimizes for
  \textbf{parameter-faithfulness}, even without the
  \(\Delta\)-components, which ensure it (\autoref{eq:delta_l2}).
\end{itemize}

The key difference between VPD and our previous work
\citep{bushnaq2025spd} is the \(\mathcal{L}_{\text{adversarial-recon}}\)
and \(\mathcal{L}_{\text{frequency-minimality}}\) losses. There are
several other, smaller differences that do not fundamentally change the
method but that we found helpful for decomposing language models. For
more details, see \autoref{sec:vpd_methods}.

We evaluate the quality of our decomposition on a number of key metrics.
For assessing the quality of a decomposition, the most important are
\(\mathcal{L}_{\text{adversarial-recon}}\) and \(L_0\) per datapoint.
For readers looking for practical advice on how to tune hyperparameters
and key optimization metrics, we provide a detailed \emph{Training
recipe for VPD} in \autoref{app:recipe}.

\section{Analyzing language model parameter
subcomponents}\label{analyzing-language-model-parameter-subcomponents}

\subsection{Target language model}\label{target-language-model}

\label{sec:language-model-details}

We trained a four-layer 67M parameter decoder-only transformer model on
an uncopyrighted subset of The Pile \citep{gao2020pile}. A summary of
the model architecture and training results can be found in
\autoref{tab:model-hyperparams} and full training details of our target
model can be found in \autoref{app:training-details}.

\begin{figure}[htbp]
\centering
\includegraphics[width=0.85\textwidth,height=0.8\textheight,keepaspectratio]{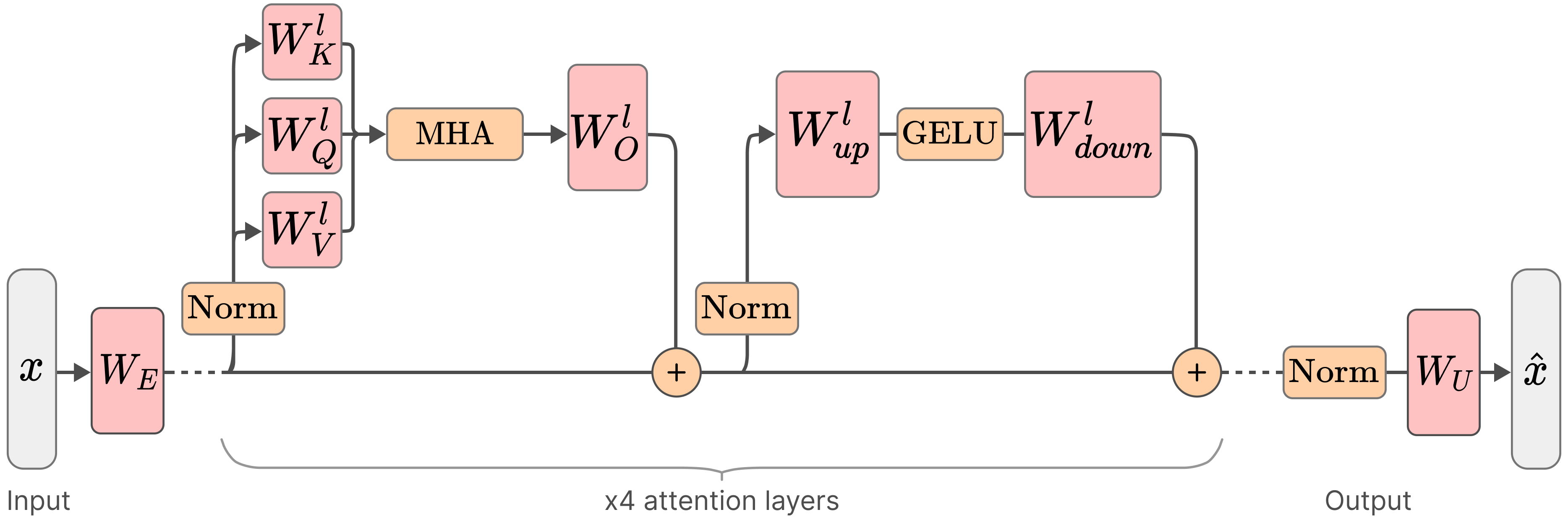}
\caption[]{Our target model is a standard decoder-only transformer language model.}\label{fig:placeholder}
\end{figure}

\begin{longtable}[]{@{}
  >{\raggedright\arraybackslash}p{(\linewidth - 2\tabcolsep) * \real{0.5000}}
  >{\raggedright\arraybackslash}p{(\linewidth - 2\tabcolsep) * \real{0.5000}}@{}}
\caption{\label{tab:model-hyperparams}}\tabularnewline
\toprule\noalign{}
\begin{minipage}[b]{\linewidth}\raggedright
Attributes of our target model
\end{minipage} & \begin{minipage}[b]{\linewidth}\raggedright
\end{minipage} \\
\midrule\noalign{}
\endfirsthead
\toprule\noalign{}
\begin{minipage}[b]{\linewidth}\raggedright
Attributes of our target model
\end{minipage} & \begin{minipage}[b]{\linewidth}\raggedright
\end{minipage} \\
\midrule\noalign{}
\endhead
\bottomrule\noalign{}
\endlastfoot
Layers & 4 \\
Residual stream dimension & 768 \\
MLP intermediate dimension & 3072 \\
Attention heads & 6 \\
Attention head dimension & 128 \\
Context length & 512 \\
Vocabulary size & 50,277 \\
Positional encoding & RoPE \citep{su2024roformer} \\
Normalization & RMSNorm \citep{zhang2019rootmeansquarelayer} \\
Activation function & GELU \citep{hendrycks2016gelu} \\
Attention type & Standard Multi-Head Attention
\citep{vaswani2017attention} \\
Tied embeddings & Yes \\
Non-embedding parameters & \textasciitilde28M \\
Total parameters (incl.~embedding) & \textasciitilde67M \\
Training dataset & The Pile \citep{gao2020pile} (subset) \\
\end{longtable}

We decomposed the 24 weight matrices in this model into a total of of
38,912 rank \(1\) subcomponents. We omitted the embedding and
unembedding matrices. The decomposition used much fewer than its full
capacity, having only \textasciitilde10,000 alive components (with a
mean causal importance greater than \(10^{-6}\)).

On average, each datapoint uses 205 subcomponents per sequence position,
representing 2.1\% of all alive subcomponents.
\autoref{tab:num-components-per-layer} shows per-layer summary
statistics for the decomposition.

\begin{longtable}[]{@{}lllll@{}}
\caption{Table: Per-layer decomposition summary statistics: Subcomponent
dictionary sizes \(C\); alive subcomponents (subcomponents with mean
causal importances above \(10^{-6}\) at the end of training); average
\(L_0\) scores of subcomponents with causal importance \(>0\) per batch
and sequence position; and fraction of all subcomponents with causal
importance \(>0\) per batch and sequence position.
\label{tab:num-components-per-layer}}\tabularnewline
\toprule\noalign{}
Layer & \(C\) & Alive & Mean L0 & L0/Alive \\
\midrule\noalign{}
\endfirsthead
\toprule\noalign{}
Layer & \(C\) & Alive & Mean L0 & L0/Alive \\
\midrule\noalign{}
\endhead
\bottomrule\noalign{}
\endlastfoot
Layer 0 & \(9728\) & \(3709\) & \(44.6\) & \(0.012\) \\
Layer 1 & \(9728\) & \(848\) & \(18.9\) & \(0.022\) \\
Layer 2 & \(9728\) & \(1943\) & \(49.5\) & \(0.025\) \\
Layer 3 & \(9728\) & \(3472\) & \(92.0\) & \(0.026\) \\
Total & \(38912\) & \(9972\) & \(205.0\) & \(0.021\) \\
\end{longtable}

\subsection{The decomposition model behaves similarly to the target
model}\label{the-decomposition-model-behaves-similarly-to-the-target-model}

\label{sec:approx-target-well}

If a decomposition method has correctly identified the mechanisms
underlying a model's computation, then activating only the mechanisms
that the method identifies as causally important on a given input should
approximately reproduce the model's behavior on that input. Conversely,
if a replacement model fails to reproduce the model's behavior, then the
decomposition has either missed important mechanisms or identified
spurious ones. Reconstruction quality is therefore a necessary (though
not sufficient) condition for a decomposition to be mechanistically
faithful.

Our parameter subcomponents capture different amounts of the target
model's performance depending on how masks are calculated
(\autoref{tab:vpd-ce-compute-compar}). One quantitative measure of
performance is cross-entropy (CE) loss on the validation set: The
decomposed model achieves between 2.72 and 3.02, depending on the type
of sampling, compared with 2.71 for the target model.

A metric that is sometimes helpful for comparison is \emph{Pretraining
Compute Recovered} \citep{gao2024scalingevaluatingsparseautoencoders},
which is the percentage of the target model's total pretraining compute
at which the target model's training curve reaches the same validation
CE loss as the reconstruction model (i.e.~a value of X\% means the
reconstruction performs no better than the target model did when only
X\% of pretraining was complete).

When we exclude the \(\Delta\)-component (which is trained to be as
causally unimportant as possible), the remaining \emph{unmasked}
parameter subcomponents recover about \(82\%\) of the pretraining
compute. When using stochastic ablations, this drops to around \(27\%.\)

\begin{longtable}[]{@{}
  >{\raggedright\arraybackslash}p{(\linewidth - 4\tabcolsep) * \real{0.4646}}
  >{\raggedleft\arraybackslash}p{(\linewidth - 4\tabcolsep) * \real{0.2020}}
  >{\raggedright\arraybackslash}p{(\linewidth - 4\tabcolsep) * \real{0.3333}}@{}}
\caption{\label{tab:vpd-ce-compute-compar}}\tabularnewline
\toprule\noalign{}
\begin{minipage}[b]{\linewidth}\raggedright
Masking mode (excluding \(\Delta\)-components)
\end{minipage} & \begin{minipage}[b]{\linewidth}\raggedleft
Validation CE Loss
\end{minipage} & \begin{minipage}[b]{\linewidth}\raggedright
Pretraining Compute Recovered (\%)
\end{minipage} \\
\midrule\noalign{}
\endfirsthead
\toprule\noalign{}
\begin{minipage}[b]{\linewidth}\raggedright
Masking mode (excluding \(\Delta\)-components)
\end{minipage} & \begin{minipage}[b]{\linewidth}\raggedleft
Validation CE Loss
\end{minipage} & \begin{minipage}[b]{\linewidth}\raggedright
Pretraining Compute Recovered (\%)
\end{minipage} \\
\midrule\noalign{}
\endhead
\bottomrule\noalign{}
\endlastfoot
\textbf{Target Model} & \textbf{2.71} & \textbf{100\%} \\
Unmasked (All masks(=)1) & 2.72 & 82.4\% \\
Stochastic Masks & 2.84 & 26.9\% \\
Rounded Masks (Mask(=)1 if CI(\textgreater)0) & 2.94 & 11.8\% \\
Rounded Masks (Mask(=)1 if CI(\textgreater)0.1) & 2.95 & 11.3\% \\
Causal Importance values (CIs) used as Masks & 2.99 & 9.4\% \\
Rounded Masks (Mask(=)1 if CI(\textgreater)0.5) & 3.02 & 8.0\% \\
\end{longtable}

Pretraining compute recovered is rarely reported, so comparisons to
other methods are difficult. Nonetheless, VPD compares favorably to the
only other method in the literature that we are aware of that reports
this metric: Top-\(k\) SAEs
\citep{gao2024scalingevaluatingsparseautoencoders} reports a pretraining
compute recovered of \(10\%\) when replacing a single layer of GPT-4
with an SAE with 16 million latents. By comparison, even though our
approach decomposes the whole model rather than just a single layer, it
recovers between \(8\%\) and \(27\%\), depending on the ablation method
used\footnote{Note, however, that comparing this metric across models
  assumes similar levels of compute optimality for both decomposed
  models, which may not be the case.}.

The table below shows KL-divergence to the target model under
adversarial masking with different numbers of adversarial optimization
steps, calculated across a batch of \(128\) of sequence length \(512\)
drawn from the evaluation set \footnote{The adversarial masks were
  calculated with Projected Gradient Descent (PGD)
  \citep{madry2018towards} optimization, sharing the same source for
  each subcomponent across the batch. For more details on the PGD loss
  evaluation metric see \autoref{sec:vpd_methods-adv}.}.

\begin{longtable}[]{@{}
  >{\raggedright\arraybackslash}p{(\linewidth - 4\tabcolsep) * \real{0.4384}}
  >{\raggedleft\arraybackslash}p{(\linewidth - 4\tabcolsep) * \real{0.1507}}
  >{\raggedright\arraybackslash}p{(\linewidth - 4\tabcolsep) * \real{0.4110}}@{}}
\caption[\label{tab:vpd-pgd-ce} While the decomposition is somewhat
robust to approximately \(20\) steps of adversarial optimization, it is
clearly not at all robust to \(160\) steps or
more.]{\label{tab:vpd-pgd-ce} While the decomposition is somewhat robust
to approximately \(20\) steps of adversarial optimization, it is clearly
not at all robust to \(160\) steps or more.\footnote{To provide some
  sense of scale, zero-ablating all of the target model's weight
  matrices gives a KL-divergence of ca. \(67.2449\).}}\tabularnewline
\toprule\noalign{}
\begin{minipage}[b]{\linewidth}\raggedright
Adversarial optimization steps \(n^{\text{adv}}\)
\end{minipage} & \begin{minipage}[b]{\linewidth}\raggedleft
KL divergence to target model
\end{minipage} & \begin{minipage}[b]{\linewidth}\raggedright
\end{minipage} \\
\midrule\noalign{}
\endfirsthead
\toprule\noalign{}
\begin{minipage}[b]{\linewidth}\raggedright
Adversarial optimization steps \(n^{\text{adv}}\)
\end{minipage} & \begin{minipage}[b]{\linewidth}\raggedleft
KL divergence to target model
\end{minipage} & \begin{minipage}[b]{\linewidth}\raggedright
\end{minipage} \\
\midrule\noalign{}
\endhead
\bottomrule\noalign{}
\endlastfoot
20 & 0.8280 & \\
40 & 1.3539 & \\
80 & 3.8381 & \\
160 & 25.2560 & \\
320 & 40.2200 & \\
\end{longtable}

However, we note that \emph{complete} adversarial robustness would not
necessarily be desirable. See \autoref{sec:vpd_recon_motivation} for
some discussion of how much adversarial robustness a decomposition ought
to exhibit to be considered mechanistically faithful.

\begin{figure}[htbp]
\centering
\includegraphics[width=\textwidth,height=0.8\textheight,keepaspectratio]{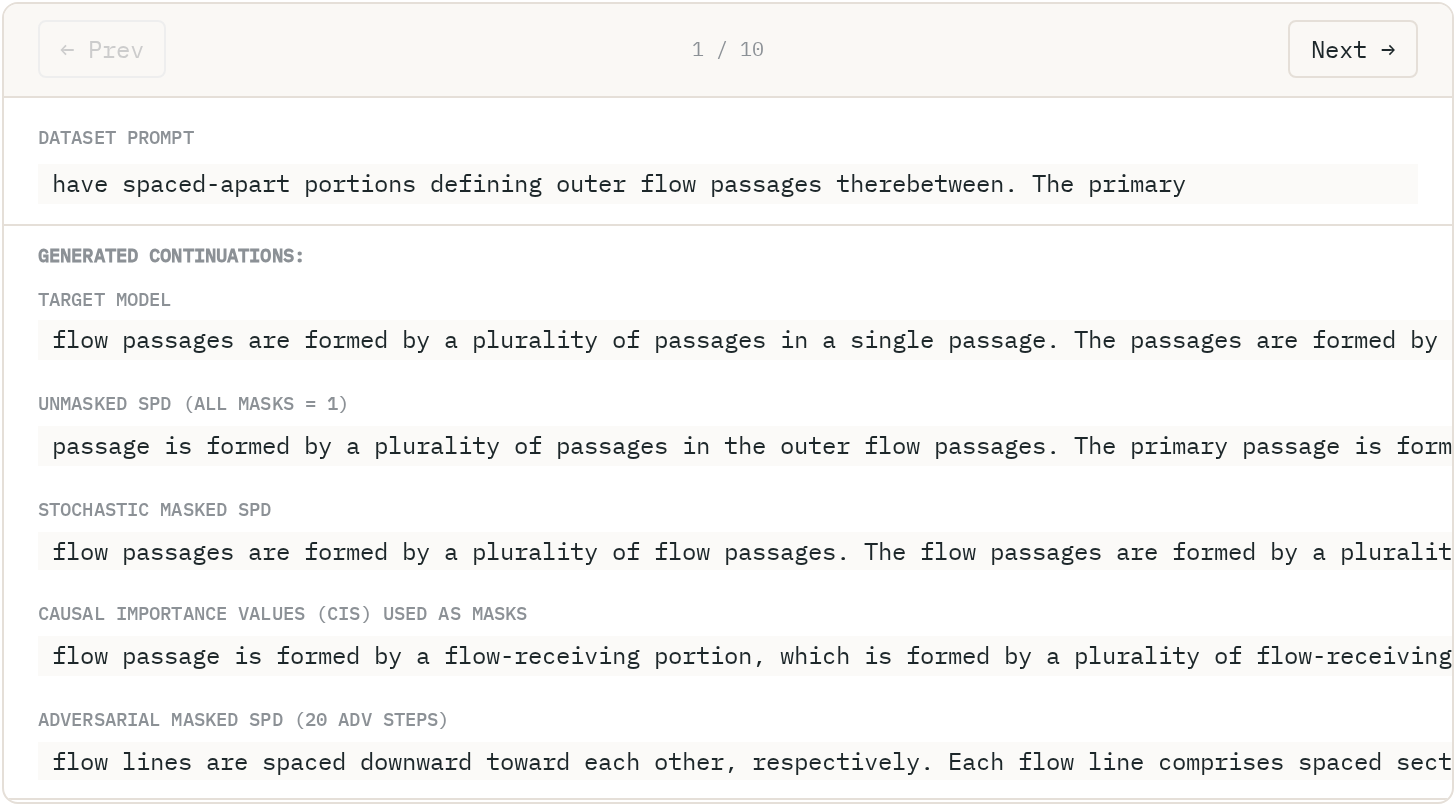}
\caption[]{Side-by-side generation comparisons across masking strategies. \emph{Static snapshot of an interactive figure; explore it at \href{https://www.goodfire.com/research/interpreting-lm-parameters\#fig:generations-showcase}{the web version}.}}\label{fig:generations-showcase}
\end{figure}

Qualitatively, the generations produced by different sampling methods
align with the above quantitative measures. The generations seem
qualitatively to produce similar behavior to the target model in most
cases (\autoref{fig:generations-showcase}).

Surprisingly, even when masks are adversarially sampled with 20 steps of
adversarial optimization, the generations are not \emph{entirely}
nonsensical. This is feasible because we only get to adversarially
sample causally unimportant parameter subcomponents.

\subsection{VPD has a better tradeoff between reconstruction versus
sparsity compared with
transcoders}\label{vpd-has-a-better-tradeoff-between-reconstruction-versus-sparsity-compared-with-transcoders}

\label{sec:decomp-model-behav-sim}

Any decomposition of a neural network faces a fundamental tradeoff
between the number of `objects' they use to reconstruct the network's
behavior and the quality of that reconstruction. If a decomposition can
use fewer objects to capture the same amount of network performance,
then that explanation is preferred according to Occam's razor, assuming
the objects use a similar amount of computational machinery.

We study the reconstruction versus sparsity tradeoffs of different
decompositions and compare the VPD model with two families of
activation-based decomposition methods: Per-layer transcoders (PLTs)
\citep{dunefsky2024transcodersinterpretablellmfeature} and cross-layer
transcoders (CLTs) \citep{lindsey2024crosscoders}, both using BatchTopK
\citep{bussmann2024batchtopk}. We simultaneously replace all 4 MLP
layers of the target model with their sparse reconstructions and measure
the resulting increase in cross-entropy loss relative to the unmodified
target model.

There isn't a straightforward apples-to-apples comparison between
transcoder latents and VPD subcomponents, so we present a number of
different comparisons (with more extensive experimental details in
\autoref{app:vpd-sparsity-acc-tradeoff}) \footnote{Comparing sparsity
  across methods requires care, because each method has structurally
  different notion of what constitutes a single active element. A CLT
  feature writes to the residual stream at every layer simultaneously,
  while a PLT latent affects only one layer. VPD subcomponents are
  scoped to individual weight matrices, and each MLP layer has two such
  matrices (\(W_{\text{in}}\) and \(W_{\text{out}}\)).}. To ensure our
conclusions are not artifacts of how we count subcomponents or latents,
we show results under three possible definitions of sparsity:

\begin{enumerate}
\def\labelenumi{\arabic{enumi}.}
\tightlist
\item
  \textbf{Average active subcomponents per module}: Active encoder
  latents for PLTs/CLTs; active subcomponents per weight matrix for VPD;
\item
  \textbf{Active subcomponents per MLP Down reconstruction}: Adjusting
  for the fact that a CLT latent affects multiple layers and that VPD
  uses two modules per MLP;
\item
  \textbf{Total active parameters}: VPD's rank-one subcomponents have
  more parameters than a PLT latent and a single CLT latent has multiple
  decoder vectors.
\end{enumerate}

We compare VPD with PLTs and CLTs trained with their standard training
losses, noting these are different objectives (VPD trains on output
reconstruction while PLTs and CLTs are trained to reconstruct
activations at each layer).

\begin{figure}[htbp]
\centering
\includegraphics[width=\textwidth,height=0.8\textheight,keepaspectratio]{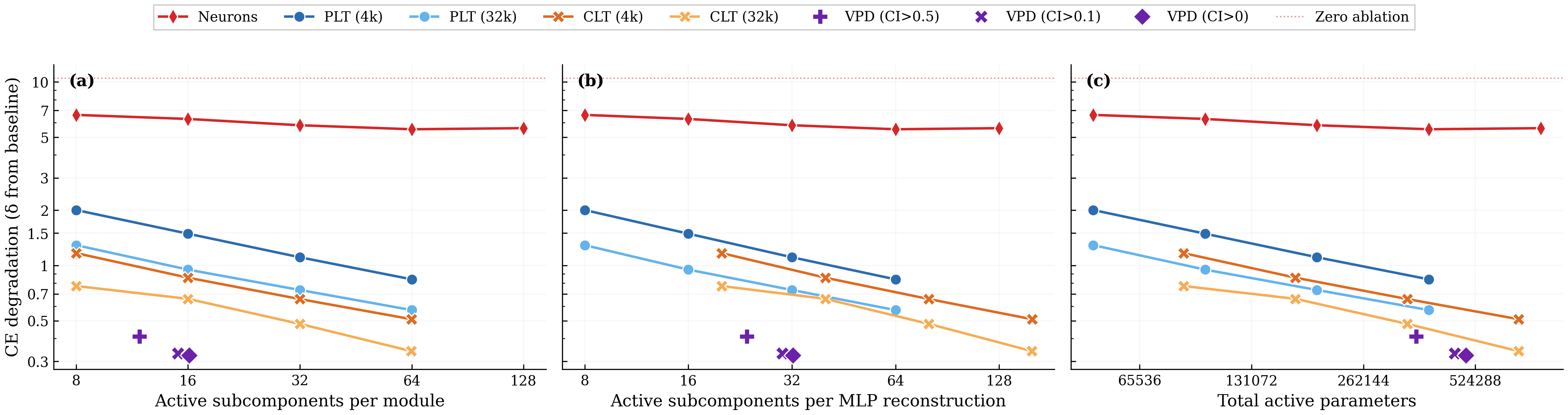}
\caption[]{CE degradation when simultaneously replacing all 4 MLP layers with sparse reconstructions from each method. \textbf{(a)} Active subcomponents per module (raw L0). \textbf{(b)} Active subcomponents per MLP reconstruction, adjusting for CLT's cross-layer writes and VPD's paired modules. \textbf{(c)} Total active parameters. VPD (purple markers) Pareto-dominates the activation-based methods under all three sparsity measures. The dashed line indicates zero-ablation (all MLP outputs set to zero). Lower is better.}\label{fig:pareto-mse}
\end{figure}

We observe that VPD performs favorably compared with activation-based
decomposition, achieving less CE degradation for a given \(L_0\) across
all three definitions of sparsity.

We noted above that VPD and the transcoders differ in training
objective. VPD is trained end-to-end, whereas activation-based
approaches are usually trained layerwise. This complicates direct
comparison and arguably makes the above analysis somewhat unfair to
activation-based methods. We address this by also comparing under
matched objectives in \autoref{app:vpd-sparsity-acc-tradeoff} and find
that VPD compares favorably to other methods: When trained and evaluated
on a range of objectives, VPD's Pareto domination disappears, but it
avoids overfitting to its particular training objective, unlike the
activation-based methods.

Additional figures and training logs for the VPD decomposition can be
found at the WandB link
\href{https://wandb.ai/goodfire/spd/runs/s-55ea3f9b}{here}.

\subsection{Parameter subcomponents are highly
interpretable}\label{parameter-subcomponents-are-highly-interpretable}

\label{sec:param-comps-interpretable}

In order to study a parameter subcomponent's role in the network's
neural algorithm, we need a definition of what it means for it to be
`active' on a given datapoint.

There are at least two reasonable definitions:

\begin{enumerate}
\def\labelenumi{\arabic{enumi}.}
\tightlist
\item
  \textbf{Causal importance}: The causal importance function is trained
  to output a value between \(0\) and \(1\) that tells us exactly how
  important a particular subcomponent is on a datapoint. It tells us if
  the subcomponent is `necessary' or `required' or `used' on that input.
  In many ways, this is a perfect definition of `active'! However, it is
  not a `local' measure of a subcomponent's activation: A subcomponent
  with a small causal importance value might interact strongly with the
  activations at a layer, only for its effect to be suppressed later by
  others. For a more `local' measure, we use the next definition.
\item
  \textbf{Subcomponent activation}: We define the subcomponent
  activation as
  \[a_c^l = ||\vec{U}^l_c|| (\vec{V}^l_c)^\top \vec{\varphi}^l,\] where
  \(\vec{\varphi}^l\) are the model's hidden activations before matrix
  \(l\) \footnote{We multiply by \(||\vec{U}^l_c||\) because neither the
    \(\vec{U}\) or \(\vec{V}\) vectors are normalized by default, and we
    therefore need to multiply by this norm to make their subcomponent
    activations comparable.}. This defines how much the activations
  interact with a given subcomponent, even if that interaction
  ultimately ends up not being causally important for the output. Due to
  superposition
  \citep{olshausen1997sparseovercomplete, goh2016decoding, elhage2022toy, Vaintrob_Mendel_Kaarel_2024, Bushnaq_Mendel_2024},
  there will be more interactions in general than there are causally
  important interactions.
\end{enumerate}

Throughout this paper, we use both definitions, highlighting which type
of activation we mean in each instance.

We find that parameter subcomponents tend to `activate' (in both senses)
for coherent categories of inputs. \autoref{fig:components-showcase}
shows some dataset examples on which each subcomponent is causally
important. It also shows the subcomponent activation in the underlines.
You can navigate the panel to explore the activations of a variety of
parameter subcomponents:

\begin{figure}[htbp]
\centering
\includegraphics[width=\textwidth,height=0.8\textheight,keepaspectratio]{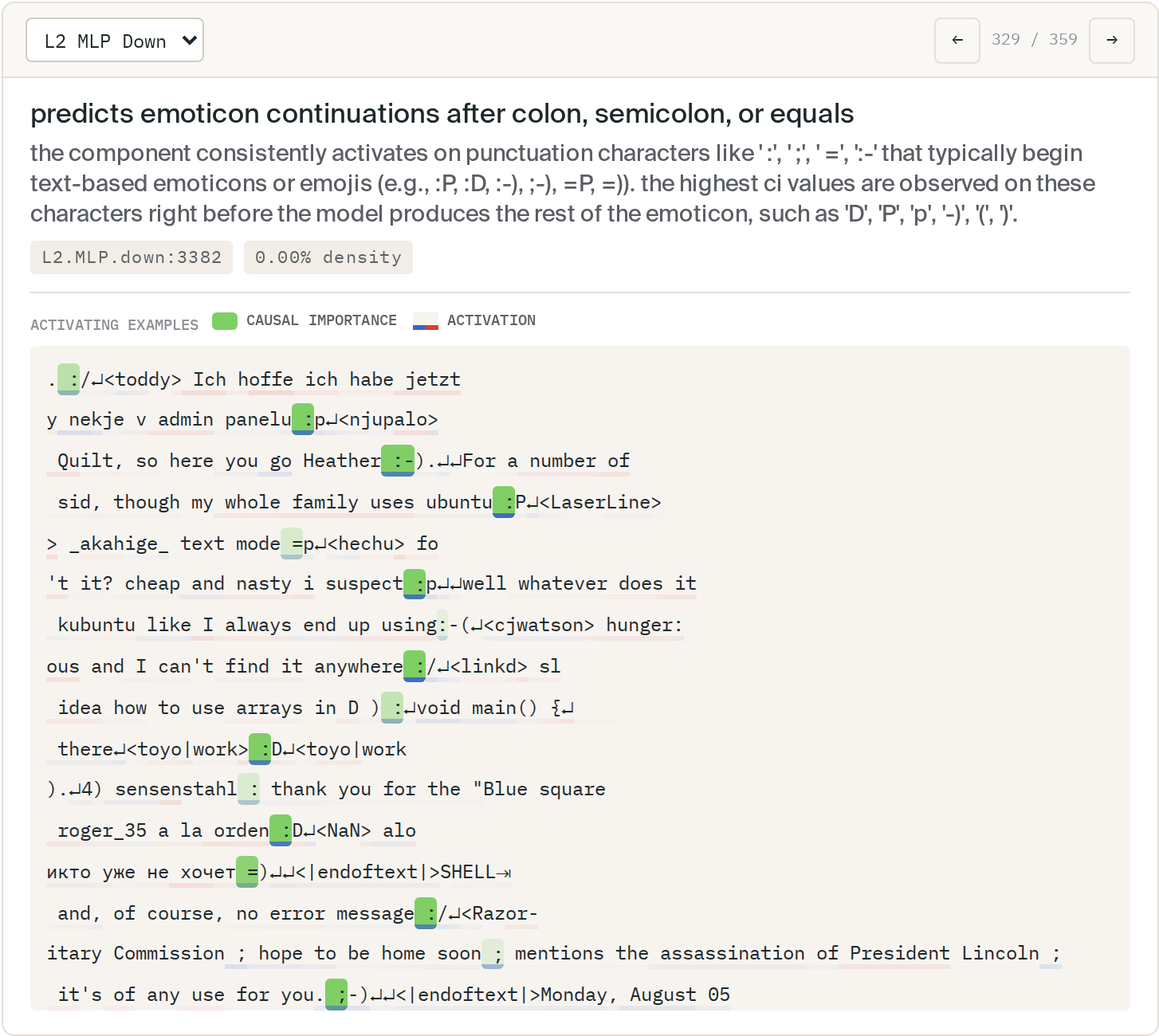}
\caption[]{Browse all VPD parameter subcomponents by weight matrix. Green highlights indicate causal importances; colored underlines show subcomponent activations. \emph{Static snapshot of an interactive figure; explore it at \href{https://www.goodfire.com/research/interpreting-lm-parameters\#fig:components-showcase}{the web version}.}}\label{fig:components-showcase}
\end{figure}

To compare how `interpretable' parameter subcomponents are relative to
transcoder latents, we can measure how semantically coherent a
subcomponent's activation patterns are using \emph{intruder detection}
\citep{chang2009reading, paulo2025evaluating}. In intruder detection, we
present an LLM-judge with a set of inputs that activate a given VPD
subcomponent or transcoder latent alongside one `intruder' example that
does not activate it. We task the LLM-judge to identify the intruder
example. It should be easier to identify the intruder among a more
semantically coherent set of inputs. In the VPD setting, we use causal
importance values in place of activation magnitudes and select intruder
examples with similar activation densities.

\begin{figure}[htbp]
\centering
\includegraphics[width=0.85\textwidth,height=0.8\textheight,keepaspectratio]{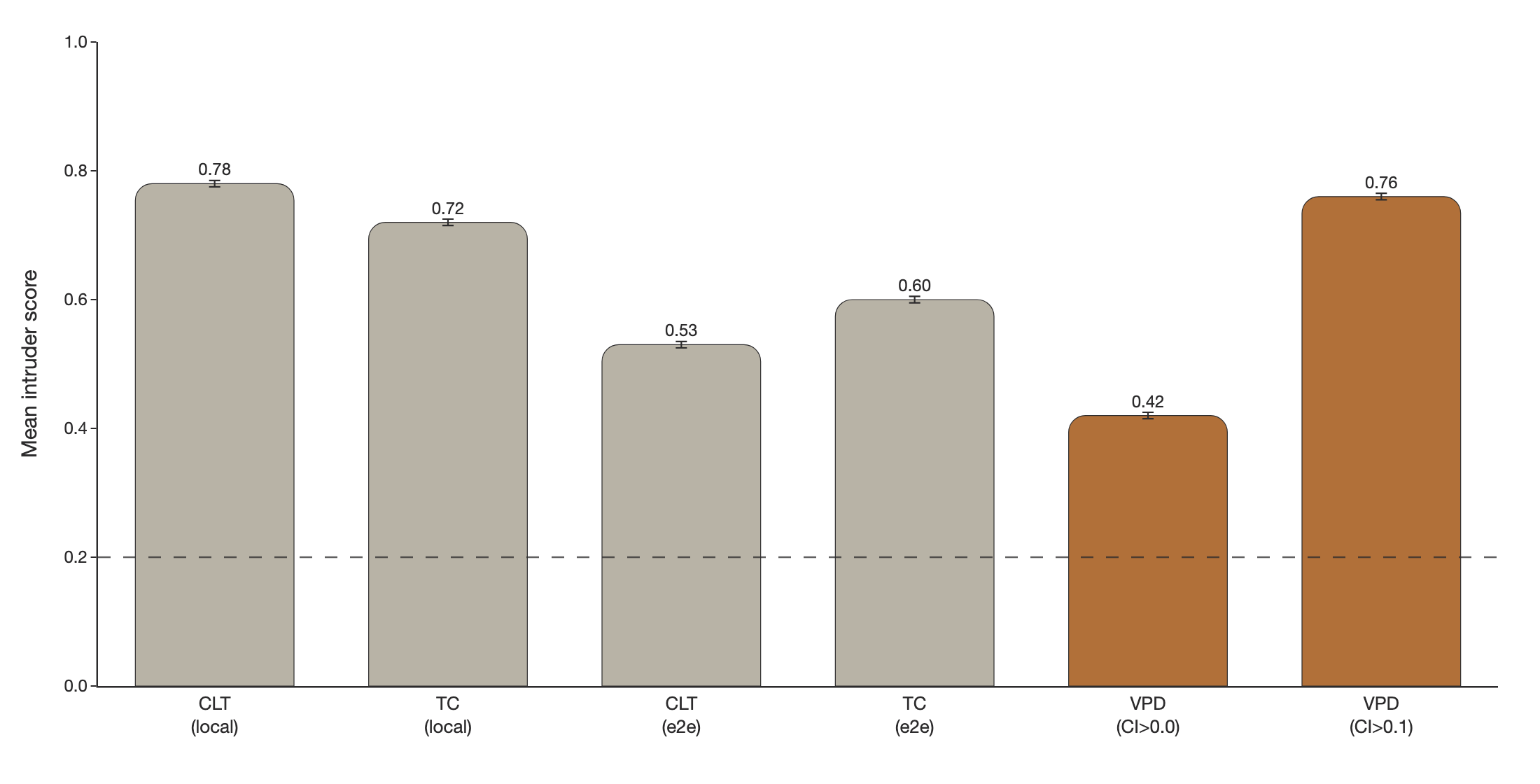}
\caption[]{Intruder detection scores for various CLT and PLT latents, and VPD subcomponents at different CI thresholds. Error bars are 95\% bootstrap CIs on the mean. Dashed line is random chance accuracy (20\%). Higher is better.}\label{fig:intruder-score}
\end{figure}

We find VPD intruder detection scores improve drastically when using CI
values thresholded with 0.1, which filters low-CI noise
\autoref{fig:intruder-score}. We think that filtering out small causal
importances is justifiable, since 0.1-rounded performance has
essentially the same performance as 0.0-rounded performance, suggesting
that very little performance is captured by subcomponents with small
activations (\autoref{tab:vpd-ce-compute-compar}).

We observe that 0.1-rounded VPD subcomponents score competitively with
CLTs and PLTs trained using a local (layerwise) MSE activation
reconstruction loss \autoref{fig:intruder-score}. VPD subcomponents are
more coherent than PLTs and CLTs that were trained end-to-end.

\subsection{VPD does not suffer from feature
splitting}\label{vpd-does-not-suffer-from-feature-splitting}

\label{sec:splitting}

Feature splitting is a well-known issue in activation-based dictionary
learning methods such as PLTs, SAEs, and CLTs
\citep{chanin2024absorptionstudyingfeaturesplitting, bricken2023monosemanticity}.
As dictionary size increases, these methods can improve sparsity and
reconstruction by replacing a `broad', reusable latent with several
narrower, more context-specific ones. In the extreme, a transcoder could
assign a unique latent to every individual datapoint in the training
set, effectively memorizing the dataset rather than uncovering reusable,
general patterns.

VPD does not suffer from this issue, either in principle or in practice.
The key reason for this is that subcomponents marked as causally
unimportant are required to be ablatable in any combination, not just
all simultaneously. The model therefore needs to be robust to variations
in parameter space along the directions of these subcomponents for all
batches and sequence positions, not just the ones on which they are
causally important. Without this constraint, the decomposition might be
able to invent overly `narrow', context-specific subcomponents that do
not actually exist in the computational structure of the original model
but that sparsely activate while reconstructing the model's behavior on
some narrow subset of the data. For example, suppose VPD attempted to
pathologically decrease \(\mathcal{L}_{\text{importance-minimality}}\)
by splitting a mechanism in the target model that ought to be
parametrised by two subcomponents into many specialised subcomponents
that lie within that mechanisms' two-dimensional subspace, each aligned
with a different training-data hidden activation vector, and marked only
one of them at a time as causally important. If we were just using the
causal importances as masks, this would reconstruct the target model's
output well. But with stochastic or adversarial masking, many of the
subcomponents not marked as causally important will be turned on as
well, making the resulting output activation vector both too large and
pointed in the wrong direction, thus ruining the reconstruction. See
\autoref{sec:vpd_recon_motivation} for further discussion.

To test empirically whether VPD does avoid feature splitting, we
incrementally increase the number of subcomponents used by different VPD
runs and count the number of ``alive'' subcomponents (subcomponents that
activate at least once every 1M tokens). We train VPD at four capacity
levels corresponding to \(0.5\times\), \(1\times\), \(2\times\),
\(4\times\) the subcomponent count of the main decomposition we study.
We compare against PLTs and CLTs at 4k and 32k dictionary sizes.

\begin{figure}[htbp]
\centering
\includegraphics[width=0.85\textwidth,height=0.8\textheight,keepaspectratio]{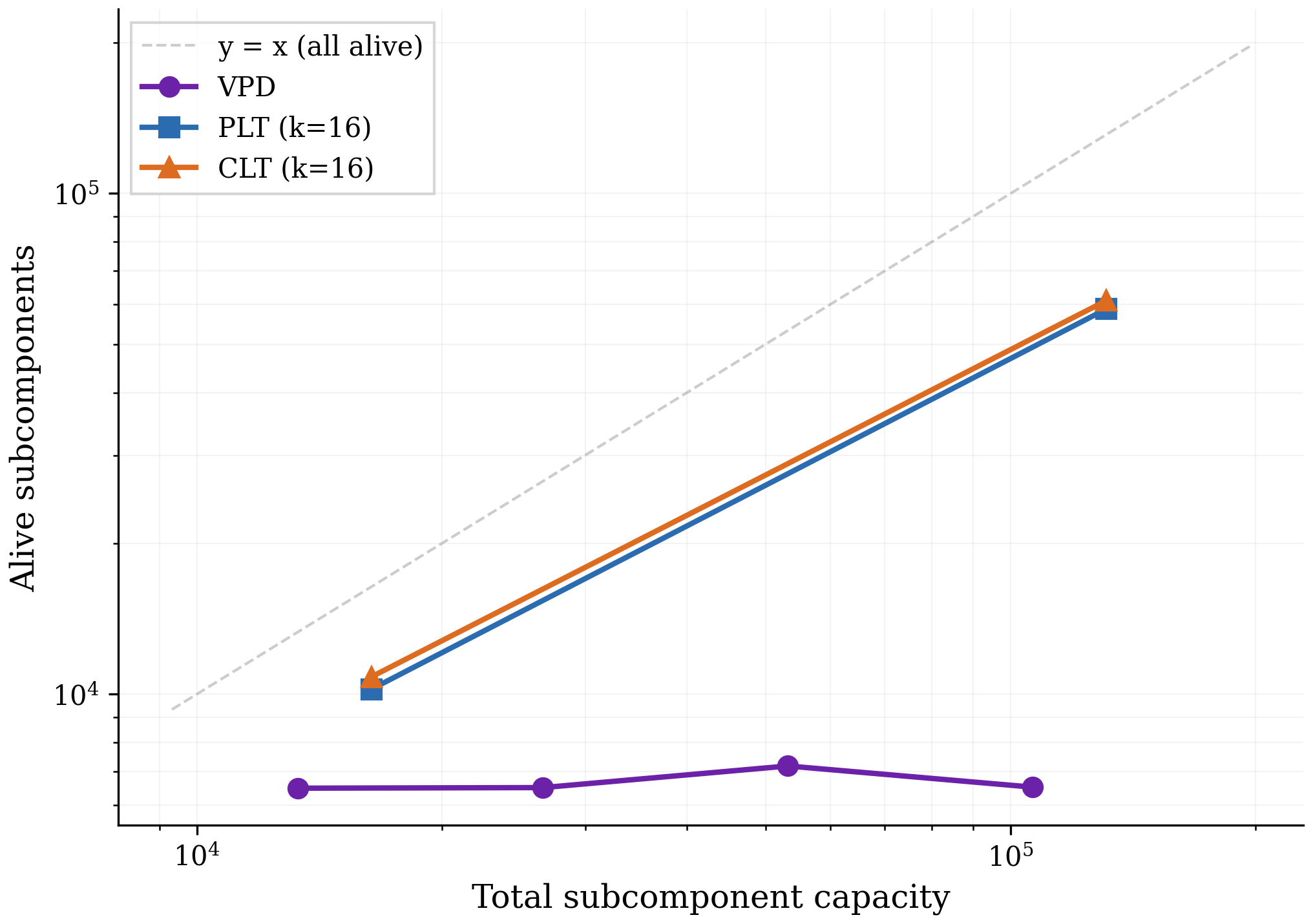}
\caption[]{Number of alive subcomponents as a function of total subcomponent capacity. PLTs and CLTs scale roughly linearly with dictionary size, staying close to the \(y = x\) line. VPD (purple) remains flat at ~6,500-7,000 alive subcomponents regardless of capacity, indicating that additional capacity is not used for feature splitting. Dashed line: \(y = x\) (all subcomponents alive).}\label{fig:feature_splitting}
\end{figure}

\autoref{fig:feature_splitting} shows that, unlike PLTs and CLTs,
increasing VPD's capacity does not increase the number of subcomponents
that the method actually uses, suggesting that feature splitting is not
a significant problem for VPD. Across all four VPD runs the sparsity and
reconstruction performance remain approximately constant, so the flat
alive count reflects unused capacity rather than a tradeoff against
sparsity or reconstruction. In
\autoref{app:confirming-feature-splitting}, we confirm that our PLTs and
CLTs are indeed splitting features rather than discovering genuinely new
ones.

While we only show results for one language model here, we have observed
the same qualitative result in every model we have decomposed with
either VPD or SPD \citep{bushnaq2025spd} despite extensive
hyperparameter sweeps, including various toy models with known ground
truth and a smaller language model trained on the SimpleStories (
\citep{finke2025parameterizedsynthetictextgeneration}) dataset.

\section{Decomposing attention behaviors that are distributed across
attention
heads}\label{decomposing-attention-behaviors-that-are-distributed-across-attention-heads}

\label{sec:attn-analysis}

Transformer language models are significant in large part because they
were the first architecture that enabled scalable sequence modelling.
The crucial component that lets transformers perform computations across
sequences is the attention layer (
\citep{vaswani2017attention, Bahdanauetal2014}).

In most prior work that studies attention layer computations, attention
heads have typically been the primary units of analysis to study
attention behaviors
\citep{vig2019multiscalevis, clark2019doesbertlookat, elhage2021mathematical, olsson2022incontextlearninginductionheads, wang2022interpretability, janiak2023polysemantic, nam2025causalheadgatingframework}.
Unfortunately for interpretability, it is possible for attention layers
to perform computations in a way that is distributed across multiple
heads \citep{jermyn2023attention, jermyn2025attention}\footnote{This
  phenomenon is sometimes called `attention head superposition'.
  However, we prefer to reserve that term for the specific case where
  the attention layer implements more computations than the number of
  heads it distributes them across, which might not happen in general.}.
It would therefore be ideal if our decomposition methods could cope with
attention computations that are distributed across heads. So far, it has
been difficult to find satisfactory activation-based decomposition
methods that can do this
\citep{jermyn2025attention, mathwin2024gated, wynroe2024qkbilinear, Kissane_Conmy_Nanda_2024, kamath2025tracing}.

Fortunately, parameter decomposition methods offer some hope: As we've
seen in \autoref{sec:param-comps-interpretable}, parameter subcomponents
seem to decompose the parameters into specialized functional units. And
since parameter subcomponents are vectors in parameter space, they can
therefore span multiple attention heads!

In this section, we demonstrate that parameter subcomponents in
attention layers are indeed interpretable, and can span multiple
attention heads (and usually do!). Focusing primarily on attention layer
1, we study three attention layer behaviors (`\emph{Previous token
behavior}', `\emph{Previous syntactic boundary movement}', and
`\emph{Detecting Existential vs.~Expletive Constructions}') and show how
parameter subcomponents distribute these computations across heads.

\subsection{Attention layer parameter subcomponents have specific
interpretable
roles}\label{attention-layer-parameter-subcomponents-have-specific-interpretable-roles}

First, we look at a few parameter subcomponents in attention layer 1. In
this layer VPD identifies different numbers of parameter subcomponents
in the \(W_Q\), \(W_K\), \(W_V\), and \(W_O\) matrices. These matrices
have 15, 48, 226, and 97 alive\footnote{These subcomponent numbers
  correspond to the number of components with mean causal importance
  above \(10^{-6}\).} components respectively, though we'll usually
present fewer for simplicity.

There are many interesting subcomponents in these matrices that
correspond to easily interpretable behaviors:

\begin{itemize}
\tightlist
\item
  \texttt{1.attn.q:308} activates on tokens related to existence or the
  verb `to be' and other `copula' verbs.
\item
  \texttt{1.attn.k:485} activates on words that predict `copula' verbs,
  such as \texttt{·there} or \texttt{·it} in ``there is/it is''.
\item
  \texttt{1.attn.k:218} activates on the word \texttt{·it} (including
  capitalized variations and variants both with and without a leading
  space)
\item
  \texttt{1.attn.k:119} activates on punctuation, spaces, brackets,
  newlines and other `interstitial' words.
\item
  \texttt{1.attn.k:290} activates on newlines and end-of-text tokens
  only.
\item
  \texttt{1.attn.v:42} activates on coordinating conjunctions, like
  \texttt{·and}, \texttt{·or}, \texttt{·but} and \texttt{·\&}.
\item
  \texttt{1.attn.v:178} activates on words related to position in time
  and, to a lesser extent, space, like \texttt{·December},
  \texttt{·South}, \texttt{·2002}, \texttt{·long} and \texttt{·far}.
\item
  \texttt{1.attn.o:983} Activates on the introductions or titles of
  texts, particularly scientific papers.
\end{itemize}

Additionally, there are some subcomponents whose role seems more related
to `sequence position' than having a particular semantic meaning:

\begin{itemize}
\tightlist
\item
  \texttt{1.attn.q:149} and \texttt{1.attn.q:497} tend to activate on
  the tokens immediately following the first token of the sequence (and,
  incidentally, reveal some of the shortcomings of our autointerp
  labelling method, which seems to have missed this!).
\item
  \texttt{1.attn.k:315}, \texttt{1.attn.k:357} and \texttt{1.attn.k:121}
  tend only to be causally important on the first few tokens of a
  sequence, though with some exceptions.
\end{itemize}

Together, these interpretations are encouraging, because they suggest
that our decomposition is identifying parts of the network that are
specialized for particular functional roles.

\subsection{Attention layer parameter subcomponents typically span
multiple
heads}\label{attention-layer-parameter-subcomponents-typically-span-multiple-heads}

We've seen evidence that attention subcomponents are specialized for
specific semantic roles, suggesting different computational functions.
Now we investigate whether these subcomponents are `located' in
particular heads.

In our model, the \(W_Q\), \(W_K\), \(W_V\), and \(W_O\) matrices are
concatenated across attention heads. But we can easily split them into
the matrices belonging to individual heads. Even though parameter
subcomponents by default span all heads in a layer, most of their `mass'
could be localized in single heads if their weights in all but one
attention heads have zero norm. But if their parameters have nonzero
norm in multiple heads, then this is weak evidence that they perform
computations across multiple heads.

\begin{figure}[htbp]
\centering
\includegraphics[width=0.85\textwidth,height=0.8\textheight,keepaspectratio]{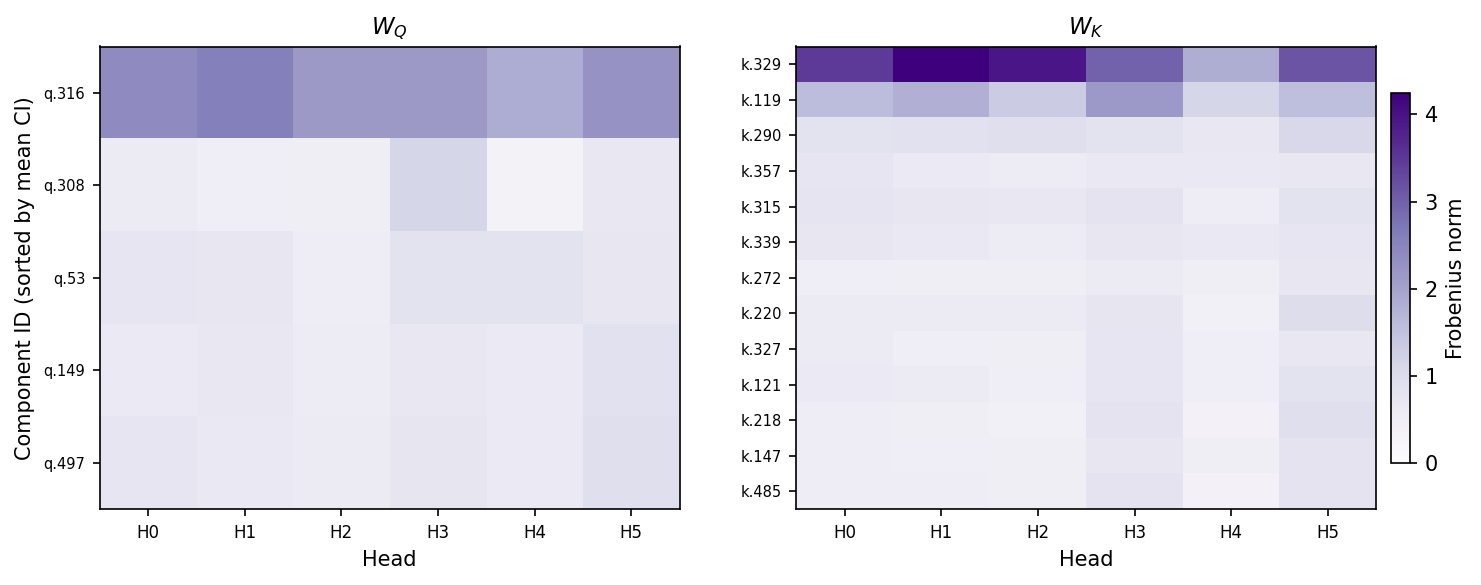}
\caption[]{The norm of the weights of each \(W_Q\) and \(W_K\) subcomponent in each head. No parameter subcomponent is exclusively localized in a single head, suggestive of computations that are distributed across attention heads.}\label{fig:qk_comp_weight_norm}
\end{figure}

We'll focus on the \(W_Q\) and \(W_K\) matrices for now. We see that, in
fact, most \(W_Q\) and \(W_K\) subcomponents have nonzero weight norm
across each head (\autoref{fig:qk_comp_weight_norm}). This suggests that
most \(W_Q\) and \(W_K\) subcomponents might perform computations in a
distributed way! The norms subcomponents of \(W_V\) and \(W_O\) matrices
seem similarly distributed across heads
(\autoref{fig:vo_comp_weight_norm})

While suggestive, this is only indirect evidence of distributed
computations. We would need to understand the computations in order to
confirm that they are indeed distributed across heads. To do this, we
will need new analysis tools. And we can make the problem slightly
easier by separately studying the two main parts of the attention layer:
The QK circuit and the OV circuit \citep{elhage2021mathematical}. We'll
focus on the QK circuit first.

\subsection{The QK circuit consists of interactions between pairs of
parameter
subcomponents}\label{the-qk-circuit-consists-of-interactions-between-pairs-of-parameter-subcomponents}

In attention layers,
\(W_Q\in \mathbb{R}^{d_{\text{model}}\times d_{\text{model}}}\) and
\(W_K\in \mathbb{R}^{d_{\text{model}}\times d_{\text{model}}}\) matrices
transform sequences of activations
\(\varphi\in \mathbb{R}^{T\times d_{\text{model}} }\) in the (normed)
residual stream to create queries (\(q = \varphi (W_Q)^\top\)) and keys
(\(k =  \varphi (W_K)^\top\)) for all heads. We can split them into the
keys and queries for each head
(e.g.~\(q = [ \varphi (W_Q^{1})^\top, \cdots , \varphi (W_Q^{H})^\top]\)).

The attention scores of head \(h\) are calculated as
\(Z^h = \varphi W_Q^{h \top} W_K^h \varphi^\top\), which are used to
calculate the head's attention pattern, \(A^h = \text{softmax} (Z^h)\).

Although the \(W_Q\) and \(W_K\) matrices are usually represented as
separate matrices, it is convenient to study them together as a single
matrix, \(W_{QK}^h = W_Q^{h \top} W_K^h\)
\citep{elhage2021mathematical}.

Prior to parameter decomposition, it was not obvious how best to further
decompose this circuit into specialized functional units. But VPD
decomposes the \(W_Q\) and \(W_K\) matrices in a sum of functionally
specialized rank-one parameter subcomponents \footnote{The \(V\)
  matrices of the subcomponents do not need \(h\) indices because they
  only read from the residual stream. The \(U\) matrices project into
  query or key space, and hence need \(h\) indices.}:

\[W_Q^h = \sum_c  \vec{U}^{h}_{Q,c} (\vec{V}_{Q,c})^\top \qquad  \qquad W_K^h = \sum_c  \vec{U}^{h}_{K,c} (\vec{V}_{K,c})^\top \]

These subcomponents are secretly also a decomposition of the QK circuit,
constructed from pairs of subcomponents of the \(W_Q\) and \(W_K\)
matrices:

\label{eq:qk-interactions} \[\begin{aligned}
W_{QK}^h &= W_Q^{h \top} W_K^h \\
&= \left( \sum_c \vec{U}^{h}_{Q,c} (\vec{V}_{Q,c})^\top \right)^\top \left( \sum_{c'} \vec{U}_{K,c'}^{h} (\vec{V}_{K,c'})^\top \right) \\
&= \sum_{c, c'} \vec{V}_{Q,c} \left( (\vec{U}_{Q,c}^{h})^\top \vec{U}_{K,c'}^h \right) (\vec{V}_{K,c'})^{\top}
\end{aligned}\]

We will use this equation to study the QK circuit, both for a form of
static (\emph{data-independent}) and dynamic (\emph{data-dependent})
analysis of the computations of the QK circuit.

We'll need to define two new metrics, one to measure the static
interaction strength between pairs of subcomponents and another to
measure how strongly a pair of subcomponents are interacting on a
particular datapoint.

\subsubsection{QK Circuit - Metric 1: Static Interaction
strength}\label{qk-circuit---metric-1-static-interaction-strength}

Although we can use \autoref{eq:qk-interactions} to understand the
\emph{static interaction strength} between subcomponents \(c\) and
\(c'\), we cannot simply use the raw term
\(\left( (\vec{U}_{Q,c}^{h})^\top \vec{U}_{K,c'}^h \right)\) for a few
reasons:

First, because both \(\vec{U}_c\) and \(\vec{V}_c\) vectors are
unnormalized, we need to scale each \(\vec{U}_c\) vector by the norm of
the corresponding \(\vec{V}_c\) vector in order to put the \(\vec{U}_c\)
vectors on the same scale.

\[||\vec{V}_{Q,c}|| \left( (\vec{U}_{Q,c}^{h})^\top \vec{U}_{K,c'}^h \right) ||\vec{V}_{K,c'}||\]

Second, we need to incorporate sequence position information. The above
equations actually leave out an important part of our transformer
language model: The Rotary Position Embedding (RoPE) rotation matrix
\citep{su2024roformer}. For transformers that use RoPE, the QK circuit
is actually:
\(W_{QK, \tau}^h = (W_Q^{h})^\top \boldsymbol{R}_{\tau} W_K^h\), where
\(\tau\) is the \emph{offset}---the difference between the sequence
position of the query and the key. The rotation matrix rotates the keys
and queries by different amounts depending on the offset. Thus we have

\[\left( ||\vec{V}_{Q,c}||  \vec{U}_{Q,c}^{h} \right)^\top \boldsymbol{R}_{\tau} \left( \vec{U}_{K,c'}^h  ||\vec{V}_{K,c'}|| \right)\]

Third, and finally, we need to know whether this interaction typically
contributes positively or negatively to the attention score. To
calculate this, we cheat slightly and import one data-dependent
statistic: The sign of the average subcomponent activation for each
subcomponent on tokens where the subcomponent is causally important.
With these three adjustments, we get the Static Interaction Strength:

\begin{equation}
\begin{gathered}
\text{StaticInteractionStrength}(c, c', \tau, h) 
  \\ =
    \Big(
      \text{sign}\left(\mathbb{E}_\varphi^{(c)} \left[\varphi\vec{V}_{Q,c} \right]\right) 
      \lVert \vec{V}_{Q,c} \rVert 
      \vec{U}^h_{Q,c}
      \Big)^\top
   \boldsymbol{R}_{\tau}  
    \Big(
      \text{sign}\left(\mathbb{E}_\varphi^{(c')} \left[\varphi \vec{V}_{K,c'}\right]\right) 
      \lVert \vec{V}_{K,c'} \rVert 
      \vec{U}^h_{K,c'}
      \Big)
\end{gathered}
\end{equation}

The Static Interaction Strength metric is not directly comparable across
heads, since each head applies a separate softmax function, making any
differences in scales or averages of interaction strength irrelevant. To
make the metric comparable across heads, we standardize it:

\[\text{StandardizedStaticInteractionStrength}(c, c', \tau, h) \\ = \frac{\text{StaticInteractionStrength}(c, c', \tau, h) - \mu_h}{\sigma_h}\]

where \(\mu_h\) and \(\sigma_h\) are the mean and standard deviation of
the Static Interaction Strengths across all \((c, c', \tau)\) for head
\(h\).

\begin{figure}[htbp]
\centering
\includegraphics[width=0.85\textwidth,height=0.8\textheight,keepaspectratio]{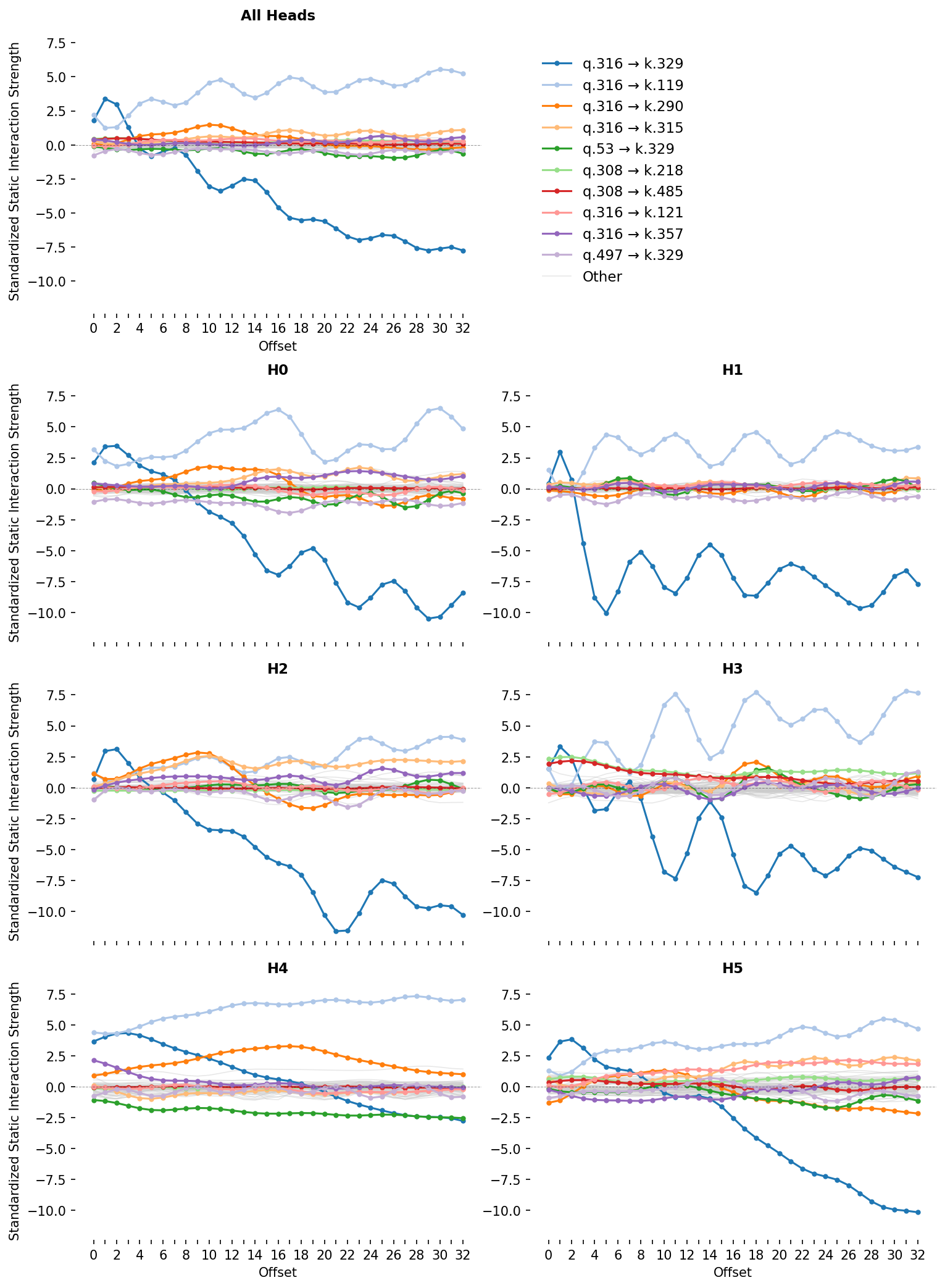}
\caption[]{The Standardized Static Interaction Strengths of pairs of parameter subcomponents in the \(Q\) and \(K\) projection matrices in each head (bottom grid) and all heads (top). The ten pairs with the largest interaction strengths at any offset are shown in color, with the rest in grey. The \texttt{1.attn.q:316} and \texttt{1.attn.k:329} pair exhibit strong positive Static Interaction Strength at early offsets, indicating this pair's involvement in cross-head previous token behavior (and, more generally, 'recent token behavior'.}\label{fig:attn_contrib_grid}
\end{figure}

For attention layer 1, we plot this metric for each pair of
subcomponents for each head and offset
(\autoref{fig:attn_contrib_grid}). We can see that for some pairs, the
Static Interaction Strength changes strongly at different offsets. This
means that, for these pairs, the same activations might have different
effects on the attention at different offsets! For others, the Static
Interaction Strengths seem independent of offset, meaning that their
effects on the attention scores are determined only by whether data that
activate them are present.

We will use this plot of Static Interaction Strength to analyze
particular attention behaviors. But before we do, we will equip
ourselves with a related metric, the Data-Dependent Interaction
Strength, which permits dynamic analysis.

\subsubsection{QK Circuit - Metric 2: Data-Dependent Interaction
Strength}\label{qk-circuit---metric-2-data-dependent-interaction-strength}

The attention patterns of each head depend on how the hidden activations
interact with the QK circuit:
\(A^h_\tau = \text{softmax} (\varphi W_{QK, \tau}^{h} \varphi^\top)\).

We can use \autoref{eq:qk-interactions} to decompose the QK circuit and
study how the activations \(\varphi\) at different timesteps \(t,t'\)
interact with each of the pairs of subcomponents:

\[\begin{aligned}
Z^h_\tau &= \varphi W_{QK, \tau}^h \varphi^\top 
&= \sum_{c, c'} \varphi \vec{V}_{Q,c} \left( (\vec{U}^{h}_{Q,c})^\top \boldsymbol{R}_{\tau} \vec{U}^h_{K,c'} \right) (\vec{V}_{K,c'})^{\top} \varphi^\top
\end{aligned}\]

Thus, the attention score at each head \(h\) and offset \(\tau\)
consists of the sum of the data's interaction with each of the
individual pairs \((c, c')\). On any input, we can therefore decompose
the attention score---and hence the attention pattern---into parts that
we can study in isolation. This lets us define a data-dependent metric
of interaction strength, which forms the basis of our dynamic analysis:

\[\begin{aligned}
\text{DataDependentInteractionStrength}(c, c', \tau, t, t', h) 
&= \left(\varphi \vec{V}_{Q,c} \left( (\vec{U}^h_{Q,c})^\top \boldsymbol{R}_{\tau} \vec{U}^h_{K,c'} \right) (\vec{V}_{K,c'})^{\top} \varphi^\top\right)_{t,t'}
\end{aligned}\]

If we broadcast this over sequence position and head, we can visualise a
subcomponent pair's interactions across a whole prompt as a stack of
per-head matrices --- and the model's full attention score \(Z\) as the
(per-head, per-position) sum of every such pair. To keep the figure
readable, we'll abbreviate the position-independent pair term as

\[\text{DataDependentInteractionStrength}(c, c', :, t, t') := \left( \varphi \vec{V}_{Q,c} \left( (\vec{U}_{Q,c})^\top \vec{U}_{K,c'} \right) (\vec{V}_{K,c'})^{\top} \varphi^\top\right)_{t,t'},\]

\begin{figure}[htbp]
\centering
\includegraphics[width=\textwidth,height=0.8\textheight,keepaspectratio]{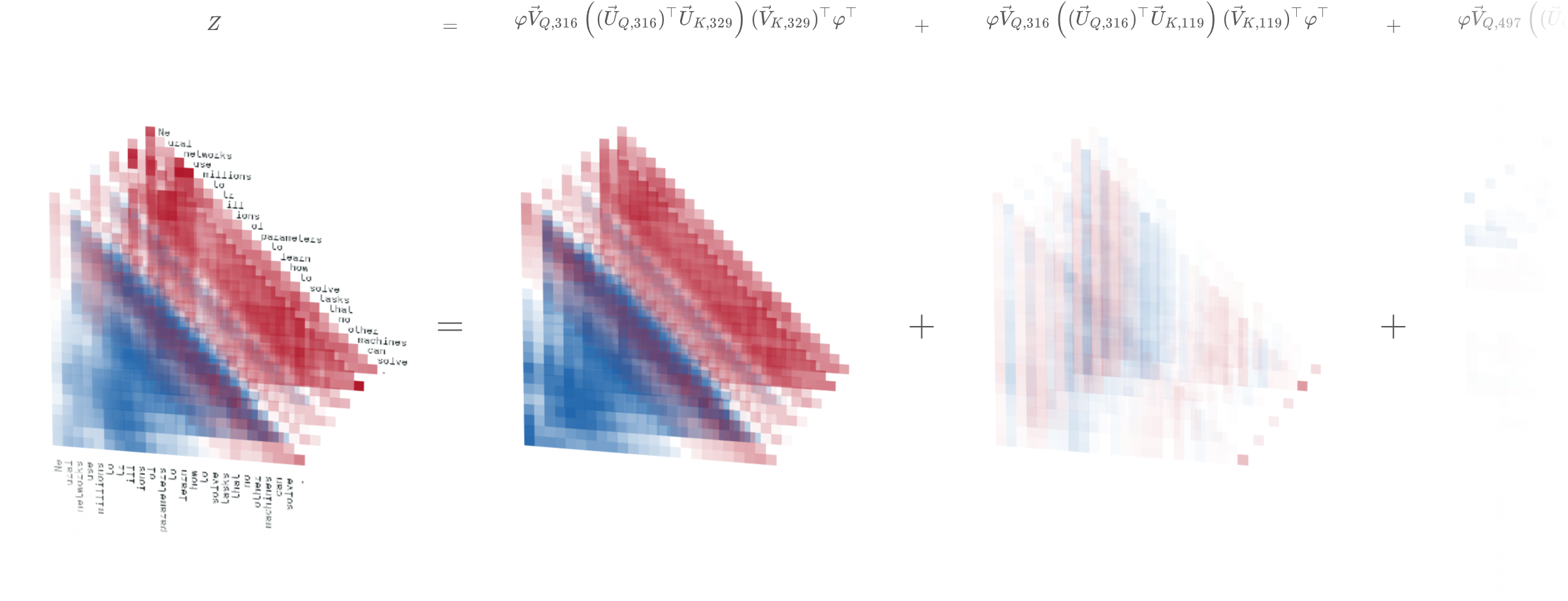}
\caption[]{Attention scores \(Z\) illustrated as a sum of Data Dependent Interaction Strengths between pairs of subcomponents. \emph{Static snapshot of an interactive figure; explore it at \href{https://www.goodfire.com/research/interpreting-lm-parameters}{the web version}.}}
\end{figure}

\begin{figure}[htbp]
\centering
\includegraphics[width=\textwidth,height=0.8\textheight,keepaspectratio]{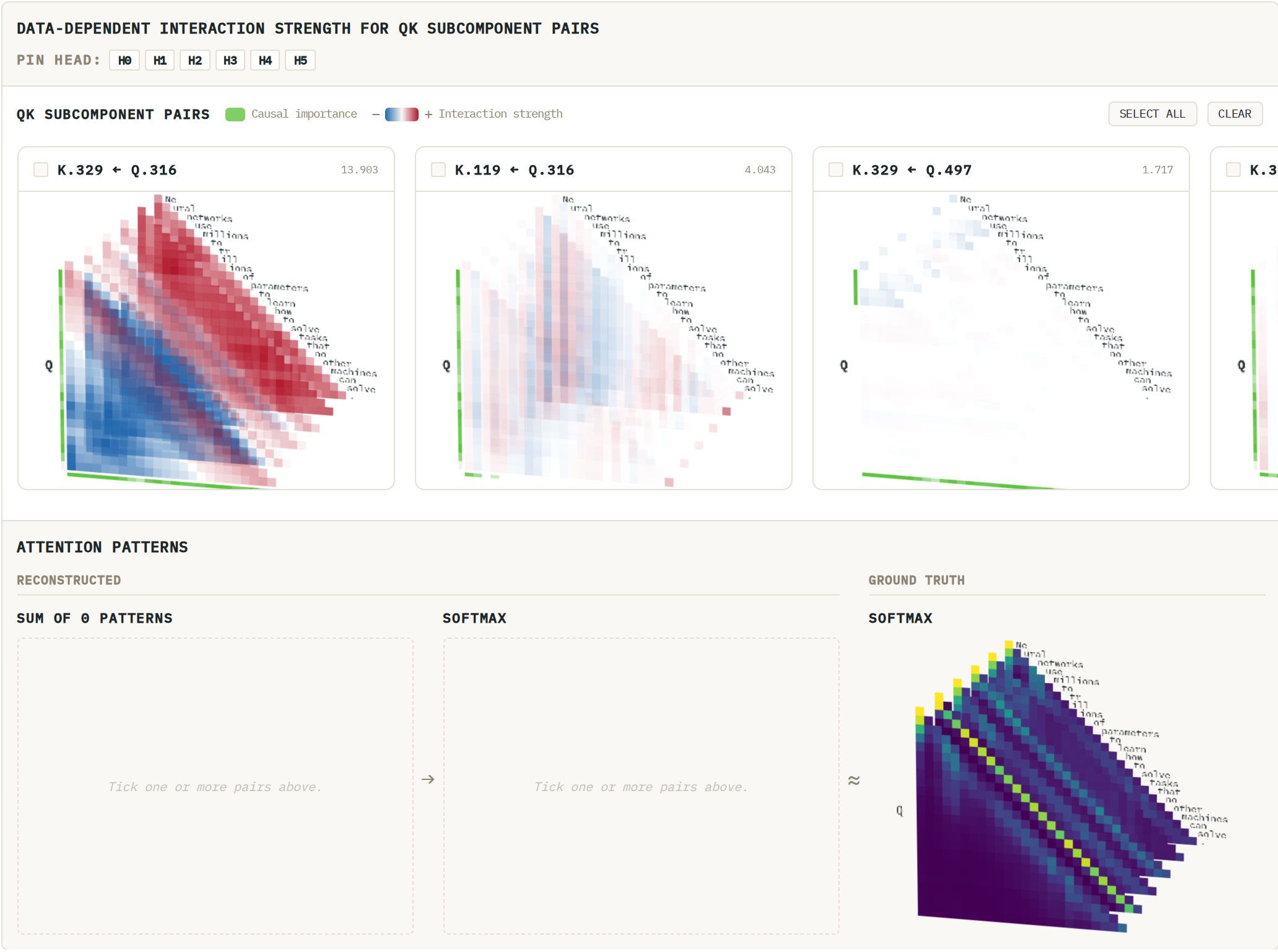}
\caption[]{The attention score consists of a sum of Data Dependent Interaction Strengths. This panel shows the same prompt as the figure above, but here you can control which pairs of subcomponents to include in the sum, allowing you to study their individual effects on the reconstructed attention score and attention pattern. \emph{Static snapshot of an interactive figure; explore it at \href{https://www.goodfire.com/research/interpreting-lm-parameters\#fig:dynamic-1}{the web version}.}}\label{fig:dynamic-1}
\end{figure}

In \autoref{fig:dynamic-1}, you can select which subcomponent
interactions to sum together and see the attention score for those
pairs. This is a very useful tool, since it splits up any given
attention pattern into the contributions of individual, functionally
distinct, subcomponent interactions.

We'll do an initial analysis of an attention behavior using only these
two QK metrics before discussing how they interact with the OV circuit.

\subsection{Decomposing attention behavior 1: Previous token
behavior}\label{decomposing-attention-behavior-1-previous-token-behavior}

\label{sec:attn-analysis-1}

\begin{figure}[htbp]
\centering
\includegraphics[width=\textwidth,height=0.8\textheight,keepaspectratio]{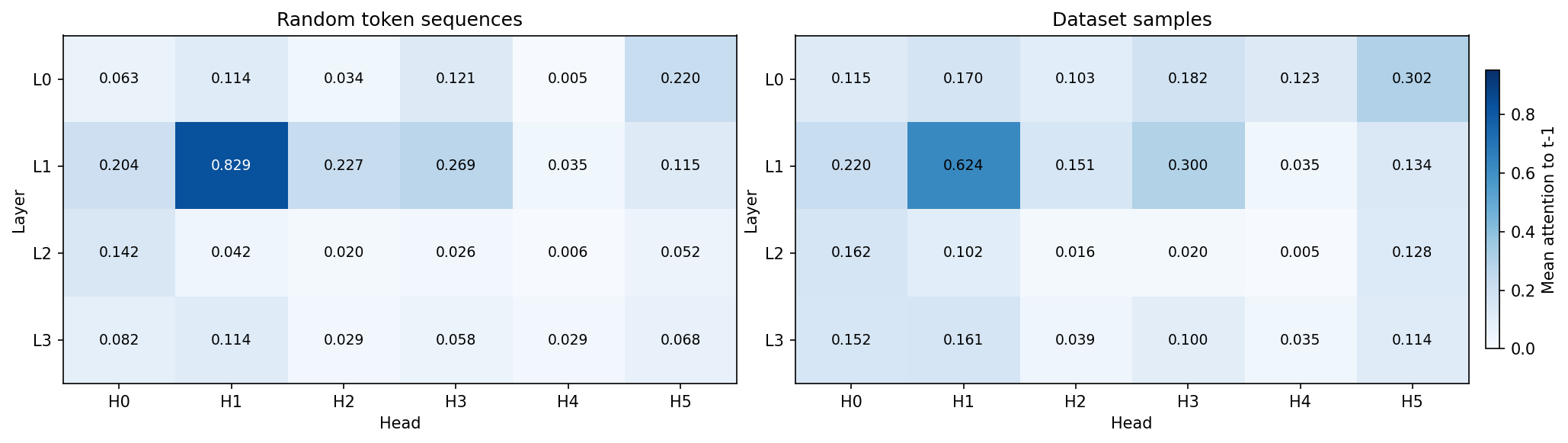}
\caption[]{Identifying the previous token head: Mean attention across multiple inputs on offset \(\tau=1\), i.e. the previous token. 
  \textbf{Left}: Average over sequences of random tokens, as per  \citep{wang2022interpretability}. \textbf{Right}: Average over sequences sampled from the dataset. The plots reveal L1H1 is the most canonical ``previous token head''. But note other heads place substantial average attention at offset \(\tau=1\).}\label{fig:prev_token_scores}
\end{figure}

Like many language models, our model has a head that, on average, places
the majority of its attention on the previous timestep
(\autoref{fig:prev_token_scores}). This is typically called a
\emph{previous token head}
\citep{clark-etal-2019-bert, elhage2021mathematical, olsson2022incontextlearninginductionheads, wang2022interpretability}
and, in our model, is head 1 in layer 1 (\textbf{L1H1}). However, L1H1
is not the only head to assign substantial probability to the previous
token; many other heads do too, including heads in the same layer as
L1H1.

Now we need to find subcomponents that might be involved in previous
token behavior and establish whether or not their computations span
multiple heads. An obvious place to start is by looking at the largest,
most frequently active subcomponents in the \(W_Q\) and \(W_K\)
matrices. Perhaps by coincidence, the largest norm subcomponents,
\texttt{1.attn.q:316} and \texttt{1.attn.k:329}, are also the most
frequently causally important (\autoref{fig:qk_comp_weight_norm})!

While most subcomponents in layer one are only active on a fraction of
tokens, both \texttt{1.attn.q:316} and \texttt{1.attn.k:329} have a CI
firing density of \(96.7\%\) and \(99.8\%\), meaning they're nearly
constantly active. Both have the largest weight norm in L1H1, which was
the head with the strongest previous token behavior
(\autoref{fig:qk_comp_weight_norm}). But they also have substantial
weight norm in other heads, suggesting they aren't exclusively located
in any particular head. Could they be responsible for cross-head
previous token behavior?

\autoref{fig:attn_contrib_grid} shows that these two subcomponents also
have very strong offset-dependent Static Interaction Strength. In
particular, their interaction is strongest at small offsets, and weak or
negative interactions at more distant offsets. This is exactly what we
would expect of two subcomponents that implement previous token behavior
or recent token behavior. This pattern holds not only in L1H1, but also
in other heads too. This is strong observational evidence that these two
subcomponents compute previous token behavior in a way that is
distributed across heads.

\begin{figure}[htbp]
\centering
\includegraphics[width=0.85\textwidth,height=0.8\textheight,keepaspectratio]{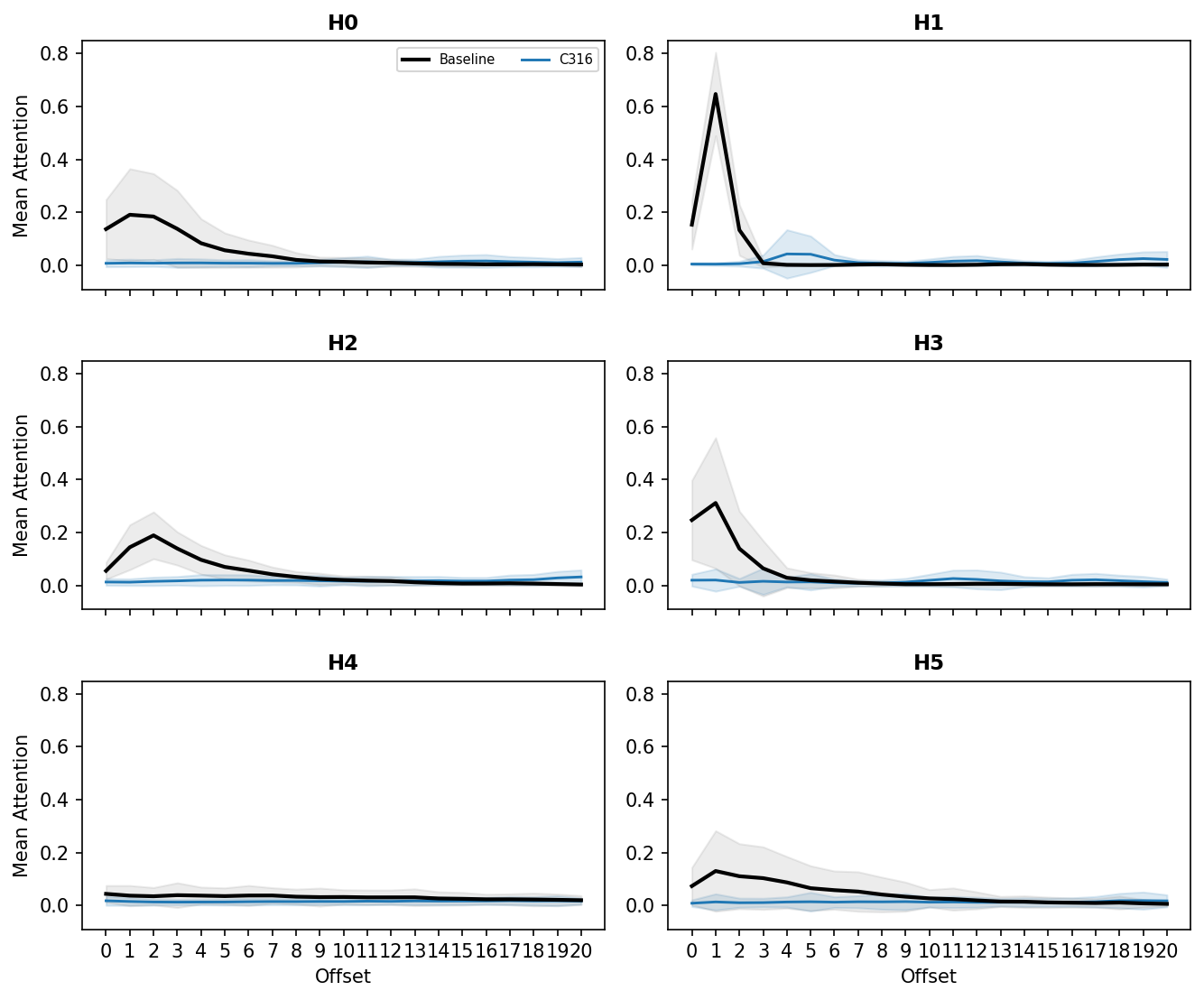}
\caption[]{Effect of ablations: Ablating \texttt{1.attn.q:316} very strongly reduces attention to tokens in the recent past across all heads that otherwise attended there strongly. The effects of ablating other W\_Q components has no distinguishable effect compared with the baseline and are therefore not shown. Here the baseline is the unablated average attention pattern.}\label{fig:attn_patterns_q_intv}
\end{figure}

We test this hypothesis causally using ablations and dynamic analysis.
When we ablate different \(W_Q\) subcomponents on a dataset of prompts,
the change in average attention is very small for most subcomponent
ablations. Only the ablation of \texttt{1.attn.q:316} results in the
large reduction of attention at recent offsets
(\autoref{fig:attn_patterns_q_intv}).

\autoref{fig:dynamic-1} shows dynamic analysis. For any of the prompts,
you can remove the contribution of the \texttt{1.attn.q:316} and
\texttt{1.attn.k:329} interaction to the attention score. Removing it
destroys the attention to tokens in the recent past across all heads
that had strong to moderate attention there.

Together, this is strong evidence that the \texttt{1.attn.q:316} and
\texttt{1.attn.k:329} interaction computes previous token behavior and
is distributed across heads.

This raises a question: What information is this attention moving from
the recent past to the current timestep? What \emph{attention values}
does this previous token behavior tend to move? Are the different heads
carrying forward information from distinct subspaces in the residual
stream? Or are they carrying redundant information, perhaps as a form of
noise robustness? To study this, we need to analyze the OV circuit, for
which we will need another metric.

\subsubsection{Previous token behavior employs non-overlapping subspaces
in the OV
circuit}\label{previous-token-behavior-employs-non-overlapping-subspaces-in-the-ov-circuit}

The OV circuit is made from the \(W_V\) and \(W_O\) matrices which
respectively read from and write to the residual stream:

\[W_{OV}^h = W_{O}^h W_{V}^h \in \mathbb{R}^{d_{\text{model}} \times d_{\text{model}}}\]

The sequence of \(T\) vectors of dimension \(d_{\text{model}}\) that the
attention layer outputs into the residual stream is computed using the
attention pattern-weighted sum of the outputs of the OV circuits at all
previous timesteps (where the attention pattern \(A^{h}\) is determined
by the QK circuit):

\[\text{AttentionLayer} (\varphi) = (A^h)^\top \varphi  (W_{OV}^h)^\top  \in \mathbb{R}^{T \times d_{\text{model}}}\]

Although \(W_{OV}^h\) is a \(d_{\text{model}} \times d_{\text{model}}\)
matrix, it only has rank \(d_{\text{head}}\). Being low rank, each head
can therefore only read from and write to a small subspace of the
residual stream. It would be useful to know if two heads read from and
write to similar subspaces.

\begin{figure}[htbp]
\centering
\includegraphics[width=\textwidth,height=0.8\textheight,keepaspectratio]{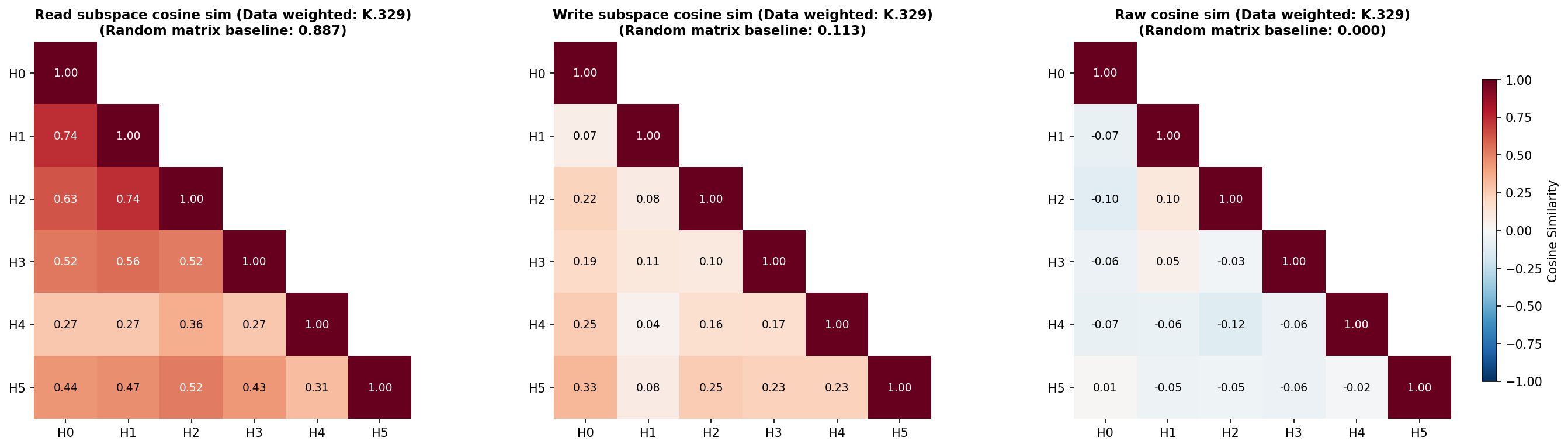}
\caption[]{Data-weighted cosine similarities between each head's \(W_{OV}^h\) read and write matrices, and the cosine similarity between each head's raw \(W_{OV}^h\). Here, data-weighting uses data where subcomponent \texttt{1.attn.k:329} is causally important.}\label{fig:prev_tok_ov_overlap_k_329}
\end{figure}

To do this, we will measure the `overlap' between the subspaces that
each head's OV circuit reads from and writes to, for which we'll use the
`Data-weighted Subspace Similarity' metric, which we construct from the
Frobenius cosine similarity of the `read subspaces' and the `write
subspaces' of each head (\autoref{fig:prev_tok_ov_overlap_k_329}). See
\autoref{app:OV-metric-data-frob} for details of how these subspaces are
constructed and for further details of this metric. We also measure the
Frobenius cosine similarity of the \(W_{OV}^h\) matrices themselves
(\autoref{fig:prev_tok_ov_overlap_k_329}). When calculating similarity,
we weight the axes of the read- and write-subspaces by how much data
variation lies in each axis, since we do not care as much about weight
similarity along axes where data do not exist or do not vary. In all
cases, we compare the measured similarities to similarities between
random, data-weighted matrices.

Most heads in layer 1, except L1H4, seem at least weakly involved with
previous token behavior, as assessed by their previous token score
(\autoref{fig:prev_token_scores}) and the offset dependence of the
Static Interaction Strength of the \texttt{1.attn.q:316} and
\texttt{1.attn.k:329} pair (\autoref{fig:attn_contrib_grid}). We
therefore should look at the overlap in the read and write subspaces of
all heads in layer 1 except L1H4.

The read subspaces of each head are close to or slightly lower than the
expected similarity of two random (data-weighted) matrices
(\autoref{fig:prev_tok_ov_overlap_k_329}). On the other hand, the write
subspaces seem close to or slightly higher than the random baseline.
These effects seem very weak, but weakly suggest a pattern of attention
heads reading from distinct subspaces but writing to slightly less
distinct subspaces.

For the head with the strongest previous token behavior, L1H1, the other
heads L1H0 and L1H2 seem to read from subspaces with similarities close
to the random baseline, but other heads read from much less similar
subspaces. When comparing the similarity of the raw \(W_{OV}^h\)
matrices, there appears to be very little deviation from levels of
overlap that would be expected of random matrices, except the comparison
between L1H1 and L1H2, which again seem to be more similar than the
random baseline. These two heads seem to write to quite different
subspaces, though.

Overall, this weakly suggests a picture that previous token behavior
spans distinct subspaces across different heads. One potential reason
for this is to be able to read more information from the residual stream
than might be readable by a single head. There appears to be very
limited, but nonzero, redundancy in how heads involved in previous token
behavior read from different subspaces, but they largely seem to write
to different subspaces.

Previous token behavior is an important behavior implemented by probably
every language model. But it is far from the only behavior implemented
in layer 1. Even in L1H1, only around 60\% of attention is on the
previous timestep (\autoref{fig:prev_token_scores}). What other
attention behaviors is this head implementing? In the next section, we
look at another behavior implemented by L1H1 in more detail, and examine
whether that behavior is also distributed across heads.

\subsection{Decomposing attention behavior 2: Previous syntax boundary
movement}\label{decomposing-attention-behavior-2-previous-syntax-boundary-movement}

\label{sec:attn-analysis-2}

Looking again at the static analysis of layer 1, we can see that L1H1
has interactions between Q and K subcomponents that seem to have quite a
different offset-dependency (\autoref{fig:attn_contrib_grid}). The
subcomponents \texttt{1.attn.q:316} and \texttt{1.attn.k:119} seem to
interact most strongly at later offsets across multiple heads, including
L1H1.

We are already familiar with \texttt{1.attn.q:316}, the query
subcomponent that is always active. The key subcomponent
\texttt{1.attn.k:119} is new: It seems to activate on brackets,
punctuation, and newlines, but also some common continuation words, such
as `the' or `and'. It is causally important on 16\% of tokens, which is
frequent, but not constantly active.

This interaction therefore involves a conditional computation: Although
\texttt{1.attn.q:316} is always active, constantly looking back in time,
the other subcomponent \texttt{1.attn.k:119} only interacts with it when
it is active.

Interestingly, \texttt{1.attn.k:119} must be active sufficiently far
back in time; otherwise, the Static Interaction Strength may not be
strong enough to contribute to the attention score. Almost every head
seems to exhibit an offset dependent interaction between subcomponents
\texttt{1.attn.q:316} and \texttt{1.attn.k:119}, suggestive of a very
distributed computation.

\begin{figure}[htbp]
\centering
\includegraphics[width=\textwidth,height=0.8\textheight,keepaspectratio]{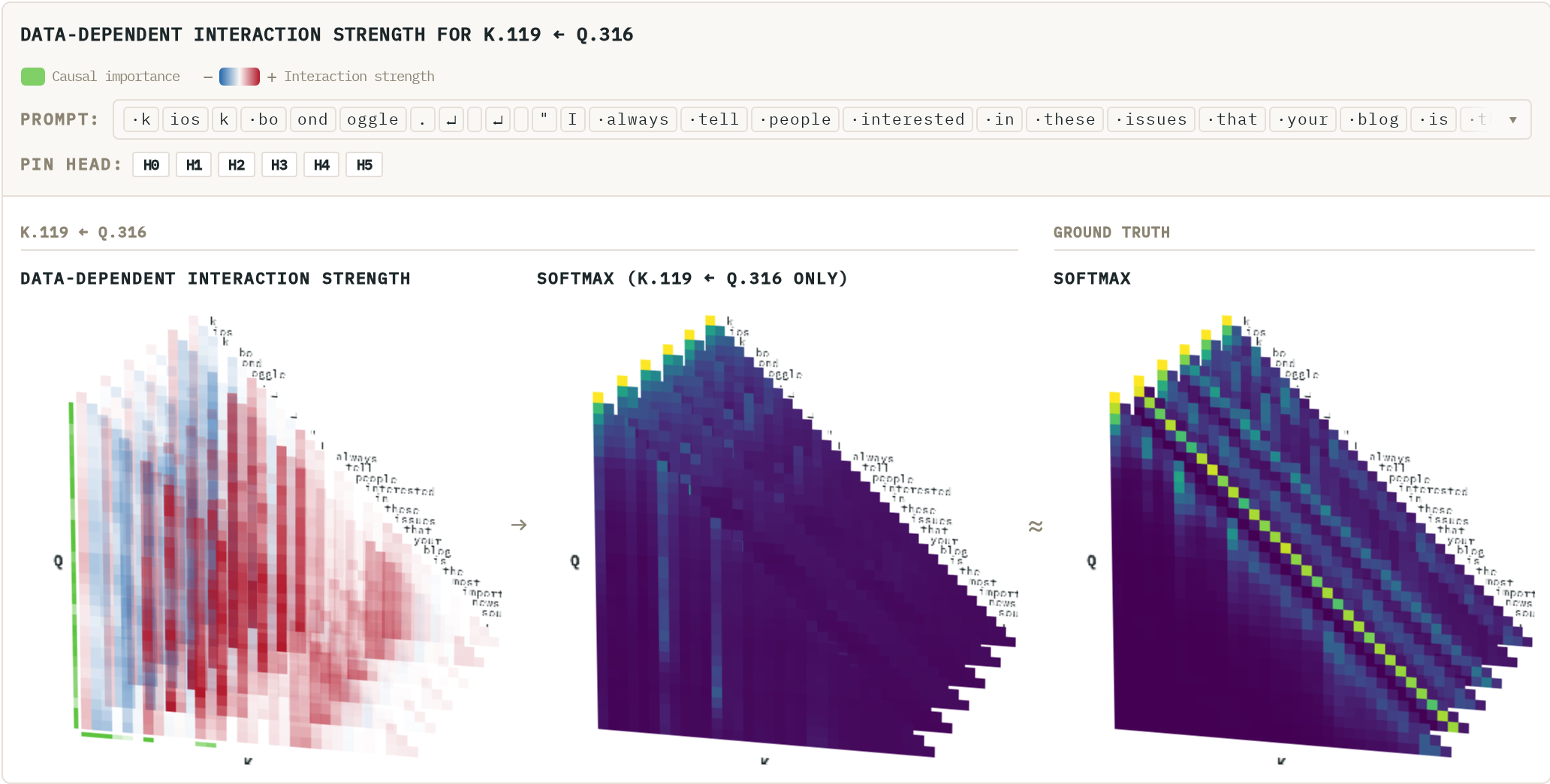}
\caption[]{Data Dependent Interaction Strength for the \texttt{1.attn.k:119}-\texttt{1.attn.q:316} interaction only, and the corresponding attention pattern if only that interaction contributed to the attention pattern. On the right hand side, the ground truth attention pattern is shown for comparison. \emph{Static snapshot of an interactive figure; explore it at \href{https://www.goodfire.com/research/interpreting-lm-parameters\#fig:dynamic-2}{the web version}.}}\label{fig:dynamic-2}
\end{figure}

Since this computation is data-dependent, we will benefit from greater
use of dynamic analysis. \autoref{fig:dynamic-2} shows the attention
patterns of all heads, but only shows the Data Dependent Interaction
Strength of the \texttt{1.attn.k:119}-\texttt{1.attn.q:316} interaction.
One prompt is shown at a time, but you can select a variety of other
prompts from the dataset in the dropdown menu.

By exploring different prompts, and inspecting the contributions of the
\texttt{1.attn.q:316} and \texttt{1.attn.k:119} interaction across all
heads, it is possible to see that this interaction contributes
significantly to the attention patterns of most heads on previous
periods, commas, and newline characters. L1H4 seems capable of
maintaining attention on these characters at quite large offsets, based
on the stronger than average vertical bars in the ground truth attention
on those tokens. Other heads seem only to have noticeable attention on
them more recently in time. This may be due to competition with other
attention score contributions from other pairs.

The activating examples of \texttt{1.attn.k:119} show firings on various
forms of punctuation, end of text tokens, newlines, latex ``\$''
symbols, brackets, etc. This suggests that this pair of subcomponents
orchestrates a syntax boundary detector with a variety of short- or
long-offset ranges. We'll call this `previous syntax boundary' movement.

This pair of subcomponents seems responsible for attention to syntax
boundary tokens at different ranges in different heads
(\autoref{fig:attn_contrib_grid}). L1H1 seems to increase self attention
upon syntax boundary tokens; L1H2 seems only mildly to attend to syntax
boundary tokens and only in the very recent past. L1H5 and L1H0 attend
to syntax boundary tokens a small number of tokens in the past. L1H4
seems to attend to syntax boundary tokens many tokens in the past. L1H3
is less clear, but seems to attend to a smaller subset of specific
syntax boundary tokens, usually with shorter offset ranges.

The QK circuit of the `previous syntax boundary movement' behavior seems
quite distributed across heads. How does it interact with the OV
circuit? We can study this by looking at probability of each key
subcomponent being active conditioned on a given value subcomponent
being active (\autoref{fig:pkv}). The value subcomponents most
associated with \texttt{1.attn.k:119} are:

\begin{itemize}
\tightlist
\item
  \texttt{1.attn.v:72}
\item
  \texttt{1.attn.v:22}
\item
  \texttt{1.attn.v:745}
\item
  \texttt{1.attn.v:919}
\item
  \texttt{1.attn.v:531}
\item
  \texttt{1.attn.v:494}
\item
  \texttt{1.attn.v:195}
\item
  \texttt{1.attn.v:612}
\item
  \texttt{1.attn.v:984}
\item
  \texttt{1.attn.v:1000}
\item
  \texttt{1.attn.v:22}
\item
  \texttt{1.attn.v:389}
\item
  \texttt{1.attn.v:188}
\item
  \texttt{1.attn.v:299}
\item
  \texttt{1.attn.v:1014}
\item
  \texttt{1.attn.v:227}
\item
  \texttt{1.attn.v:946}
\item
  \texttt{1.attn.v:340}
\item
  And some with weaker associations (\autoref{fig:pkv}).
\end{itemize}

As in the case of previous token behavior, the data-weighted OV circuits
(where we weight the similarity using dataset examples and tokens where
\texttt{1.attn.k:119} is causally important) do not seem to read from
very similar residual stream subspaces
(\autoref{fig:prev_tok_ov_overlap_k_119}), though they seem to write to
somewhat more similar subspaces than would be expected in random
matrices. The OV circuit subcomponents that subcomponent
\texttt{1.attn.k:119} seems to overlap strongest with are associated
with other punctuation and syntax boundary-like tokens across seemingly
all heads, in both the read and the write matrices
(\autoref{app:ov-alignment-k119}).

To understand why the model is carrying forward information about the
previous syntax boundary, we would need to know how the values are being
used downstream. But it is possible to surmise at least part of its
function: It is useful to know what the previous syntax boundary tokens
are in order to perform tasks like closing opened brackets; knowing
whether a list is a bullet list or dashed list; or knowing if a token is
within or outside of a quotation; and more.

\subsection{Decomposing attention behavior 3: Detecting Existential
vs.~Expletive
Constructions}\label{decomposing-attention-behavior-3-detecting-existential-vs.-expletive-constructions}

\label{sec:attn-analysis-3}

Both of the above attention behaviors (\autoref{sec:attn-analysis-1} and
\autoref{sec:attn-analysis-2}) have involved \(W_Q\) or \(W_K\)
subcomponents where one is `always active'. Although the vast majority
of the attention scores in this layer seem to involve at least one of
these subcomponents, it would be interesting to study an even more
conditional behavior.

We'll investigate an attention behavior involving the \(W_Q\)
subcomponent \texttt{1.attn.q:308}.

This subcomponent appears to activate on a subset of \textbf{copula}
verbs. Examples of copula verbs include:

\begin{itemize}
\tightlist
\item
  ``To be'' (``she \textbf{is}'', ``it \textbf{was}'', ``What
  \textbf{were}'', ),
\item
  Verbs related to sensory appearance (``it certainly \textbf{seems}'',
  ``she \textbf{appeared} as though'', ``they \textbf{looked} like''),
  and
\item
  Verbs related to state (``we \textbf{remain}'', ``it \textbf{becomes}
  readily apparent'', ``there \textbf{exists}'').
\end{itemize}

Grammatically, copula verbs behave as \emph{linking verbs}: They connect
a subject (``it'', ``she'', ``there'', ``we'', etc.) to a description or
complement, rather than expressing an action. They are relatively
ubiquitous throughout English, so it makes sense that even a small
language model would learn computations involving them.

\subsubsection{\texorpdfstring{Subcomponent \texttt{1.attn.q:308}
activates on a subset of copula
verbs}{Subcomponent 1.attn.q:308 activates on a subset of copula verbs}}\label{subcomponent-1.attn.q308-activates-on-a-subset-of-copula-verbs}

Although \texttt{1.attn.q:308} has a large subcomponent activation on
copula verb tokens, it is noteworthy that it is not causally important
on all instances them. Here are several prompts containing copula verb
tokens on which \texttt{1.attn.q:308} is not causally important on some
tokens despite having high subcomponent activation:

\begin{figure}[htbp]
\centering
\includegraphics[width=\textwidth,height=0.8\textheight,keepaspectratio]{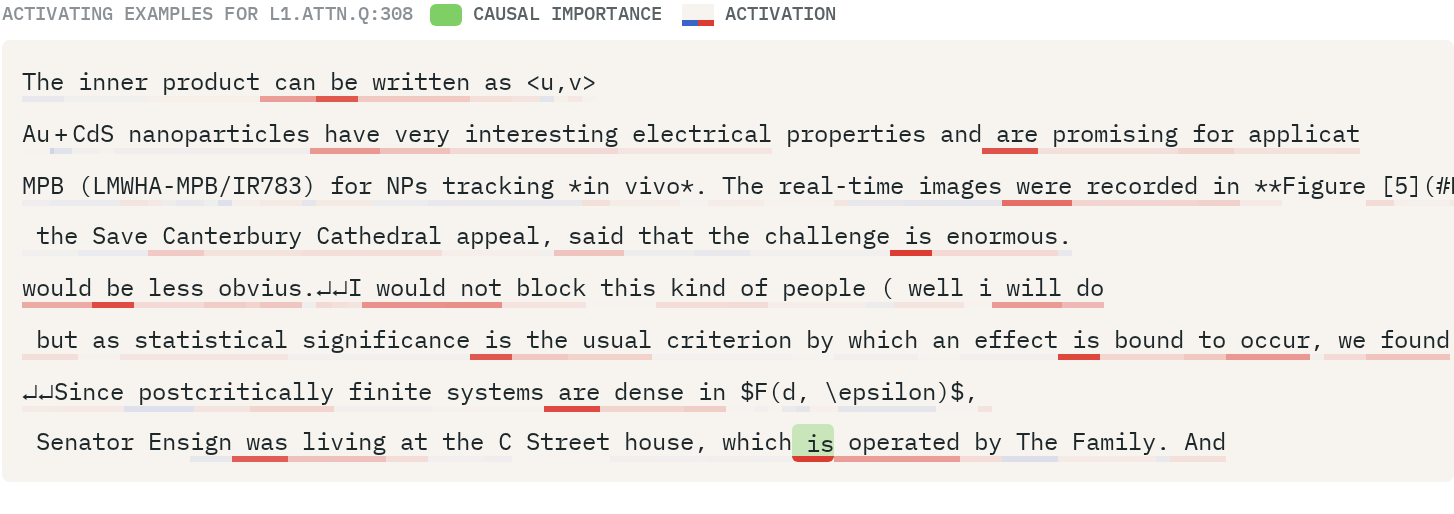}
\caption[]{\emph{Static snapshot of an interactive figure; explore it at \href{https://www.goodfire.com/research/interpreting-lm-parameters}{the web version}.}}
\end{figure}

By contrast, here are a few prompts where \texttt{1.attn.q:308}
\emph{is} causally important on copula verb tokens:

\begin{figure}[htbp]
\centering
\includegraphics[width=\textwidth,height=0.8\textheight,keepaspectratio]{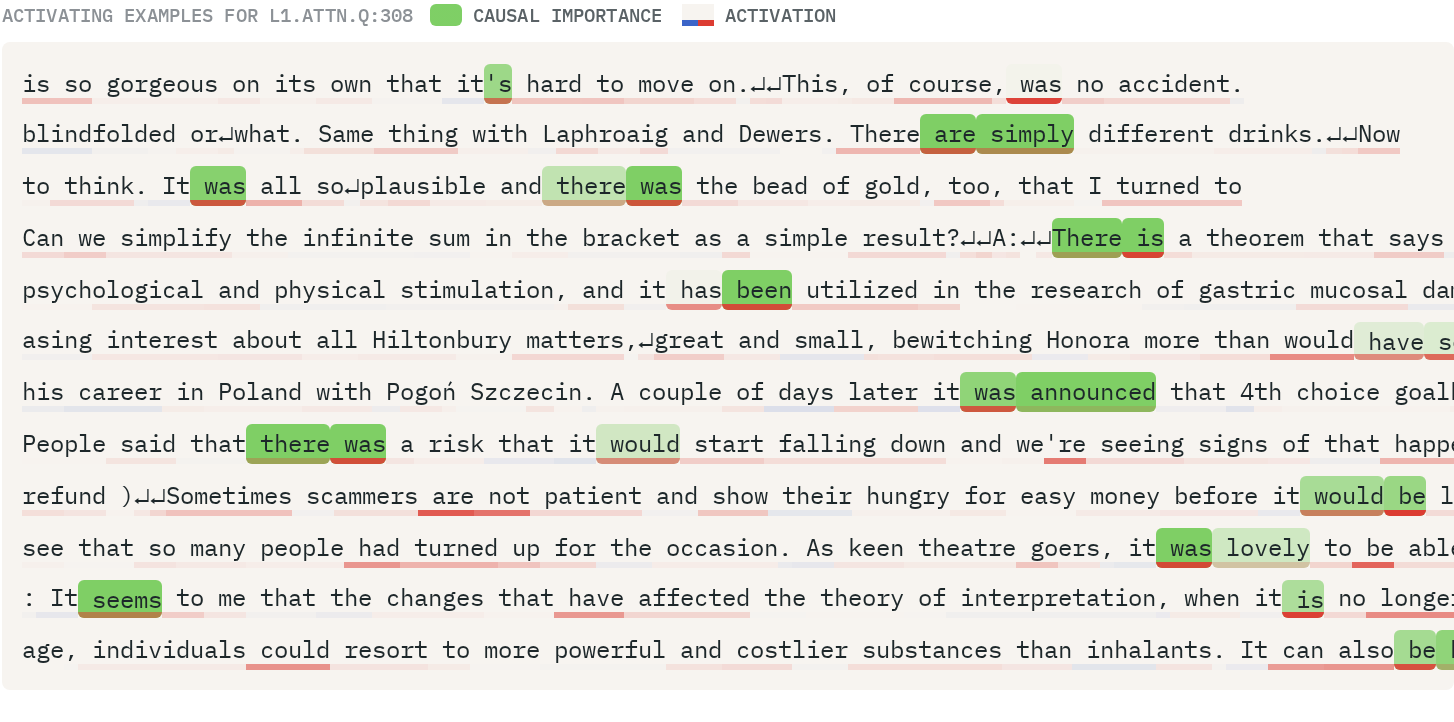}
\caption[]{\emph{Static snapshot of an interactive figure; explore it at \href{https://www.goodfire.com/research/interpreting-lm-parameters}{the web version}.}}
\end{figure}

By studying the difference between these two sets, it is possible to
notice a pattern: Although \texttt{1.attn.q:308} has a large positive
subcomponent activation on most instances of copula verbs, the cases
where it is causally important are typically when it is preceded by
\texttt{it}, \texttt{there}, \texttt{here} (as in ``it is'', ``there
is'', ``there are'', ``here is'', ``makes it seem'') and related tokens.

Constructions like these have specific linguistic terms:
\textbf{Existential} and \textbf{expletive constructions}, which use
``there'' and ``it'' in particular senses:

\begin{itemize}
\item
  The `\textbf{existential ``there''}': Where ``there'' is used to make
  assertions about the existence of something. \emph{Examples:
  ``\textbf{There is} a problem'', ``\textbf{there wasn't} enough'',
  ``\textbf{there seems to be} several'', ``\textbf{there exists}'',
  ``\textbf{there have been} few attempts'', ``\textbf{there remains} a
  number of''}
\item
  The `\textbf{expletive ``it''}': Where ``it'' is used as a dummy
  subject, with no real referent. \emph{Examples: ``\textbf{It is}
  unusual'', ``\textbf{It appears} likely'', ``\textbf{It was} found
  that'', ``\textbf{It dawned} on him that''\footnote{Note that, even
    though ``dawned'' is not a copula, we can still use the `expletive
    ``it''\,' with it.}, ``\textbf{It looks} like''}
\end{itemize}

Even though \texttt{1.attn.q:308} has a high subcomponent activation on
most copula verbs, it is usually not causally important (with some
exceptions) when the copula verb is preceded by personal pronouns
(e.g.~``she'', ``he'', ``they''):

\begin{figure}[htbp]
\centering
\includegraphics[width=\textwidth,height=0.8\textheight,keepaspectratio]{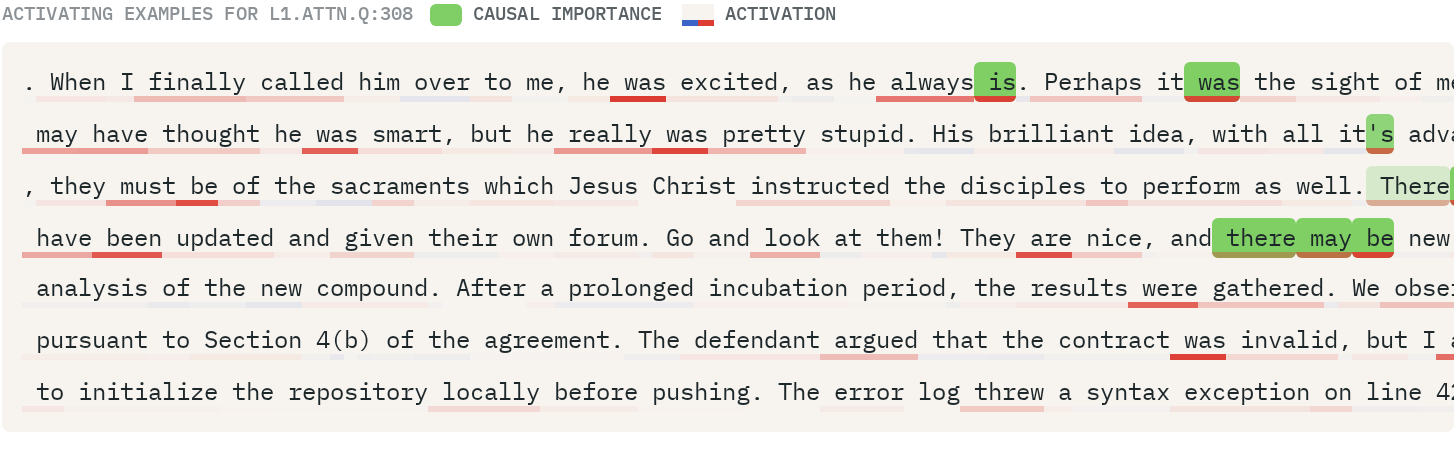}
\caption[]{\emph{Static snapshot of an interactive figure; explore it at \href{https://www.goodfire.com/research/interpreting-lm-parameters}{the web version}.}}
\end{figure}

How could this be? The main way that QK subcomponents can influence
downstream computations (and hence have causal importance) is by
influencing attention. \texttt{1.attn.q:308} having a large subcomponent
activation is insufficient for attention. There needs to be a key
subcomponent that aligns with \texttt{1.attn.q:308} (i.e.~has a high
Static Interaction Strength) that also has a high subcomponent
activation in order for a Q-K subcomponent pair to have high Data
Dependent Interaction Strength, and hence to contribute significantly to
the attention pattern. There must therefore be \(W_K\) subcomponents
that \texttt{1.attn.q:308} `looks for' in the past that, if present,
give this subcomponent its causal importance.

\subsubsection{\texorpdfstring{Subcomponent \texttt{1.attn.q:308}
interacts with two specific \(W_K\)
subcomponents}{Subcomponent 1.attn.q:308 interacts with two specific W\_K subcomponents}}\label{subcomponent-1.attn.q308-interacts-with-two-specific-w_k-subcomponents}

We'll start looking for this interaction by looking at whether there are
subcomponents that have a high Static Interaction Strength with
\texttt{1.attn.q:308}.

Looking again at \autoref{fig:attn_contrib_grid}, we can see that
\texttt{1.attn.q:308} has strongest offset-dependent Static Interaction
Strength with two \(W_K\) subcomponents, namely \texttt{1.attn.k:218}
and \texttt{1.attn.k:485}. These two interactions are strongest in L1H3,
but also in L1H5.

Incidentally, the norm plot (\autoref{fig:qk_comp_weight_norm}) supports
the idea that \texttt{1.attn.q:308} is primarily located in L1H3, and
secondarily in L1H5, since the weight norm is largest in those two heads
and negligible elsewhere.

These two \(W_K\) subcomponents (\texttt{1.attn.k:218} and
\texttt{1.attn.k:485}) seem to be causally important on related, but
semantically distinct, tokens, which we explore in detail in the
following sections.

\subsubsection{\texorpdfstring{Subcomponents \texttt{1.attn.k:218} and
\texttt{1.attn.q:308} make an ``it + copula verb''
detector}{Subcomponents 1.attn.k:218 and 1.attn.q:308 make an ``it + copula verb'' detector}}\label{subcomponents-1.attn.k218-and-1.attn.q308-make-an-it-copula-verb-detector}

\texttt{1.attn.k:218} seems to have high subcomponent activation on any
instance of the word ``it'', including capitalized variants. It is also
causally important on any instance of the word ``it'', but its causal
importance tends to be higher on instances of the `expletive
``it''\,'\footnote{This includes edges cases like the `meterological
  it', as in ``It is raining.''}.

The phrase ``it is'' is often an `expletive ``it''\,' followed by a
copula. But it may also be an `anaphoric pronoun ``it''\,' followed by a
copula, as in ``It is mine''. It turns out that \texttt{1.attn.k:218} is
causally important on both types of ``it is''. But it is not causally
important for expressions involving other pronouns followed by copulas,
such as ``he is'', ``they are'', etc. It therefore seems that this pair
of subcomponents interact to implement an ``it + copula verb'' detector,
including both `expletive ``it''\,' and `anaphoric pronoun ``it''\,'
followed by a copula

We can see its Data Dependent Interaction Strengths in the figure below.
The interaction strengths are strongest in L1H3, with a small amount in
L1H5, with essentially none in any other head. The attention patterns
reveal that the \texttt{1.attn.q:308} subcomponent `looks back in time'
from copula verbs, and has high Data-Dependent Interaction Strength with
\texttt{1.attn.k:218} if it finds it. If it does, it usually contributes
enough to the attention score that it becomes causally important.

\begin{figure}[htbp]
\centering
\includegraphics[width=\textwidth,height=0.8\textheight,keepaspectratio]{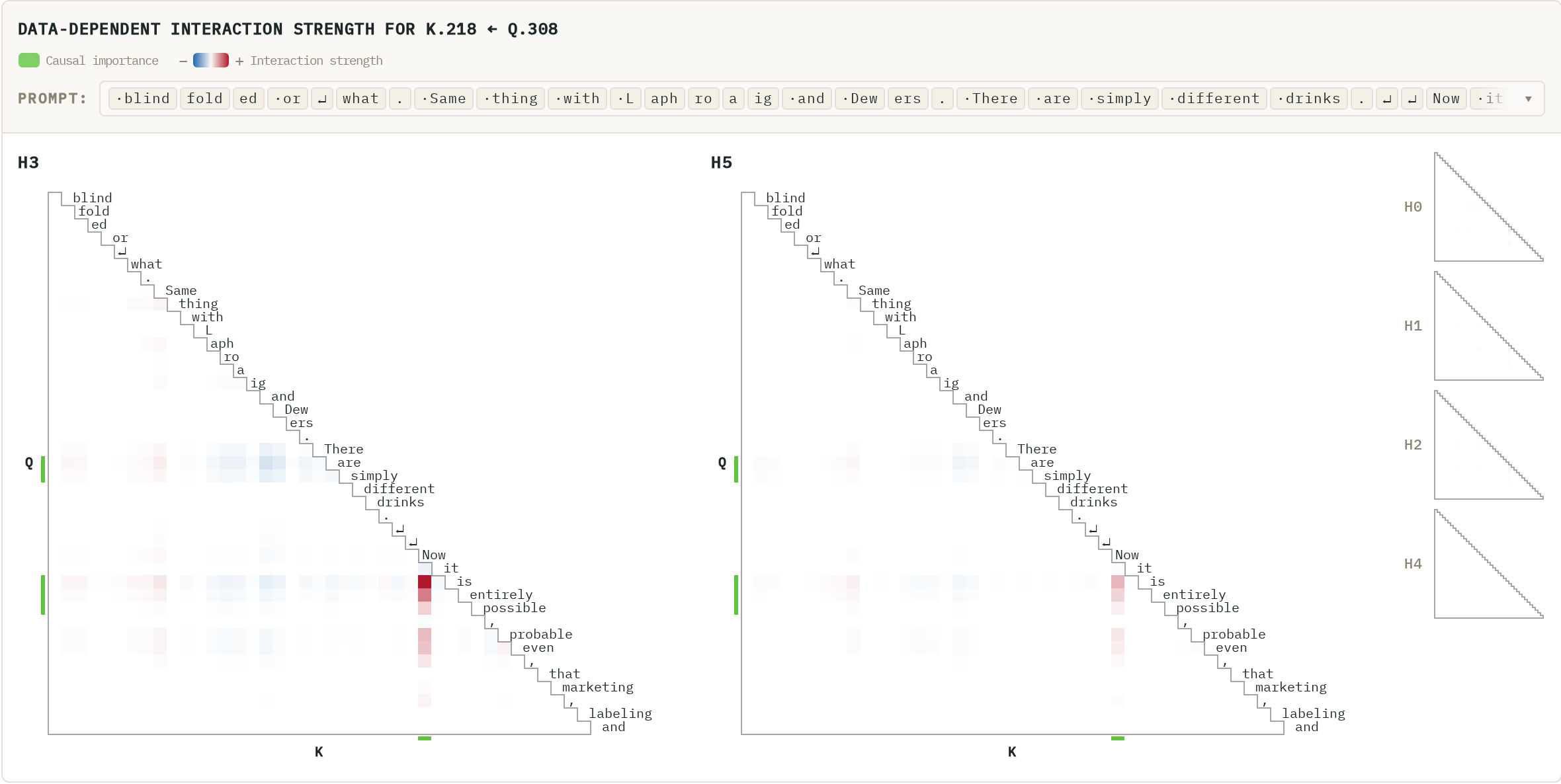}
\caption[]{\emph{Static snapshot of an interactive figure; explore it at \href{https://www.goodfire.com/research/interpreting-lm-parameters}{the web version}.}}
\end{figure}

It turns out that it is quite an overzealous ``it + copula verb''
detector. It often produces high Data Dependent Interaction Strength
even at quite large offsets, even when the ``it'' and the copula verb
are not related to each other. For an example, see the prompts below
where a copula verb late in the prompt attends back to an unrelated
\texttt{it} token in an earlier sentence:

\begin{figure}[htbp]
\centering
\includegraphics[width=\textwidth,height=0.8\textheight,keepaspectratio]{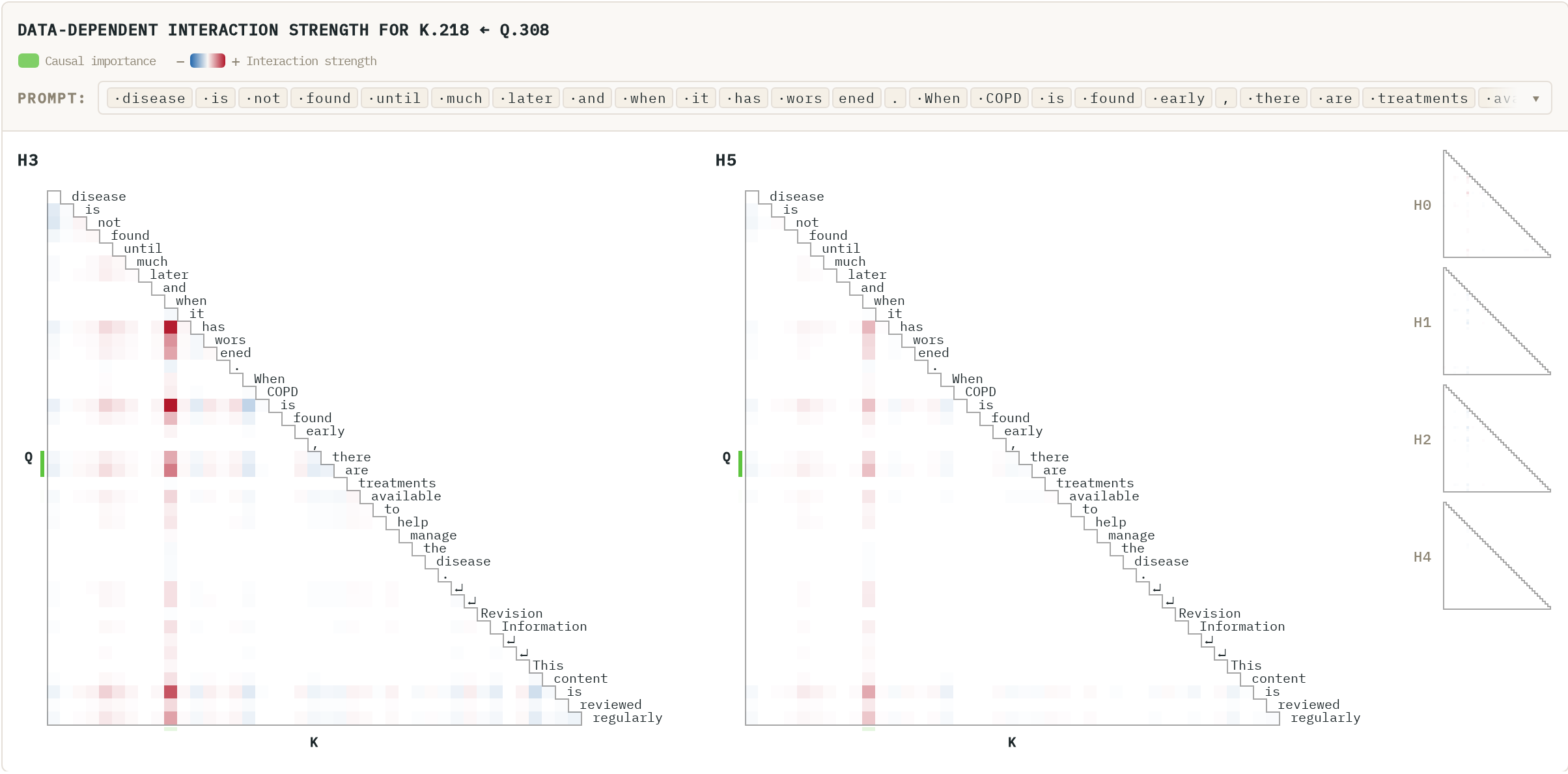}
\caption[]{\emph{Static snapshot of an interactive figure; explore it at \href{https://www.goodfire.com/research/interpreting-lm-parameters}{the web version}.}}
\end{figure}

\subsubsection{\texorpdfstring{The
\texttt{1.attn.k:485}-\texttt{1.attn.q:308} interaction plays a mostly
overlapping role to the \texttt{1.attn.k:218}-\texttt{1.attn.q:308}
interaction}{The 1.attn.k:485-1.attn.q:308 interaction plays a mostly overlapping role to the 1.attn.k:218-1.attn.q:308 interaction}}\label{the-1.attn.k485-1.attn.q308-interaction-plays-a-mostly-overlapping-role-to-the-1.attn.k218-1.attn.q308-interaction}

The other subcomponent with which \texttt{1.attn.q:308} has a strong
interaction is \texttt{1.attn.k:485}. It has strongest subcomponent
activation on the word ``there'', but also activates for ``here'' and
``it'' (and all their capitalized variants). It tends to be causally
important when any of these words is followed by a copula verb.

\begin{figure}[htbp]
\centering
\includegraphics[width=\textwidth,height=0.8\textheight,keepaspectratio]{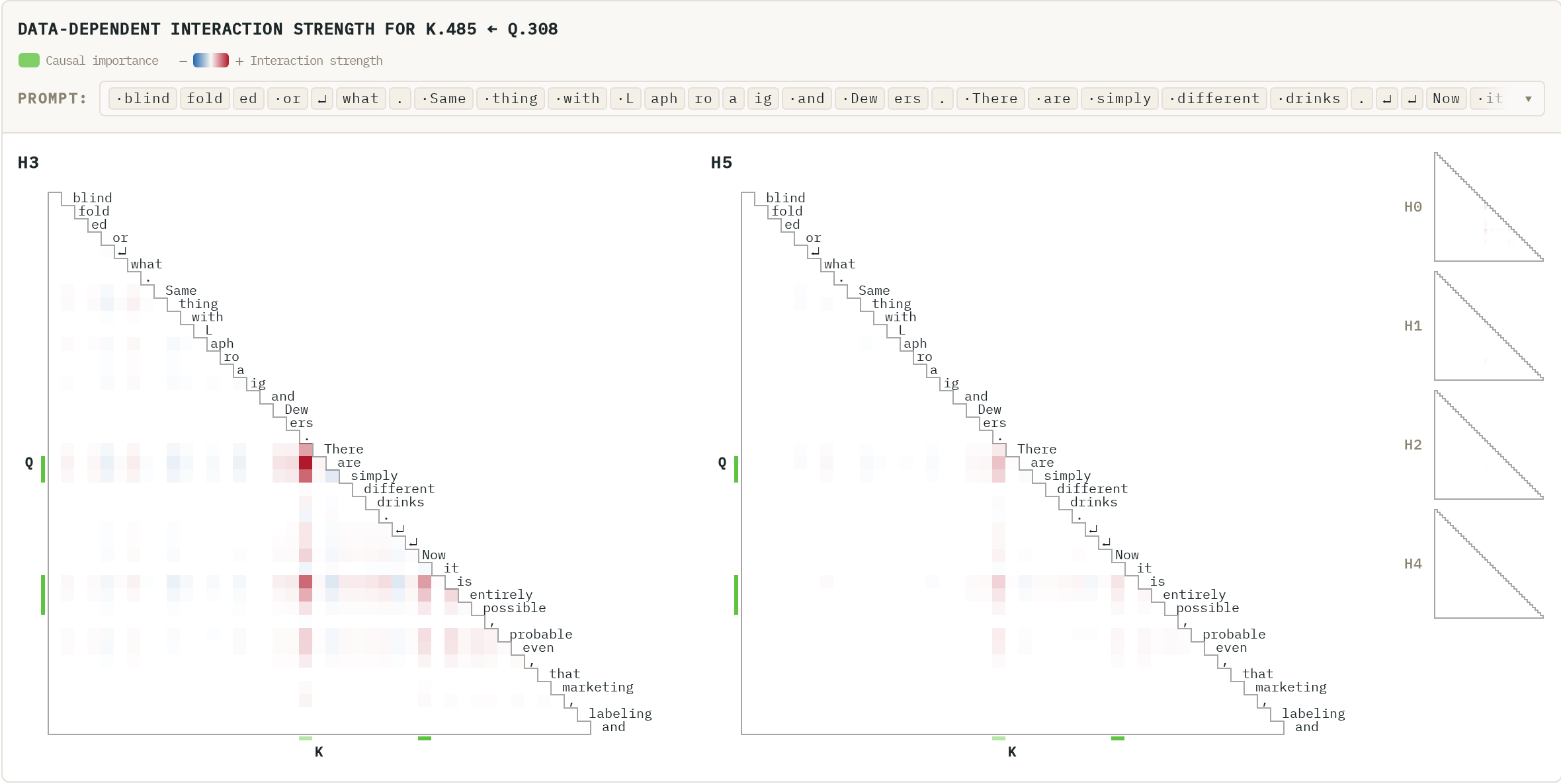}
\caption[]{\emph{Static snapshot of an interactive figure; explore it at \href{https://www.goodfire.com/research/interpreting-lm-parameters}{the web version}.}}
\end{figure}

Together, this indicates that the interaction between
\texttt{1.attn.k:485} and \texttt{1.attn.q:308} causes attention to
\textbf{existential constructions}, such as ``There is'', ``Here are'',
``There exists'', as well as \textbf{expletive constructions} (which we
studied in detail in the previous subsection). This means that its
function overlaps with the function of the \texttt{1.attn.k:218} and
\texttt{1.attn.q:308} interaction, which also detects expletive
constructions.

However, the \texttt{1.attn.k:485}-\texttt{1.attn.q:308} interaction
contributes relatively less attention to expletive constructions
compared with the interaction between \texttt{1.attn.k:218} and
\texttt{1.attn.q:308}. For example, in the prompt below, the
\texttt{1.attn.k:485}-\texttt{1.attn.q:308} interaction misses the
`expletive ``it''\,' in ``make it probable'' while
\texttt{1.attn.k:218}-\texttt{1.attn.q:308} detects it and causes
attention to it.

\begin{figure}[htbp]
\centering
\includegraphics[width=\textwidth,height=0.8\textheight,keepaspectratio]{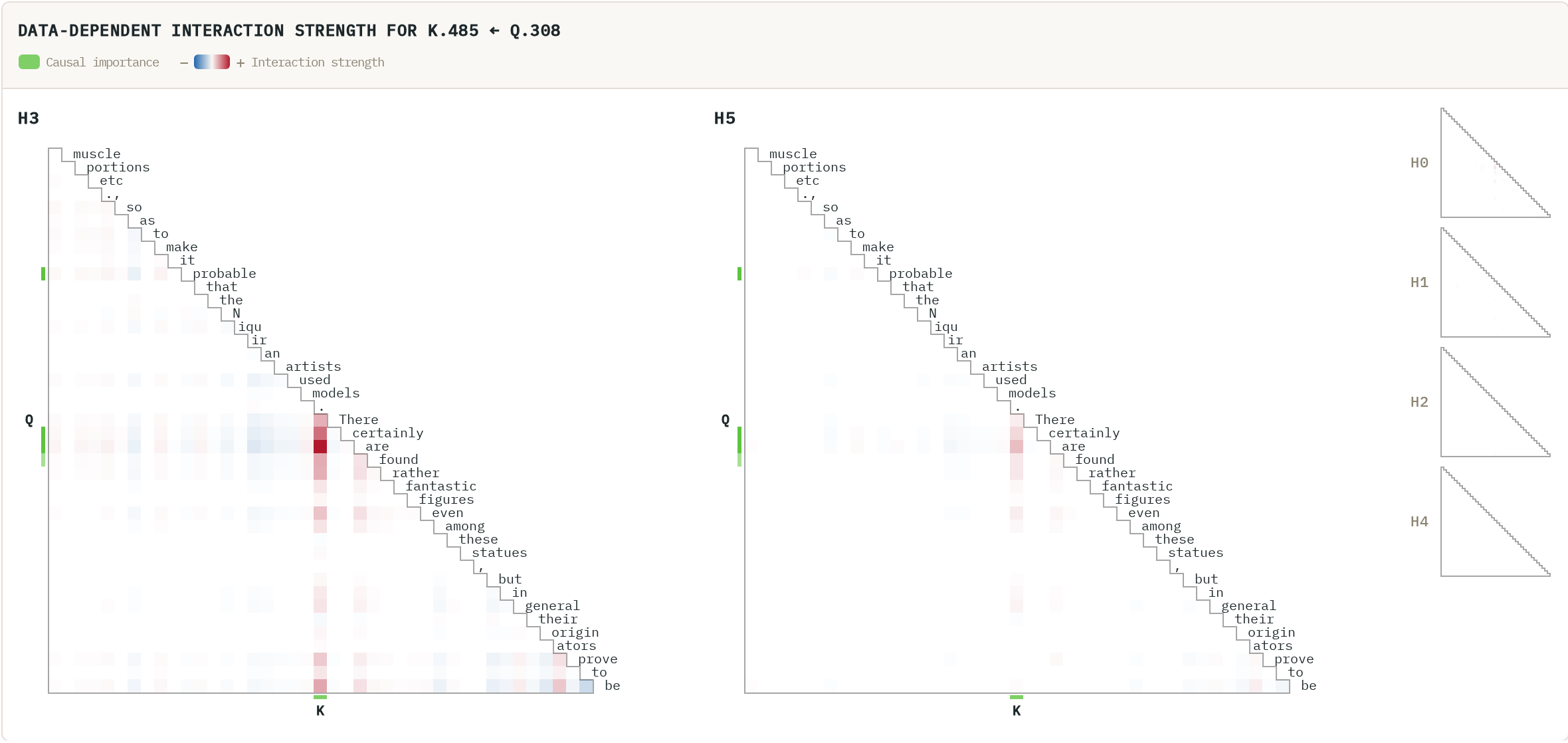}
\caption[]{\emph{Static snapshot of an interactive figure; explore it at \href{https://www.goodfire.com/research/interpreting-lm-parameters}{the web version}.}}
\end{figure}

\begin{figure}[htbp]
\centering
\includegraphics[width=\textwidth,height=0.8\textheight,keepaspectratio]{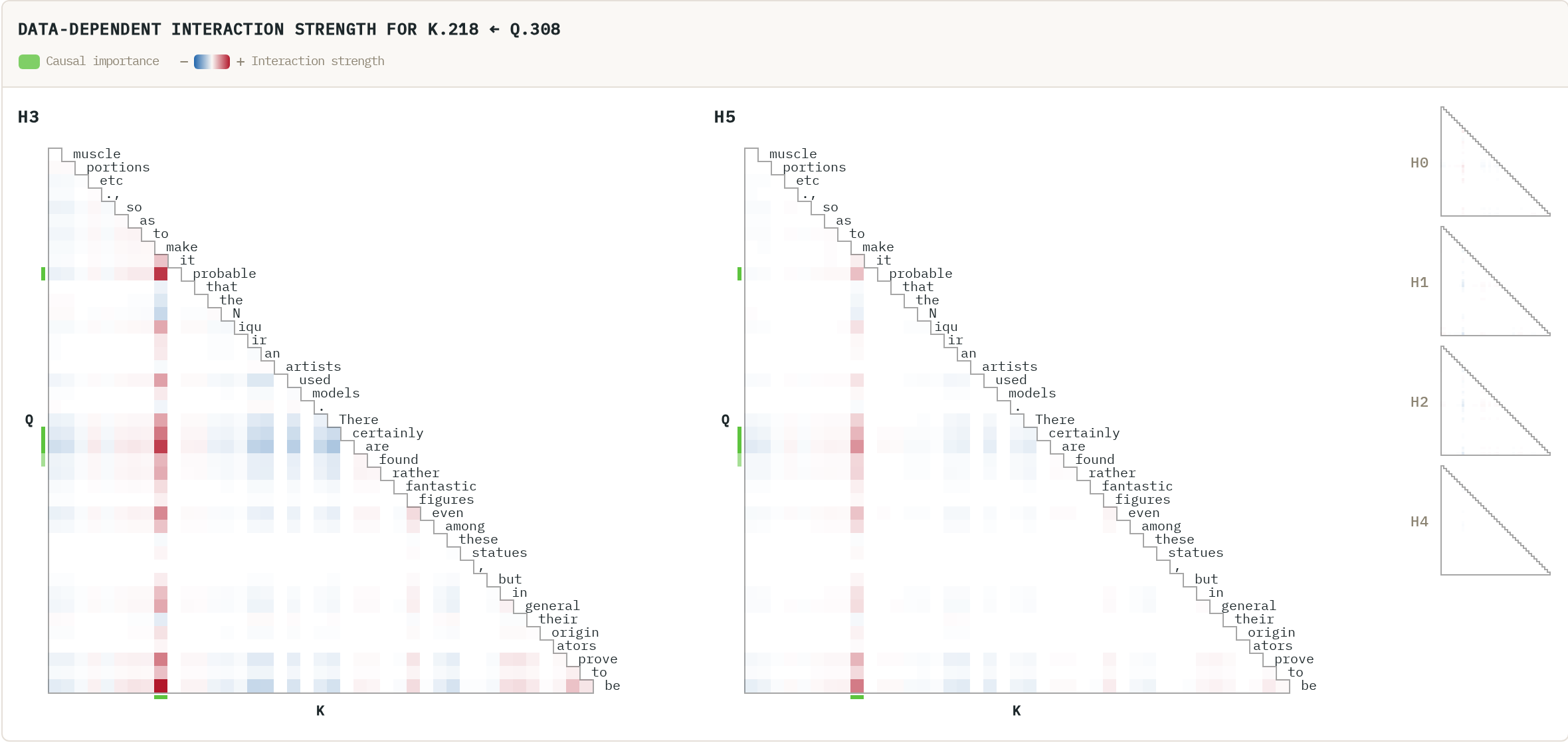}
\caption[]{\emph{Static snapshot of an interactive figure; explore it at \href{https://www.goodfire.com/research/interpreting-lm-parameters}{the web version}.}}
\end{figure}

These two interactions thus both play overlapping, but somewhat
specialized roles in detecting what type of construction a copula verb
is in.

\subsubsection{Both QK subcomponent interactions have similar OV
circuits}\label{both-qk-subcomponent-interactions-have-similar-ov-circuits}

Their overlapping, but slightly distinct, roles are reflected by their
OV circuits.

If either \texttt{1.attn.k:218} and \texttt{1.attn.k:485} are causally
important, the \(W_V\) subcomponents with the highest probability of
also being causally important are (\autoref{fig:pkv}):

\begin{itemize}
\item
  \texttt{1.attn.v:744}
\item
  \texttt{1.attn.v:180}
\item
  \texttt{1.attn.v:946}
\item
  \texttt{1.attn.v:649}
\end{itemize}

However, both \(W_K\) subcomponents do not have identical relationships
with all \(W_V\) subcomponents. Subcomponent \texttt{1.attn.v:448} seems
only to have a high conditional probability of being causally important
with \texttt{1.attn.k:485}, not \texttt{1.attn.k:218}. Combined, these
values seem to be carrying both grammatical and `content' information.
It's worthwhile noting that these \(W_V\) subcomponents are not
localized to particular heads, and therefore their information may be
mediated via more than one head (\autoref{fig:vo_comp_weight_norm}).

On a normative level, why does the model learn these two behaviors and
implement them in this way? On one level, the answer is somewhat
obvious: These constructions (existential, expletive, anaphoric) tend to
be followed by different types of text, which therefore demands
different kinds of predictions. On another level, it feels likely that a
better model could have implemented better detectors. To determine
whether layer 1 is simply too early in the model for a `cleaner'
implementation, or whether the model is simply too small, would require
further investigation. We leave those investigations, as well as studies
of how these overlapping, but separable, detectors influence downstream
computations, to future work.

We have barely scratched the surface of the extent and complexity of
attention computations of even this small model. Nonetheless, we are
excited by the possibilities for understanding attention computations
opened up by decomposing attention layer parameters into parameter
subcomponents. We believe the breadth of this analysis could be
massively increased and note there is significant room for increasing
the depth analysis that use parameter subcomponents to decompose and
understand attention. We have not, for instance, studied how parameter
subcomponents could interact across attention layers, perhaps forming
structures akin to `virtual attention heads', but decomposed into their
constituent parameter subcomponents.

\section{Interpreting circuits of parameter
subcomponents}\label{interpreting-circuits-of-parameter-subcomponents}

\label{sec:circuits}

So far, we have studied parameter subcomponents individually, or one
attention layer at a time, looking at how they combine within a single
attention layer to produce behaviors like previous-token movement and
previous-syntactic-boundary movement. But the outputs of a language
model are computed using many layers in series. In this section, we use
parameter subcomponents to understand at least some aspects of the
target model's internal computations from the input embedding all the
way to the output on a few different prompts.

To make sense of these multi-step computations, we need a way to study
how information flows between parameter subcomponents throughout the
entire model. We do this by calculating attributions, which measure the
strength of the interaction between causally important subcomponents on
particular prompts. The resulting attribution graphs let us trace, on
individual prompts, how information moves between subcomponents across
layers. In particular, we use gradient attributions, but use
stop-gradients on every node other than the source and target so that we
measure only the `direct' effects of one subcomponent on another
(\autoref{sec:attr-calcs}).

It should be noted that using gradients in this way `abstracts away' the
complexity of non-linear interactions between subcomponents by
summarizing them into a single number. As a result, such attributions
are only `local' measures of interaction strength; their value depends
on the particular datapoint that we measure them on. Many works have
pointed out issues (such as saturated softmax functions in attention
layers) that can cause such local attributions to be unrepresentative of
more `global' measures
\citep{kramar2024atpefficientscalablemethod, jafari2025relpfaithfulefficientcircuit}.
In order to identify more `global' measures of interaction strength, we
would need to better characterize the nonlinear relationships between
parameter subcomponents. This is an important research priority, and one
that we've already begun exploring, but not something that this paper
covers in detail. We do nonetheless provide analysis that suggests
parameter subcomponents of MLP matrices, despite not being directly
selected to have simple interactions, tend toward it anyway
(\autoref{app:interactions-gis-vs-coact}).

\subsection{Attribution calculations}\label{attribution-calculations}

\label{sec:attr-calcs}

To calculate attributions between two subcomponents, we leverage
gradients. In particular, we calculate the gradients between each
``subcomponent activation'',
\(a^l_c = (\vec{V}^l_c)^\top \vec{\varphi}^l\). However, we do not
always simply use \(\frac{\partial a_{c}}{\partial a_{c'}}\), the
partial derivative of the target subcomponent activation \(a_{c}\) with
respect to the source subcomponent activation. The partial derivative
measures the influence of \(a_{c'}\) on \(a_{c}\) through both
\emph{direct} and \emph{indirect} pathways. Understanding the direct
effects of a subcomponent give us the clearest mechanistic picture of
its role in the network's neural algorithm. We therefore need an
attribution method that can distinguish between direct and indirect
effects, unlike the partial derivative
\(\frac{\partial a_{c}}{\partial a_{c'}}\). But, complicating matters
further, in models with residual streams a subcomponent's direct effects
are not limited only to those in the immediate next layer. The direct
effects may skip many layers!

\begin{figure}[htbp]
\centering
\includegraphics[width=0.85\textwidth,height=0.8\textheight,keepaspectratio]{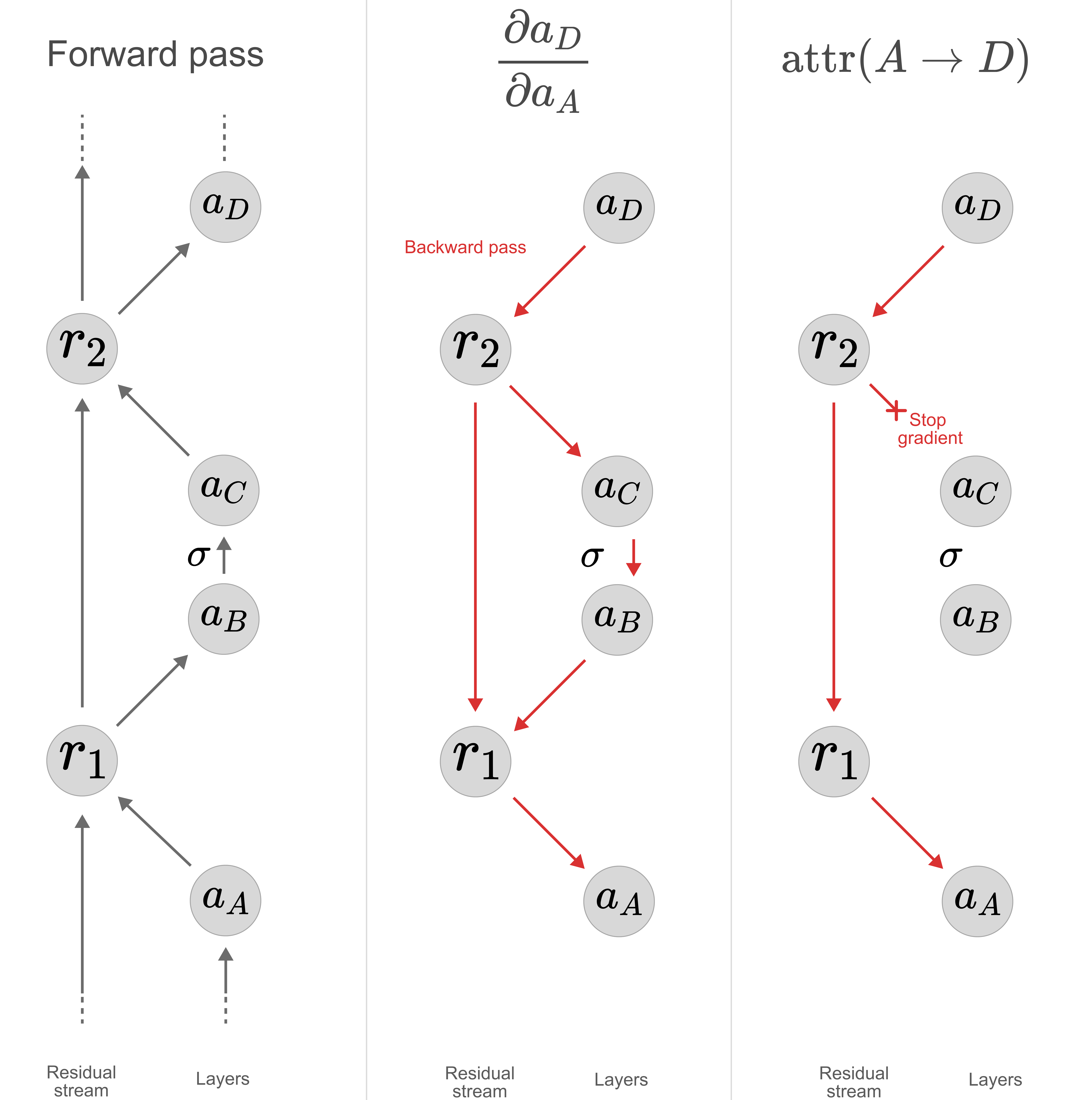}
\caption[]{To exclude indirect effects (i.e. effects that one parameter subcomponent has on another that are mediated by intermediate parameter subcomponents), we stop the gradients flowing through all subcomponents that are not the source subcomponent.}\label{fig:attr-graph-expl}
\end{figure}

Instead of using the partial derivative
\(\frac{\partial a_{c}}{\partial a_{c'}}\), we use the fact that we can
control how gradients flow on the backwards pass. We take the partial
derivative \(\frac{\partial a_{c}}{\partial a_{c'}}\), but we stop the
gradients flowing through all subcomponents that are not the source
subcomponent (\autoref{fig:attr-graph-expl}). This avoids measuring
their effects on the target node, including the indirect effects of the
source node that flow through them.

This derivative approximates how sensitive the target node is to the
source node. Our attribution multiplies this ``sensitivity'' by the
strength of the activation of the source node in order to measure its
overall influence. Additionally, we do not want to include causally
unimportant nodes in our attributions, and therefore multiply the
resulting term by the source subcomponent's causal importance:

\[\text{attr}(c' \to c) = \left( \frac{\partial a_c}{\partial a_{c'}} \right)^* \cdot a_{c'} \cdot g_{c'}\]

where the \(*\) around the partial derivative denotes stopped gradients
on non-source subcomponents.

For more details on our gradient attributions, see
\autoref{app:gradient_attributions}.

\subsection{Pruning for specific
behaviors}\label{pruning-for-specific-behaviors}

\label{sec:attr-graph-post-proc}

Most prompts, even simple ones, tend to activate hundreds of parameter
subcomponents, which is too many to analyze at once!

We can further reduce the number of subcomponents we need to analyze by
keeping only those subcomponents involved in computing some particular
output behavior on a prompt that we are interested in.

To do this, we use two types of pruning methods, yielding two types of
attribution graph. We'll use both types in our analysis below:

\begin{itemize}
\tightlist
\item
  \textbf{Attribution graph type 1: Pruning with adversarial sampling}:
  Suppose, on the prompt \texttt{The} \texttt{·princess} \texttt{·lost}
  \texttt{·her} \texttt{·crown} \texttt{.}, we wanted to analyze how the
  model successfully predicts \texttt{·her}. We would therefore only be
  interested in subcomponents that were involved in computing this
  specific prediction at this specific sequence position, which is a
  smaller subset of subcomponents than the set used to predict all
  tokens at all sequence positions. We can therefore find \emph{new}
  causal importances to identify only those subcomponents by minimizing
  a set of masks while demanding that cross-entropy reconstruction loss
  on the label \texttt{·her} on the sequence position for \texttt{·lost}
  remains good. This is different from standard VPD training, which
  instead used KL-divergence to all the target model's output
  probabilities on all sequence positions of the prompt. As in VPD
  training, we optimize causal importances under both stochastic and
  adversarial mask sampling to try to ensure that the resulting graphs
  are mechanistically faithful. For details about this technique, see
  \autoref{app:posthoc_ci}.
\end{itemize}

One might wonder whether adversarial sampling is actually necessary for
mechanistic faithfulness for this post-hoc pruning. After all, the
parameter subcomponents are now frozen and only the causal importances
can change, so the optimization has much fewer degrees of freedom to
create spurious graphs that score well on the loss. To investigate this,
we use the second type of attribution graph:

\begin{itemize}
\tightlist
\item
  \textbf{Attribution graph type 2: Pruning with only causal importances
  as masks (no stochastic or adversarial sampling)}: We can also make
  attributions graphs by making a new set of causal importances. But in
  this case, we minimizing the causal importances without stochastic or
  adversarial sampling. We still demand that cross-entropy
  reconstruction loss on the label \texttt{·her} on the sequence
  position for \texttt{·lost} remains good, but this is much easier
  because we no longer need to be robust to stochastic or adversarial
  sampling of the causal importance masks. optimize the graph only for
  the nodes that are necessary for making a specific prediction (as in
  the previous type of graph).
\end{itemize}

In the case studies, we will see that graph type 2 often looks
interpretable and contain far fewer subcomponents than the adversarially
pruned graphs (type 1). This is somewhat misleading: Because they are
not robust to adversarial ablations of parameter subcomponents, type 2
graphs will yield mechanistically unfaithful accounts of the network's
computations. They claim that many components are causally unimportant
(ablatable), when in fact they are important (unablatable under
adversarial sampling), making them look much simpler and more
superficial than would a mechanistically faithful account of the the
network's computations. Further confirming the mechanistic
unfaithfulness of the non-adversarially pruned graphs, they often score
much better on the task than the actual target model, reaching near
\(100\%\) accuracy.\footnote{We also found that graphs pruned with only
  stochastic sampling---but no adversarial sampling---often seemed to be
  mechanistically unfaithful as well, but we do not show these results
  here.} Importantly, we believe that this issue is likely to apply in
any setting in which masking-without-adversarial-sampling is used to
identify subsets of nodes (or `subnetworks') in large causal graphs that
are important for downstream tasks, which includes a large fraction of
the neural network subgraph identification literature to date.

The most important type of graph that we study below is type 1 (Pruning
with adversarial sampling), but we will use type 2 to demonstrate the
necessity of choosing subcomponents such that the subgraph is robust to
adversarial sampling, if our aim is to give mechanistically faithful
accounts of the network's computations.

\subsection{Case study 1: Gendered possessive
pronoun}\label{case-study-1-gendered-possessive-pronoun}

\label{sec:case-studies-pronoun}

On the prompt \texttt{The} \texttt{·princess} \texttt{·lost}
\texttt{·her} \texttt{·crown} \texttt{.} the target model correctly
predicts with high probability (\(0.586\)) that \texttt{·her} follows
\texttt{·lost}. This requires recognizing that a possessive pronoun is
likely to come next, remembering that the previous token was
\texttt{·princess}, and knowing that princesses are predominantly
associated with female pronouns. How does the model perform this task?

We can use attribution graphs to follow the flow of information between
parameter subcomponents and see what information is processed and by
which parameters.

\begin{figure}[htbp]
\centering
\includegraphics[width=\textwidth,height=0.8\textheight,keepaspectratio]{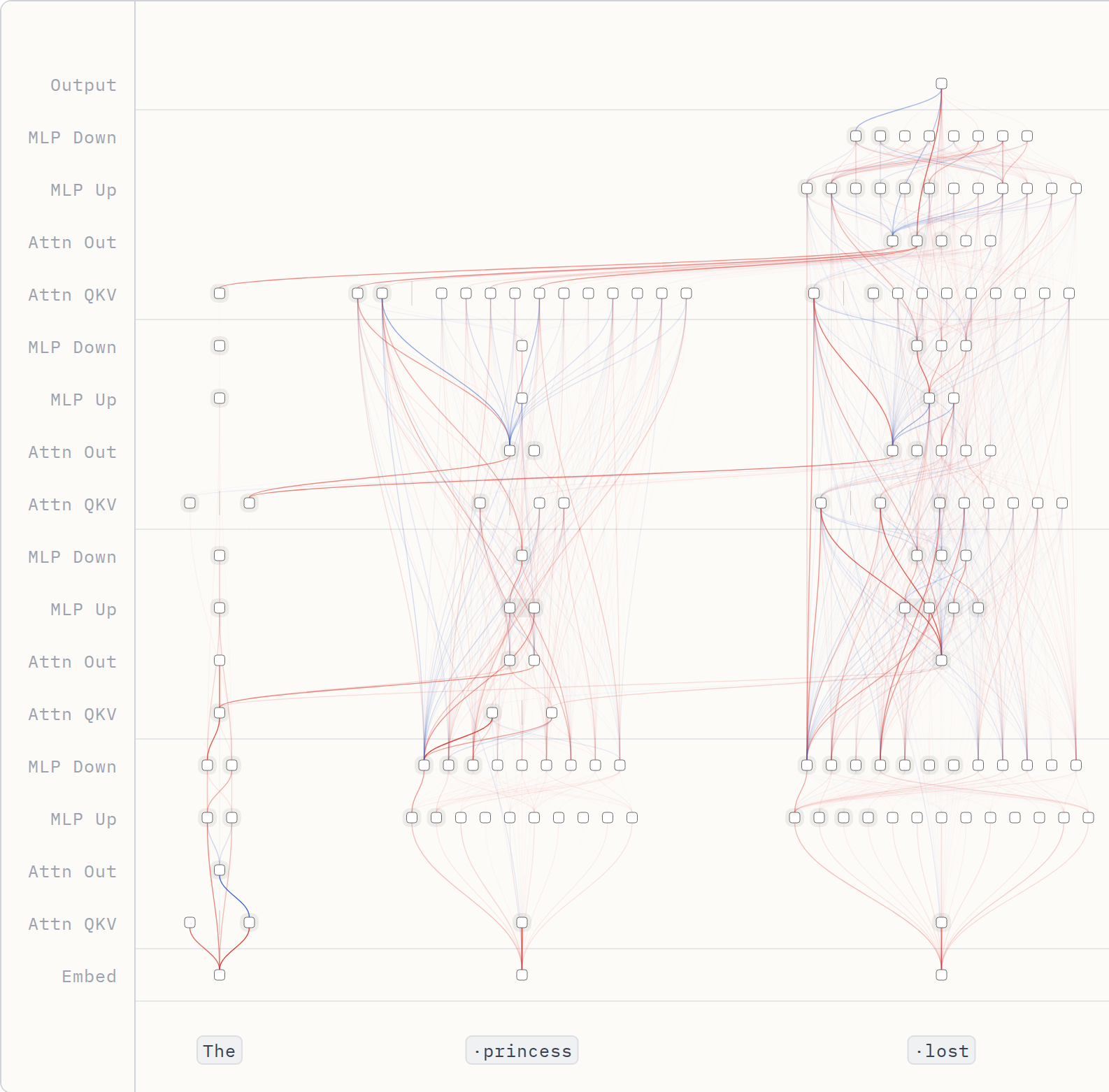}
\caption[]{Attribution graph for predicting \texttt{·her} on the prompt ``The princess lost her crown.'', pruned with adversarial sampling.\protect\footnotemark{} There are 150 subcomponents in the graph. \emph{Static snapshot of an interactive figure; explore it at \href{https://www.goodfire.com/research/interpreting-lm-parameters\#graph:princess}{the web version}.}}\label{graph:princess}
\end{figure}
\footnotetext{Coefficient \(0.5\) for cross-entropy reconstruction with stochastic sampling, coefficient \(0.5\) for cross entropy with \(4\) steps of PGD, lr \(1\), importance minimality coefficient \(0.09\), \(p=0.3\), \(2000\) optimization steps.}

Attribution graph for the prompt \texttt{The} \texttt{·princess}
\texttt{·lost} \texttt{·her} \texttt{·crown} \texttt{.} after
adversarial pruning, keeping only the subcomponents that matter for
predicting the output \texttt{·her} after \texttt{·lost}.\footnote{Using
  \(2000\) optimization steps, cross-entropy reconstruction with
  stochastic sampling, loss coefficient \(0.5\), cross entropy with
  \(4\) steps of PGD at lr\(=1\), coefficient \(0.5\), and importance
  minimality loss coefficient \(0.09\), \(p=0.3\).} The graph has a
total of 150 subcomponents. The target model assigns probability
\(0.586\) to the output \texttt{·her}. Causal importance masking with
the nodes in this graph increases that probability to \(1.000\) and
stochastic masking increases it to \(0.999\). However, adversarial
masking decreases the probability on the output \texttt{·her} to
\(0.443\), which indicates that this graph still isn't quite capturing
all the relevant computation going on in the model.

Working backward from the output, we will see that the top two positive
attributions to the output node \texttt{·her} in the graph come from two
different computational pathways.

\textbf{Pathway 1}

This pathway appears to carry information about the `femaleness' of the
\texttt{·princess} token forward in time to make the pronoun prediction
\texttt{·her}. Working backward from output node to input nodes:

The largest positive attributions to the output token \texttt{·her} is
from a layer 3 attention output subcomponent labeled
\texttt{3.attn.o:2:281}. Ablating it out of the target model changes the
top prediction to \texttt{·his}. That subcomponent, in turn, receives
its largest attribution edges from a subcomponent of the attention layer
3's \(W_K\) at the \texttt{·princess} sequence position, which is
causally important on almost every token (\texttt{3.attn.k:1:145}), and
a subcomponent of the attention layer 3's \(W_V\), likewise on the
\texttt{·princess} sequence position, labeled \texttt{3.attn.v:1:676}.

The \(W_V\) subcomponent in turn receives its top attribution from
\texttt{0.mlp.down:1:3473}, which appears to be polysemantic. It is
active on various female names and other words and sentences associated
with or about women, but also in a range of other contexts, perhaps
particularly scientific ones. Its top attribution comes from a
subcomponent of the layer 0 MLP Up projection matrix labeled
\texttt{0.mlp.up:1:327}, which then connects straight to the
\texttt{·princess} input embedding.

In summary, this pathway appears to carry a femaleness attribute from
the \texttt{·princess} sequence position to the \texttt{·lost} sequence
position using the layer 3 attention. The relevant key and query
subcomponents almost always fire, indicating that this attention routing
happens as part of the generic previous token behavior.

\textbf{Pathway 2}

The second largest positive attributions to the output \texttt{·her} is
from the layer 2 MLP Down projection subcomponent
\texttt{2.mlp.down:2:773}. It seems to also be causally important when
the model is about to predict an object pronoun, among other things
(though this detail seems to have been missed by its autointerp label
\texttt{2.mlp.down:2:773}).

The strongest attribution to this subcomponent, in turn, comes from a
layer 2 MLP Up subcomponent labeled \texttt{2.mlp.up:2:401}\footnote{It
  indeed appears to be causally important primarily on tokens that are
  verbs. Notably, whether a token is classified as a verb for this
  purpose is context-dependent. For example, in the sentence
  \texttt{I\textquotesingle{}d} \texttt{·like} \texttt{·to} \texttt{·do}
  \texttt{·something} \texttt{·like} \texttt{·this}, the subcomponent
  has high activation (\(12.7\), \(19.1\)) and causal importance \(1.0\)
  on \texttt{do} and the first \texttt{like} token, but low activation
  (\(2.9\)) and causal importance \(0\) on the second \texttt{like}
  token.}.

This subcomponent receives attribution from a diverse set of
verb-related layer 0 MLP subcomponents, such as \texttt{0.mlp.up:2:3063}
and \texttt{0.mlp.down:2:1189}, which then connect to the \texttt{·lost}
embedding.

In summary, this pathway appears to upweight object pronoun predictions
based on detecting the verb \texttt{·lost} in the input.

The top two pathways in the adversarially pruned graph suggest two core
mechanisms: one which moves the femaleness attribute of
\texttt{·princess} over to the next token via attention layer 3, and
another which detects the verb \texttt{·lost} via MLP layer 2 and
suggests that an object pronoun might follow.

\begin{figure}[htbp]
\centering
\includegraphics[width=\textwidth,height=0.8\textheight,keepaspectratio]{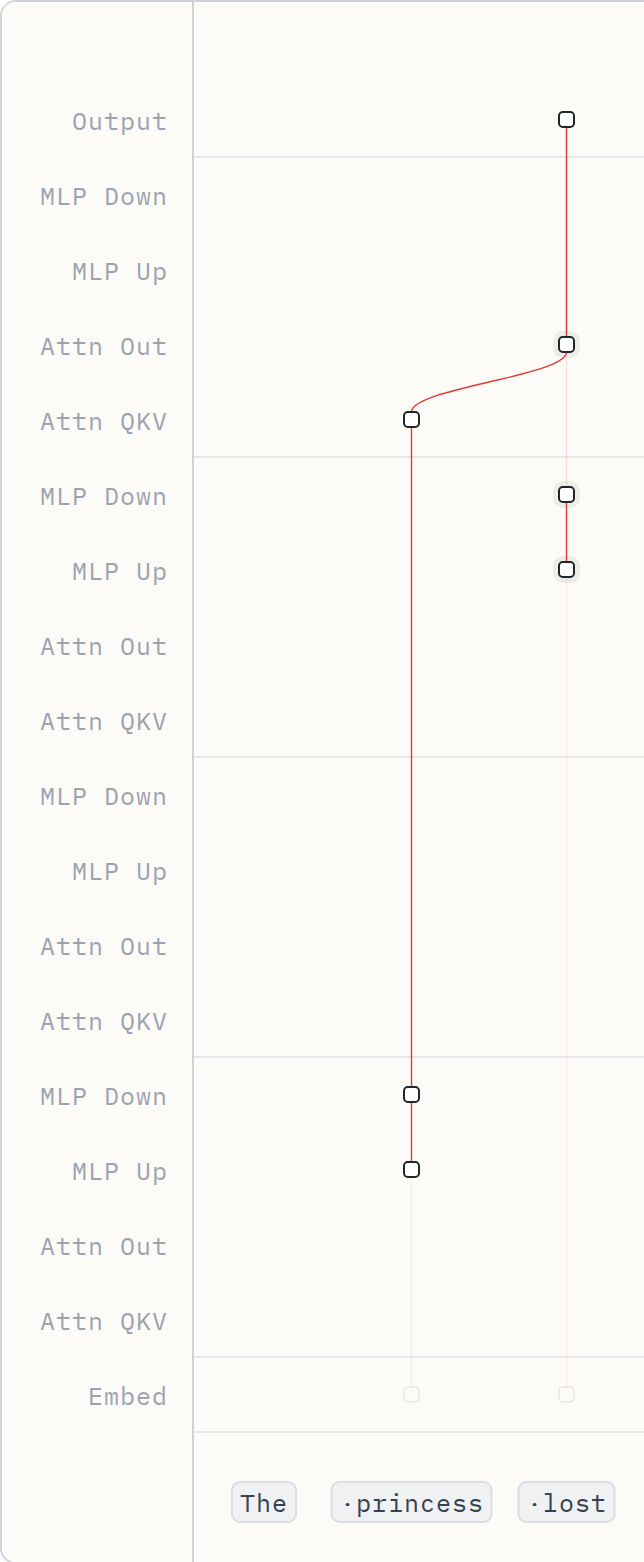}
\caption[]{Attribution graph for predicting \texttt{·her} on the prompt ``The princess lost her crown.'', pruned with causal importance masking.\protect\footnotemark{} \emph{Static snapshot of an interactive figure; explore it at \href{https://www.goodfire.com/research/interpreting-lm-parameters\#graph:princess_ci_masked}{the web version}.}}\label{graph:princess_ci_masked}
\end{figure}

If we prune the graph for high probability on \texttt{·her} using only
the causal importances as masks, neglecting adversarial robustness, we
recover a graph of just six subcomponents
(\autoref{graph:princess_ci_masked}), which corresponds almost exactly
to the most attributed subcomponents in these same two top pathways.

\footnotetext{Coefficient \(1.0\) for cross-entropy reconstruction with causal importance masking, importance minimality coefficient \(1.0\), \(p=0.3\), \(2000\) optimization steps.}

Attribution graph for predicting \texttt{·her} on the prompt
\texttt{The} \texttt{·princess} \texttt{·lost} \texttt{·her}
\texttt{·crown} \texttt{.}, pruned with causal importance
masking.\footnote{Using a cross-entropy reconstruction with causal
  importance masking, importance minimality coefficient \(1.0\),
  \(p=0.3\), \(2000\) optimization steps.} There are 6 subcomponents in
total, forming two distinct pathways. The subcomponents in these two
pathways correspond almost exactly to the most strongly attributed
subcomponents in the two top pathways of the much larger adversarially
pruned graph depicted in \autoref{graph:princess}.

\textbf{Pathway 1:} From the \texttt{·princess} embedding to
subcomponents in the layer 0 MLP Up and Down projection matrices labeled
\texttt{0.mlp.up:1:327} and \texttt{0.mlp.down:1:3473}, to a
subcomponent in the layer 3 attention value matrix labeled
\texttt{3.attn.v:1:676} to a subcomponent in the layer 3 attention
output matrix on the \texttt{·lost} sequence position labeled
\texttt{3.attn.o:2:281}.

\textbf{Pathway 2:} From the \texttt{·lost} embedding to subcomponents
in the layer 2 MLP Up and Down projection matrices labeled
\texttt{2.mlp.up:2:401} and \texttt{2.mlp.down:2:773}.

For more discussion of the subcomponents in these pathways, see
\autoref{graph:princess}.

The target model assigns probability \(0.586\) to the output
\texttt{·her}. Causal importance masking with the six subcomponents in
this small graph increases this probability to \(0.895\), and stochastic
masking based on the causal importances in this graph increases it even
more, up to \(0.969\).

One might then falsely suppose that these six subcomponents perform all
the important computation for this pronoun prediction task and the other
subcomponents are, if anything, just a hindrance. But evaluation with
adversarial masking based on the causal importances in this
graph\footnote{With \(4\) PGD optimization steps at learning rate \(1\).}
drops the probability on the \texttt{·her} prediction down to
\(<0.0005\), revealing that this isn't true at all.

This confirms that these six subcomponents are sufficient for
reproducing the desired output. This much smaller graph even generalises
to slightly different prompts: On the input \texttt{The} \texttt{lady}
\texttt{·lost} \texttt{·her} \texttt{·crown} \texttt{.}, a forward pass
using only the six subcomponents in the small graph at the exact same
sequence positions also recovers the target model's \texttt{·her}
prediction\footnote{With output probability \(0.895\) under causal
  importance masking, and \(0.275\) under stochastic masking.}. But the
lack of adversarial robustness in the smaller graph confirms that it
does not provide anything close to a full account of the relevant
computation going into the model's prediction.\footnote{Pruning with
  stochastic masking doesn't perform any better. A graph for the
  princess prompt we pruned with stochastic masking ended up with \(14\)
  subcomponents in total, and still assigned probability \(<0.0005\)
  under adversarial masking.} All 150 subcomponents in
\autoref{graph:princess} likely play some role --- otherwise the
optimization would have pruned them. While these six subcomponents
suffice to put high probability on \texttt{·her}, they fail to suppress
other computational pathways that would predict different outputs. We do
not attempt to fully understand the complete graph here.

How similar are the structures of graphs for slightly different prompts?
To investigate this, we run the same analysis on the prompt \texttt{The}
\texttt{·prince} \texttt{·lost} \texttt{·his} \texttt{·crown}
\texttt{.}, where the target model predicts \texttt{·his} with
probability \(0.512\). We recover similar, but not identical results.

\begin{figure}[htbp]
\centering
\includegraphics[width=\textwidth,height=0.8\textheight,keepaspectratio]{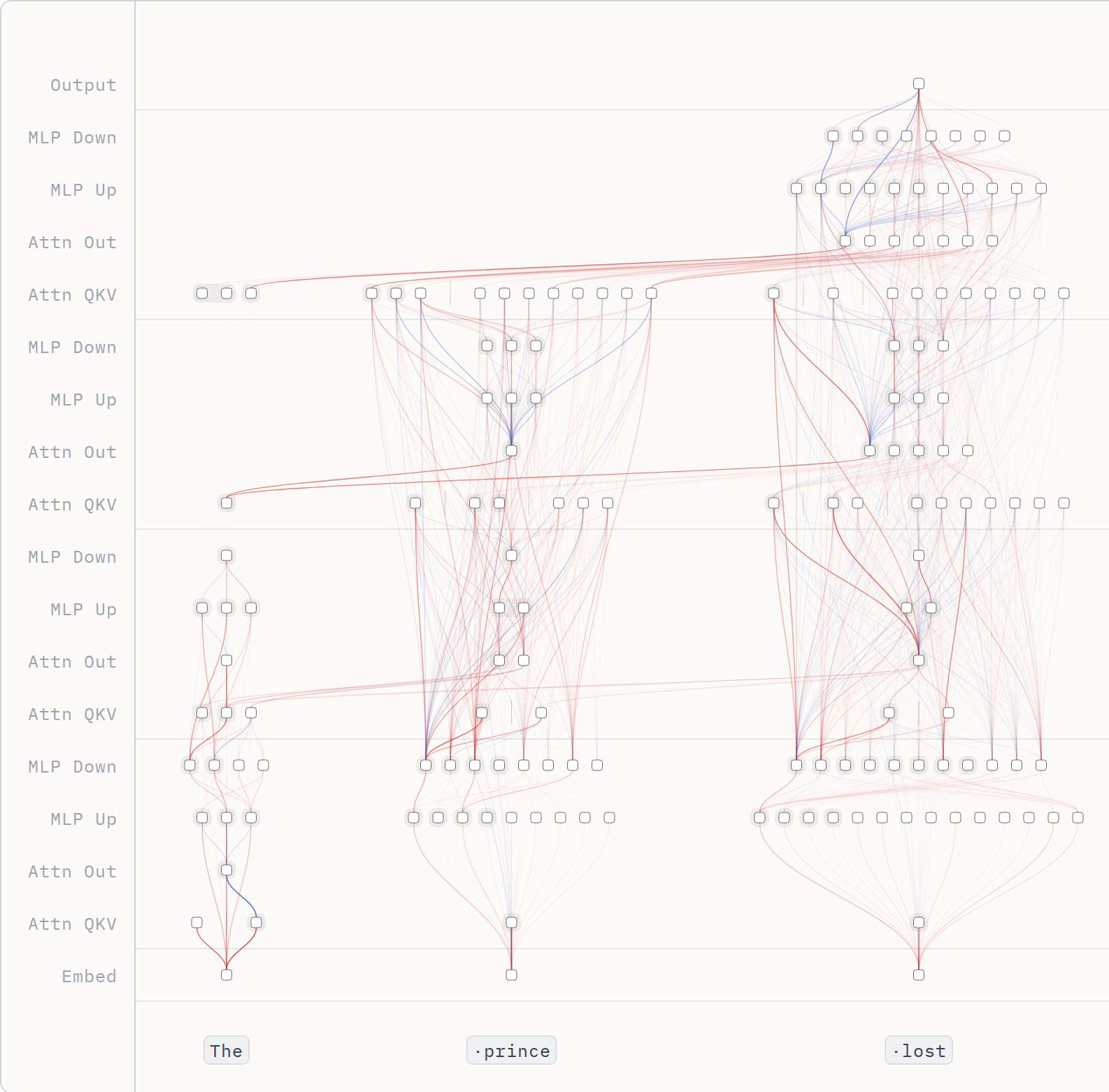}
\caption[]{Attribution graph for predicting \texttt{·his} on the prompt ``The prince lost his crown.'', pruned with adversarial sampling.\protect\footnotemark{} There are 160 subcomponents in the graph. The target model assigns probability \(0.512\) to \texttt{·his}. \emph{Static snapshot of an interactive figure; explore it at \href{https://www.goodfire.com/research/interpreting-lm-parameters\#graph:prince-full}{the web version}.}}\label{graph:prince-full}
\end{figure}
\footnotetext{Coefficient \(0.5\) for cross-entropy reconstruction with stochastic sampling, coefficient \(0.5\) for cross entropy with \(4\) steps of PGD, lr \(1\), importance minimality coefficient \(0.05\), \(p=0.3\), \(2000\) optimization steps.}

Attribution graph for predicting \texttt{·his} on the prompt
\texttt{The} \texttt{·prince} \texttt{·lost} \texttt{·his}
\texttt{·crown} \texttt{.}, pruned with adversarial sampling.\footnote{Here
  we use coefficient \(0.5\) for the cross-entropy reconstruction loss
  with stochastic sampling, coefficient \(0.5\) for cross entropy loss
  with \(4\) steps of PGD, lr \(1\), importance minimality coefficient
  \(0.05\), \(p=0.3\), \(2000\) optimization steps.} The graph has a
total of 160 subcomponents. The target model assigns probability
\(0.512\) to \texttt{·his}. Causal importance masking with the nodes in
this graph increases that probability to \(1.000\) and stochastic
masking increases it to \(0.998\). However, adversarial
masking\footnote{With 4 PGD optimization steps at learning rate \(1\).}
decreases the probability on \texttt{·his} to \(0.383\), which indicates
that this graph still isn't quite capturing all the relevant computation
going on in the model.

The graph is structurally similar to the adversarially pruned graph for
the \texttt{·princess} prompt in \autoref{graph:princess}. 95 of the 150
subcomponents in that graph also show up at the same sequence position
in this graph, including the subcomponents we discussed that form a
pathway for upweighting object pronoun predictions based on detecting
the verb \texttt{·lost} in the input. However, as we might expect, the
subcomponents for moving the femaleness attribute to the next sequence
position is not present here.

As with the princess prompt, pruning with CI masking instead of
adversarial masking recovers a much smaller graph of just six
subcomponents organised into two pathways that is sufficient to compute
the \texttt{·his} prediction, but isn't adversarially robust at all.

\begin{figure}[htbp]
\centering
\includegraphics[width=\textwidth,height=0.8\textheight,keepaspectratio]{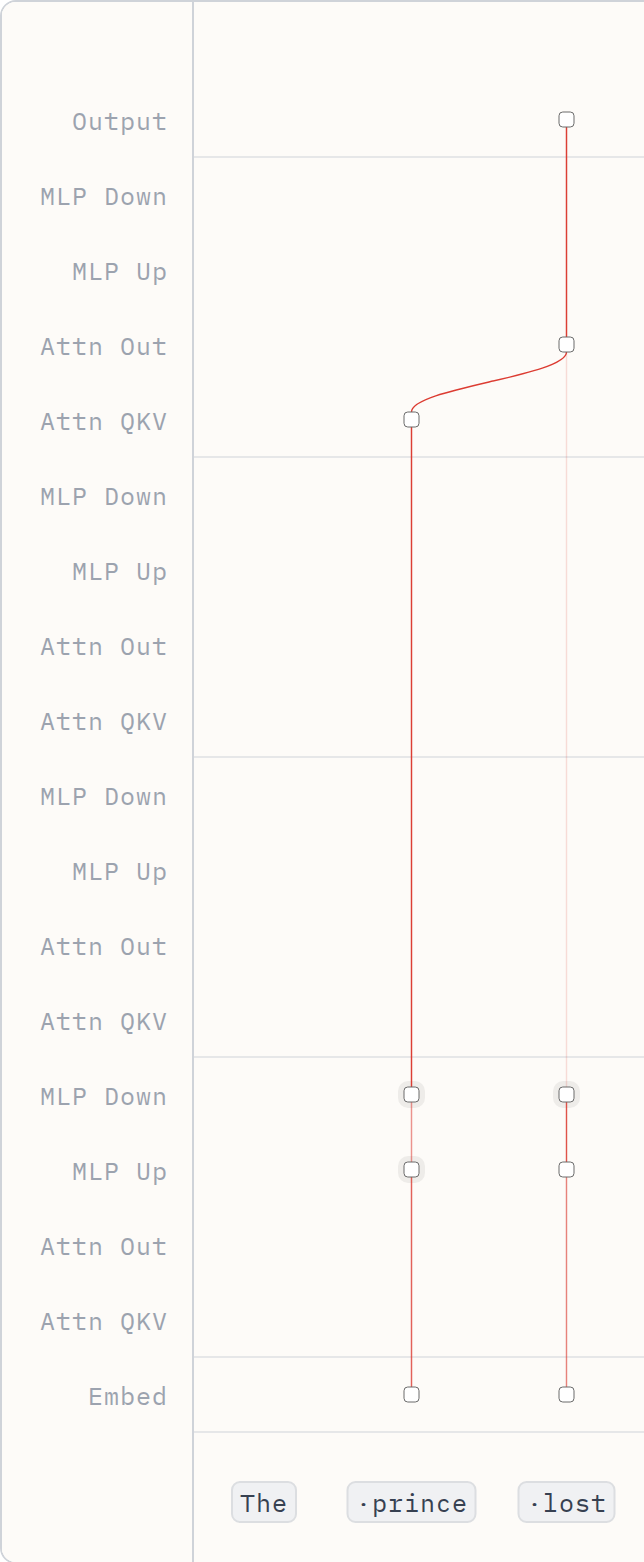}
\caption[]{Attribution graph for predicting \texttt{·his} on the prompt ``The prince lost his crown.'', pruned with causal importance masking. \emph{Static snapshot of an interactive figure; explore it at \href{https://www.goodfire.com/research/interpreting-lm-parameters\#graph:prince-minimal}{the web version}.}}\label{graph:prince-minimal}
\end{figure}

Attribution graph for predicting \texttt{·his} on the prompt
\texttt{The} \texttt{·prince} \texttt{·lost} \texttt{·his}
\texttt{·crown} \texttt{.}, pruned with causal importance
masking.\footnote{coefficient \(1.0\) for cross-entropy reconstruction
  with causal importance masking, importance minimality coefficient
  \(1.0\), \(p=0.3\), \(2000\) optimization steps.}

The six subcomponents in this graph form two pathways, mirroring the two
pathways in the graph for the princess prompt.

\textbf{Pathway 1}: From the \texttt{·prince} embedding to subcomponents
in the layer 0 MLP Up and Down projection matrices labeled
\texttt{0.mlp.up:1:2822} and \texttt{0.mlp.down:1:3455}, to
subcomponents in the layer 3 attention value and output matrices labeled
\texttt{3.attn.v:1:1010} and \texttt{3.attn.o:2:776}.

Compared to the four subcomponents in the corresponding core pathway
moving the female attribute from \texttt{·princess} to \texttt{·lost} in
\autoref{graph:princess}, these four subcomponents seem less gender
specific, firing in both male and female contexts, though more often
male ones. This suggests a mechanism under which male pronoun prediction
is the default unless actively contradicted. Reinforcing this
hypothesis, running the princess prompt with just the six subcomponents
in this graph results in the model predicting \texttt{·his} rather than
\texttt{·her}.

\textbf{Pathway 2}: From the \texttt{·lost} embedding to a subcomponent
in the layer 0 MLP Up projection matrix labeled
\texttt{0.mlp.up:2:3063}, to a subcomponent in the layer 0 MLP Down
projection matrix labeled \texttt{0.mlp.down:2:1189}. These two
subcomponents also formed part of the second core pathway for the
\texttt{·princess} prompt we discussed before, see
\autoref{graph:princess}.

We stress again that the above is far from a complete account of the
meaningful computation going on in the model for these input prompts. We
have merely traced out the flow of information between a subset of
subcomponents that are sufficient for computing the output, which is
much smaller than the subset of subcomponents that are actually involved
in computing the output.

\subsection{Case study 2: Bracket
closing}\label{case-study-2-bracket-closing}

\label{sec:case-studies-bracket}

On the prompt \texttt{\textless{}} \texttt{u} \texttt{,} \texttt{v}
\texttt{\textgreater{}} the target model correctly predicts that
\texttt{\textgreater{}} follows \texttt{v}, assigning probability
\(0.547\). This requires the model to remember that, earlier in the
sentence, \texttt{\textless{}} opened a bracket that now needs to be
closed. How does the model perform this task?

\begin{figure}[htbp]
\centering
\includegraphics[width=\textwidth,height=0.8\textheight,keepaspectratio]{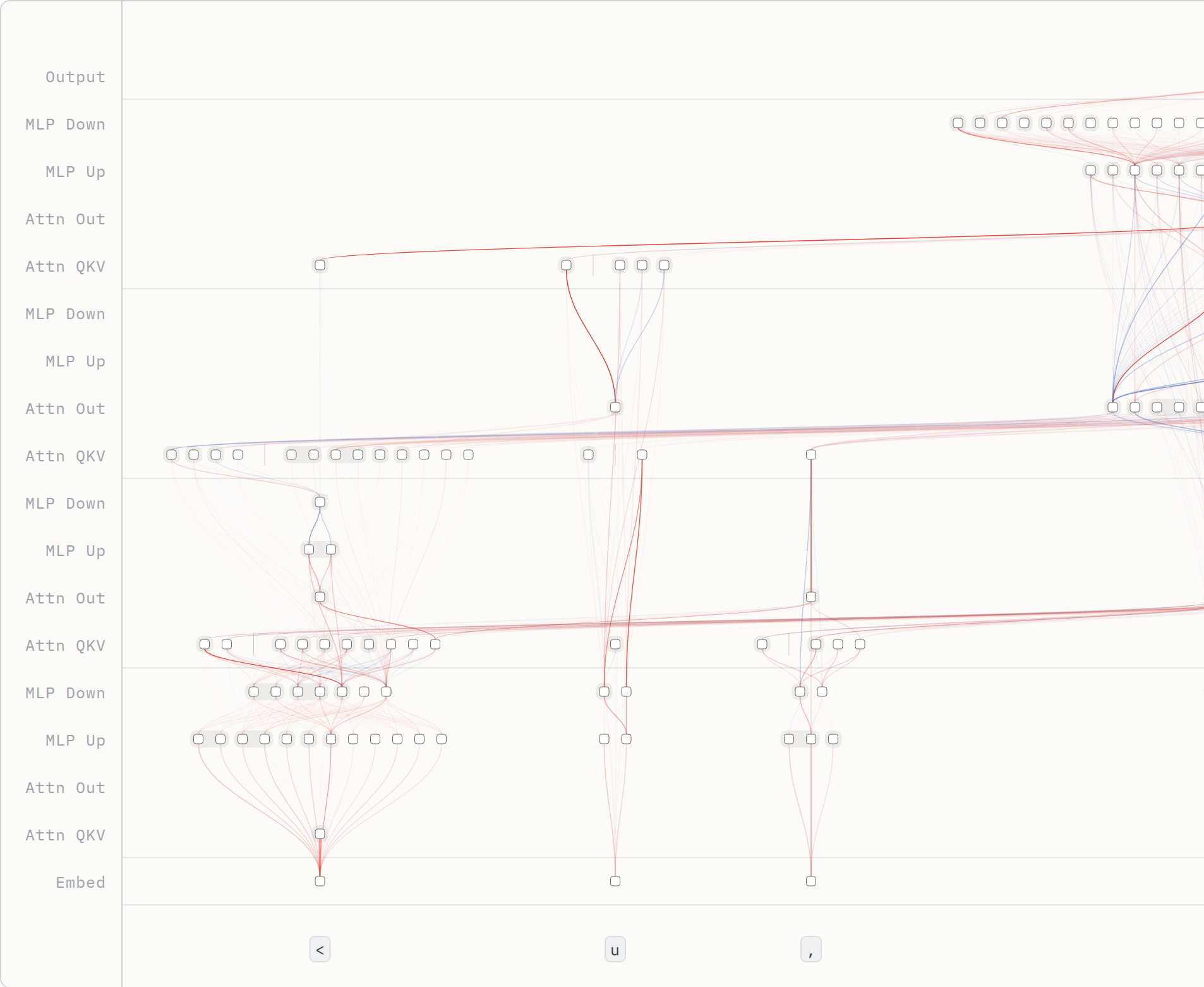}
\caption[]{Attribution graph for predicting \texttt{>} after \texttt{v} on the prompt \texttt{<} \texttt{u} \texttt{,} \texttt{v} \texttt{>}, pruned with adversarial sampling. \emph{Static snapshot of an interactive figure; explore it at \href{https://www.goodfire.com/research/interpreting-lm-parameters\#graph:bracket}{the web version}.}}\label{graph:bracket}
\end{figure}

Attribution graph for predicting \texttt{\textgreater{}} after
\texttt{v} on the prompt \texttt{\textless{}} \texttt{u} \texttt{,}
\texttt{v} \texttt{\textgreater{}}, pruned with adversarial
sampling.\footnote{Coefficient \(0.5\) for cross-entropy reconstruction
  with stochastic sampling, coefficient \(0.5\) for cross entropy with
  \(4\) steps of PGD, lr \(1\), importance minimality coefficient
  \(0.05\), \(p=0.3\), \(4000\) optimization steps.} The target model
predicts \texttt{\textgreater{}} after \texttt{v} with probability
\(0.547\).

Most of the 158 subcomponents in the graph appear to be specialized for
predicting closing delimiters, closing angled brackets more
specifically, or closing angled brackets in particular, spanning large
subspaces within the model.

The two largest positive attributions to the output
\texttt{\textgreater{}} come from:

\begin{enumerate}
\def\labelenumi{\arabic{enumi}.}
\item
  A layer 3 MLP Down projection matrix subcomponent labeled
  \texttt{3.mlp.down:3:1414}
\item
  A layer 2 MLP Down projection matrix subcomponent labeled
  \texttt{2.mlp.down:3:1560}
\end{enumerate}

Ablating these two subcomponents out of the target model severely
degrades the \texttt{\textgreater{}} prediction, lowering the
probability from \(0.547\) to \(0.158\) and \(0.243\) for individual
ablations, and to \(0.046\) for joint ablation. The model instead
reassigns probability mass to other delimiters such as \texttt{)},
\texttt{\_}, \texttt{,} or \texttt{)\$}, suggesting that the model still
knows there is an open delimiter to close, but not that it is a right
angled bracket in particular.

These subcomponents must rely on information about the open angled
bracket received from the previous sequence position. We can see in the
graph that information is carried from the \texttt{\textless{}} position
to the \texttt{v} position through attention at layers 1, 2, and 3. In
the following, we will give a brief survey of the attention
subcomponents involved in this transfer.

\textbf{Layer 1 attention summary}

As we will see in the following, interpretations of the query and key
subcomponents in layer 1 suggest that information about the preceding
open angled bracket is moved from the \texttt{\textless{}} sequence
position to the \texttt{v} sequence position in this layer as a result
of both generic previous token behavior, and as part of a behavior that
moves information at formatting boundaries to following sequence
positions.

Ablating the layer 1 attention output subcomponents out of the target
model on the \texttt{v} sequence position degrades performance severely,
with the model now assigning just \(0.015\) probability to
\texttt{\textgreater{}} instead of \(0.547\). Its top logit instead
becomes \texttt{\textless{}\textbar{}endoftext\textbar{}\textgreater{}},
with probability \(0.056\).

Similarly, ablating the layer 1 attention output subcomponents out of
the graph reduces the probability the adversarially masked forward pass
puts on \texttt{\textgreater{}} down to \(0.021\). However, with causal
importance masking, the probability assigned to \texttt{\textgreater{}}
stays at \(\approx 1.000\). This once again indicates that using naive
masking schemes to infer causality can be very misleading, and
adversarial sampling can help us avoid underestimating the number of
subcomponents involved in the target model's computation.

\textbf{Layer 1 attention query and key matrices}

\begin{itemize}
\tightlist
\item
  A single query subcomponent labeled \texttt{1.attn.q:3:316} on the
  \texttt{v} sequence position. This indicates that the relevant query
  at this layer is triggered as part of the generic previous token
  behavior.
\item
  Two key subcomponents on the \texttt{\textless{}} sequence position,
  labeled \texttt{1.attn.k:0:329} and \texttt{1.attn.k:0:119}. This
  indicates that the \texttt{\textless{}} sequence position is attended
  to in this layer partially as part of generic previous token behavior,
  and partially as part of a behavior that moves information at
  formatting boundaries.
\item
  The key subcomponent labeled \texttt{1.attn.k:2:329} is also kept on
  the \texttt{,} sequence position, indicating that the relevant
  information there is attended to purely as part of generic previous
  token behavior.
\end{itemize}

\textbf{Layer 1 attention value matrix}

There are eight value subcomponents on the \texttt{\textless{}} sequence
position.

\begin{itemize}
\tightlist
\item
  Two subcomponents labeled \texttt{1.attn.v:0:22} and
  \texttt{1.attn.v:0:984}, which also appear on the \texttt{,} position,
  as one might expect since they seem related to a wider set of
  delimiter syntax that also includes commas.
\item
  Three subcomponents labeled \texttt{1.attn.v:0:249},
  \texttt{1.attn.v:0:788} and \texttt{1.attn.v:0:102}, which appear more
  specialized to angled brackets in particular. Their activations and
  causal importances also tend to be much lower for closing angled
  brackets than opening angled brackets. \texttt{1.attn.v:0:102} is also
  part of a larger component that also has two subcomponents in the
  layer 2 attention value matrix of this graph. Subcomponents in this
  component all seem to be causally important primarily on various left
  angle brackets, like \texttt{\textless{}},
  \texttt{\textgreater{}\textless{}/}, \texttt{\}\ \textless{}} etc.
\item
  One subcomponent labeled \texttt{1.attn.v:0:474} fires on angled
  brackets, again more strongly for opening angled brackets, but also a
  few other delimiter types, such as \texttt{:} after \texttt{A} in the
  context of a Q\&A.
\end{itemize}

\textbf{Layer 1 attention value matrix} (\emph{continued})

On the \texttt{\textless{}} sequence position:

\begin{itemize}
\tightlist
\item
  One subcomponent labeled \texttt{1.attn.v:0:504}, which fires on
  opening brackets more generally, including e.g.~\texttt{\{},
  \texttt{{[}}, and variations like \texttt{\textbackslash{}\^{}\ \{},
  as well as some delimiters like \texttt{;}, though apparently only in
  technical and math heavy contexts, and a few closing brackets like
  \texttt{);}, \texttt{\}}. Again, the subcomponents' activation on
  these closing brackets is notably lower than on the opening brackets.
\item
  Finally, one subcomponent labeled \texttt{1.attn.v:0:571}, which is
  active almost exclusively on the first or first few tokens in a
  sequence.
\end{itemize}

On the other two sequence positions:

\begin{itemize}
\tightlist
\item
  There are three subcomponents on the \texttt{,} sequence position. Two
  also appear on the \texttt{\textless{}} sequence position, see
  previous page. The third is labeled \texttt{1.attn.v:2:299}.
\item
  A subcomponent labeled \texttt{1.attn.v:1:428} is the subcomponent in
  the layer 1 attention on the \texttt{u} sequence position.
\end{itemize}

\textbf{Layer 1 attention output matrix}

There are five subcomponents in the layer 1 attention output matrix on
the \texttt{v} sequence position:

\begin{itemize}
\tightlist
\item
  One subcomponent, labeled \texttt{1.attn.o:3:899}, appears to be
  active primarily whenever an open left angled bracket
  (\texttt{\textless{}}, \texttt{.\textless{}}, etc.) has not been
  closed yet, or when the previous token was a backslash
  (\texttt{\textbackslash{}}, \texttt{\$\textbackslash{}}, etc.).
\item
  Another subcomponent, labeled \texttt{1.attn.o:3:91}, seems to be
  active on and everywhere between separators and delimiters like commas
  or semicolons in lists, and various brackets in math or code.
\item
  A subcomponent labeled \texttt{1.attn.o:3:300}, which seems to
  likewise activate primarily on tokens between delimiters, in this case
  seemingly exclusively various kinds of brackets in latex or code.
\item
  One subcomponent, labeled \texttt{1.attn.o:3:187} appears to be active
  on any markup, HTML or other code and, seemingly to a somewhat lesser
  extent, on latex.
\item
  The final subcomponent, labeled \texttt{1.attn.o:3:362}, was somewhat
  difficult for us to make sense of. It fires on short text passages in
  succession, as if it is predicting something from the moment some left
  delimiter is seen until some other right delimiter is hit, but we
  could not determine from the examples what those delimiters are.
\end{itemize}

\textbf{Layer 2 attention summary}

Judging by the attribution lines in the graph, layer 2 seems to attend
to information at the \texttt{\textless{}} sequence position from the
\texttt{v} sequence position in part because the information received at
the previous attention layer triggering a more closing-delimiter
specific query that searches for a preceding opening-delimiter key. So,
the two layers do not just operate in parallel, they also at least
partially compose in series.

Just as with layer 1, ablating the layer 2 attention output
subcomponents on the \texttt{v} position out of the target model
severely degrades performance. The model then still expects some kind of
bracket, but not an angled bracket in particular. For example, the
probability it assigns to \texttt{)} increases from \(0.079\) to
\(0.279\), the probability it assigns to \texttt{{]}} increases from
\(0.015\) to \(0.075\), and the probability it assigns to \texttt{);}
increases from \(0.004\) to \(0.052\). The probability it assigns to
\texttt{\textgreater{}} decreases from \(0.547\) to \(0.02\). This
indicates that the information carried by the value and output
subcomponents in this attention layer is important for distinguishing
which specific kind of left bracket needs to be closed with sufficient
confidence.

\textbf{Layer 2 attention query and key matrices}

\begin{itemize}
\tightlist
\item
  There are two query subcomponents on the \texttt{v} sequence position.
  The first is labeled \texttt{2.attn.q:3:270}, the second
  \texttt{2.attn.q:3:279}. They receive high positive attribution from
  both the layer 0 MLP Down projection subcomponents and the layer 1
  attention output subcomponents. Specifically, the latter subcomponent
  receives high positive attribution from the layer 1 attention output
  subcomponent \texttt{1.attn.o:3:187} and a little from
  \texttt{1.attn.o:3:899}. This suggests that this query is partially
  triggered by the received closed angled bracket information from the
  layer 1 attention, as part of a compositional pathway involving two
  attention layers in series.
\item
  There are four key matrix subcomponents on the \texttt{\textless{}}
  sequence position. Two, labeled \texttt{2.attn.k:0:197} and
  \texttt{2.attn.k:0:347} fire on various opening brackets such as
  \texttt{\textless{}}, \texttt{(} and \texttt{{[}}, as well as other
  delimiters like opening quotation marks, \texttt{\$} in latex,
  \texttt{**} and variations of these created by the tokeniser, like
  \texttt{{[}@}, \texttt{(*}, \texttt{\_\{}, \texttt{!{[}} and such. The
  third is labeled \texttt{2.attn.k:0:204}, and the final one
  \texttt{2.attn.k:0:206}.
\end{itemize}

\textbf{Layer 2 attention value matrix}

There are nine value subcomponents on the \texttt{\textless{}} sequence
position:

\begin{itemize}
\tightlist
\item
  Two, labeled \texttt{2.attn.v:0:121} and \texttt{2.attn.v:0:484} are
  part of the same ``left angled brackets'' component that also had a
  subcomponent in the layer 1 attention value matrix of the graph at
  this same sequence position.
\item
  Another two, labeled \texttt{2.attn.v:0:234} and
  \texttt{2.attn.v:0:961} are part of another component consisting of
  four subcomponents that seem to fire on left angled braces, but also
  left curly braces, opening quotation markers, and the start of links.
\item
  The other five subcomponents are labeled \texttt{2.attn.v:0:22},
  \texttt{2.attn.v:0:65}, \texttt{2.attn.v:0:473},
  \texttt{2.attn.v:0:394} and \texttt{2.attn.v:0:927}, and likewise
  variously fire on left angled brackets, left brackets in general, left
  delimiters somewhat more generally, and in one case both left and
  right delimiters. Some of them are also causally important on the
  tokens after left delimiters as well, as though they are responding to
  the delimiters information being carried forward from the previous
  sequence position.
\end{itemize}

\textbf{Layer 2 attention output matrix}

There are fourteen attention output subcomponents on the \texttt{v}
sequence position:

\begin{itemize}
\tightlist
\item
  Two, labeled \texttt{2.attn.o:3:161} and \texttt{2.attn.o:3:433} seem
  to fire whenever there are unclosed left delimiters, particularly left
  angled brackets, but left round, curly or boxy brackets.
\item
  Eight subcomponents, labeled (\texttt{2.attn.o:3:963},
  \texttt{2.attn.o:3:855}, \texttt{2.attn.o:3:359},
  \texttt{2.attn.o:3:722}, \texttt{2.attn.o:3:878},
  \texttt{2.attn.o:3:218}, \texttt{2.attn.o:3:529}, and
  \texttt{2.attn.o:3:1000}) seem to fire inside or on angled brackets or
  on other markup and xml related closing and syntax elements like
  e.g.~\texttt{","}, `\texttt{{[}@}\ldots{}\texttt{{]}}' and one appears
  to be active inside brackets in latex code.
\item
  One subcomponent, labeled \texttt{2.attn.o:3:286} is active inside
  angled brackets, but also on what appear to be chat messages, with
  particularly high magnitude activations on the line breaks in these
  messages.
\item
  Two subcomponents, labeled \texttt{2.attn.o:3:495} and
  \texttt{2.attn.o:3:121} appears to be more generally active active on
  contexts like latex, math, computer science, code and foreign language
  text.
\item
  The final subcomponent is \texttt{2.attn.o:3:735}
\end{itemize}

\textbf{Layer 3 attention summary}

There are fewer subcomponents in the layer 3 attention of the graph than
at the previous two layers. Judging by their labels, this layer attends
to the \texttt{\textless{}} and \texttt{u} sequence position from the
\texttt{v} sequence positions as part of generic previous token
behavior.

This attention layer seems less crucial to the overall computation than
layers 1 and 2. Ablating its attention output subcomponents, save for
the one labeled \texttt{3.attn.o:3:806}, only lowers the probability on
\texttt{\textgreater{}} from \(0.547\) to \(0.498\). Ablating this
output ``bias'' subcomponent does essentially destroy performance ---
likely due to the central role of this subcomponent in setting typical
activation sizes, since it has very high attributions to many downstream
nodes, rather than any sophisticated computational role. Notably, the
same is not true of the layer 2 attention output, which also has a
subcomponent labeled \texttt{2.attn.o:3:735}: ablating all output
subcomponents in layer 2 of the graph save for that one still reduces
the probability on \texttt{\textgreater{}} under adversarial sampling to
less than \(0.001\).

\textbf{Layer 3 attention matrices}

\begin{itemize}
\tightlist
\item
  There is only one query subcomponent on the \texttt{v} sequence
  position, labeled \texttt{3.attn.q:3:334}
\item
  There is one value subcomponent on the \texttt{v} sequence position,
  indicating that it is part of a self-attention mechanism in this
  layer: \texttt{3.attn.v:3:120}. In slight contradiction to its
  autointerp label it also seems to sometimes activate slightly on
  tokens outside latex math mode, though only in the sort of text that
  may typically also feature latex, or on Latex-related tokens such as
  \texttt{Eq}, \texttt{Appendix}, \texttt{proof}, and \texttt{Newton}.
\item
  There are no key subcomponents on the \texttt{\textless{}} sequence
  position, and only one value subcomponent, \texttt{3.attn.v:1:677},
  which is also kept on the \texttt{u} sequence position. It is causally
  important on more than \(25\%\) of tokens, firing mostly on
  delimiters, ``syntactic glue words'' like \texttt{and}, \texttt{the},
  \texttt{a}, \texttt{is}, \texttt{would}, \texttt{of}, \texttt{on},
  \texttt{to} and to a lesser extent text following right after
  delimiters and these connective words.
\item
  There is one key subcomponent on the \texttt{u} sequence position,
  labeled \texttt{3.attn.k:1:145}. This indicates that the relevant
  information is moved from \texttt{u} to \texttt{v} as part of generic
  previous token behavior.
\item
  There are three value subcomponents on the \texttt{u} sequence
  position: The first, \texttt{3.attn.v:1:677}, is also kept on the
  \texttt{\textless{}} sequence position. The two others are labeled
  \texttt{3.attn.v:1:76} and \texttt{3.attn.v:1:95}.
\item
  There are three attention output subcomponents on the \texttt{v}
  sequence position, labeled \texttt{3.attn.o:3:283},
  \texttt{3.attn.o:3:398}, and \texttt{3.attn.o:3:806}.
\end{itemize}

Notably, in the attention layer 2 of this graph, information about the
open bracket seems to be moved from the \texttt{\textless{}} position to
the \texttt{v} position, partly due to information previously received
from the \texttt{\textless{}} position in attention layer 1. This
triggers a query that is specific to closing-delimiters (such as
\texttt{\textgreater{}}), which searches for a preceding
opening-delimiter (such as \texttt{\textless{}}) key.

\begin{figure}[htbp]
\centering
\includegraphics[width=\textwidth,height=0.8\textheight,keepaspectratio]{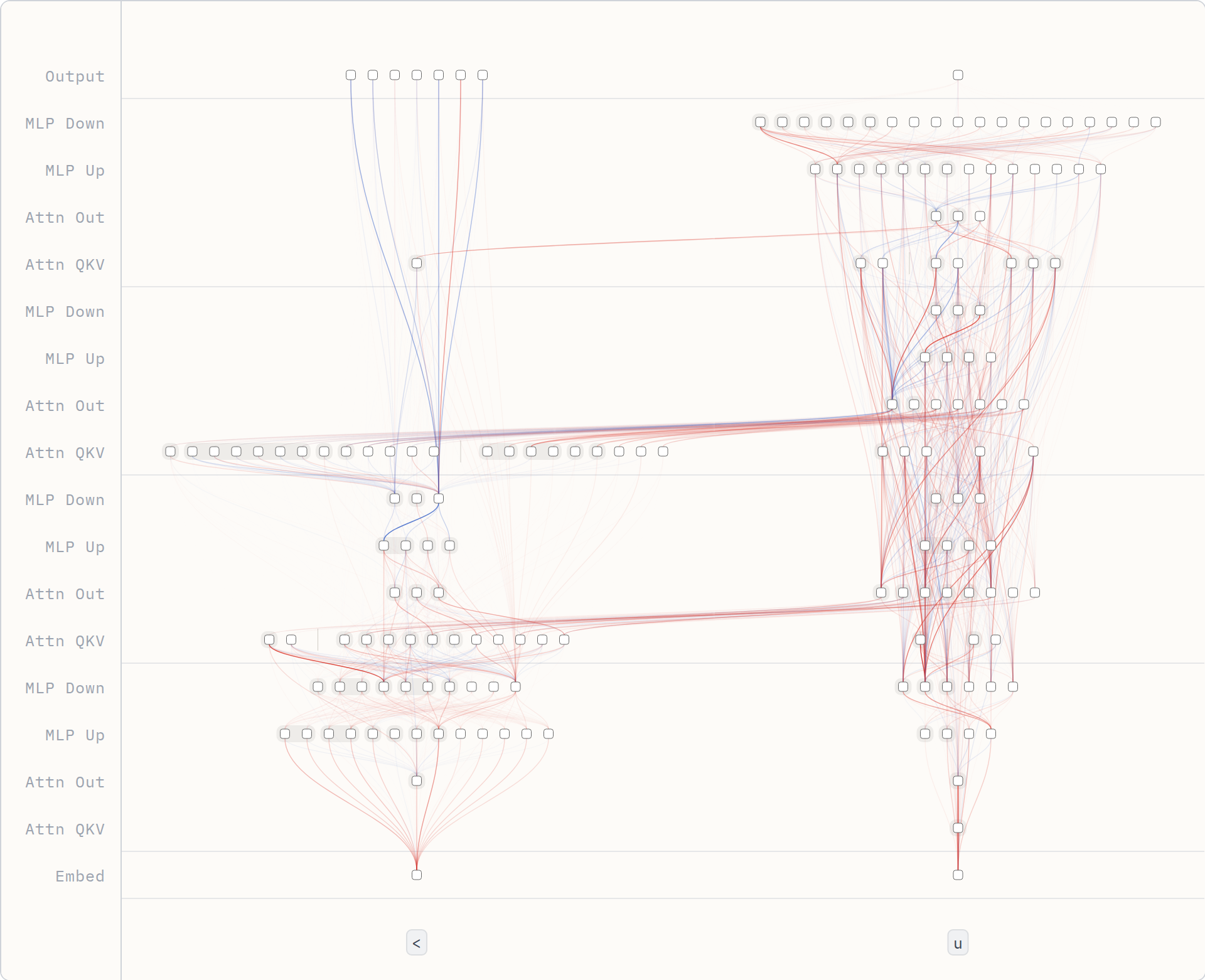}
\caption[]{Attribution graph for predicting \texttt{>} on the prompt \texttt{<} \texttt{u} \texttt{,} \texttt{v} \texttt{>} after \texttt{u}, pruned with adversarial sampling.\protect\footnotemark{} There are 162 subcomponents in the graph. The target model assigns probability \(0.119\) to \texttt{>}. \emph{Static snapshot of an interactive figure; explore it at \href{https://www.goodfire.com/research/interpreting-lm-parameters\#graph:bracket_u}{the web version}.}}\label{graph:bracket_u}
\end{figure}

Since the \(W_Q\) subcomponents used in attention layer 1 and 3 appear
to be generically always-active rather than triggering in response to
preceding commas, and the queries in layer 2 do not appear to only
trigger conditional on a comma at the previous token either, one might
wonder whether the model would also predict a closing
\texttt{\textgreater{}} right after \texttt{u}. It turns out that it
does --- predicting \texttt{\textgreater{}} as its top logit after
\texttt{u} as well, though with lower confidence (\(0.119\)
vs.~\(0.547\) after \texttt{v}). \autoref{graph:bracket_u} shows a
structurally similar graph for this prediction, but lacking the
subcomponents active on \texttt{u} and \texttt{,} in the attention
layers. This suggests the longer context reinforces the math context and
thus the likelihood of a closing bracket. Interestingly, the model does
not predict a closing bracket after \texttt{,}, suggesting it recognises
that the comma indicates the statement inside the bracket is not yet
complete.

\footnotetext{Coefficient \(0.5\) for cross-entropy reconstruction with stochastic sampling, coefficient \(0.5\) for cross entropy with \(4\) steps of PGD, lr \(1\), importance minimality coefficient \(0.1\), \(p=0.3\), \(4000\) optimization steps.}

\begin{figure}[htbp]
\centering
\includegraphics[width=\textwidth,height=0.8\textheight,keepaspectratio]{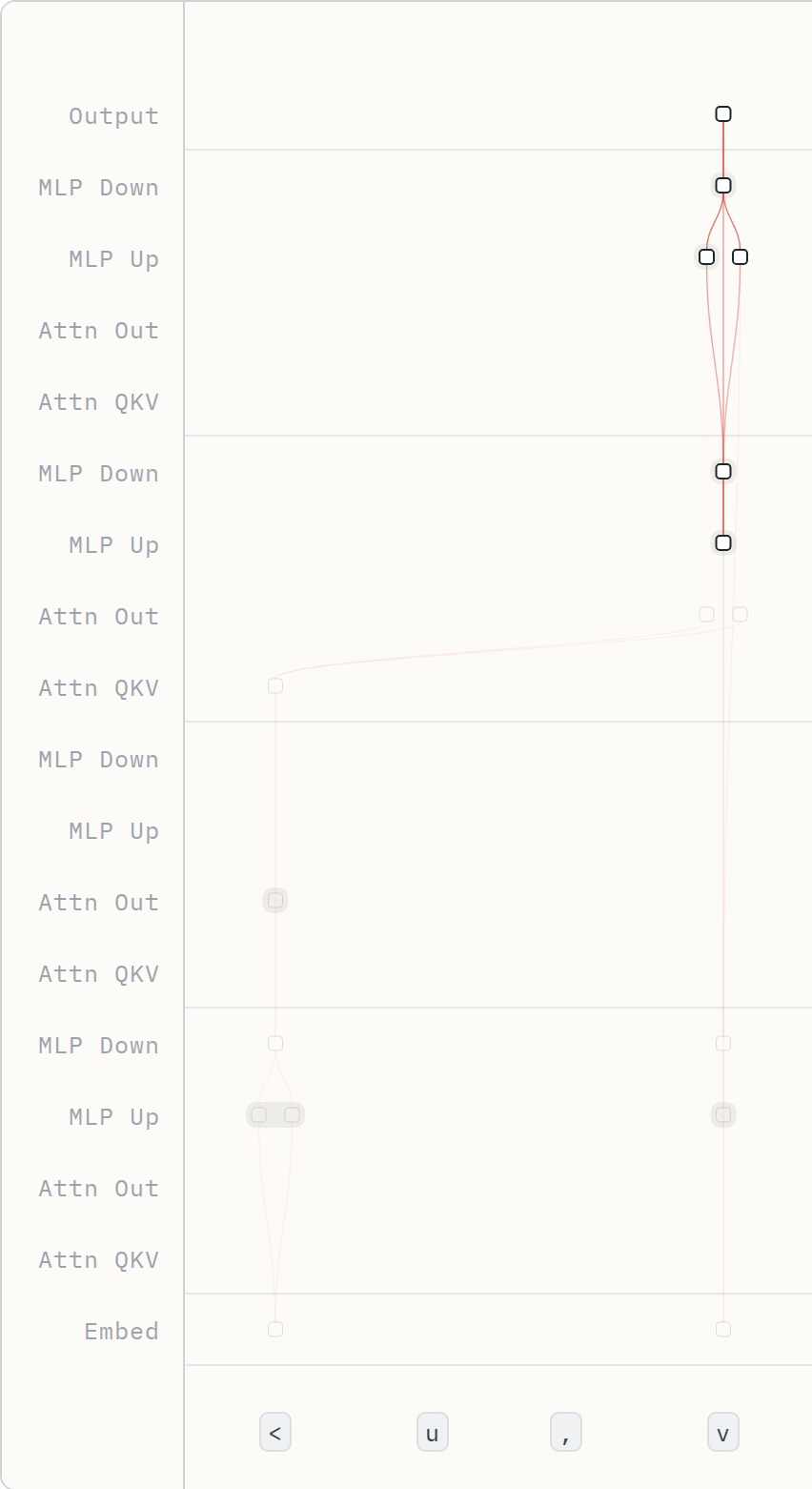}
\caption[]{Attribution graph for predicting \texttt{>} on the prompt \texttt{<} \texttt{u} \texttt{,} \texttt{v} \texttt{>}, pruned with causal importance masking. \emph{Static snapshot of an interactive figure; explore it at \href{https://www.goodfire.com/research/interpreting-lm-parameters\#graph:bracket_ci}{the web version}.}}\label{graph:bracket_ci}
\end{figure}

Pruning with CI masking instead of adversarial masking recovers a much
smaller graph of just 14 subcomponents (\autoref{graph:bracket_ci}). It
predicts \texttt{\textgreater{}} correctly under CI masking but fails
completely under adversarial masking, giving a very incomplete account
of the computation. Nevertheless, it highlights some core pathways.

Attribution graph for predicting \texttt{\textgreater{}} after
\texttt{v} on the prompt \texttt{\textless{}} \texttt{u} \texttt{,}
\texttt{v} \texttt{\textgreater{}}, pruned with causal importance
masking.\footnote{Coefficient \(1.0\) for cross-entropy reconstruction
  with causal importance masking, importance minimality coefficient
  \(0.1\), \(p=0.3\), \(2000\) optimization steps.} There are 14
subcomponents in the graph.

The two largest direct positive attributions to the output
\texttt{\textgreater{}} come from:

\begin{enumerate}
\def\labelenumi{\arabic{enumi}.}
\item
  A layer 3 MLP Down projection subcomponent labeled
  \texttt{3.mlp.down:3:1414}. It receives attribution from one MLP Up
  projection subcomponents labeled \texttt{3.mlp.up:3:2565} and one
  subcomponent labeled \texttt{3.mlp.up:3:1051}, with the latter firing
  on a more general set of closing delimiters.
\item
  \texttt{2.mlp.down:3:1560}, a layer 2 MLP Down projection
  subcomponent. It connects strongly to the two layer 3 MLP Up
  subcomponents mentioned above in addition to the output, suggesting
  that the layer 2 and 3 MLP pathways are partially interlinked in
  series rather than parallel and independent. It receives attribution
  from an MLP Up projection subcomponent labeled
  \texttt{2.mlp.up:3:2151}.
\end{enumerate}

Ablating these two MLP Down projection subcomponents out of the target
model severely degrades the \texttt{\textgreater{}} prediction, lowering
the probability from \(0.547\) to \(0.158\) and \(0.243\) for individual
ablations, and to \(0.046\) for joint ablation. The model instead
reassigns probability mass to other delimiters such as \texttt{)},
\texttt{\_}, \texttt{,} or \texttt{)\$}, suggesting they are important
for singling out a right angled bracket in particular.

These MLP subcomponents receive information about the open angled
bracket from layer 2 attention output subcomponents labeled
\texttt{2.attn.o:3:855} and \texttt{2.attn.o:3:878}, which in turn
receive from a subcomponent of the layer 2 attention value matrix at the
\texttt{\textless{}} sequence position labeled \texttt{2.attn.v:0:473} .
This subcomponent receives information directly from the
\texttt{\textless{}} embedding, as well as through three layer 0 MLP
subcomponents, labeled \texttt{0.mlp.down:0:3069},
\texttt{0.mlp.up:0:2149}, and \texttt{0.mlp.up:0:2643}.

While the 14-component graph highlights the core pathways, the full
graph in \autoref{graph:bracket} makes clear that the actual computation
is far more intricate.

Given how few subcomponents our decomposition has in total (ca. 10,000
alive in the whole model) it is perhaps remarkable how many of them
appear to be dedicated to moving around and processing information for
predicting closing delimiters of various kinds. This may be partially
due to delimiter closing being one of perhaps relatively few prediction
tasks that is simple enough for a model of this size to perform well.

\section{Editing a language model's parameters by hand to modify its
neural
algorithm}\label{editing-a-language-models-parameters-by-hand-to-modify-its-neural-algorithm}

\label{sec:model-editing}

One of the subtasks in language modelling is predicting tokens that
comprise emoticons, such as
\texttt{:\textasciigrave{}\textasciigrave{})},
\texttt{:\textasciigrave{}\textasciigrave{}-\textasciigrave{}\textasciigrave{}(},
\texttt{X\textasciigrave{}\textasciigrave{}D},
\texttt{;\textasciigrave{}\textasciigrave{})}, or
\texttt{=\textasciigrave{}\textasciigrave{})}.

We used the decomposition to perform a simple edit to the model's
learned algorithm: Manually modifying a single rank-1 subcomponent to
make the model predict that all emoticons are surprised-face emoticons.

The challenge here is to make models predict the token \texttt{o}, as in
a surprised-face emoticon \texttt{:} \texttt{o}, with high probability
without substantially altering the model's behavior in other,
non-emoticon contexts. Because
\texttt{:}/\texttt{;}/\texttt{X}/\texttt{=} (etc.) tokens can be used in
many non-emoticon contexts, this rewrite can't be achieved with a
token-level remapping; we have to rewrite the algorithm that the model
applies to its hidden activations.

We find that multiple subcomponents in the MLP Down matrix of layer 2
specifically activate on the first characters in emoticons with low or
zero activations elsewhere, including on these same tokens in other
contexts:

\begin{itemize}
\tightlist
\item
  \texttt{2.mlp.down:1672}
\item
  \texttt{2.mlp.down:2359}
\item
  \texttt{2.mlp.down:2623}
\item
  \texttt{2.mlp.down:3290}
\item
  \texttt{2.mlp.down:3327}
\item
  \texttt{2.mlp.down:3382}
\end{itemize}

We picked one of these subcomponents, \texttt{2.mlp.down:2359}, as our
target for editing. Our edit leverages the idea that each subcomponent,
being a rank-1 matrix \(\vec{U}^l_c (\vec{V}^l_c)^\top\) has one `read'
direction and one `write' direction, which are its right and left
singular vectors respectively. We changed the `write' direction of the
subcomponent so that, when it activates, it writes very strongly to the
same direction as the \texttt{o} token in the model's unembedding
matrix.

We performed this edit by replacing the subcomponent's write vector
\(\vec{U}^l_c\) with \(-\alpha\vec{u}_o / \lVert \vec{u}_o \rVert\),
where \(\vec{u}_o\) is the unembedding direction for the token
o\footnote{The new write vector points opposite to \(\vec{u}_o\) because
  the read activation \((\vec{V}^l_c)^\top \vec{x}\) is itself negative
  at firing positions for \texttt{2.mlp.down:2359}; the two negatives
  cancel, so the residual stream gains a positive multiple of
  \(\vec u_o\) and the logit for o rises.}. The new weight matrix thus
becomes

\[{W^l}' = W^l - \left(\vec{U}^l_c + \alpha\frac{\vec{u}_o}{\lVert \vec{u}_o \rVert}\right)\vec{V^l_c}^\top.\]

To measure the amount of undesired off-target effects caused by the
edit, we use two metrics, which characterize off-target effects in
slightly different ways, one measuring effects on tokens that are
potentially computationally `nearby' to our edit, and the other
measuring all changes:

\begin{itemize}
\tightlist
\item
  \(D_{\text{KL},\text{Surrounding}}\): The KL-divergence between the
  target model and the edited model on the \(20\) tokens before and
  after a token on which \texttt{2.mlp.down:2359} is causally important;
\item
  \(D_{\text{KL},\text{Global}}\): The KL-divergence between the target
  model and the edited model on all tokens on which
  \texttt{2.mlp.down:2359} is not causally important, sampled from the
  whole dataset.
\end{itemize}

As baselines for comparison, we trained two conventional LoRA adapters
for the MLP Down projection matrix in layer 2. The LoRAs were trained to
convergence on \(n\) dataset examples (\(n=10\) or \(947\)). The
training dataset examples consisted of the token on which the
subcomponent \texttt{2.mlp.down:2359} is causally important and the 20
tokens before and after. They were trained both (a) to predict an
\texttt{o} after the emoticon's initial token (e.g.~\texttt{:}) and (b)
to minimize the off target effects. Concretely, for (a), each LoRA was
trained with a cross-entropy loss to predict the \texttt{o} label after
the token on which the subcomponent is causally important. For (b),
off-target effects were minimized using a KL divergence term (weighted
by the off-target effect penalty coefficient, \(\lambda\)) between the
output logits of the target model and the logits of the edited model on
the rest of the tokens in the example\footnote{Here, the LoRA training
  loss is equivalent to both \(D_{\text{KL},\text{Surrounding}}\) and
  \(D_{\text{KL},\text{Global}}\) (due to how the datapoints are
  selected).}.

\begin{figure}[htbp]
\centering
\includegraphics[width=\textwidth,height=0.8\textheight,keepaspectratio]{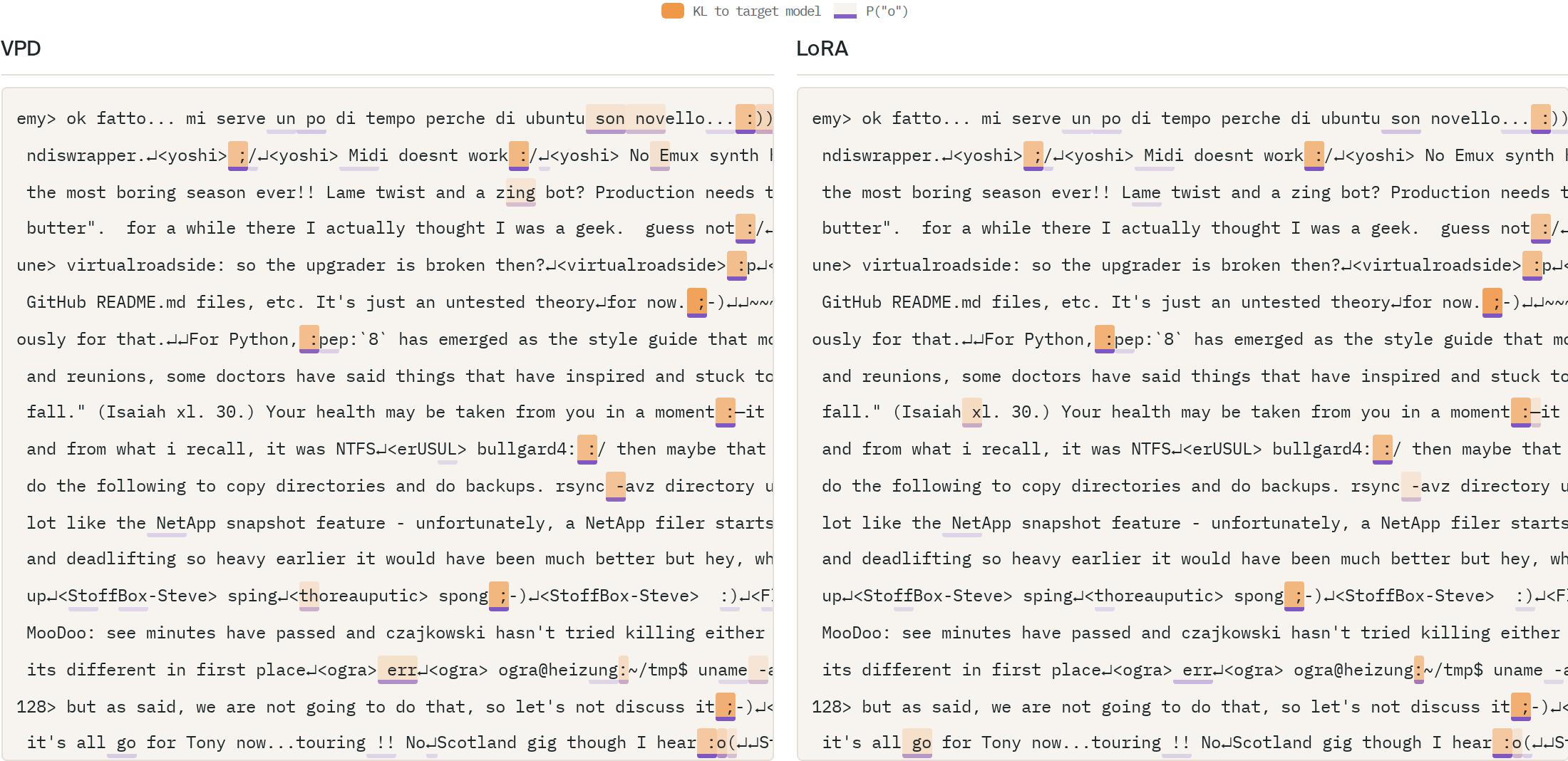}
\caption[]{Comparing VPD-based edits with a LoRA-augmented target model trained to complete emoticons with \texttt{o}. The orange background shows the per-token KL divergence between the edited model and the target model. The purple underline shows the probability assigned to \texttt{o}. The left panel shows the direct VPD-based manual edit, obtained by adding the unembedding vector for \texttt{o} to the emoticon subcomponent's \(\vec{U}\) vector (scaled by prefactor \(\alpha=3\)). The right panel shows a LoRA-augmented target model trained on \(n=947\) examples, each consisting of a token the emoticon subcomponent was causally important on and the \(20\) tokens either side of it. \emph{Static snapshot of an interactive figure; explore it at \href{https://www.goodfire.com/research/interpreting-lm-parameters\#fig:model-editing-heatmap}{the web version}.}}\label{fig:model-editing-heatmap}
\end{figure}

To give a sense of the effectiveness of the VPD edits and the LoRA
baseline, \autoref{fig:model-editing-heatmap} shows the per-token KL
divergences to the target model for both the VPD edit and the LoRA
baseline.

\begin{figure}[htbp]
\centering
\includegraphics[width=\textwidth,height=0.8\textheight,keepaspectratio]{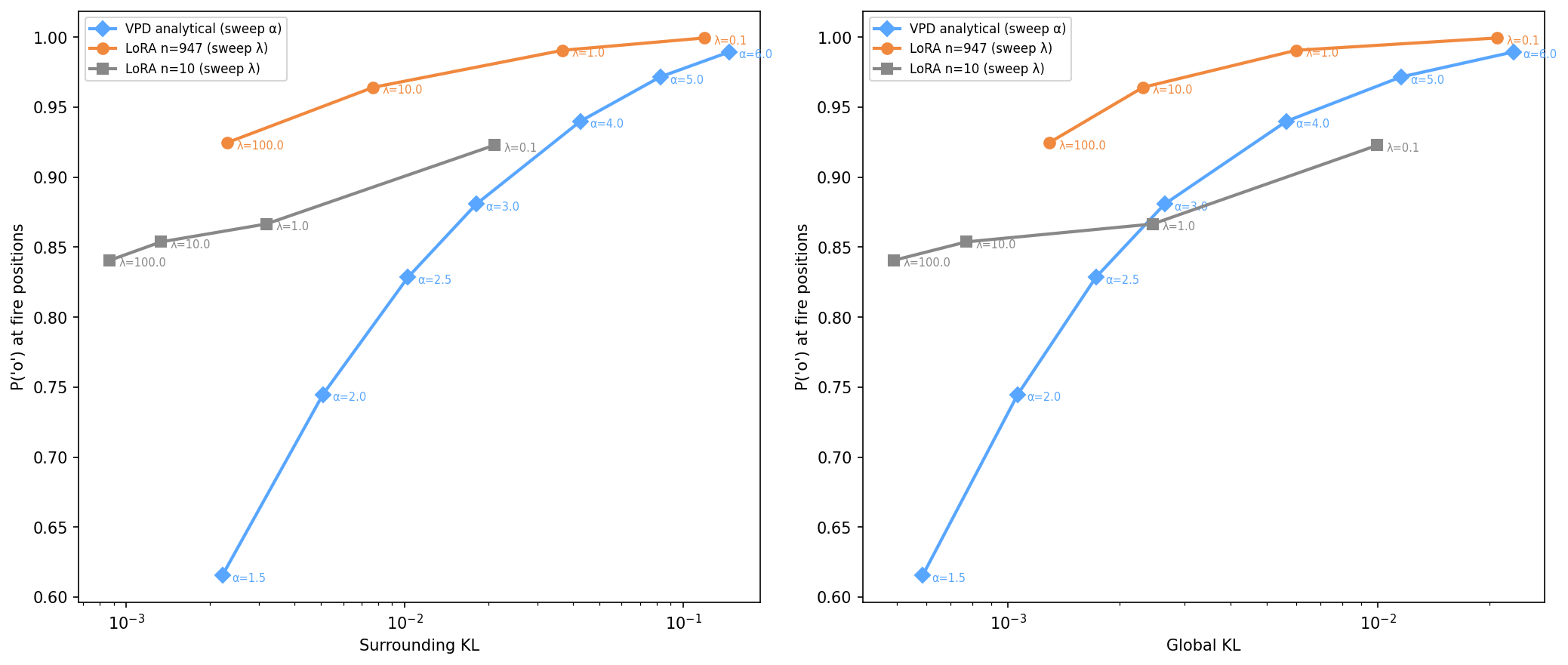}
\caption[]{Model editing for emoticon completions,  LoRA vs. manual subcomponent edit. Manual edits were performed by adding the unembedding vector for \texttt{o} to the \(\vec{U}\) vector of the emoticon subcomponent with different prefactors \(\alpha\). LoRAs were trained on \(n=10\) and \(n=947\) examples, each consisting of a token the emoticon subcomponent was causally important on, and the \(20\) tokens immediately preceding and following it, with a KL-regularisation term weighted by \(\lambda\). The y-axis shows the average probability the edited model assigns to \texttt{o} on tokens the emoticon subcomponent is active on. The x-axis in the left plot shows  \(D_{\text{KL},\text{Surrounding}}\), the KL divergence between the edited model and the target model on the \(20\) other tokens immediately preceding and following tokens the emoticon subcomponent is causally important on, across a holdout set of \(50\) examples. The x-axis in the right plot shows \(D_{\text{KL},\text{Global}}\), the average KL-divergence between the edited model and the target model on all other tokens across samples from the whole dataset.}\label{fig:model-editing-pareto}
\end{figure}

In \autoref{fig:model-editing-pareto} we vary both \(n\) (the number of
training dataset examples) and \(\lambda\) (the off-target effect
penalty coefficient) for our LoRA baselines. We plot the trade-off
between the probability of predicting an \texttt{o} versus off target
effects. We compare the LoRAs with our manual edit with different scale
factors \(\alpha\) for the \texttt{o} unembedding vector added to the
subcomponents' \(\vec{U}\) vector of \texttt{2.mlp.down:2359}.

LoRAs trained on just \(n=10\) examples outperform the manual edit on
\(D_{\text{KL},\text{Surrounding}}\), the setting they were trained on,
but not on \(D_{\text{KL},\text{Global}}\). LoRAs trained with \(n=947\)
examples outperform the manual edit on both
\(D_{\text{KL},\text{Surrounding}}\) and
\(D_{\text{KL},\text{Global}}\).

While this is a promising result, we stress that this is a very
preliminary investigation. The method we used to edit the subcomponent,
adding the appropriate unembedding vector, was simply the first
interpretable editing technique we tried. Other editing techniques might
work better. For example, although this edit clearly affects the output
in the intended way, there is another layer in between our edited layer
and the output, which may lead to some of our edits' off target effects.
We may be able to do better by choosing a direction that maximally
avoids affecting the computations of the intermediate layer while still
projecting strongly onto the \texttt{o} token in the unembedding matrix.
This may help to close the gap between the performance of our edit and
the performance of the LoRA.

On the other hand, the example is cherry picked. We deliberately chose
this task because the model seemed to have a small number of
subcomponents related exclusively to emoticon prediction. We
nevertheless conclude that VPD shows some promise for model editing in
cases where correctly labeled data for training a LoRA is difficult to
obtain, or where it is desirable for the edit to be somewhat
interpretable. We think that there are very likely ways to leverage
parameter decomposition to do much better editing than we have here in
this proof of concept.

\section{Discussion}\label{discussion}

\label{sec:discussion}

At this point, it is worth reflecting on what our parameter
decomposition approach has actually bought us with regard to the
highest-level goals of our field:

In mechanistic interpretability, we aim to reverse engineer the
computational machinery of neural networks. In particular, we want to
know how that machinery takes inputs, computes hidden representations,
performs computations on those hidden representations, and finally
computes its output behavior. Concretely, this means that the objects we
want to understand are the computational graphs of neural networks and
how they interact with data. To make this as manageable as possible,
we'd like to understand small parts of these computational graphs of
short description length in isolation, yet have our explanations
aggregate together so that, eventually, we can come to understand the
entire network as a whole.

In the following sections, we discuss how VPD makes progress toward
these goals, or how it does not.

\subsection{Parameter decomposition makes fewer assumptions about neural
networks' representations than other
methods}\label{parameter-decomposition-makes-fewer-assumptions-about-neural-networks-representations-than-other-methods}

\label{sec:discussion-computation}

Parameter decomposition methods are less opinionated than other popular
decomposition methods about the `form' of the computation that we expect
to find in neural networks. Sparse dictionary learning methods, such as
transcoders or cross-layer transcoders, effectively train replacement
models to substitute for parts of the original model, using different
architectures with significantly wider layers that are often also
composed of different non-linear units. These replacement architectures
can have greater representational capacity than the original model, and
consequently correspond to entirely different function classes. As a
result, they can implement computations that would not be representable
using the original model architecture. To ensure these replacement
models are mechanistically faithful to the computations in the model, we
therefore need to have a somewhat clear idea of the form of those
computations \emph{ex ante}, and use that knowledge to choose the right
replacement model architecture and hyperparameters. Otherwise, the
replacement models might learn a different internal structure entirely.
Feature splitting is one example of this issue.

In contrast, parameter decomposition methods always stay within the
confines of the original model architecture. Parameter components are
effectively just a different set of coordinates for representing the
target model's parameters. Every combination of parameter components
corresponds to a valid weight vector we can implement in the original
model architecture. In a sense, we're letting the network itself tell us
what the form of its computation is instead of guessing a particular
form in advance.

\subsection{Explanations of attribution graphs are not explanations of
computational
graphs}\label{explanations-of-attribution-graphs-are-not-explanations-of-computational-graphs}

A full explanation of a network's behavior should amount to an
end-to-end algorithm that is essentially equivalent to the algorithm
implemented by the target network. In other words, it should be possible
to represent the explanation as a \emph{computational graph} that is
mechanistically faithful to the computational graph of the original
network, which is typically expressed in terms of its neurons, weights
matrices, nonlinearities, etc.

By contrast, in this paper, we used VPD to produce \emph{attribution
graphs} rather than computational graphs. It is not possible to compute
the model output on a datapoint using only the attribution graph without
access to the original model itself. An attribution graph can track how
strongly any given upstream node in a computational graph influenced any
given downstream node, which is useful for understanding the flow of
information in the graph, but it does not represent the functional
relationship between upstream and downstream nodes. This means we have
not yet explained the network's computational graph; the explanations of
the model's behavior we provide here are incomplete. Additionally,
attribution methods such as the gradient attributions we used in this
paper also have some well-known issues that can lead them to misjudge
the magnitude of the influence one node in a graph has on another
\citep{neel2022attribution, kramar2024atpefficientscalablemethod}. For
example, if an attention head in a model has a saturated softmax,
gradient attributions through it will tend to systematically
underestimate the effect of ablating the upstream node on the downstream
node. Despite these limitations, we think attribution graphs are still
useful as a basic picture of how information flows between VPD
subcomponents on a forward pass, and have been used to similar effect
for other decomposition methods, such as CLTs
\citep{ameisen2025circuit}.

In future work we aim to deepen our study of full computational graphs
by studying in detail the interactions of VPD subcomponents at
nonlinearities, such as MLP neuron activation functions. For some
preliminary investigations into characterising nonlinear interactions
between subcomponents at MLP neurons, see
\autoref{app:interactions-gis-vs-coact}. This preliminary analysis
suggests that parameter subcomponents may tend toward simpler nonlinear
interactions than the worst case scenario
(\autoref{app:interactions-gis-vs-coact}), which is encouraging for the
feasibility of this direction, but it is still far from definitive
evidence. Speculatively, we suspect that this property arises because
it's hard for lots of subcomponents to be ablatable if lots of
subcomponents are interacting.

\subsection{Robustness to adversarial ablations permits aggregation of
explanations}\label{robustness-to-adversarial-ablations-permits-aggregation-of-explanations}

\label{sec:vpd_recon_motivation}

One of the central promises of ablation-based parameter decomposition is
that explanations of a model's behavior on individual datapoints, given
in terms of causally important parameter components and their
interactions, can be aggregated into more global explanations of its
behavior across the full distribution. To illustrate why this admittedly
strict requirement is necessary, consider the following spurious
decomposition of a hypothetical weight matrix in a model:

For every data point \(x\), we make up a unique low-rank component
\(\theta_x\), and assign it causal importance \(1\) on \(x\) and \(0\)
for every other input. We pick the parameters of \(\theta_x\) such that
the resulting model exactly matches the final output of the original
model: \(f(x\vert \theta_x)=f(x\vert \theta)\).\footnote{To ensure our
  auxiliary loss \(\mathcal{L}_{\text{Delta-L2}}\) is also \(0\), we
  just make up one more component
  \(\theta_{X+1}:=\theta-\sum^X_{x=1} \theta_x\) so that the sum
  \(\sum^{X+1}_{x=1} \theta_x\) equals the target model parameter
  vector, and assign it causal importance \(0\) on every data point.}
This decomposition would perfectly reconstruct the original model output
on every training datapoint, but the resulting components would be
spurious and completely unrelated to the mechanistic structure of the
target network's learned algorithm. We did not even need to refer to the
target model's internals to construct them! They amount to a giant
lookup table of the training dataset, and won't generalise to new data
points or tell us anything about how the original model actually
computed its outputs.

Requiring that the causally unimportant parameter components can be
ablatable in any combination rather than just all together excludes
counterexamples like this, because it ensures that components do not
interfere with the computation on data points they are \emph{not}
causally important on. This prevents the decomposition from ``splitting
up'' general computational machinery in the target model into large sets
of specialized components that each just memorise a particular
input-output pair.

More generally, this stricter requirement ensures that \emph{local
descriptions} of the model's behavior on single data points (or small
subsets of the dataset) in terms of their causally important parameter
components will correctly aggregate into more \emph{global descriptions}
of the network's behavior over larger subsets of the dataset in the way
we expect: If we explain the network's behavior on two data points
\(x_1\) and \(x_2\) using two different parameter vectors
\(\sum_{i\in S_1} \theta_i, \sum_{i\in S_2} \theta_i\), formed from two
subsets of the parameter components \(S_1, S_2\), a parameter vector
formed by the union of both subsets
\(\sum_{i\in S_1 \cup S_2} \theta_i\) will still compute approximately
the same output on both datapoints:

\[f(x_1\vert \sum_{i\in S_1} \theta_i) \approx f(x_1\vert \sum_{i\in S_1 \cup S_2} \theta_i) \quad\text{ AND }\quad f(x_2\vert \sum_{i\in S_2} \theta_i) \approx f(x_2\vert \sum_{i\in S_1 \cup S_2} \theta_i).\]

In principle, one could start from individual datapoint explanations and
incrementally combine them --- first into explanations of the model's
behavior on narrow sub-distributions (such as bracket closing or pronoun
prediction), then into broader and broader accounts, eventually
approaching a complete reverse engineering of the model. It remains
unclear whether our current decomposition is sufficiently adversarially
robust for this purpose. Our primary uncertainty is that it is unclear
how much adversarial robustness is necessary for `local' explanations to
aggregate into `global' ones.

\textbf{How much robustness is necessary?}

If we do not have enough adversarial robustness, then we lose the
ability to aggregate explanations of parts of the model into a coherent
whole. However, if we are too strict in our demands for robustness to
adversarial ablations, it is sometimes possible to exclude
decompositions we would intuitively regard as valid, because the
adversary can systematically exploit random interference noise in
`unused' circuitry to change the network output. In
\autoref{sec:vpd_methods-adv}, we point out a theoretical toy case in
which strictly demanding full adversarial robustness causes this
problem. This would seem to put us in a difficult spot! How much
robustness do we need to demand for our explanation to be
mechanistically faithful? How much robustness is actually too much, and
would exclude short descriptions of network behavior we would like to
regard as valid?

We do not currently have a fully satisfying answer to this question, but
we suggest that a reasonable approach may be to ground the answer in
practical considerations: What combinations of (sub)component ablations
might we realistically want to perform when using VPD to understand or
edit a given model? And over which subsets of the data would we want to
investigate the behavior of the resulting ablated models? So long as the
decomposition is robust enough that it is unlikely for any of the
ablations we end up performing in practice to be in the non-robust set
for any model input we care about, the lack of complete robustness may
not be relevant to us. Even if we do end up encountering a component
ablation the decomposition is not robust to, the problems caused by this
may be limited if they only apply to a few data points and the edited
model is still behaving as we would expect for the vast majority of
inputs\footnote{Ultimately, we think this practical mindset also sheds
  some light on how we can think about the theoretical toy case we
  mentioned above, where the adversarial sampler exploits unstructured
  noise in `inactive' circuits to change the output: Those `inactive'
  circuits really are \emph{somewhat} involved in computing the model's
  outputs. It's just that their involvement is quite limited, and only
  becomes relevant in exponentially rare edge cases that involve
  ablating very particular sets of components on very particular data
  points. So, a shorter description of the model's behavior in terms of
  the much smaller number of `active' components really isn't completely
  mechanistically faithful, but it is \emph{mostly} mechanistically
  faithful. Dropping the inactive circuits from the description retains
  its predictive power for almost all cases we might care about while
  drastically decreasing its length, and that trade-off is usually worth
  it to us.}.

\subsection{Interpretability demands Minimum Description Length, but of
what kind of
description?}\label{interpretability-demands-minimum-description-length-but-of-what-kind-of-description}

Most interpretability methods aim to explain neural networks using short
descriptions. Sparse dictionary learning methods aim to find minimal
length descriptions of the activations of a network on each forward pass
using sparse coding schemes, where the descriptions are the indices of
active elements in a dictionary and their activations. VPD aims to find
minimal length descriptions of the parameters of a network that are
causally important on each forward pass, where the descriptions are the
indices of causally important parameter subcomponents and the parameter
subcomponents themselves.

However, as discussed above, we ultimately want to understand neural
networks' computations: The nonlinear transformations they apply to
inputs and hidden activations to produce their outputs. To do this, we
want to obtain short descriptions of the computational graphs of the
networks' forward passes. Minimizing the description length of the
parameters used on a forward pass is not the same thing as minimizing
the description length of the forward pass itself. It is merely a proxy
measure. The rank constraint and frequency minimality loss
(\(\mathcal{L}_{\text{frequency-minimality}}\)) we use are just one
possible set of proxies for encouraging parameter subcomponents to be
computationally simple objects, and we have no reason to believe that
they are the optimal choice. While the frequency minimality loss at
least addresses one way the rank of weight matrices can fail to measure
computational complexity, there may be other failure modes we have not
yet identified. More broadly, we lack a principled, general-purpose
measure of the computational simplicity of a parameter subcomponent, and
developing one remains an open problem.

\subsection{Other limitations and future
work}\label{other-limitations-and-future-work}

\label{sec:limitations}

Beyond the limitations outlined above, we identify a range of other
shortcomings and avenues for future work:

\textbf{Scaling to larger models and non-language models.} VPD has
permitted the scaling of parameter decomposition methods to the
67M-parameter model we decomposed here. But this is still a small model.
We would like to continue scaling the method. In addition to engineering
improvements for greater compute efficiency, we suggest that
improvements to the adversarial samplers and causal importance
functions, which we discuss more below, may be helpful for this goal.
Beyond scale, we are also interested in applying VPD to other domains,
such as vision models, multi-modal models, and biological foundation
models.

\textbf{Studying the biology of language model parameters.} Even without
scaling, there is important further work to be done at models at the
current scale. Our work explored only a small number of the model's
behaviors. Even the behaviors we explored could be studied in more
depth. We would like to see more work that uses parameter decomposition
as a tool for language model `biology' \citep{lindsey2025biology}.

\textbf{Our decomposition is not as adversarially robust as we would
like.} As shown in \autoref{tab:vpd-pgd-ce}, while the decomposition is
at least somewhat robust to \(\approx 20\) steps of adversarial
optimization (KL divergence \(0.83\)), robustness degrades rapidly with
more optimization steps, reaching a KL divergence of \(40.2\) at \(320\)
steps. This means that there exist sets of subcomponent ablations
involving only causally unimportant subcomponents that drastically alter
the model's output. As discussed in \autoref{sec:vpd_recon_motivation},
we do not necessarily expect or even desire complete robustness to
arbitrarily many steps of adversarial optimization. However, we suspect
that substantially higher levels of adversarial robustness are still
achievable. In the future, we would like to see work that improves
adversarial sampling and mask parameterizations. Our current adversarial
sampler uses a relatively primitive form of projected gradient descent
(PGD) to find worst-case ablation masks. We think it should be possible
to improve the performance of this sampler. For example, we might be
able to identify particularly important subspaces of masking space for
the sampler to focus on, such as the subspace spanned by the sums of
causally important subcomponents on other data points in the same batch.

\textbf{Better causal importance functions.} The causal importance
function \(\Gamma\) is currently implemented as a vanilla transformer
that takes as input the target model's hidden activations concatenated
across layers into a single vector. This is a relatively simple
architecture for a task that requires predicting the ablatability of
every subcomponent at every sequence position, and we suspect that more
sophisticated architectures might produce more accurate causal
importance predictions.

Another potential improvement may be to use continuous cut-off scales
instead of binary causal importances. Currently, our causal importance
functions classifies subcomponents in a largely binary manner: Either
they are causally important for computing the network's output, or they
are not. However, in reality, subcomponents lie on a more continuous
scale of affecting the output to a larger or smaller degree. The more we
care about low description length relative to output reconstruction, the
more subcomponents we will want to drop from our description of the
forward pass, starting with those that affect the final output the
least. To account for this, we might train a function that predicts
\emph{cut-off scales} on the Pareto frontier between output
reconstruction and description length instead of fixed causal
importances. This way, a single decomposition could provide a variable
resolution scale for describing the forward pass, ranging from short and
simplified descriptions of the network's computation involving just the
most important subcomponents, to longer but more accurate descriptions
involving more subcomponents, all the way up to descriptions which
recover the target model's performance completely.\footnote{The causal
  importance functions already enable this to an extent through
  fractional causal importance values, see \autoref{fig:pareto-mse}, but
  they are not really trained with this application in mind.}

\textbf{Our clustering method is blind to multi-sequence position
circuits} VPD decomposes weight matrices into rank-one subcomponents,
which must then be clustered into full parameter components that span
multiple weight matrices (\autoref{app:clustering}). Our clustering
algorithm is based on minimizing description length, but it currently
only uses correlations between causal importances on the same sequence
position. This ignores possible compression based on cross-sequence
position correlations. For example, \(Q\) and \(K\) components in an
induction head might never operate on a computation at the same sequence
position.

\textbf{Our clustering method has not been carefully tuned.} Our
MDL-based clustering algorithm has a key hyperparameter \(\alpha\) that
controls the trade-off between the number of components and their
complexity. We did not sweep this hyperparameter particularly carefully.
This was not a priority because individual subcomponents already proved
to be fairly interpretable on their own, but it means the parameter
components we report may not reflect the best possible grouping.

\textbf{Clustering post-hoc can make the adversarial sampler stricter
than is necessary} Demanding that a model's outputs do not vary under
adversarial ablation of two causally unimportant rank one subcomponents
is a somewhat stricter condition than demanding that its outputs do not
vary under adversarial ablation of those two rank one subcomponents
clustered into a single rank two component. The former condition samples
a two-dimensional subspace of ablation masks, the latter only a
one-dimensional subspace. Thus, clustering subcomponents into components
post-hoc effectively makes our reconstruction losses stricter than they
technically need to be. It is currently unclear whether this causes
substantial issues in practice. If it does, future parameter
decomposition methods may be able to compress parameter description
length further than VPD does by integrating the clustering phase into
subcomponent training, allowing us to use clustering information to
inform the ablation mask sampling.

\textbf{Automated reverse engineering of parameter components would
enable broader interpretations.} In our case studies
(\autoref{sec:case-studies-pronoun},
\autoref{sec:case-studies-bracket}), we manually traced information flow
through small parts of the attribution graphs for a few specific prompts
and behaviors. Building a full picture of how a model computes its
outputs will require scaling up this kind of analysis considerably, to
more prompts and on more paths through their graphs. We aim to do this
using automated interpretability methods.

\textbf{Data-subset decompositions may reduce the cost and scope of
decompositions} Rather than decomposing the model with respect to the
full training distribution, one could apply VPD to a specific data
subset, recovering only the subcomponents relevant to that subset. This
would not surface all the model's subcomponents, but it might be
substantially cheaper and more practical for more narrowly targeted
investigations or editing.

\textbf{Better model editing is likely possible.} Our model editing
experiment (\autoref{sec:model-editing}) demonstrated a proof-of-concept
in which we modified a single subcomponent's left singular vector to
change the model's emoticon predictions. More ambitious editing could,
for example, attempt to make more complicated behavioral changes, or
could make edits that avoid off target effects by taking the geometric
structure of other parameter components into account. Hybrid approaches
that combine the interpretability of parameter subcomponents with the
optimization power of LoRA --- for instance, training a low-rank
adaptation with left or right singular vectors restricted to those of
specific subcomponents --- could also yield edits that are both more
performant and more interpretable than either approach alone.

\textbf{Interpreting parameter gradients to understand training and
finetuning.} Parameter subcomponents are directions in parameter space,
so it is possible to project any parameter gradient into the basis
defined by the decomposition's subcomponents. This could allow us to
express each gradient update to a model as a combination of upweighting,
downweighting, or modifying existing subcomponents, as well as creating
new ones outside the span of the existing subcomponent subspace. Since
individual parameter subcomponents are interpretable, this may give us
some idea of what each training step is teaching the model. Similarly,
one could apply VPD to only a `weight diff' obtained by finetuning,
instead of the `diff' from single gradient updates.

\subsection{Related work}\label{related-work}

\label{sec:related-work}

\subsubsection{Ablation-based parameter
decomposition}\label{ablation-based-parameter-decomposition}

VPD is built primarily on prior parameter decomposition methods, namely
attribution-based parameter decomposition (APD)
\citep{braun2025interpretabilityparameterspaceminimizing} and stochastic
parameter decomposition (SPD) \citep{bushnaq2025spd}. These papers
introduced most of the core ideas used by our method, including (a) the
idea that networks could be decomposed into sparsely used functional
units consisting of vectors in parameter space that sum to the
parameters of the target model, and (b) causal importances can be
identified using a causal importance network and ablations. SPD lacked
adversarial sampling scheme that would make the causal importances
robust to adversarial ablations, as well as the additional loss to
encourage computational simplicity, here implemented as the
frequency-minimality loss. Those works also focused primarily on toy
models, rather than language models trained on natural data. Other work
\citep{christensen2025decomposition} did apply SPD to parts of a larger
model, but did not decompose a whole language model, and lacked the
crucial extra losses as Bushnaq et al \citep{bushnaq2025spd}.

\subsubsection{Identifying computational subgraphs in architectural unit
basis}\label{identifying-computational-subgraphs-in-architectural-unit-basis}

Much work in interpretability views neural networks as computational
graphs and circuits as computational subgraphs that have a particular
function \citep{wang2022interpretability, conmy2024towards}. The
identification of subgraphs has been approached through a range of
methods, including using learned masks, ablations, or the use of
attributions to identify ablatable network components.

Some of the work that identifies subgraphs learns explicit
differentiable masks
\citep{csordas2021neuralnetsmodularinspecting, decao2021sparseinterventionslanguagemodels}
is loosely analogous to our causal importance functions. But these
methods use the learned masks as the actual ablations, rather than to
parameterize an ablation procedure. It is very unlikely, therefore, that
the masks are robust to adversarial ablation (where, e.g.~the masked
parameters are only partly ablated, which should be equivalent to full
ablation if those parameters were actually causally important) and hence
unlikely that the `subnetworks' found by those works are mechanistically
faithful. Those works also learned masks for sets of datapoints, rather
than single datapoints, as in our work. Additionally, the masks learned
by those works were aligned with the parameter unit basis, unlike in our
work where the parts of the parameters that are ablated are not
necessarily aligned with the parameter unit basis. Later work
\citep{conmy2024towards} adapted the mask-learning procedure of
\citep{decao2021sparseinterventionslanguagemodels} to identify subgraphs
where each node could be tested for its importance on a task, which is
assessed by ablations, namely activation patching. Activation patching
involves replacing a nodes activation with a choice of baseline, such as
the zero, mean, random, or other baseline. Our work operates on
parameters, and therefore avoids the need to choose a baseline in
activation space.

\subsubsection{Identifying computational subgraphs using learned
decompositions}\label{identifying-computational-subgraphs-using-learned-decompositions}

Much of the above work operates on architectural components of networks,
such as the neuron unit-basis, parameter unit-basis, whole MLP layers,
or whole attention heads
\citep{csordas2021neuralnetsmodularinspecting, decao2021sparseinterventionslanguagemodels, conmy2024towards, wang2022interpretability}.
But neural computations may not be aligned with those bases, and
therefore the subgraphs they identify may involve components that are
polysemantic (cite polysem references) and thus not yield accounts of
neural computation that are maximally parsimonious. Like our work,
existing work aims to address this issue by learning decompositions of
neural networks from which to make more easily interpretable subgraphs
(though see cite transluce paper, which argues that the neuron basis was
not as unparsimonious as previously thought).

Most similar to ours is the line of work that involves training CLTs and
building attribution graphs for them, thus enabling accounts of
computation that are not necessarily aligned with individual neurons or
layers
\citep{ameisen2025circuit, lindsey2025biology, kamath2025tracing}. CLTs
build on per-layer transcoders
\citep{dunefsky2024transcodersinterpretablellmfeature, ge2024localglobal}.
In contrast to our work, CLTs and transcoders decompose activations,
which are the results of computations, rather than parameters, which
learn to implement the computations (through interactions with the
nonlinearities). Additionally, while Kamath et al.
\citep{kamath2025tracing} built on CLTs to extend their attribution
graphs to attention layers, their approach did not identify ways to
decompose attention layers into functional units that may be distributed
across heads. In our work, our parameter subcomponents learn specialized
functional roles and also span multiple heads by default.

In addition to these topics, our work builds on broader foundations,
including sparse dictionary learning, causal mediation analysis,
interpretability of neural network parameters, automated circuit
discovery, and other topics. We refer readers to our previous papers for
deeper discussion of prior work on related topics
\citep{braun2025interpretabilityparameterspaceminimizing, bushnaq2025spd}.

\section{Conclusion}\label{conclusion}

On the surface, neural network weights may seem like large inscrutable
matrices of floating point numbers, and the computations they implement
like monolithic, irreducible transformations between high-dimensional
vector spaces. Parameter decomposition methods offer a lens with which
these matrices can be decomposed and their computational roles
scrutinized. We are very excited that now, with VPD, it is possible to
decompose the parameters of non-toy models (such as language models)
which solve tasks using neural algorithms that we do not yet know how to
design ourselves. This represents an important step beyond the
capabilities of previous parameter decomposition methods
\citep{bushnaq2025spd, braun2025interpretabilityparameterspaceminimizing}.
We have shown that VPD can be used for many of the major
interpretability tasks (such as constructing interpretable attribution
graphs for circuit analysis) that have so far been achieved with other
methods, such as CLTs. It also addresses a number of important
shortcomings of other methods, not least the ability to decompose
attention layers into functionally specialized units and study their
interactions. However, we think it is likely that parameter
decomposition methods will require further improvement, as they are
scaled to larger models, and as unforeseen pathologies with the current
methods are revealed with time. Even if key parts of the method turn out
to require rethinking, we believe future iterations of it will continue
to resemble VPD in spirit.

We think parameter decomposition may open up new affordances, not just
for mechanistic interpretability, but for deep learning in general. We
need to understand neural algorithms in terms of their parameters before
we can design whole neural networks --- whole minds --- that have more
of the qualities we want and fewer of those we do not. We think the
ability of VPD to decompose networks' parameters into minimal, simple,
mechanistically faithful parts represents a meaningful step toward that
vision.

\section{Contributions statement}\label{contributions-statement}

\label{sec:contributions-statement}

\textbf{Research iteration}

Our method underwent significant iteration throughout development,
changing many times in response to experimental results. LB, OCG, LS,
and DB were primarily responsible for driving forward various iteration
cycles, with NH responsible for some cycles. DB and LB tuned
hyperparameters for various methods throughout the length of the
project. LB did early method and hyperparameter iteration to get
adversarial losses working on toy models and an earlier model trained on
SimpleStories.

\textbf{Conceptualisation}

LB conceptualised the adversarial reconstruction loss and its
implementation via projected gradient descent (PGD) on sources, with
some input from LL. OCG came up with using persistence in the
adversarial training loss and did hyperparameter optimization for it. DB
conceptualised the part of the current adversarial loss which does
several steps of warmup of the persistent sources for each outer loss
step. LS identified the pathological bisemanticity of subcomponent
activations that helped to motivate the addition of a `computational
simplicity' penalty. LB, based on discussions with LL and external
collaborators as well as empirical iteration, conceptualized the
frequency-minimality loss and did most of the testing and tuning for it.
LB conceptualized the new lower-leaky sigmoid after discussion with LS.
LL conceptualised the sign exception on the straight-through estimator
after LB noticed a problem with the previous version. LB conceptualized
delta components and did the early testing for them. NH came up with the
idea for subset routing and ran the first experiments with it. LS
conceptualized the parameter faithfulness warmup and did some
experimental investigation into its usefulness. NH also contributed
p-annealing and other method optimizations and evaluations that were
useful for assessing the value of modifications to the method. OCG
designed the current causal importance function architecture, as well as
the shared\_mlp, global\_shared\_mlp, and vector gate MLP architectures
used in earlier versions. LS did an initial implementation of the global
causal importance function. LB conceptualised post-hoc causal importance
optimization and post-hoc adversarial optimization restricted to base
graph nodes, and did most of the hyperparameter tuning for post-hoc
causal importances. NH contributed p-annealing, subset reconstruction
losses, and other methods optimizations. LB conceptualised using
subcomponent activations on top of causal importances for
interpretability.

\textbf{Clustering}

LB conceptualised the first form of the clustering algorithm, including
the MDL framing, initial MDL loss function, hierarchical merging,
stopping based on MDL minimum, and picking alpha based on coactivation
threshold. MI developed the algorithm further, with inputs from NH, LB,
and LS. NH helped MI on clustering, primarily conceptually. LB did some
of the empirical iteration to pick a clustering for the paper. OCG and
DB optimized the clustering implementation for efficiency.

\textbf{Attributions and analysis}

LB did much of the conceptualisation work for the attributions used in
the paper (including gradient stopping), with input from OCG, DB, and
LS. LS conceptualized the dataset attributions. LS and LB jointly
conceptualised the nonlinear interaction metric. LS ran initial
investigations into nonlinear interactions on an older language model,
and LB ran the nonlinear interaction experiments used in the paper. LS
was responsible for the analysis of attention behaviors and the
geometric consistency seed analysis. LB did the first circuits stories
on the simple stories model and the two circuits stories in this paper.

\textbf{Model editing}

OCG did early explorations of model editing. LB contributed early
conceptualisation for model editing. OCG and LB together did the final
version of the model editing experiment in the paper.

\textbf{Comparisons and evaluations}

OCG was primarily responsible for autointerp pipeline and intruder
detection comparisons. BB trained the per-layer and cross-layer
transcoders used for comparisons to VPD, did the evaluation and analysis
of the reconstruction performance comparing VPD to transcoders, and did
the feature splitting analysis.

\textbf{Target Model pretraining}

DB was responsible for model pretraining. LS helped train target models
on the Pile dataset.

\textbf{Engineering and infrastructure}

OCG and DB equally managed the codebase and the implementations of the
various methods.

\textbf{Visualization and interactive figures}

OCG was primarily responsible for the internal visualization app and for
the interactive figures in the paper. DB helped with the internal
visualisation app and the attribution graph visualisation. LS and LB
contributed some features to the visualization app. LS designed and made
various didactic figures used in the paper.

\textbf{Writing}

LS planned the paper and wrote initial drafts of some sections. LB wrote
initial drafts for the two biostories, methods sections on frequency
minimality loss, mechanistic faithfulness, and adversarial loss, the
nonlinear interactions section, model editing section, parts of the
discussion section, training recipe, and most of the mathematical
sections in the appendix. MI wrote an initial draft of the paper section
on clustering. BB drafted the section comparing VPD to transcoders and
drafted the feature splitting section. OCG was primarily responsible for
web development and for the interactive figures, with contributions from
others. DB helped with editing.

\textbf{Project management and mentorship}

LS was responsible for overall management of the project and planning
the paper. LS was the main point of contact for MI, NH, and BB and gave
input on their work throughout the collaboration. LB and DB also gave
input on their work.

\subsubsection{Funding declarations}\label{funding-declarations}

Linda Linsefors was supported by a grant from Coefficient Giving during
her work on the project.

\subsubsection{Acknowledgements}\label{acknowledgements}

We extend our sincere thanks to several individuals for their inputs to
our work.

We are especially grateful to Tom McGrath for ongoing conversations,
support, and feedback at various stages throughout the project, and to
many other colleagues at Goodfire, including Dron Hazra, Eric Ho, Curt
Tigges, Thomas Fel, Sheridan Feucht, Usha Bhalla, and Michael Jae Byun.

We express our gratitude to Chris Olah for extended written
correspondence about nonlinear interactions between subcomponents. We
are also grateful to Joshua Batson, Emmanuel Ameisen, Thomas Conerly,
Brian Chen, Jeff Wu, Harish Kamath, Stefan Heimersheim, and Asher
Parker-Sartori for their in-depth technical engagement, and to Connor
Watts, Dan Murfet, and Eric Ho whose collective feedback greatly
improved drafts of our paper.

We thank Param Luhadiya for exploratory work on the frequency-minimality
penalty, and Markus Salmela, Finn McDonnell, and Ed Allison for their
feedback on how to explain our work to a general audience. We also thank
Cory Kendrick for inputs on research management, and also thank Asher
Parker-Sartori, Jack Peck, and Antoine Vigouroux for technical
discussions and experimental work adjacent to the content of the paper.

\subsubsection{Citation}\label{citation}

\begin{verbatim}
@misc{bushnaq2026interpreting,
  title={Interpreting Language Model Parameters},
  author={Bushnaq, Lucius and Braun, Dan and Clive-Griffin, Oliver and
    Bussmann, Bart and Hu, Nathan and Ivanitskiy, Michael and Linsefors,
    Linda and Sharkey, Lee},
  journal={Technical Report},
  institution={Goodfire and MATS},
  month={April},
  year={2026},
}
\end{verbatim}

\appendix

\section{Methods}\label{methods}

\label{sec:vpd_methods}

Here, we expand on some aspects of adVersarial Parameter Decomposition
(VPD) in more detail. See \autoref{sec:method} for an introduction to
VPD.

\subsection{\texorpdfstring{\(\Delta\)-L2
penalty}{\textbackslash Delta-L2 penalty}}\label{delta-l2-penalty}

\label{sec:vpd_delta_l2}

The \(\Delta\)-components are different from normal subcomponents we
train. Their rank can be greater than \(1\), meaning they can be more
complicated objects than regular subcomponents. We thus have a
particular interest in ensuring that they do not contain computations
that affect the model's outputs. Theoretically, since we define the
causal importances of \(\Delta\)-components to always be zero, the
stochastic and adversarial losses should ensure that this is the case.
But in practice our reconstruction losses are not perfect, so we
additionally encourage the \(\Delta\)-components to be small with an
auxiliary MSE loss:

\[\mathcal{L}_{\text{Delta-L2}}=\frac{1}{N}\sum^L_{l=1}\sum_{i,j}\left(\Delta^l_{i,j}\right)^2=\frac{1}{N}\sum^L_{l=1}\sum_{i,j}{\left( W^{l}_{i,j}- \sum^C_{c=1} U^l_{i,c} V^l_{j,c}\right)}^2.\]

Here, \(N\) is the total number of decomposable model parameters.

\subsection{Causal Importance Function
Architecture}\label{causal-importance-function-architecture}

\label{sec:vpd_ci_function}

The causal importance function \(\Gamma\) maps the target model's hidden
activations to per-subcomponent causal importances. It is a single,
shared network that jointly computes causal importances for all
subcomponents across all weight matrices in the target model.

\textbf{Inputs}

Let \(L\) denote the number of weight matrices being decomposed, and let
\(\vec{h}^l_{b,t} \in
\mathbb{R}^{d_l}\) denote the input hidden activation to weight matrix
\(l\) of the target model at batch element \(b\) and sequence position
\(t\). Each activation vector is independently RMS-normalized, and the
normalized vectors are concatenated to form the input:

\[\vec{\varphi}_{b,t} = \left[
 \operatorname{RMSNorm}(\vec{\varphi}^1_{b,t}) \;|\; \cdots \;|\;
 \operatorname{RMSNorm}(\vec{\varphi}^L_{b,t})
 \right] \in \mathbb{R}^{D},
 \quad D = \sum_{l=1}^{L} d_l.\]

\textbf{Input projection}

The concatenated activation vector is linearly projected to the
transformer's \(d_{\mathrm{model}}\) dimension:

\[\vec{z}^{(0)}_{b,t} = W_{\mathrm{in}} \vec{h}_{b,t} + \vec{b}_{\mathrm{in}},
 \quad W_{\mathrm{in}} \in \mathbb{R}^{d_{\mathrm{model}} \times D}, \;
 \vec{b}_{\mathrm{in}} \in \mathbb{R}^{d_{\mathrm{model}}}.\]

\textbf{Transformer layers}

The projected activations are processed by \(N\) pre-norm transformer
layers. Each layer \(n \in
\{1, \ldots, N\}\) applies bidirectional multi-head self-attention
followed by a feedforward network, each with a residual connection:

\[\begin{aligned}
\vec{\hat{z}}^{(n)}_{b,t} &= \vec{z}^{(n-1)}_{b,t} +
\operatorname{Attn}\!\left(
\operatorname{RMSNorm}\!\left(\vec{z}^{(n-1)}_{b,t}\right)
\right)\!, \\
\vec{z}^{(n)}_{b,t} &= \vec{\hat{z}}^{(n)}_{b,t} +
\operatorname{FFN}\!\left(
\operatorname{RMSNorm}\!\left(\vec{\hat{z}}^{(n)}_{b,t}\right)
\right)
\end{aligned}\]

where \(\operatorname{Attn}\) denotes multi-head scaled dot-product
attention with Rotary Position Embeddings (RoPE) \citep{su2024roformer},
applied bidirectionally (i.e., without a causal mask) across all \(T\)
sequence positions; and \(\operatorname{FFN}\) is a two-layer
feedforward network with GELU activation:

\[\operatorname{FFN}(\vec{z}) = W_2 \operatorname{GELU}(W_1 \vec{z} + \vec{b_1}) + \vec{b_2},
 \quad W_1 \in \mathbb{R}^{d_{\mathrm{ff}} \times d_{\mathrm{model}}}, \;
 W_2 \in \mathbb{R}^{d_{\mathrm{model}} \times d_{\mathrm{ff}}}.\]

\textbf{Output head}

After the final transformer layer, a linear output head projects back to
the total number of subcomponents:

\[\vec{z}^{(N+1)}_{b,t} = W_{\mathrm{out}} \vec{z}^{(N)}_{b,t} + \vec{b}_{\mathrm{out}},
 \quad W_{\mathrm{out}} \in \mathbb{R}^{C_{\mathrm{total}} \times d_{\mathrm{model}}}, \;
 C_{\mathrm{total}} = \sum_{l=1}^{L} C_l.\]

The output is partitioned according to each matrix's subcomponent count
\(C_l\).

\textbf{Leaky hard sigmoids}

Theoretically, the causal importance for subcomponent \(c\) in matrix
\(l\) is obtained simply by clamping the outputs of the final
transformer layer \(\vec{z}^{(N+1)}_{b,t}\) to the interval\([0,1]\)
with a hard sigmoid function:

\[g^l_{b,t,c} = \Gamma(\vec{\varphi}_{b,t})^l_c=\sigma_{\mathrm{H}}\!\left(\vec{z}^{(N+1)}_{b,t}\right),
 \quad \sigma_{\mathrm{H}}(\vec{z}) = \mathrm{clamp}(\vec{z}, 0, 1),\]

However, in practice, the flat regions in a hard sigmoid function can
lead to dead gradients for inputs below \(0\) or above \(1\). To avoid
this, we use leaky hard sigmoids instead. Specifically, we use
\emph{lower-leaky} hard sigmoids \(\sigma_{H,\text{lower}}(\vec{z})\)
for the causal importance used to create the masks for the actual
forward passes for the \(\mathcal{L}_{\text{stochastic-recon}}\) and
\(\mathcal{L}_{\text{stochastic-recon-layerwise}}\) losses, and we use
\emph{upper-leaky} hard sigmoids \(\sigma_{H,\text{upper}}(\vec{z})\) in
the importance minimality loss
\(\mathcal{L}_{\text{importance-minimality}}\) and the frequency
minimality loss \(\mathcal{L}_{\text{frequency-minimality}}\).

The lower-leaky hard sigmoid \(\sigma_{H,\text{lower}}(\vec{z})\) has a
forward pass identical to a regular hard sigmoid, but below \(0\) it
uses a straight-through gradient estimator: Gradients pass through for
\(\vec{z} \leq 0\) scaled by a leak coefficient \(\alpha = 0.01\) when
the incoming gradient is negative, preventing subcomponents from
becoming permanently deactivated. The upper-leaky hard sigmoid
\(\sigma_{H,\text{upper}}(\vec{z})\) is identical to a regular hard
sigmoid for \(\vec{z} \leq 1\), but has a slope of \(0.01\) above
\(1.0\).

We use a straight-through estimator for the lower-leaky hard sigmoid
instead of actually modifying the slope on the forward pass to avoid
creating subcomponent masks smaller than zero. We restrict the
straight-through estimator to apply only to negative gradients to
prevent entries of \(\Gamma(\vec{\varphi}_{b,t})^l_c\) from updating to
become ever more negative indefinitely.

This is in contrast to \citep{bushnaq2025spd}, where the lower-leaky
hard sigmoid did have an actual slope of \(0.01\) below \(0\) on the
forward pass. We made this change because we discovered that negative
masks actually led to instabilities. For example, we found that the
spurious subcomponent splitting observed for too-high importance
minimality loss coefficients depicted in Figure 8 of that paper largely
disappears if the straight-through estimator is used instead.

\textbf{Hyperparameters}

\autoref{tab:ci-hyperparams} lists the hyperparameters used for the
causal importance function \(\Gamma\) in our experiments.

\begin{longtable}[]{@{}ll@{}}
\caption[\label{tab:ci-hyperparams} \emph{Table: Hyperparameters for the
causal importance function \(\Gamma\)}.]{\label{tab:ci-hyperparams}
\emph{Table: Hyperparameters for the causal importance function
\(\Gamma\)}\footnote{Recall that the target model is a 4-layer
  Llama-style transformer with \(d_{\mathrm{model}} = 768\) and
  \(d_{\mathrm{intermediate}} = 3072\), decomposed across \(L = 24\)
  weight matrices (6 per layer: \texttt{c\_fc}, \texttt{down\_proj},
  \texttt{q\_proj}, \texttt{k\_proj}, \texttt{v\_proj},
  \texttt{o\_proj}), yielding a total of \(C_{\mathrm{total}} = 38,912\)
  subcomponents and an input dimension of \(D = 27,648\)}.}\tabularnewline
\toprule\noalign{}
\textbf{Parameter} & \textbf{Value} \\
\midrule\noalign{}
\endfirsthead
\toprule\noalign{}
\textbf{Parameter} & \textbf{Value} \\
\midrule\noalign{}
\endhead
\bottomrule\noalign{}
\endlastfoot
CI model dimension (\(d_{\mathrm{model}}\)) & 2048 \\
Transformer layers (\(N\)) & 8 \\
Attention heads & 16 \\
Head dimension & 128 \\
FFN hidden dimension (\(d_{\mathrm{ff}}\)) & 8192 \\
Positional encoding & RoPE (base \(= 10,000\), max length \(= 512\)) \\
Attention & Bidirectional (no causal mask) \\
Activation function & Leaky hard sigmoid (\(\alpha = 0.01\)) \\
\end{longtable}

\subsection{Reconstruction losses}\label{reconstruction-losses}

\label{sec:recon}

\subsubsection{Formal Setup}\label{formal-setup}

\label{sec:vpd_opt-mech-faith-setup}

Ablation-based parameter decomposition methods, at their core,
instantiate this definition of mechanistic faithfulness by using their
causal importance functions (\autoref{sec:opt-minimality}) to estimate
how ablatable each parameter subcomponent is on a given datapoint. They
then actually do an ablation and train the model with ablated parameters
to approximate the same output as the unablated model. Crucially, the
ablations may be full \emph{or partial}.

Formally, we define ablation masks \(m^l_{b,t,c}(r)\in[0,1]\) for each
subcomponent at each each batch index \(b\) and sequence position \(t\).
These masks define new weight matrices \(W^{\prime l}_{b,t}(r)\) which
can take the place of the original model matrices \(W^l\):\footnote{For
  simplicity, we omit the addition of the \(\Delta\)-component masking
  term \(m^l_{b,t,C+1} \Delta^l_{i,j}\) to this sum.}

\begin{equation}
\begin{aligned}
    W^{\prime l}_{b,t}
  (r)
  :=
    \sum^C_{c=1}
    \vec{U}^l_c
      m^l_{b,t,c}
        (r)
    (\vec{V}^l_c)^\top
  \end{aligned}
\end{equation}

Crucially, the masks are not the causal importances, \(g^l_{b,t,c}\).
Instead, the masks are given by

\[m^l_{b,t,c}(r) :=g^l_{b,t,c}+(1-g^l_{b,t,c})r^l_{b,t,c},\]

where \(r^l_{b,t,c} \in [0, 1]\) is called a `source'. This means that
if a subcomponent's causal importance is \(1\), the only possible value
of its mask is \(1\), whereas if the causal importance is \(0\), its
mask can take any value between \(0\) and \(1\). The causal importance
of the \(\Delta\)-components \(\Delta^l\) is always zero.

Concretely, when computing the output vector of matrix \(l\) at batch
index \(b\) and sequence position \(t\), we replace the original weight
matrix \(W^l\) with \(W^{\prime l}_{b,t}(r)\), which is constructed from
the masks at that specific position. This means that during a single
forward pass through the network, different linear transformations are
applied at each sequence position, determined by which subcomponents are
masked on vs.~off at that position. In the idealised setting, we then
demand that, for \emph{all possible joint combinations} of sources
\(r\in {[0,1]}^{L\times B \times T \times C+1}\), the resulting masked
weight matrices yield outputs that approximately match those of the
original model at every batch index and every output sequence position:

\begin{equation}
\forall r
  :
    f(\vec{x}_b \vert 
        W^{\prime 1}_{b}(r),\dots,W^{\prime L}_{b}(r)
      )
  \approx 
  f(\vec{x}_b\vert W^1,\dots,W^L).
\label{eq:subcomponents}
\end{equation}

where \(f(\vec{x}_b\vert W^1,\dots,W^L)\) is the sequence of output
vectors produced by the target model for input sequence \(\vec{x}_b\).
This definition of ablatability lies at the heart of how VPD and other
ablation-based parameter decomposition methods try to ensure that the
subcomponents and causal importances they provide are mechanistically
faithful to the original network.

\subsubsection{Stochastic reconstruction
losses}\label{stochastic-reconstruction-losses}

\label{sec:vpd_methods-stoch}

We can use an output reconstruction loss to train the masked model's
output to approximate the target model's. Unfortunately, to ensure we
satisfy \autoref{eq:subcomponents}, we would need to do this for
\emph{all possible values of}
\(r\in {[0,1]}^{L\times B\times T \times C+1},\) which is a high
dimensional continuous interval, making such a loss impossible to
compute exactly.

However, a key insight of Bushnaq et al. \citep{bushnaq2025spd} was that
it is possible to \emph{approximately} minimize reconstruction loss on
all values in that interval using a finite number \(S\) of uniform
random samples \(r^{l,(s)}_{b,t,c} \sim \mathcal{U}(0,1)\) for every
sequence index \(t\) and every batch index \(b\). These samples can be
used to create stochastic masks
\(m^l_{b,t,c} \sim \mathcal{U}(g^l_{b,t,c}, 1)\), and minimize
reconstruction loss on that finite number of samples.

This leads to the \emph{stochastic reconstruction loss}:

\begin{equation}
\begin{aligned}
  \mathcal{L}_{\text{stochastic-recon}}
  &=
  \frac{1}{S}
  \sum^{S}_{s=1}
  \frac{1}{B}
  \sum^{B}_{b=1}
    D
    \Big(
        f(
          \vec{x}_b
          \vert
            W
        )
      ,
        f(
          \vec{x}_b
          \vert
            W'_b(
                r^{(s)}
            )
        )
    \Big)
   \\
  \end{aligned}
\end{equation}

where \(D\) is an appropriate divergence measure in the space of model
outputs, such as KL-divergence or mean squared error. In practice, we
find that using one sample (\(S=1\)) produces similar training behavior
as using more samples.

In practice, for better convergence, we train by sampling masks for
randomly chosen subsets of the model's weight matrices instead of all
matrices simultaneously. See the next section for details.

\subsubsection{Stochastic Subset reconstruction
loss}\label{stochastic-subset-reconstruction-loss}

\label{sec:vpd_subset_recon}

\citep{bushnaq2025spd} found that using a reconstruction loss which
samples stochastic masks

\[\begin{aligned}
&m^l_{b,t,c}(r^{\text{stoch}}):=g^l_{b,t,c}+\left(1-g^l_{b,t,c}\right)r^{\text{stoch},l}_{b,t,c}\\
&r^{\text{stoch},l}_{b,t,c} \sim \mathcal{U}(0,1)
\end{aligned}\]

for all target model matrices \(l\) simultaneously\footnote{Simply
  called ``stochastic reconstruction loss'' in that paper, but here we
  reserve that term for the formulation that ends up in the training
  loss.}

\[\begin{aligned}
\mathcal{L}_{\text{stochastic-recon-all}}&=\frac{1}{B}\sum^B_{b=1} D \left( f\left(\vec{x}_b\vert W^1,\dots,W^L\right), f\left(\vec{x}_b\vert {W'}^1\left(r^{\text{stoch}}\right),\dots, {W'}^L\left(r^{\text{stoch}}\right)\right) \right) \\
\end{aligned}\]

together with a layer-wise stochastic reconstruction loss which samples
stochastic masks for one target model matrix at a time

\[\begin{aligned}
\mathcal{L}_{\text{stochastic-recon-layerwise}}=\frac{1}{L}\sum^L_{l=1}\frac{1}{B}\sum^B_{b=1} D \Big(f\left(\vec{x}_b\vert W^1,\dots,W^L\right), f\left(\vec{x}_b\vert W^1,\dots,W'^l(r^{\text{stoch}}),\dots,W^L\right) \Big) \\
\end{aligned}\]

performed better than training either
\(\mathcal{L}_{\text{stochastic-recon-all}}\) or
\(\mathcal{L}_{\text{stochastic-recon-layerwise}}\) alone, due to
covering a somewhat more structurally diverse set of ablation. However,
layer-wise reconstruction loss requires one forward-pass for every
matrix in the model we decompose, which is expensive. For VPD training,
we unify \(\mathcal{L}_{\text{stochastic-recon-all}}\) and layerwise
stochastic reconstruction loss
\(\mathcal{L}_{\text{stochastic-recon-layerwise}}\) into a single
stochastic reconstruction loss. For every sequence position and batch
index, we independently sample a number \(\in\{1,\dots,L\}\), where
\(L\) is the number of weight matrices in the target model. We draw that
many of the target model's weight matrices, sample stochastic masks for
only those, and perform a forward pass replacing those matrices with the
masked ones. This is no more computationally expensive than
\(\mathcal{L}_{\text{stochastic-recon-all}}\), and covers more
structurally diverse ablations than layer-wise stochastic reconstruction
losses, since it includes subsets of single matrices as well as the
whole set as special cases.

Although this reconstruction loss on its own is enough to succeed in
many toy settings, our attempts to apply that method at larger scales
(such as language models) revealed several pathologies that we missed.
We had under-appreciated the importance of reconstruction under
worst-case ablation masking which we address in the next section.

\subsubsection{Adversarial reconstruction
loss}\label{adversarial-reconstruction-loss}

\label{sec:vpd_methods-adv}

VPD additionally optimizes for \emph{adversarial ablatability} of
parameter subcomponents that are causally unimportant on a datapoint,
which is a stricter criterion than \emph{stochastic ablatability}.

In the limit of infinite samples and perfect reconstruction,
\(\mathcal{L}_{\text{stochastic-recon}}\) loss would perfectly
approximate our desired condition from \autoref{eq:subcomponents}. But
we don't have time to draw infinite samples. And
\autoref{eq:subcomponents} requires that the masked model approximates
the target model well for \emph{all} possible values of \(r\), not just
on average. Thus, if the reconstruction loss isn't exactly zero, which
will essentially always be the case in practice, stochastic sampling can
greatly underestimate the worst-case reconstruction error for values of
\(r\) that are sampled adversarially to maximize reconstruction loss. We
found that training without an adversarial sampling scheme produces
decompositions for which adversarial sampling can find values of \(r\)
that have worse-than-random reconstruction loss, which is not permitted
under \autoref{eq:subcomponents} (See also \autoref{fig:adv-vs-no-adv}).

VPD therefore introduces an adversarial loss to help ensure this
property more: Instead of sampling the sources \(r\) randomly, they are
sampled by an adversarial optimizer to be as bad as possible.

The optimization objective of the adversarial optimizer is maximizing
the reconstruction loss on the masked forward pass:

\[\begin{aligned}
\mathcal{L}_{\text{adversarial-recon}}:=\frac{1}{B}\sum^B_{b=1} D \Big(f(\vec{x}_b\vert W^1,\dots, W^L), f\left(\vec{x}_b\vert W'^1(r^{{\text{adv}}}),\dots,W'^L(r^{{\text{adv}}})\right) \Big)
\end{aligned}\]

by optimizing adversarial sources \(r^{{\text{adv}},l}_{b,t,c}\) for the
masks \(m^l_{b,t,c}(r^{\text{adv}})\):

\[\begin{aligned}
m^l_{b,t,c}(r^{\text{adv}}) &:=g^l_{b,t,c}+(1-g^l_{b,t,c})r^{\text{adv},l}_{b,t,c}\\
W'^l_{b,t,i,j}(r^{{\text{adv}}})&:=\sum^C_{c=1} U^l_{i,c} m^l_{b,t,c}(r^{\text{adv}}) V^l_{j,c}
\end{aligned}\]

for subcomponent \(c\) of target model matrix \(l\) on batch index \(b\)
at sequence position \(t\). The optimizer we use is projected gradient
ascent \citep{bertsekas1999nonlinear, madry2018towards}, clamping the
sources \(r^{\text{adv}}_{b,t,c}\) to the interval \([0,1]\) at every
update step to ensure that the masks \(m^l_c(x,t,r^{\text{adv}})\) stay
between \(0\) and \(1\). The sources for the \(\Delta\)-components'
masks (see \autoref{sec:method-components}) are treated identically to
those used for the regular subcomponents, i.e.~they are also
adversarially optimized.

\begin{equation}
\begin{aligned}
  \mathcal{L}_{\text{adversarial-recon}}
  &=
      \max_{r^{\text{adv}}}
    \frac{1}{B}
    \sum^{B}_{b=1}
      D
      \Big(
          f(
            \vec{x}_b
            \vert
              W
          )
        ,
          f(
            \vec{x}_b
            \vert
              W'_b(
                  r^{ \text{adv} }
              )
          ),
      \Big)
  \end{aligned}
\label{eq:adv_recon}
\end{equation}

\textbf{Complete adversarial robustness seems too strict}

However, if the adversarial sampler were completely unconstrained, it
would actually be too strict: Some decompositions that we would
intuitively regard as valid would be effectively excluded by it. For
example, in many theoretical toy models of circuits in superposition
\citep{hanni2024mathematicalmodelscomputationsuperposition, bushnaq2024circuits, linsefors2025circuits}
models can contain more circuits than neurons, only some of which are
used by the model on any given forward pass. However, the inactive
circuits each still contribute some small interference ``noise'' to the
computation. Since this noise is uncorrelated between superposed
circuits, its overall size remains small enough that the interference
doesn't ``break'' the computation. We would like to consider these
inactive circuits not to be causally important since the model is in
some sense not really using them to compute the output. But if we chose
the absolute worst-case \(r^{\text{adv}}_{c}\) in such a model (which we
can do if we have a completely unconstrained adversarial sampler), we
could, for example, ablate all inactive circuits which contribute noise
with a negative sign, but keep all inactive circuits which contribute
noise with a positive sign. This would vastly increase the overall size
of the noise and thus change the final output of the model!

In general, we want the adversarial sampler to penalise
\emph{systematic} defects in the decomposition, where a particular
choice of ablation masks changes the model output on many data points
even though it shouldn't. But we do not want the sampler to exploit
random noise by finely tuning its choice of ablations to particular data
points. This is because in practice, when using the decomposition to
understand or edit the target model, we usually care about the behavior
of particular subcomponent maskings over multiple data points, rather
than the behavior of all possible maskings on single data points.

For example, suppose we wanted to edit the target model for some
practical purpose, like erasing some of its knowledge about biology. We
could therefore apply a mask to some of the model's subcomponents that
are causally important in biology contexts, but not other contexts.
Ideally, the resulting model should still behave the same way for all
inputs on which those subcomponents were not causally important. This
mask would be very unlikely to be exactly tuned to random noise in the
activations of some other input. And even if it did happen to be so
tuned, then this would merely cause the edited model to behave
unexpectedly on the input that the mask happened to be tuned to, and
thus not be a very effective adversarial mask on other inputs. But if
the decomposition was \emph{systematically} defective, we might have a
realistic chance of picking a mask that causes the edited model to
behave differently than the target model on many inputs not related to
biology. This would be an effective adversarial mask that would hurt the
model editing more broadly.

Thus, in order to force the adversarial sampler to rely on systematic
flaws in the decomposition instead of fine-tuning to individual data
points, we restrict it to use the same \(r^{\text{adv}}_c\) on all
elements in a batch. This approach is somewhat related to
universal/shared adversarial perturbation methods that optimize one
perturbation across many inputs
\citep{shafahi2019universaladversarialtraining, moosavidezfooli2017universaladversarialperturbations, Mummadi_2019_ICCV}.

Ideally, we might like to use the same sources for the whole data set,
but this would be too computationally expensive in training. In
practice, we thus use two different sampling schemes for
\(r^{\text{adv},l}_{c}\) source schemes for evaluation and training.

\textbf{Persistent PGD (PPGD) adversarial reconstruction loss for
training:}

For training, we optimize a single set of sources
\(r^{\text{adv},l}_{b,t,c}\) that persists across batches, with \(b\)
ranging across the batch index and \(t\) across sequence position. On
every batch, the adversarial Adam optimizer performs \(n_{\text{adv}}\)
update steps on the adversarial sources \(r^{\text{adv}}_{b,t,c}\),
trying to maximise the adversarial loss
\(\mathcal{L}_{\text{adversarial-recon}}\) (In this paper, we used
\(n_{\text{adv}}=3\)).

This persistent adversarial source optimization is a heuristic in the
same broad family as efficient adversarial-training methods that
amortize inner maximization by reusing or accumulating adversarial
perturbations across outer optimization steps, such as
\citep{shafahi2019universaladversarialtraining, zheng2020efficientadversarialtrainingtransferable}.

\textbf{PGD adversarial reconstruction loss for evaluation:}

Continuously updating a single set of persistent adversarial sources is
more computationally efficient, but not principled. Hypothetically, the
VPD optimizer might trap the adversarial optimizer in some local
extremum at some point during training, rendering the adversarial loss
useless. Thus, for evaluation, we use a new set of adversarial sources
\(r^{ \text{adv},l}_c\) for every evaluation batch, but use more
adversarial optimization steps per batch \(n_{\text{adv}}\) than we do
in training.

\subsection{Frequency minimality loss}\label{frequency-minimality-loss}

\label{sec:vpd_frequency_penalty}

Suppose some rank-1 subcomponent \(U_1 (\vec{V}_1)^\top\) in a model
parametrizes two unrelated circuits \(A\) and \(B\), which are rarely
used to compute the model's output at the same batch and sequence
position. We would like VPD to break up this subcomponent into two
subcomponents,
\(\vec{U}_1 (\vec{V}_1)^\top=\vec{U}_A (\vec{V}_A)^\top + \vec{U}_B (\vec{V}_B)^\top\),
with \(\vec{U}_A (\vec{V}_A)^\top\) containing the weights for circuit
\(A\), and \(\vec{U}_B (\vec{V}_B)^\top\) containing the weights for
circuit \(B\). Our loss \(\mathcal{L}_{\text{importance-minimality}}\)
will not incentivise this, because either \(\vec{U}_A (\vec{V}_A)^\top\)
or \(\vec{U}_B (\vec{V}_B)^\top\) will be causally important whenever
\(\vec{U}_1 (\vec{V}_1)^\top\) is, so
\(\sum_{b,t}\vert g_{b,t,1}\vert^p\leq \sum_{b,t}(\vert g_{b,t,A}\vert^p+\vert g_{b,t,B}\vert^p)\).
One way to break up subcomponents like \(\vec{U}_1 (\vec{V}_1)^\top\) is
introducing an additional loss penalty that is very slightly
\emph{superlinear} in causal importance frequency, i.e.~penalizing a
subcomponent that is causally important half of the time more heavily
than two subcomponents that are each active a quarter of the time.

This leaves the question of what precise functional form this
superlinear penalty should take. We ultimately opted for a term that
grows approximately as \(\sum^L_{l=1}\sum^C_{c=1}f^l_c \log_2(f^l_c)\)
with causal importance frequency
\(f^l_c:=\frac{1}{BT}\sum^B_{b=1}\sum^T_{t=1}\vert g^l_{b,t,c}\vert^0\).
This was largely motivated by empirical iteration, though we provide
some theoretical motivation for the log scaling below, based on
minimizing mechanistic description length per data point: The effective
description length of subcomponents in bits (weakly) grows with
\(\log_2(f^l_c)\), because subcomponents that activate more frequently
effectively need to be specified to higher precision to maintain good
output reconstruction.

The normalisation \(\frac{1}{BT}\) inside the \(\log_2\) argument can be
absorbed into the importance minimality loss term via the relation
\(\log_2(f^l_c)=\log_2(\sum^B_{b=1}\sum^T_{t=1}\vert g^l_{b,t,c}\vert^0)-\log_2(BT)\).
Adding a \(1.0\) inside the \(\log_2\) for numeric stability and using
\(L_p\) norm in place of \(L_0\) then yields

\[\begin{aligned}
\mathcal{L}_{\text{frequency-minimality}}=\frac{1}{BT}\sum^L_{l=1}\sum^B_{b'=1}\sum^T_{t'=1}\sum^C_{c=1}\vert g^l_{b',t',c}\vert^p \log_2(1+\sum^B_{b=1}\sum^T_{t=1} \vert g^l_{b,t,c}\vert^p)\,.
\end{aligned}\]

\subsubsection{Information theory
motivation}\label{information-theory-motivation}

\label{app:frequency_penalty_motivation}

Here, we provide an information theoretic motivation for the functional
form of \(\mathcal{L}_{\text{frequency-minimality}}\) based on
minimizing description length per data point: In a fixed dictionary of
subcomponents, subcomponents that are more frequently causally important
effectively need to be specified to more bits of precision to
reconstruct the model's outputs accurately.

In the idealized setting, subcomponents are vectors of real numbers. In
reality, we instead store them as vectors of finite precision floats.
This quantisation effectively induces a discrepancy \(\delta^l_c\) in
parameter space between the ideal parameter vector for subcomponent
\(c\) in matrix \(l\), and our floating point approximation of it. At
sufficiently high float precision, the expected size of this discrepancy
will scale as \(\approx a_1 2^{-b^l_c}\), where \(b^l_c\) is a bit count
and \(a_1\) is some constant. Suppose we want to keep the impact of this
discrepancy on our decomposition low. Specifically, we want the number
of bits \(b^l_c\) to be large enough for the KL divergence between the
VPD forward pass outputs and the target model forward pass outputs
summed over the batch to stay below some fixed \(\epsilon>0\). How large
will we need to make \(b^l_c\) as a function of \(\epsilon\) to achieve
this?

Over a batch of \(B\) inputs of sequence length \(T\), a subcomponent
will be causally important with some frequency
\(f^l_c:=\frac{\sum^{B,T}_{b,t=1}\vert g^l_{b,t,c}\vert^0}{B T}\). For
simplicity, we assume that applying some small perturbation of size
\(\delta\) along the direction of a subcomponent in parameter space does
not change the model output at all on data points where
\(g^l_{b,t,c}=0\), but increases the KL divergence to the original model
outputs by some \(p(\delta)\) on data points where \(g^l_{b,t,c}=1\),
where \(p\) is an analytic function that is approximately the same for
every subcomponent and every data point. Then, the increase to the total
loss summed over all \(B T\) data points from adding a perturbation
\(\delta\) to subcomponent \(c\) is of approximate size
\(\approx \sum^{B,T}_{b,t=1}\vert g^l_{b,t,c}\vert^0 p(\delta)\). This
yields the inequality

\[\begin{aligned}
&\log_2(p(\delta))+\log_2(\sum^{B}_{b=1}\sum^{T}_{t=1}\vert g^l_{b,t,c}\vert^0)<\log_2(\epsilon)\\
\end{aligned}\]

Since \(p\) is an analytic function, for sufficiently small \(\delta\),
it can be Taylor approximated to leading order as \(a_2 \delta^n\) with
some \(n\in\{1,2,\dots\}\). Inserting this approximation yields:

\[\begin{aligned}
b^l_c&>\frac{1}{n}\log_2(\sum^{B}_{b=1}\sum^{T}_{t=1}\vert g^l_{b,t,c}\vert^0)-\frac{\log_2(\epsilon)}{n }+\frac{\log_2(a_2)}{n}+\log_2(a_1)\\
\end{aligned}\]

So, the required bit precision \(b^l_c\) for the parameters of a
subcomponent grows approximately linearly with the logarithm of that
subcomponent's number of causal importance activations across the
dataset \(\log_2(f^l_c)\). If we use a fixed dictionary of subcomponents
to describe how the model computes its outputs, the mechanistic
description length of our descriptions summed over a batch will thus
have a term that scales as
\(\approx \sum^L_{l=1} \sum^C_{c=1} f^l_c \log_2(f^l_c)\). Substituting
the definition
\(f^l_c=\frac{\sum^{B,T}_{b,t=1}\vert g^l_{b,t,c}\vert^0}{B T}\) and
absorbing the \(-\log_2(BT)\) term into the importance minimality loss
yields \(\mathcal{L}_{\text{frequency-minimality}}\).

\subsection{p-annealing}\label{p-annealing}

The \(L^p\) quasi-norm in the importance minimality loss
\(\mathcal{L}_{\text{importance-minimality}}\) and frequency minimality
loss \(\mathcal{L}_{\text{frequency-minimality}}\) (\autoref{eq:minimal}
and \autoref{eq:freq_minimality}) serves as a smooth surrogate for the
\(L_0\) `norm', with smaller \(p\) yielding a tighter approximation.
However, \citep{bushnaq2025spd} found that optimization is substantially
easier at larger \(p\) values like \(p = 2\). We therefore linearly
anneal \(p\) over the course of training, starting from the
easy-to-optimize \(p_0 = 2.0\) and decreasing to
\(p_{\mathrm{final}} = 0.4\):

\[p(t) = p_0 + (p_{\mathrm{final}} - p_0) \cdot \frac{t}{t_{\max}},\]

where \(t\) is the current training step and \(t_{\max}\) is the total
number of steps. In our experiments, annealing begins at the start of
training and proceeds linearly over the full run
(\(t \in [0, t_{\max}]\)).

\subsection{A training recipe for VPD}\label{a-training-recipe-for-vpd}

\label{app:recipe}

In this section, we offer practical guidance for applying VPD to other
language models, based on our experience training with the model studied
in this paper, as well as a range of other toy models. See
\autoref{app:training-details} for the hyperparameters used in the
decomposition studied in this paper.

\textbf{Evaluation metrics.}

To assess whether a VPD decomposition has converged to a satisfactory
solution, we recommend tracking the following primary metrics:

\begin{enumerate}
\def\labelenumi{\arabic{enumi}.}
\tightlist
\item
  \textbf{PGD reconstruction loss} (adversarial masks, freshly
  initialized at each step): The most important metric. This evaluates
  reconstruction quality under adversarially chosen masks optimized
  independently for each batch. The setting we want is
  \texttt{shared\_across\_batch}, see \autoref{sec:vpd_methods-adv} for
  why. This is stricter than the persistent adversarial loss used during
  training and is our primary indicator of mechanistic faithfulness. For
  deeper models, more adversarial steps may be needed. As a rough
  heuristic, we keep
  \(n_{\text{adv}} \cdot \text{lr}_{\text{adv}} \approx 2\); if
  increasing the number of steps, decrease the learning rate
  proportionally so the adversarial optimizer can tune more precisely.
  For discussion on how much adversarial optimization exactly our causal
  importances should be robust to, see \autoref{sec:discussion}.
\item
  \textbf{\(L_0\) per data point}: The average number of subcomponents
  with nonzero causal importance on a data point. This should be tracked
  relative to the rank of the original weight matrices. For a
  transformer, MLP matrices typically have rank \(d_{\text{resid}}\);
  the \(L_0\) should be significantly smaller than this for the
  decomposition to be providing a useful simplification. Note that
  \(L_0\) typically starts high and decreases steadily over training due
  to \(p\)-annealing (see below), so early in training the importance
  minimality loss value is a better predictor of what the final \(L_0\)
  will be.
\end{enumerate}

Additionally, we often monitor \textbf{Stochastic reconstruction loss},
because it indicates performance under the average permitted masking as
opposed to worst-case maskings, \textbf{unmasked reconstruction loss}
(all masks set to \(1.0\), excluding the \(\Delta\)-components), because
it indicates the extent to which the sum of all subcomponents is
identical to the target model and \textbf{CI-masked reconstruction loss}
(using the causal importance values directly as masks) as well as
\textbf{Rounded CI-masked reconstruction loss} (as CI-masked but all
causal importance greater than zero are rounded to \(1.0\)) because they
indicate performance when keeping exactly those subcomponents deemed
causally important. Note though that the latter two are only useful
indicators because VPD does not directly optimize for them: It would be
(and in practice is) trivial to achieve almost perfect reconstruction
for these two maskings if we included them in the training loss. But
this would not indicate that our decomposition was actually capturing
more of the target model's computation, because these metrics are not
robust to ``cheating'' in the way the adversarial, and to a lesser
extent stochastic reconstruction losses are.

\textbf{Training Loss terms.}

VPD training uses the following loss terms, each of which requires its
own loss coefficient. We discuss considerations for tuning these below.

\begin{enumerate}
\def\labelenumi{\arabic{enumi}.}
\tightlist
\item
  \textbf{Adversarial reconstruction loss}
  (\(\mathcal{L}_{\text{PPGD recon}}\), coefficient \(0.5\)): This is
  the persistent PGD loss described in \autoref{sec:vpd_methods-adv}.
  Making the adversarial optimizer cheap yet effective is nontrivial.
  The adversarial learning rate usually needs to be tuned and depends on
  the regular learning rate. For the other hyperparameters of the
  adversarial optimizer, we recommend using the defaults described in
  \autoref{app:training-details}: an Adam optimizer with
  \(\beta_1 = 0.5\), \(\beta_2 = 0.99\), constant learning rate with
  short warmup, per-batch-per-position source scope, and updating the
  sources \(n_{\text{adv}}=3\) times for each outer step (in our
  implementation, we do two inner ``warmup'' steps and then apply the
  outer loss step which also updates the sources). For smaller models,
  fewer adversarial steps per training step may suffice; for larger,
  especially deeper, models may need more steps (and a correspondingly
  lower adversarial learning rate). We usually keep this loss
  coefficient fixed to \(0.5\), setting the scale for the other losses.
\item
  \textbf{Stochastic reconstruction loss}
  (\(\mathcal{L}_{\text{stochastic-recon}}\), coefficient \(0.5\)): This
  loss primarily prevents the optimization from stalling early in
  training, and secondarily prevents it from over-focusing on worst-case
  ablations at the expense of average-case reconstruction quality. We
  keep the coefficients for the two reconstruction losses equal and
  normalized to \(\frac{1}{2}\) each. We usually keep this loss
  coefficient fixed to \(0.5\), setting the scale for the other losses.
\item
  \textbf{Importance minimality loss}
  (\(\mathcal{L}_{\text{importance-minimality}}\)): This is typically
  one of the most sensitive hyperparameters and often requires tuning.
  The \(p\)-norm exponent is annealed linearly from \(p_0 = 2.0\) to
  \(p_{\text{final}} = 0.4\) over the full training run. We recommend
  keeping this annealing schedule fixed and tuning the coefficient
  instead. Setting the coefficient too high leads to collapsed
  decompositions with poor reconstruction; too low leads to
  decompositions where too many subcomponents are simultaneously active.
\item
  \textbf{Frequency minimality loss}
  (\(\mathcal{L}_{\text{frequency-minimality}}\)): The coefficient for
  this term also requires some tuning, but interacts with the importance
  minimality coefficient: increasing the frequency minimality loss
  coefficient effectively increases sparsity pressure, so it may be
  necessary to lower the importance minimality loss coefficient to
  compensate. As a starting point, we suggest setting the frequency
  minimality loss coefficient at roughly \(0.5\times\) the importance
  minimality coefficient, unless problems are observed. Too low a
  coefficient tends to produce fewer, overly polysemantic subcomponents.
\item
  \textbf{\(\Delta\)-component L2 penalty}
  (\(\mathcal{L}_{\text{Delta-L2}}\)): This penalizes the MSE between
  the sum of subcomponents and each target weight matrix. In practice,
  this coefficient is not very sensitive. We recommend increasing it by
  factors of \(10\) from a conservative starting point until the
  unmasked reconstruction loss becomes negligibly small. It is safe to
  overshoot the coefficient considerably, though making it excessively
  large can still impair optimization.
\end{enumerate}

\textbf{Subcomponent count \(C\).} The number of subcomponents per
weight matrix is not extremely sensitive. It should be set large enough
for the optimization to capture all the subcomponents that are present.
If unsure, we recommend erring on the side of too many subcomponents,
then inspecting the spectrum of log mean causal importances (averaged
over a batch) at the end of an exploratory run. There is typically a
sharp cutoff in this spectrum separating ``alive'' from ``dead''
subcomponents, which reveals how many subcomponents are actually in
use\footnote{We've found the log mean causal importance spectra much
  more valuable as a measure of the number of ``dead subcomponents''
  compared to counting the number of datapoints on which a subcomponent
  fires at all. There are often some very small firings that aren't
  meaningful, making choosing a cutoff difficult.}. The optimization
tends to work best when \(C\) is larger than needed, but not excessively
so---roughly within a factor of \(2\) of the true number of
subcomponents appears to work well.

\textbf{Causal importance function} For decomposing transformer models,
we recommend using \texttt{global\_shared\_transformer} as the causal
importance function. This is itself a transformer model, which receives
the concatenated hidden activations of the target model as input, and
produces causal importances for all subcomponents as output. We
typically choose the depth of this transformer to be within
\(\frac{1}{2}-2\) times the depth of the target model, though we have
not investigated this hyperparameter as much as some others. We choose
the residual stream to be wider than that of the target model since it
needs to accommodate all of its hidden activations. For this paper, we
used \(2048\) compared to \(768\) for the target model. As is somewhat
standard, we usually choose the MLP width to be approximately four times
the width of the residual stream.

\textbf{Summary} Applying the method to a new model usually requires
adjusting

\begin{enumerate}
\def\labelenumi{\arabic{enumi}.}
\tightlist
\item
  The importance minimality loss coefficient.
\item
  The learning rate
\item
  The adversarial learning rate
\item
  The frequency minimality loss coefficient
\item
  The number of subcomponents \(C\)
\item
  The Delta L2 penalty loss coefficient.
\end{enumerate}

In our experience, the first three typically require the most extensive
tuning. For larger models, the size of the model used for the causal
importance function will likely need to be increased as well. The number
of adversarial steps and the adversarial learning rate may also require
adjustment.

\subsection{Training Details and
Hyperparameters}\label{training-details-and-hyperparameters}

\label{app:training-details}

Below we list the hyperparameters involved in training and decomposing
the 4-layer Pile model.

\textbf{Target model training.}

Target model training artifacts can be found on WandB
(\href{https://wandb.ai/goodfire/spd/runs/t-9d2b8f02/files/final_config.yaml}{config},
\href{https://wandb.ai/goodfire/spd/runs/t-9d2b8f02/files/model_step_99999.pt}{checkpoint},
\href{https://wandb.ai/goodfire/spd/runs/t-9d2b8f02}{run logs}).

The target model architecture is described in
\autoref{sec:language-model-details} and
\autoref{tab:model-hyperparams}. It was trained on a subset of The Pile
\citep{gao2020pile} for \(100,000\) steps with batch size \(1024\) and
context length \(512\). We used Adam \citep{kingma2017adam} with
learning rate \(3 \times 10^{-4}\) (cosine decay to \(10\%\)), weight
decay \(0.1\), gradient clipping at \(1.0\), and \(600\) warmup steps.
Training used \texttt{bfloat16} mixed precision and
\texttt{torch.compile}.

\textbf{VPD training.}

Decomposition artifacts can be found on WandB
(\href{https://wandb.ai/goodfire/spd/runs/s-55ea3f9b/files/final_config.yaml}{config},
\href{https://wandb.ai/goodfire/spd/runs/s-55ea3f9b/files/model_400000.pt}{checkpoint},
\href{https://wandb.ai/goodfire/spd/runs/s-55ea3f9b}{run logs}).

VPD decomposes 24 weight matrices (6 per layer: \texttt{c\_fc},
\texttt{down\_proj}, \texttt{q\_proj}, \texttt{k\_proj},
\texttt{v\_proj}, \texttt{o\_proj}) into subcomponents with
Delta-components enabled. Training ran for \(400,000\) steps with batch
size \(64\) on the same Pile dataset with context length \(512\). The
\(U,V,\) and CI function parameters were jointly optimized with AdamW
(weight decay \(0\)), initial learning rate \(5 \times 10^{-5}\) with
cosine decay to \(10\%\) of the initial value. \(U,V\) gradients were
clipped at norm \(0.01\). One stochastic mask sample (\(S=1\)) was drawn
per step. Faithfulness warmup ran for \(400\) steps (AdamW, lr
\(= 10^{-3}\), weight decay \(0\)), optimizing only the \(U,V\)
parameters against \(\mathcal{L}_{\text{Delta-L2}}\) before the main
training loop. The output divergence measure \(D\) is KL divergence
throughout.

The causal importance function \(\Gamma\) is a shared bidirectional
transformer (architecture described in \autoref{tab:ci-hyperparams}). It
takes RMS-normalized concatenations of all 24 pre-weight activations
(total input dimension \(D = 27,648\)) and outputs
\(C_{\mathrm{total}} = 38,912\) causal importance values via a leaky
hard sigmoid (\(\alpha = 0.01\)).

The \(p\)-norm exponent in both
\(\mathcal{L}_{\mathrm{importance\text{-}minimality}}\) and
\(\mathcal{L}_{\mathrm{frequency\text{-}minimality}}\) is linearly
annealed from \(p_0 = 2.0\) to \(p_{\mathrm{final}} = 0.4\) over the
full training run.

\textbf{Adversarial reconstruction.}

To optimize the persistent sources in the persistent PGD adversarial
loss, an Adam optimizer with \(\beta_1 = 0.5\), \(\beta_2 = 0.99\) and
learning rate \(0.01\) (constant schedule with \(2.5\%\) warmup) was
used. Sources are scoped per batch element per sequence position
(i.e.~each individual batch element and sequence position has its own
source), and each source receives \(2\) warmup PGD steps per training
step before the final loss computation. Stochastic and adversarial
reconstruction losses both use uniform-\(k\)-subset routing, where a
random subset of the 24 weight matrices is masked on each step.

\textbf{Combined minimality loss in code.}

For efficiency, in the training code
\(\mathcal{L}_{\mathrm{importance\text{-}minimality}}\) and
\(\mathcal{L}_{\mathrm{frequency\text{-}minimality}}\) are implemented
as a single fused term per layer, which factors their shared
per-component sum:

\[\mathcal{L}_{\mathrm{minimality}} \;=\; \frac{1}{BT} \sum^{L}_{l=1} \sum^{C}_{c=1} \left[\, s^l_c \;+\; \beta \, s^l_c \, \log_2\!\left(1 + s^l_c\right) \right],
\qquad
s^l_c \;=\; \sum^{B}_{b=1} \sum^{T}_{t=1} \vert g^l_{b,t,c} + \epsilon\vert^{p}.\]

Here \(\beta = 0.5\) is the frequency minimality weight and \(\epsilon\)
is a small constant for numerical stability. This is functionally
equivalent to summing
\(\mathcal{L}_{\mathrm{importance\text{-}minimality}} + \beta \,\mathcal{L}_{\mathrm{frequency\text{-}minimality}}\)
as defined in \autoref{eq:minimal} and \autoref{eq:freq_minimality}; we
fuse them because both terms depend on the same per-component sum
\(s^l_c\), so the fused form avoids recomputing it.

\textbf{Loss terms and coefficients.}

\autoref{tab:vpd-loss-coefficients} lists all loss terms and their
coefficients.

\begin{longtable}[]{@{}
  >{\raggedright\arraybackslash}p{(\linewidth - 4\tabcolsep) * \real{0.3333}}
  >{\raggedright\arraybackslash}p{(\linewidth - 4\tabcolsep) * \real{0.3333}}
  >{\raggedright\arraybackslash}p{(\linewidth - 4\tabcolsep) * \real{0.3333}}@{}}
\caption{\label{tab:vpd-loss-coefficients} \emph{VPD loss terms and
their coefficients. The importance minimality loss uses \(p\)-annealing
from \(2.0\) to \(0.4\). In practice we implement
\(\mathcal{L}_{\mathrm{importance\text{-}minimality}}\) and
\(\mathcal{L}_{\mathrm{frequency\text{-}minimality}}\) as a single fused
term with an inner weight \(\beta = 0.5\) on the frequency part (see
above); this is functionally equivalent to the two losses with the
coefficients shown. All reconstruction losses use KL
divergence.}}\tabularnewline
\toprule\noalign{}
\begin{minipage}[b]{\linewidth}\raggedright
\textbf{Loss term}
\end{minipage} & \begin{minipage}[b]{\linewidth}\raggedright
\textbf{Reference}
\end{minipage} & \begin{minipage}[b]{\linewidth}\raggedright
\textbf{Coefficient}
\end{minipage} \\
\midrule\noalign{}
\endfirsthead
\toprule\noalign{}
\begin{minipage}[b]{\linewidth}\raggedright
\textbf{Loss term}
\end{minipage} & \begin{minipage}[b]{\linewidth}\raggedright
\textbf{Reference}
\end{minipage} & \begin{minipage}[b]{\linewidth}\raggedright
\textbf{Coefficient}
\end{minipage} \\
\midrule\noalign{}
\endhead
\bottomrule\noalign{}
\endlastfoot
\(\mathcal{L}_{\text{Delta-L2}}\) (auxiliary loss; a.k.a.
parameter-faithfulness) & \autoref{eq:delta_l2} & \(10^{7}\) \\
\(\mathcal{L}_{\mathrm{stochastic\text{-}recon\text{-}subset}}\)
(stochastic KL) & \autoref{eq:random_recon} & \(0.5\) \\
\(\mathcal{L}_{\mathrm{adversarial\text{-}recon\text{-}subset}}\)
(persistent PGD KL) & \autoref{eq:adv_recon} & \(0.5\) \\
\(\mathcal{L}_{\mathrm{importance\text{-}minimality}}\) (\(\ell_p\) on
CI values) & \autoref{eq:minimal} & \(2 \times 10^{-4}\) \\
\(\mathcal{L}_{\mathrm{frequency\text{-}minimality}}\) (superlinear CI
frequency penalty) & \autoref{eq:freq_minimality} &
\(1 \times 10^{-4}\) \\
\end{longtable}

\textbf{Subcomponent counts.}

\autoref{tab:vpd-subcomponent-counts} lists the number of subcomponents
\(C\) we give to each module at initialization.

\begin{longtable}[]{@{}ll@{}}
\caption{\label{tab:vpd-subcomponent-counts} \emph{Number of
subcomponents per module type at initialization.}}\tabularnewline
\toprule\noalign{}
\textbf{Module type} & \textbf{Subcomponents (\(C\)) per layer} \\
\midrule\noalign{}
\endfirsthead
\toprule\noalign{}
\textbf{Module type} & \textbf{Subcomponents (\(C\)) per layer} \\
\midrule\noalign{}
\endhead
\bottomrule\noalign{}
\endlastfoot
\texttt{c\_fc} (MLP Up-projection, \(768 \times 3072\)) & 3072 \\
\texttt{down\_proj} (MLP Down-projection, \(3072 \times 768\)) & 3584 \\
\texttt{q\_proj} (query projection, \(768 \times 768\)) & 512 \\
\texttt{k\_proj} (key projection, \(768 \times 768\)) & 512 \\
\texttt{v\_proj} (value projection, \(768 \times 768\)) & 1024 \\
\texttt{o\_proj} (output projection, \(768 \times 768\)) & 1024 \\
\textbf{Total per layer} & 9728 \\
\textbf{Total (4 layers)} & 38912 \\
\end{longtable}

The CI function architecture is shown in \autoref{tab:ci-hyperparams}.

The training and evaluation losses achieved by the primary training run
studied in this paper are listed in \autoref{tab:vpd-eval-losses} and
\autoref{tab:vpd-train-losses} respectively.

\begin{longtable}[]{@{}ll@{}}
\caption{\label{tab:vpd-train-losses} \emph{Training losses (Measured at
final step).}}\tabularnewline
\toprule\noalign{}
Loss & Value \\
\midrule\noalign{}
\endfirsthead
\toprule\noalign{}
Loss & Value \\
\midrule\noalign{}
\endhead
\bottomrule\noalign{}
\endlastfoot
Total & \(24.62\) \\
Delta-L2 (MSE) & \(0.00000240\) \\
StochasticReconSubsetLoss (KL) & \(0.2419\) \\
PersistentPGDReconLoss (KL) & \(0.5733\) \\
ImportanceMinimalityLoss & \(1102.0\) \\
\end{longtable}

\begin{longtable}[]{@{}ll@{}}
\caption{\label{tab:vpd-eval-losses} \emph{Evaluation reconstruction
losses.}}\tabularnewline
\toprule\noalign{}
Loss & Value \\
\midrule\noalign{}
\endfirsthead
\toprule\noalign{}
Loss & Value \\
\midrule\noalign{}
\endhead
\bottomrule\noalign{}
\endlastfoot
StochasticReconSubsetLoss (KL) & \(0.2381\) \\
PGDReconLoss (KL) & \(0.9268\) \\
StochasticHiddenActsReconLoss (MSE) & \(0.4130\) \\
CIHiddenActsReconLoss (MSE) & \(0.8464\) \\
\end{longtable}

\textbf{CLT/PLT WandB links}

\begin{longtable}[]{@{}
  >{\raggedright\arraybackslash}p{(\linewidth - 4\tabcolsep) * \real{0.3333}}
  >{\raggedright\arraybackslash}p{(\linewidth - 4\tabcolsep) * \real{0.3333}}
  >{\raggedright\arraybackslash}p{(\linewidth - 4\tabcolsep) * \real{0.3333}}@{}}
\toprule\noalign{}
\begin{minipage}[b]{\linewidth}\raggedright
Used in
\end{minipage} & \begin{minipage}[b]{\linewidth}\raggedright
What
\end{minipage} & \begin{minipage}[b]{\linewidth}\raggedright
WandB link
\end{minipage} \\
\midrule\noalign{}
\endhead
\bottomrule\noalign{}
\endlastfoot
\autoref{fig:pareto-mse}; \autoref{fig:splitting-heatmap} & PLT/CLT,
local-MSE objective, \(k \in \{8, 16, 32, 64\}\) &
\href{https://wandb.ai/mats-sprint/pile_local_sweep_jose}{dict\_4k},
\href{https://wandb.ai/mats-sprint/pile_local_sweep_jose_32k}{dict\_32k} \\
\autoref{fig:pareto-e2e} & PLT/CLT, end-to-end KL objective,
\(k \in \{8, 16, 32, 64\}\), three training modes (\texttt{cascading} =
error-propagating, \texttt{parallel} = clean-input, \texttt{independent}
= single-layer) &
\href{https://wandb.ai/mats-sprint/pile_e2e_sweep_jose}{dict\_4k},
\href{https://wandb.ai/mats-sprint/pile_e2e_sweep_jose_32k}{dict\_32k} \\
\autoref{tab:seed-mmcs} --- PLT/CLT seed runs & 5 seeds \(\times\) \{PLT
local-MSE, PLT e2e-independent, CLT local-MSE, CLT e2e-parallel\},
\(k = 16\), 4k dict &
\href{https://wandb.ai/mats-sprint/pile_multiseed_jose2}{multiseed} \\
\autoref{tab:seed-mmcs} --- VPD seed runs & 5 VPD seed runs (otherwise
identical to the main decomposition) &
\href{https://wandb.ai/goodfire/spd?nw=n9l0amrrudc}{VPD multiseed} \\
\autoref{tab:seed-mmcs} --- hidden-activation aux-loss VPD & VPD trained
with an auxiliary stochastic-forward-pass hidden-activation MSE loss &
\href{https://wandb.ai/goodfire/spd/runs/s-aa4fec0a}{VPD hidden-act
run} \\
\autoref{fig:feature_splitting}; \autoref{fig:splitting-heatmap} & VPD
capacity sweep (\(0.5\times\), \(1\times\), \(2\times\), \(4\times\)
subcomponents); \(1\times\) is the main run above &
\href{https://wandb.ai/goodfire/spd/workspace?nw=ckmtpmd21yl}{capacity\_sweep} \\
\autoref{fig:adv-vs-no-adv} & No-adversarial-loss control run; otherwise
identical training configuration to the main decomposition &
\href{https://wandb.ai/goodfire/spd/runs/s-05ef623e}{VPD
no-adversarial-loss run} \\
\end{longtable}

All activation-based comparison runs target the same
\href{https://wandb.ai/goodfire/spd/runs/t-9d2b8f02}{t-9d2b8f02} model
and use \(\text{LR} = 3 \times 10^{-4}\), batch size \(4096\), sequence
length \(512\), \(500\)M tokens of the Pile, with BatchTopK activation.

\textbf{SimpleStories decomposition}

We also trained and decomposed a 2-layer model with the same
architecture as above on the SimpleStories dataset
\citep{finke2025parameterizedsynthetictextgeneration}. A wandb link for
that decomposition is
\href{https://wandb.ai/goodfire/spd/runs/s-eab2ace8}{here}. We found
that the attribution graphs and components were cleaner here, though are
of course restricted to a much narrower data distribution.

\subsection{Clustering Subcomponents into
Components}\label{clustering-subcomponents-into-components}

\label{app:clustering}

VPD decomposes each weight matrix \(W_l\) into a sum of rank-one
\emph{subcomponents}:
\(W_l \approx \sum_{c} \vec{U}^l_c (\vec{V}^l_c)^\top\). While each
subcomponent only spans a single weight matrix, a full \emph{parameter
component} could span the entire parameter space, potentially involving
subcomponents from multiple weight matrices. We therefore need a method
to identify which subcomponents across different weight matrices should
be grouped together into coherent parameter components.

\subsubsection{Minimum Description Length
Clustering}\label{minimum-description-length-clustering}

We frame the clustering problem using the \emph{Minimum Description
Length} (MDL) principle, which states that the shortest description of
the data is the best one. The goal is to find a single grouping of
subcomponents that minimizes the total cost of describing the causally
important components both for each data point in isolation and for a
whole dataset.

Consider a partition of \(n\) subcomponents into \(k\) groups
\(\{\theta_1, \ldots, \theta_k\}\) forming candidate components. For
each component \(\theta_i\), we define its causal importance
\(g_{b,t,i} \in \{0, 1\}\) at sequence position \(t\) on batch index
\(b\) to be the OR of the causal importances of all subcomponents in
that group.\footnote{This is a somewhat conservative assumption. If
  \(n\) individual parameter vectors are ablatable in any combination,
  their sum is guaranteed to also be ablatable.} We use a threshold
\(\tau\) on the causal importances of individual subcomponents to
guarantee they are binary, rounding values \(\leq \tau\) to \(0\) and
values \(>\tau\) to \(1\) (we use \(\tau = 0.01\) by default).

Given a batch of \(B\) sequences of length \(T\), we compute a
\emph{coactivation matrix} that measures how often pairs of components
are causally important together:

\[s_{i,j} := \sum^B_{b=1} \sum^T_{t=1} g_{b,t,i}g_{b,t,j}\]

The diagonal values \(s_i := s_{i,i}\) are the summed causal importances
for component \(i\) across batch and sequence.

Our MDL cost for the current grouping is then:

\[\mathcal{L}_{\text{MDL}}
 = \sum_{i=1}^{k} s_i \left( \log_2(k)
 + \alpha \cdot r(\theta_i) \right)\]

where \(r(\theta_i) = |\theta_i|\) denotes the \emph{rank} of component
\(\theta_i\) summed across its constituent matrices.

This cost has an intuitive interpretation: each time a group is causally
important, we must encode the components for:

\begin{enumerate}
\def\labelenumi{\arabic{enumi}.}
\tightlist
\item
  A counterparty to whom we are transmitting a whole batch or dataset of
  mechanistic descriptions. In this case, we transmit the full
  dictionary of components first\footnote{This is a one-time cost that
    becomes negligible for sufficiently large datasets, so we do not
    include it in \(\mathcal{L}_{\text{MDL}}\).}, then transmit
  \(\sum_{i=1}^{k} s_i \log_2(k)\) bits on each data point to single out
  the indices of the causally important components.
\item
  Counterparties to whom we are transmitting the mechanistic description
  of each single datapoint in isolation. In that case, we transmit the
  raw floating point values of the rank-one matrices comprising causally
  important components, so our description length will be proportional
  to \(\sum_{i=1}^{k} s_i r(\theta_i)\).
\end{enumerate}

The hyperparameter \(\alpha\) controls how much we care about the
average description length of the matrices we need to inspect to
understand how the target model computes its output on any one data
point, which matters to us because we assume that causal graphs with
longer description lengths tend to be harder for us to understand. On
the other hand, the \(\log_2(k)\) quantifies the description length of
the sets of components involved in calculating the model output for each
input across a whole dataset. We care about this description length
because we assume that if the same set of components is used on
different data points, it will be easier for us to unify and generalise
our separate explanations of the model's behavior on many different
inputs into a single explanation of the model's behavior on all those
inputs.

\subsubsection{Stochastic Hierarchical
Merging}\label{stochastic-hierarchical-merging}

We use a stochastic hierarchical clustering algorithm that starts with
each subcomponent in its own group and iteratively merges pairs to
reduce the MDL cost. At each iteration, we compute the \emph{merge cost}
for combining groups \(\theta_i\) and \(\theta_j\). Let
\(s_\Sigma = \sum_i s_i\) be the total activation count. Then the change
in MDL cost from merging is:

\[\begin{aligned}
\Delta\mathcal{L}(\theta_i, \theta_j)
&= \underbrace{
(s_\Sigma - s_i - s_j) \log_2 \frac{k-1}{k}
}_{\text{dictionary reduction}} \\
&+ \underbrace{
s_{i,j} \log_2(k-1) - s_i \log_2 (k) - s_j \log_2 (k)
}_{\text{index encoding}} \\
&+ \underbrace{
\alpha \left( s_{i,j} \cdot r(\theta_{i,j}) - s_i \cdot r(\theta_i) - s_j \cdot r(\theta_j) \right)
}_{\text{rank penalty}}
\end{aligned}\]

where \(r(\theta_{i,j})\) is the rank of the merged group summed across
its constituent matrices. For simplicity, we approximate that
\(r(\theta_{i,j}) \approx r(\theta_i) + r(\theta_j)\).

Naively, one might greedily select the pair
\((i^*, j^*) = \arg\min_{i < j} \Delta\mathcal{L}(\theta_i, \theta_j)\)
and merge them, but this risks getting stuck in local minima. To allow
for more exploration of the space of possible clusterings, we use
stochastic selection: instead of always choosing the minimum-cost pair,
we sample from all pairs using a probability distribution that
exponentially decays with higher cost. Specifically, we rank all
candidate merge pairs by their cost \(\Delta\mathcal{L}\) in ascending
order and assign each pair a probability that decays exponentially in
its rank:\\
\[P \propto \exp(-\gamma \cdot J), \quad J = 0, 1, \ldots, \tbinom{k}{2} - 1\]

where \(J = 0\) corresponds to the lowest-cost pair and \(\gamma > 0\)
is a decay rate controlling exploration. Setting \(\gamma \to \infty\)
recovers greedy selection, while \(\gamma \to 0\) gives uniform
sampling. We sample efficiently via the inverse CDF: letting
\(N = \binom{k}{2}\) be the number of candidate pairs and
\(u \sim \text{Uniform}(0,1)\), the sampled rank is\\
\[J = \left\lfloor \frac{-\log\bigl(1 - u(1 - e^{-\gamma N})\bigr)}{\gamma} \right\rfloor.\]

where \(\lfloor \dots \rfloor\) is the floor function rounding down to
the nearest integer. In our experiments, we use \(\gamma = 0.2\), which
concentrates most probability mass on the top few candidates while
maintaining meaningful probability on roughly the five lowest-cost
merges. This stochastic selection allows the algorithm to escape local
minima that greedy merging would get trapped in, while still strongly
preferring merges that reduce the MDL cost.

We run the hierarchical clustering algorithm until all subcomponents
have been merged into a single component. Then, we find the iteration at
which the marginal change in description length from merging
\(\Delta\mathcal{L}\) crossed \(\Delta\mathcal{L}=0\), and use the
clusters at that iteration as our components.

\subsubsection{Choosing alpha}\label{choosing-alpha}

As an intuition pump for choosing the \(\alpha\) hyperparameter in
practice, consider two rank-1 components \(\theta_1, \theta_2\) with
causal importances that are exactly zero or one on all data points,
where component \(\theta_2\) is causally important with some probability
conditional on \(\theta_1\) being causally important:
\(\mathrm{co}(\theta_2\mid \theta_1):=\Pr(\theta_2\text{ important}\mid \theta_1\text{ important})\).
If the total dictionary size is large enough that we can approximate
\(\log_2(k-1)\approx \log_2(k)\), and the summed causal importances
across the batch and sequence for both components are equal
(i.e.~\(s_1=s_2\)), the mdl loss will be lowered by merging these two
components into one if \[\alpha
< \frac{\mathrm{co}(\theta_2\mid \theta_1)}{1-\mathrm{co}(\theta_2\mid \theta_1)}\cdot \frac{\log_2(k)}{2}.\]

\subsection{Automated subcomponent
labeling}\label{automated-subcomponent-labeling}

\label{app:autointerp}

We describe the procedure here used to produce the subcomponent labels.

\textbf{Evidence collection.} For each subcomponent, we run a
forward-pass harvest over the training distribution and record (i) the
firing density of the subcomponent, (ii) a random sample of
\emph{activating examples} (token windows in which the subcomponent's
causal importance exceeds a threshold \(\tau_{\text{ci}} = 0.1\)), and
(iii) per-token co-occurrence statistics for \emph{input} tokens (the
token at the firing position) and \emph{output} tokens (the model's
next-token distribution at the firing position). Each activating example
is a window of \(20\) tokens of context on either side of the firing
position (\(41\) tokens total), truncated at sequence boundaries. We
sample up to \(30\) activating examples per subcomponent.

\textbf{Prompt.} For each subcomponent we construct a single prompt
containing:

\begin{enumerate}
\def\labelenumi{\arabic{enumi}.}
\tightlist
\item
  A short description of VPD
\item
  A note distinguishing \emph{causal importance} (CI, the mask value,
  which is what we care about) from \emph{inner activation} (the dot
  product of the input with the read direction \(V\), scaled by the
  write-direction norm). The prompt instructs the labeler to weight CI
  heavily and treat low-CI/high-act positions as background.
\item
  A note on the sign convention: negating both \(\vec{u_c}\) and
  \(\vec{v_c}\) leaves the rank-1 matrix unchanged, so the absolute sign
  of the inner activation is arbitrary, but the \emph{relative} sign
  within a subcomponent is meaningful (positive- and negative-act
  clusters can correspond to two distinct roles).
\item
  The subcomponent's layer (``MLP up-projection in the 2nd of 4
  blocks'', etc.) and firing density.
\item
  The top-recall and top-PMI tables for output and input tokens.
\item
  The \(30\) activating examples of the form
  \texttt{{[}{[}token\ (ci:X,\ act:Y){]},\ ...{]}} (consecutive firings
  are grouped). Annotations are shown only for firing positions, since
  CI and act are noisy on non-firing tokens.
\item
  The task: return an \(8\) word \texttt{label} and a short
  \texttt{reasoning} summary.
\end{enumerate}

\textbf{Model and decoding.} All labels were produced by
\texttt{google/gemini-3.1-pro-preview} \citep{deepmind2025gemini3} at
\texttt{reasoning\_effort=medium}, accessed through the OpenRouter API.

\textbf{Coverage.} We labeled the \(10{,}000\) subcomponents with the
highest firing density on the harvested distribution. This covers all
subcomponents that fire often enough for the harvested sample of
activating examples to be informative; subcomponents below this cutoff
fire too rarely to support reliable labeling.

\section{Results}\label{results}

\label{sec:app:results}

\subsection{End-to-end transcoders}\label{end-to-end-transcoders}

\label{app:vpd-sparsity-acc-tradeoff}

In \autoref{sec:decomp-model-behav-sim}, we showed that VPD
Pareto-dominates MSE-trained PLTs and CLTs under all three sparsity
measures (\autoref{fig:pareto-mse}). However, that advantage may
partially reflect a difference in training signal, since VPD optimizes
end-to-end on the output distribution while the transcoders optimize
layer-wise MSE. Here we control for this by training all
activation-based methods with the same end-to-end KL-divergence
objective as VPD.

\textbf{Training and evaluation protocols}

\label{app:mode-mismatch}

When we replace all MLP layers simultaneously, there is an important
design choice: should each layer's encoder see the \emph{clean} residual
stream (as computed by the original model) or the \emph{modified}
residual stream (which includes reconstruction errors from earlier
layers)? We call these the \textbf{\emph{clean-input}} and
\textbf{\emph{error-propagating}} evaluation protocols, respectively. A
third option, \textbf{\emph{single-layer}}, replaces only one MLP at a
time, with all other layers left unmodified. For a perfectly
mechanistically faithful reconstruction --- one that exactly replicates
each MLP's computation --- these three protocols would produce similar
results.

\begin{figure}[htbp]
\centering
\includegraphics[width=\textwidth,height=0.8\textheight,keepaspectratio]{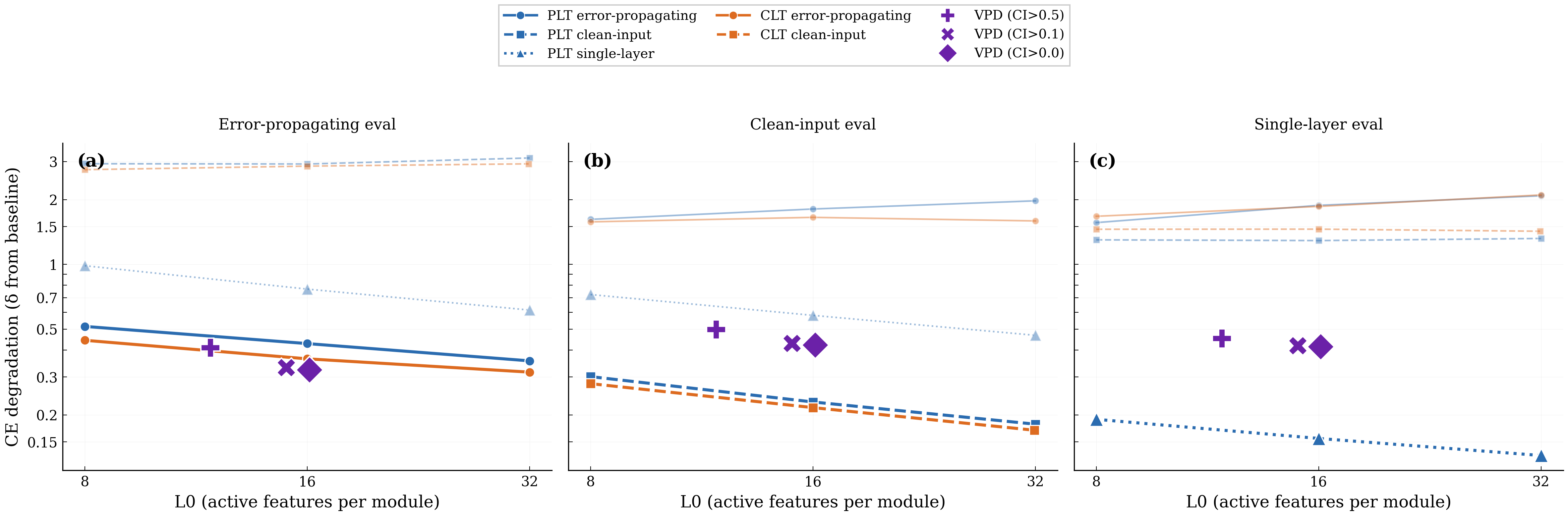}
\caption[]{CE degradation vs. L0 (active features per module) for end-to-end KL-trained methods under three evaluation protocols. \textbf{(a)} Error-propagating: each encoder sees the modified residual stream. \textbf{(b)} Clean-input: each encoder sees the clean residual stream. \textbf{(c)} Single-layer replacement, averaged over layers. PLTs (blue) and CLTs (orange) perform well in their training mode but degrade by 5-20x in the mismatched mode. VPD (purple markers) is relatively stable across all three protocols. Linestyle indicates training mode: solid = error-propagating, dashed = clean-input, dotted = single-layer.}\label{fig:pareto-e2e}
\end{figure}

We train separate sweeps of BatchTopK PLTs and CLTs
(\(k \in \{8, 16, 32\}\)) in clean-input and error-propagating mode, as
well as single-layer-trained PLTs. All use KL divergence on the output
logits as the training loss, matching VPD. Each model is then evaluated
under all three protocols. \autoref{fig:pareto-e2e} shows the results.

\textbf{Activation-based methods are brittle to mode mismatch.}

The activation-based methods exhibit severe brittleness to evaluation
mode mismatch. In the matched setting, error-propagating-trained PLTs
achieve CE degradation as low as \(\delta = 0.32\), and
clean-input-trained PLTs reach \(\delta = 0.18\) at \(k=32\)
(\autoref{fig:pareto-e2e}b). But when evaluated in the \emph{mismatched}
setting, these same models degrade catastrophically: clean-input-trained
models evaluated in error-propagating mode suffer
\(\delta \approx 2.9\)---\(3.5\), roughly an order of magnitude worse.
The pattern is symmetric: error-propagating-trained models fail in
clean-input evaluation (\(\delta \approx 1.6\)---\(2.2\)). CLTs exhibit
the same pattern. The gap between matched and mismatched performance is
a factor of \(3\)---\(20\times\).

This brittleness reveals that e.g.~a PLT trained in error-propagating
mode does not simply learn to approximate each MLP's input-output
function. Instead, it learns a replacement model that \emph{jointly}
accounts for both the MLP's true computation and the systematic
reconstruction errors introduced by the PLTs in earlier layers. This is
a compensatory strategy rather than a mechanistically faithful
approximation of the original target model.

Single-layer-trained PLTs, which each see only the clean residual stream
for their own layer, are the most robust of the activation-based
methods, and perform best in the single-layer replacement setting
(\(\delta \approx 0.13\)---\(0.19\)). However, when all four
single-layer-trained PLTs are inserted simultaneously, they still
exhibit meaningful degradation (\(\delta \approx 0.56\)---\(0.99\)),
because each was trained in isolation and cannot account for
reconstruction errors accumulating from other layers.

\textbf{VPD is stable across protocols.}

VPD's CE degradation, by contrast, is relatively consistent across all
three evaluation protocols. At CI(\textgreater)0, VPD achieves
\(\delta \approx 0.32\)--\(0.42\) regardless of whether it is evaluated
in error-propagating, clean-input, or single-layer mode. This arises
because VPD's stochastic and adversarial masking during training already
exposes the decomposition to a rich diversity of partial ablation
patterns: on each training step, a random subset of subcomponents across
random subsets of weight matrices are partially masked, which naturally
covers patterns resembling both error-propagating and clean-input
replacement as special cases. More fundamentally, VPD's subcomponents
sum to the original weight matrices, and the masked forward pass uses
the same architecture and nonlinearities as the target model. A VPD
reconstruction is therefore not a different function approximating the
MLP, but rather a subset of the MLP's computations.

That said, VPD does not achieve the lowest CE degradation in every
individual setting. In matched-mode evaluation, the best
activation-based models outperform it (e.g., clean-input PLTs at
\(k=16\) reach \(\delta \approx 0.23\) vs.~VPD's
\(\delta \approx 0.42\)). We view this as the expected cost of
faithfulness: a model specifically optimized to compensate for a
particular error pattern will naturally outperform one that has not
learned such compensation.

\subsection{Confirming feature splitting in PLTs and CLTs
geometrically}\label{confirming-feature-splitting-in-plts-and-clts-geometrically}

\label{app:confirming-feature-splitting}

To investigate whether the PLTs and CLTs are indeed splitting features
rather than discovering genuinely new ones, we match features between
models of different sizes. For each pair of models, we count what
fraction of alive objects, latents for PLTs/CLTs and subcomponents for
VPD, in one model have more than one match among the alive subcomponents
of the other model, averaged across layers. We match objects by
calculating the cosine similarity between their output vectors (decoder
vector for PLT/CLT; down-projection \(\vec{U}\) vector for VPD) and
consider a cosine similarity \(> 0.5\) a match. Results are
qualitatively stable across cosine similarity thresholds in
\([0.3, 0.7]\). An object with multiple matches in a target model is
evidence that the target model has split what the source model
represents as a single feature.

\begin{figure}[htbp]
\centering
\includegraphics[width=0.85\textwidth,height=0.8\textheight,keepaspectratio]{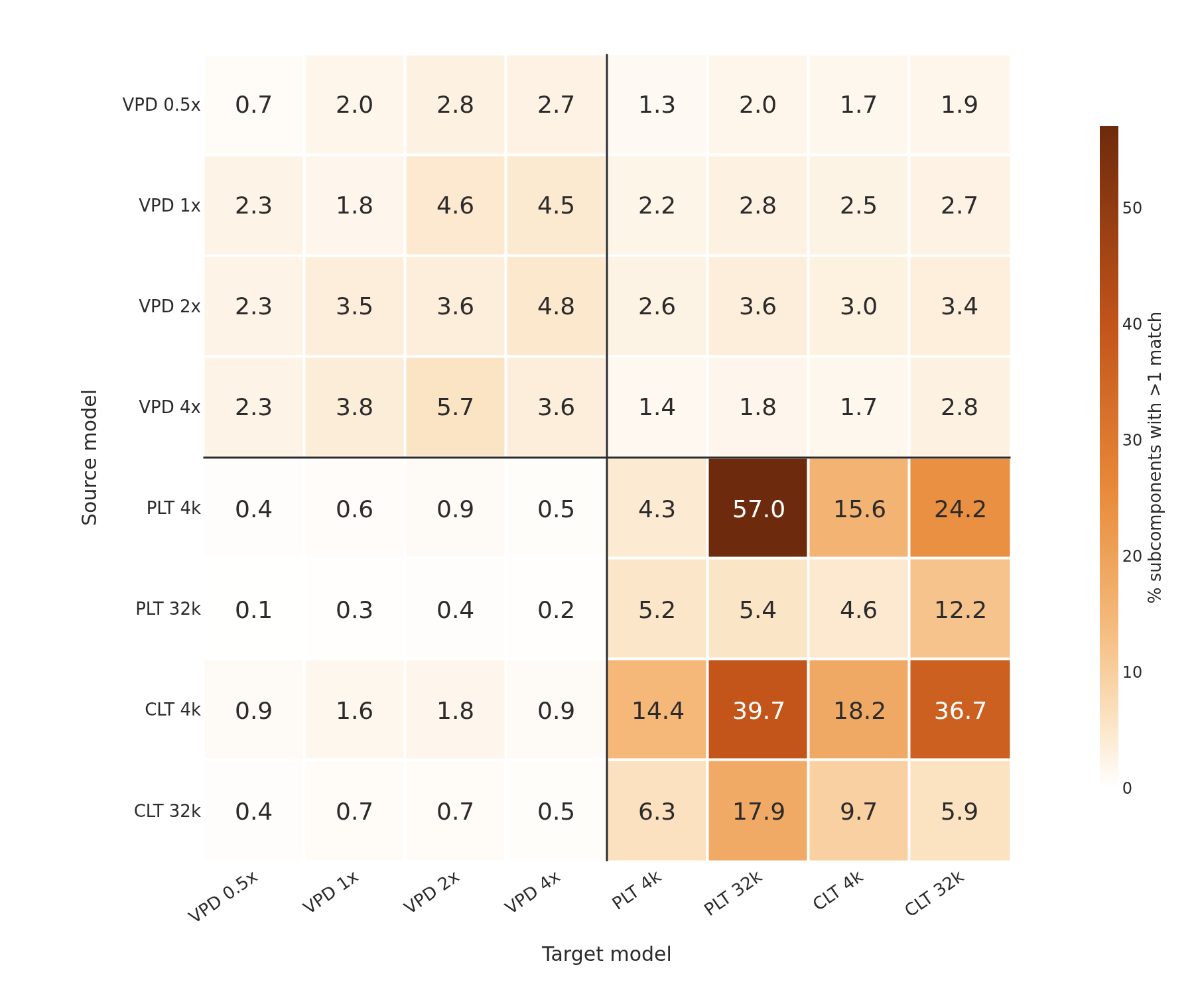}
\caption[]{Cross-model latent/subcomponent splitting (decoder vector cosine similarity \(> 0.5\)). Each cell shows the percentage of alive subcomponents or latents in the source model (row) that have more than one match in the target model (column). VPD shows low cosine similarity across models. PLTs and CLTs show high mutual cosine similarity, suggesting substantial splitting.}\label{fig:splitting-heatmap}
\end{figure}

The heatmap confirms that the proportion of features that have multiple
decoder matches in a version with more subcomponents is higher in PLTs
and CLTs. For example, 57.0\% of the components in 4k PLT have more than
one match in the 32k PLT. On the other hand, only 2.7\% of the
subcomponents in the 0.5x VPD model have more than one match in 4x VPD
model.

\subsection{Geometric consistency across
seeds}\label{geometric-consistency-across-seeds}

\label{app:seed-stability}

In mechanistic interpretability, it is common to look for the
`mechanisms' or `features' that a network uses in its computations.
There's an implication here: That there is a fixed, ground truth set of
objects that we're looking for (``\emph{the} mechanisms''). How true is
this? And how would we measure how close we are to finding the right
objects?

One approach is to run a decomposition method with different random
seeds or using different hyperparameters. If the approaches converge to
the same results despite these differences, this is suggestive that they
converged to the `right' set of objects.

Previous work has used mean max cosine similarity (MMCS) to measure this
similarity quantitatively \citep{Sharkey_Braun_Millidge_2022}. Suppose
we perform two decompositions using the same method, but with different
random seeds. Given these two sets of transcoder latents or VPD
subcomponents, we calculate the cosine similarity between the objects in
each set, and find the most similar for each, and the take the average
cosine similarity between those maximally similar pairs. High MMCS means
that decompositions are similar across seeds.

Since VPD is trained using an end-to-end (e2e) loss we compare it with
transcoders trained with an e2e loss (\autoref{tab:seed-mmcs}). We find
that the MMCS of the transcoder latents is similar or slightly worse
than the MMCS of VPD U and V vectors. But PLTs and CLTs are usually not
trained with an e2e loss; they are usually trained to reconstruct
activations at each layer (i.e.~a `local MSE' loss). VPD does not
typically train with a hidden activation reconstruction loss; if it
reconstructed hidden activations perfectly, it would be constructing
activations that are not relevant for performance and merely correspond
to `superposition noise'. Despite not training on hidden activation
reconstruction loss, the constellation of other loss functions results
in a hidden activation reconstruction loss that is similar, albeit
slightly higher, than if we do minimize it directly (Stochastic forward
pass hidden activation MSE: 0.33 vs.~0.41). When transcoders are trained
using their typical training loss (local MSE), their MMCS are much
better than VPD.

\begin{longtable}[]{@{}ll@{}}
\caption{\label{tab:seed-mmcs}}\tabularnewline
\toprule\noalign{}
Method & Cross-seed Mean Max Cos Sim \\
\midrule\noalign{}
\endfirsthead
\toprule\noalign{}
Method & Cross-seed Mean Max Cos Sim \\
\midrule\noalign{}
\endhead
\bottomrule\noalign{}
\endlastfoot
VPD U vectors & 0.4808 \\
VPD V vectors & 0.5156 \\
PLT (e2e) & 0.4390 \\
CLT (e2e, parallel) & 0.3468 \\
TC (local MSE) & 0.8063 \\
CLT (local MSE) & 0.6078 \\
VPD rank-1 (V@U) & 0.2826 \\
VPD components (rank-N, cross-model) & 0.3181 \\
(Baseline) VPD at init U vectors & 0.1263 \\
(Baseline) VPD at init V vectors & 0.1300 \\
(Baseline) VPD at init rank-1 (V@U) & 0.0122 \\
\end{longtable}

Overall, we're uncertain how much emphasis to put on these similarities.
While it is naturally appealing to think that there is a single
`correct' decomposition, we are not sure that this intuition fully
accounts for the extent of the degeneracy in neural networks. One of the
reasons that neural networks are so good at learning is the sheer amount
of degeneracy they seem to have: It is easier to find a good solution in
a space where there are many good solutions! It seems quite possible
that, even though we place a number of constraints on the solution that
VPD looks for, there is not just one set of ground truth mechanisms, but
in fact an entire space of optimal parameter components that are
nonetheless mechanistically faithful! The same is true of dictionary
learning approaches. While (all else equal) cross seed consistency is a
desirable property of a decomposition method, other properties such as
mechanistic faithfulness are probably closer to what we want our methods
to achieve.

\subsection{Stochastic vs.~adversarial training
loss}\label{stochastic-vs.-adversarial-training-loss}

\label{app:decomp-stats}

\begin{figure}[htbp]
\centering
\includegraphics[width=0.85\textwidth,height=0.8\textheight,keepaspectratio]{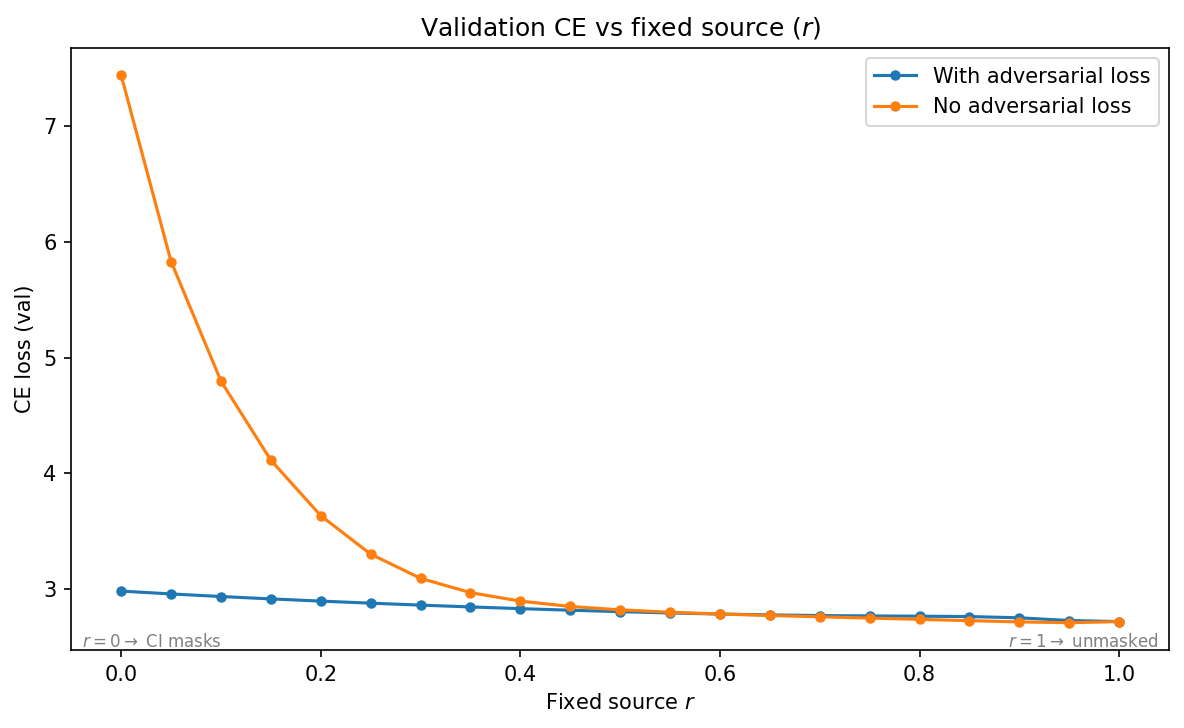}
\caption[]{Comparison between a decomposition with and without adversarial loss. The training configuration is otherwise identical. The CE loss is especially improved for small values of \(r\).}\label{fig:adv-vs-no-adv}
\end{figure}

The adversarial loss greatly improves the decomposition performance for
small source (\(r\)) values (\autoref{fig:adv-vs-no-adv}).

\subsection{OV circuit weight norms}\label{ov-circuit-weight-norms}

Subcomponents in the \(W_V\) and \(W_O\) matrices are spread across
multiple heads, despite having specialized semantic roles. This is
suggestive evidence of attention computations that are distributed
across heads.

\begin{figure}[htbp]
\centering
\includegraphics[width=0.85\textwidth,height=0.8\textheight,keepaspectratio]{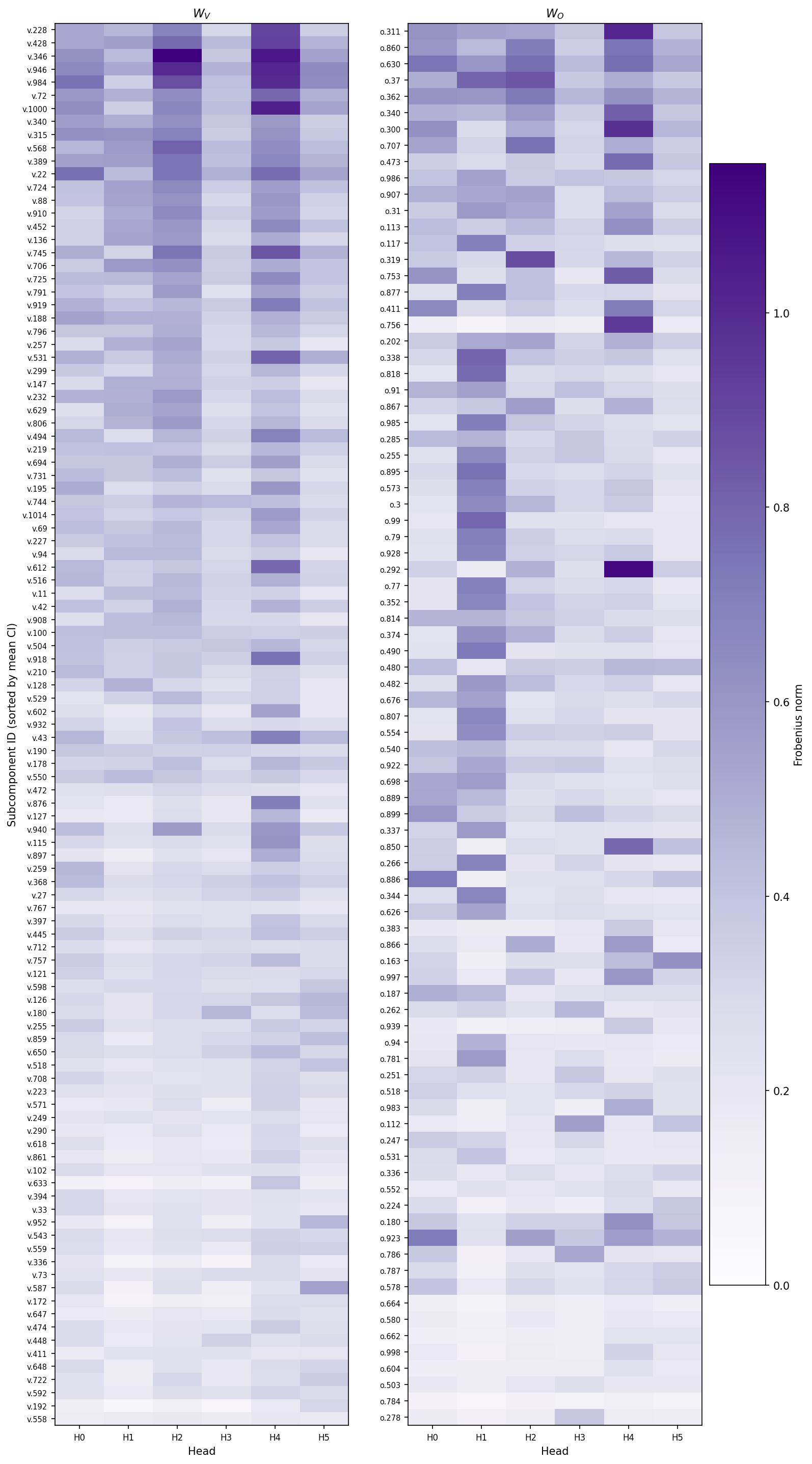}
\caption[]{The norm of the weights of each \(W_V\) and \(W_O\) subcomponent in each head.}\label{fig:vo_comp_weight_norm}
\end{figure}

\subsection{OV circuit metric: Data-weighted Frobenius cosine
similarity}\label{ov-circuit-metric-data-weighted-frobenius-cosine-similarity}

\label{app:OV-metric-data-frob}

To study the OV circuit across multiple heads, it is helpful to think of
\(W_{OV}^h\) in terms of its singular value decomposition:
\(W_{OV}^h = \boldsymbol{L} \boldsymbol{S} \boldsymbol{R}^\top\). Now,
we construct two new matrices for each \(W_{OV}^h\) matrix:

\[M^{\text{read}}_h = ({W_{OV}^h})^\top W_{OV}^h = \boldsymbol{R} \boldsymbol{S}^2 \boldsymbol{R}^\top\]
\[M^{\text{write}}_h = W_{OV}^h ({W_{OV}^h})^\top = \boldsymbol{L} \boldsymbol{S}^2 \boldsymbol{L}^\top\]

We can study how much each head reads and writes to the same subspace by
comparing the similarity between the \(M^{\text{read or write}}_h\)
matrices of different heads. We compare them using a metric called the
\textbf{Frobenius cosine similarity}, which is a cosine similarity
metric for matrices:

\[ S(M_a, M_b) = \frac{\langle M_a, M_b \rangle_F}{\|M_a\|_F \|M_b\|_F} \]

We will also measure the Frobenius cosine similarity between the raw
\(W_{OV}^h\) matrices of each head, since it is possible that even
though matrices might read from and write to similar subspaces, their
singular vectors might be paired differently.

How should we understand this metric? On an intuitive level, we can
think of a given \(W_{OV}^h\) matrix's read- or write-subspace as a
\(d_{\text{head}}\)-dimensional ellipsoid in
\(\mathbb{R}^{d_{\text{model}}}\) space, where the axes of the ellipsoid
are the scaled right or left singular vectors of \(W_{OV}^h\) matrix
respectively. The Frobenius cosine similarity measures how much the
read- or write-ellipsoid of one head overlaps with another head's. If
the ellipsoids perfectly overlap, then the Frobenius cosine similarity
is 1. If they exist in entirely non-overlapping subspaces, then their
Frobenius cosine similarity is 0. For comparison purposes, we'll compare
the Frobenius cosine similarities with a random matrix baseline. This
will help us understand whether the model has learned to use more or
less overlapping subspaces than would be expected for a pair of random
matrices of the same size and dimension.

\textbf{Weighting subspaces by data variation}

However, the raw Frobenius cosine similarity between these matrices may
potentially be misleading. The network does not use every subspace
equally. Some subspaces might not contain much of the activations.
Unless our metric accounts for how much of the activations lie within
the subspaces that the \(W_{OV}\) matrices read from and write to, we
may get a misleading sense of how similar a pair of heads is. We should
therefore weight different dimensions according to the amount of
activation variation that exists along that axis.

To do this, we form the \textbf{data-weighted} value matrix for each
head. For a dataset of activations \(\varphi\), we perform PCA to get
the principal axes of variation \(\bar{\varphi}\):

\[\bar{\varphi} = \varphi - \vec{1} (\vec{\mu})^\top, \qquad \bar{\varphi} = \bar{U} \bar{S} \bar{Z}^\top,\]

\[\bar{U} \in \mathbb{R}^{T \times d_{\text{model}}}, \quad \bar{S} \in \mathbb{R}^{d_{\text{model}} \times d_{\text{model}}}, \quad \bar{Z}^\top \in \mathbb{R}^{d_{\text{model}} \times d_{\text{model}}}\]

where \(\vec{\mu} = \frac{1}{T}\sum^T_{t=1} \vec{\varphi}_t\). We then
project \(W_{OV}^h\) onto the data's principal axes of variation and
scale each axis by the corresponding singular value, yielding the
data-weighted value projection matrix for head \(h\):

\[W_{OV}^{h, \varphi} = W_{OV}^h \bar{Z} \bar{S}.\]

We can now construct data-weighted read and write Gram matrices
(\(M^{\varphi, \text{read}}_h\) and \(M^{\varphi, \text{write}}_h\)) as
described above for the data-\emph{un}weighted case. We can then use the
Frobenius cosine similarity between them to understand how similarly the
OV circuit of each head reads and writes the actual data that it sees.

We can also use this approach to \emph{selectively} study how the OV
circuit interacts with particular QK pairs. If we filter the dataset
such that it contains only datapoints where the associated K
subcomponent is causally important, then we can understand whether those
pairs are moving similar or dissimilar value information in each head!

We should note that the pair of subcomponents involved in previous token
Behavior 1 in \autoref{sec:attn-analysis-1} are almost always active,
and so we don't get to benefit from this QK-based filtering approach.
But for Behavior 2 in \autoref{sec:attn-analysis-1} we study a behavior
that is conditionally active, where it will be beneficial to understand
how similar the OV circuits in each head behave only when that QK
subcomponent interaction is active.

We are now equipped enough to return to our analysis of previous token
behavior and study its OV circuit to establish whether its heads attend
to similar or distinct residual stream subspaces.

\subsection{Expected Frobenius Cosine Similarity of Random Low-Rank Gram
Matrices}\label{expected-frobenius-cosine-similarity-of-random-low-rank-gram-matrices}

\label{app:expected_frob_proof}

In this section, we derive an approximation for the expected Frobenius
cosine similarity between the Gram matrices of two randomly initialized
attention heads. We compare it to an empirical test, and find good
agreement.

\subsubsection{Standard (unweighted) Frobenius cosine
similarity}\label{standard-unweighted-frobenius-cosine-similarity}

Let
\(W_a, W_b \in \mathbb{R}^{d_{\text{head}} \times d_{\text{model}}}\) be
the value projection matrices for two attention heads, where
\(d_{\text{model}}\) is the residual stream dimension
(\(d_{\text{model}}\)) and \(d_{\text{head}}\) is the head dimension. We
initialize the elements of \(W_a\) and \(W_b\) independently from a
standard normal distribution, \(\mathcal{N}(0, 1)\).

We define the corresponding Gram matrices as \(M_a = W_a^\top W_a\) and
\(M_b = W_b^\top W_b\). Notice that while
\(M_a, M_b \in \mathbb{R}^{d_{\text{model}} \times d_{\text{model}}}\),
their rank is bounded by \(d_{\text{head}}\).

Because \(W_a\) has elements drawn from \(\mathcal{N}(0, 1)\), its Gram
matrix \(M_a\) follows a standard Wishart distribution with
\(d_{\text{head}}\) degrees of freedom, denoted as
\(M_a \sim \mathcal{W}_{d_{\text{model}}}(d_{\text{head}}, I_{d_{\text{model}}})\).

The Frobenius cosine similarity between \(M_a\) and \(M_b\) is defined
as:
\[S(M_a, M_b) = \frac{\langle M_a, M_b \rangle_F}{\|M_a\|_F \|M_b\|_F} = \langle \frac{M_a}{\|M_a\|_F}, \frac{M_b}{\|M_b\|_F} \rangle_F\]
Since \(M_a\) and \(M_b\) are independent:
\[\mathbb{E}[S(M_a, M_b)] =\langle \mathbb{E}\big[\frac{M_a}{\|M_a\|_F}\big], \mathbb{E}\big[\frac{M_b}{\|M_b\|_F}\big] \rangle_F=\|\mathbb{E}\big[\frac{M}{\|M\|_F}\big]\|^2_F\]
The Wishart distribution is rotationally invariant, so we know that
\(\mathbb{E}[\frac{M}{\|M\|_F}]\) is a multiple of the identity matrix,
and hence
\[\mathbb{E}[S(M_a, M_b)] =\frac{1}{d_{\text{model}}}\left(\operatorname{tr}(\mathbb{E}[\frac{M}{\|M\|_F}])\right)^2= \frac{1}{d_{\text{model}}}\left(\mathbb{E}\big[\frac{\operatorname{tr}(M)}{\sqrt{\operatorname{tr}(M^2)}}\big]\right)^2\]

Now, we will approximate the numerator \(\operatorname{tr}(M)\) and
denominator \(\sqrt{\operatorname{tr}(M^2)}\) using the fact that traces
of high-dimensional matrices concentrate around their expectation value:

\textbf{Approximating the numerator}

The expectation of a Wishart matrix
\(\mathcal{W}_{d_{\text{model}}}(d_{\text{head}}, I_{d_{\text{model}}})\)
is \(d_{\text{head}} I_{d_{\text{model}}}\), so

\[\begin{aligned}
\operatorname{tr}(M)\approx d_{\text{head}} d_{\text{model}}
\end{aligned}\]

\textbf{Approximating the denominator}

To approximate \(\operatorname{tr}(M^2)=\|M\|_F^2\), we sum the expected
squared values of all elements in the Gram matrix. Let \(M_{ij}\) be the
entry in the \(i\)-th row and \(j\)-th column.
\[\|M\|_F^2 = \sum_{i=1}^{d_{\text{model}}} \sum_{j=1}^{d_{\text{model}}} M_{ij}^2 = \sum_{i=1}^{d_{\text{model}}} M_{ii}^2 + \sum_{i \neq j} M_{ij}^2\]

\begin{enumerate}
\def\labelenumi{\arabic{enumi}.}
\tightlist
\item
  \textbf{Diagonal Elements}: The diagonal elements are
  \(M_{ii} = \sum^{d_{\text{head}}}_{r=1} W_{ri}^2\). Since
  \(W_{ri} \sim \mathcal{N}(0, 1)\), each \(M_{ii}\) follows a
  Chi-squared distribution with \(k\) degrees of freedom
  (\(\chi^2_{d_{\text{head}}}\)). The mean of a
  \(\chi^2_{d_{\text{head}}}\) variable is \(d_{\text{head}}\), and its
  variance is \(2d_{\text{head}}\). Using the identity
  \(\mathbb{E}[X^2] = \operatorname{Var}(X) + (\mathbb{E}[X])^2\), we
  get \(\mathbb{E}[M_{ii}^2] = 2d_{\text{head}} + d^2_{\text{head}}\).
  Since there are \(d_{\text{model}}\) diagonal elements, their total
  contribution is
  \(d_{\text{model}}(d^2_{\text{head}} + 2d_{\text{head}})\).
\item
  \textbf{Off-Diagonal Elements}: For the off-diagonal elements
  \(M_{ij} = \sum^{d_{\text{head}}}_{r=1} W_{ri} W_{rj}\) with
  \(i \neq j\), \(W_{ri}\) and \(W_{rj}\) are independent standard
  normal variables, so their product has a mean of \(0\) and a variance
  of \(1\). The sum of \(d_{\text{head}}\) such independent terms has a
  mean of \(0\) and a variance of \(d_{\text{head}}\), so we get
  \(\mathbb{E}[M_{ij}^2] = \operatorname{Var}(M_{ij}) + (\mathbb{E}[M_{ij}])^2 = d_{\text{head}} + 0 = d_{\text{head}}\).
  There are \(d_{\text{model}}(d_{\text{model}}-1)\) off-diagonal
  elements, so their total contribution is
  \(d_{\text{model}}(d_{\text{model}}-1)d_{\text{head}}\).
\end{enumerate}

Combining the diagonal and off-diagonal contributions yields:
\[\begin{aligned}
\|M\|_F^2 &\approx d_{\text{model}}d_{\text{head}}(d_{\text{model}} + d_{\text{head}} + 1)
\end{aligned}\]

\textbf{Final Expected Baseline}

Substituting the expected inner product and the expected squared norm
back into our similarity approximation:
\[\mathbb{E}[S(M_a, M_b)] \approx \frac{1}{d_{\text{model}}}\frac{d^2_{\text{head}} d^2_{\text{model}}}{d_{\text{model}}d_{\text{head}}(d_{\text{model}} + d_{\text{head}} + 1)} = \frac{d_{\text{head}}}{d_{\text{model}} + d_{\text{head}} + 1}\]

For our specific architecture, the residual stream dimension is
\(d_{\text{model}} = 768\) and the head dimension is
\(d_{\text{head}} = 128\). Plugging these values into the derived
formula gives an approximation of the expected random baseline for the
subspace overlap:
\[\mathbb{E}[S(M_a, M_b)] \approx \frac{128}{768 + 128 + 1} = \frac{128}{897} \approx 0.1427\]

Thus, we have \(\approx 0.1427\) as the expected baseline for the
Frobenius cosine similarity between two randomly initialized heads of
this dimension and rank.

This value exactly matches an empirical random baseline computed via
Monte Carlo simulation:

\subsubsection{Empirical: Standard (unweighted) Frobenius cosine
similarity}\label{empirical-standard-unweighted-frobenius-cosine-similarity}

We generate 1000 pairs of random matrices
\(W_a, W_b \in \mathbb{R}^{d_{\text{head}} \times d_{\text{model}}}\)
with i.i.d. standard normal entries, compute their Gram matrices
\(M = W^\top W\), and calculate the Frobenius cosine similarity
\(\frac{\text{tr}(M_a M_b)}{\lVert M_a \rVert_F \lVert M_b \rVert_F}\)
for each pair. The mean across pairs gives the expected overlap between
matrices with no structural relationship. The empirical result exactly
matched the theoretical result proved above (0.1427).

\subsubsection{Data-weighted Frobenius cosine
similarity}\label{data-weighted-frobenius-cosine-similarity}

For the data-weighted case, there is no analytical solution, so we use
the same Monte Carlo procedure, except that we right-multiply each
random matrix by \(\bar{Z}\bar{S}\) (the right singular vectors scaled
by singular values from the mean-centered data) before computing Gram
matrices. This ensures the baseline reflects the anisotropy of the
residual stream because, in a low-rank data distribution, even unrelated
matrices may exhibit elevated subspace overlap.

\subsection{Layer 1 K and V subcomponent
relations}\label{layer-1-k-and-v-subcomponent-relations}

\begin{figure}[htbp]
\centering
\includegraphics[width=0.85\textwidth,height=0.8\textheight,keepaspectratio]{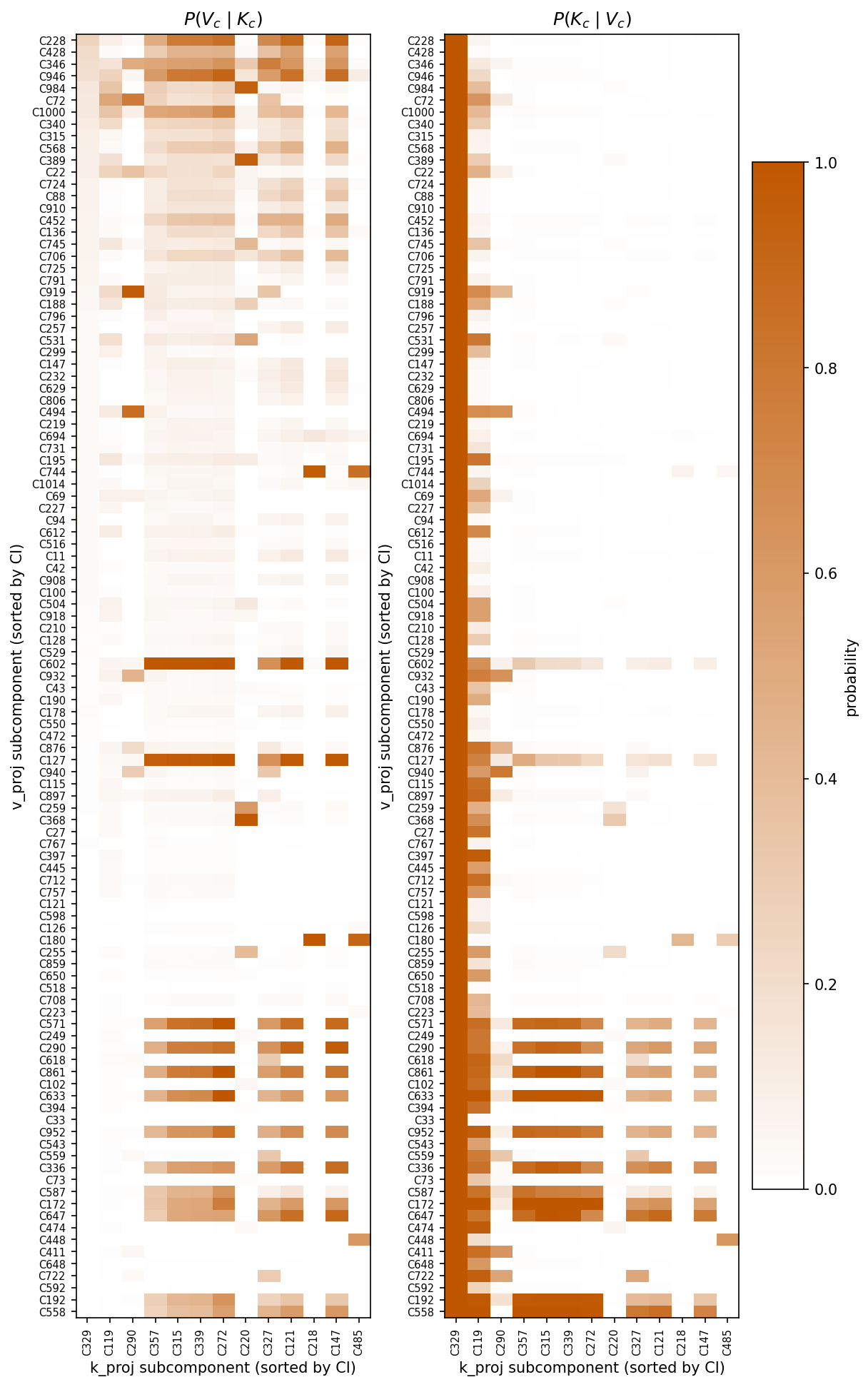}
\caption[]{The probability of each K subcomponent being active when a given V subcomponent is active, and vice versa. This tells us what \(W_K\) subcomponents are responsible for moving information from particular kinds of \(W_V\) subcomponents. The \texttt{1.attn.k:329} subcomponent is always active, and therefore moves all kinds of \(W_V\) subcomponents.}\label{fig:pkv}
\end{figure}

\begin{figure}[htbp]
\centering
\includegraphics[width=\textwidth,height=0.8\textheight,keepaspectratio]{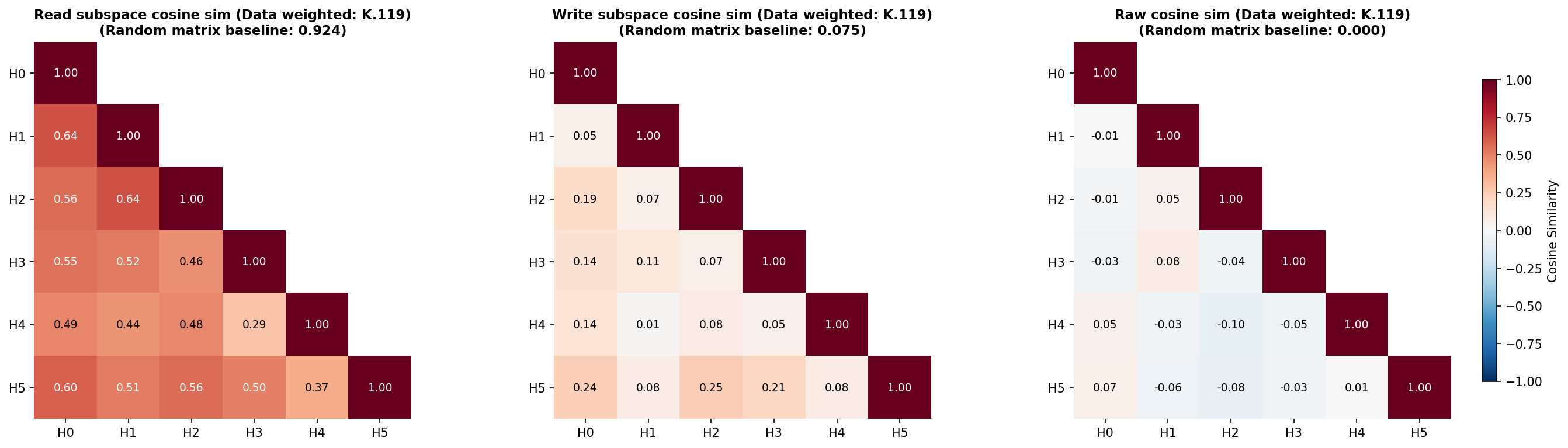}
\caption[]{Data-weighted cosine similarities between each head's \(W_{OV}^h\) read- and write matrices, and the cosine similarity between each head's raw \(W_{OV}^h\). Here, data-weighting uses data where subcomponent \texttt{1.attn.k:119} is causally important.}\label{fig:prev_tok_ov_overlap_k_119}
\end{figure}

\subsection{Layer 1 O, V subcomponents most aligned with attention heads
on data where Layer 1 K.119 is causally
important}\label{layer-1-o-v-subcomponents-most-aligned-with-attention-heads-on-data-where-layer-1-k.119-is-causally-important}

\label{app:ov-alignment-k119}

Here we list, for each attention head, the top-5 V subcomponents
(read-aligned) and top-5 O subcomponents (write-aligned) whose
subcomponents are most aligned with that head's OV circuit.

Here, alignment \(= ||W v_{c}^{\text{scaled}}||\) (read) or
\(||W^\top u_{c}^{\text{scaled}}||\) (write), where vectors are scaled
by the norm of the other factor in the rank-1 decomposition.

Note that this can exclude several subcomponents not in the top-5 that
nonetheless have meaningful alignment.

\paragraph{Head 0}\label{head-0}

\textbf{Read-aligned V subcomponents (top 5)}

\begin{longtable}[]{@{}
  >{\raggedright\arraybackslash}p{(\linewidth - 6\tabcolsep) * \real{0.2000}}
  >{\raggedright\arraybackslash}p{(\linewidth - 6\tabcolsep) * \real{0.2000}}
  >{\raggedright\arraybackslash}p{(\linewidth - 6\tabcolsep) * \real{0.3667}}
  >{\raggedright\arraybackslash}p{(\linewidth - 6\tabcolsep) * \real{0.2333}}@{}}
\toprule\noalign{}
\begin{minipage}[b]{\linewidth}\raggedright
Rank
\end{minipage} & \begin{minipage}[b]{\linewidth}\raggedright
Comp
\end{minipage} & \begin{minipage}[b]{\linewidth}\raggedright
Alignment
\end{minipage} & \begin{minipage}[b]{\linewidth}\raggedright
Label
\end{minipage} \\
\midrule\noalign{}
\endhead
\bottomrule\noalign{}
\endlastfoot
1 & v.22 & 74.9453 & punctuation, syntax, and formatting tokens \\
2 & v.984 & 68.3169 & fires on punctuation and symbols \\
3 & v.1000 & 64.4708 & fires on punctuation, delimiters, and structural
boundaries \\
4 & v.346 & 62.6574 & distinguishes function words (positive) and
content words (negative) \\
5 & v.568 & 59.1917 & fires on word prefixes and partial words \\
\end{longtable}

\textbf{Write-aligned O subcomponents (top 5)}

\begin{longtable}[]{@{}
  >{\raggedright\arraybackslash}p{(\linewidth - 6\tabcolsep) * \real{0.2000}}
  >{\raggedright\arraybackslash}p{(\linewidth - 6\tabcolsep) * \real{0.2000}}
  >{\raggedright\arraybackslash}p{(\linewidth - 6\tabcolsep) * \real{0.3667}}
  >{\raggedright\arraybackslash}p{(\linewidth - 6\tabcolsep) * \real{0.2333}}@{}}
\toprule\noalign{}
\begin{minipage}[b]{\linewidth}\raggedright
Rank
\end{minipage} & \begin{minipage}[b]{\linewidth}\raggedright
Comp
\end{minipage} & \begin{minipage}[b]{\linewidth}\raggedright
Alignment
\end{minipage} & \begin{minipage}[b]{\linewidth}\raggedright
Label
\end{minipage} \\
\midrule\noalign{}
\endhead
\bottomrule\noalign{}
\endlastfoot
1 & o.923 & 231.5484 & first token of the sequence \\
2 & o.411 & 203.4490 & code, markup, and technical formatting syntax \\
3 & o.630 & 191.1107 & punctuation, symbols, and syntax in technical
text \\
4 & o.753 & 181.5993 & closing parentheses and brackets in code and
math \\
5 & o.300 & 173.9110 & code and structured text syntax/indentation \\
\end{longtable}

\paragraph{Head 1}\label{head-1}

\textbf{Read-aligned V subcomponents (top 5)}

\begin{longtable}[]{@{}
  >{\raggedright\arraybackslash}p{(\linewidth - 6\tabcolsep) * \real{0.2000}}
  >{\raggedright\arraybackslash}p{(\linewidth - 6\tabcolsep) * \real{0.2000}}
  >{\raggedright\arraybackslash}p{(\linewidth - 6\tabcolsep) * \real{0.3667}}
  >{\raggedright\arraybackslash}p{(\linewidth - 6\tabcolsep) * \real{0.2333}}@{}}
\toprule\noalign{}
\begin{minipage}[b]{\linewidth}\raggedright
Rank
\end{minipage} & \begin{minipage}[b]{\linewidth}\raggedright
Comp
\end{minipage} & \begin{minipage}[b]{\linewidth}\raggedright
Alignment
\end{minipage} & \begin{minipage}[b]{\linewidth}\raggedright
Label
\end{minipage} \\
\midrule\noalign{}
\endhead
\bottomrule\noalign{}
\endlastfoot
1 & v.984 & 59.9101 & fires on punctuation and symbols \\
2 & v.346 & 56.6177 & distinguishes function words (positive) and
content words (negative) \\
3 & v.22 & 52.5358 & punctuation, syntax, and formatting tokens \\
4 & v.1000 & 52.4579 & fires on punctuation, delimiters, and structural
boundaries \\
5 & v.946 & 45.8153 & distinguishes content words from function
words/symbols \\
\end{longtable}

\textbf{Write-aligned O subcomponents (top 5)}

\begin{longtable}[]{@{}
  >{\raggedright\arraybackslash}p{(\linewidth - 6\tabcolsep) * \real{0.2000}}
  >{\raggedright\arraybackslash}p{(\linewidth - 6\tabcolsep) * \real{0.2000}}
  >{\raggedright\arraybackslash}p{(\linewidth - 6\tabcolsep) * \real{0.3667}}
  >{\raggedright\arraybackslash}p{(\linewidth - 6\tabcolsep) * \real{0.2333}}@{}}
\toprule\noalign{}
\begin{minipage}[b]{\linewidth}\raggedright
Rank
\end{minipage} & \begin{minipage}[b]{\linewidth}\raggedright
Comp
\end{minipage} & \begin{minipage}[b]{\linewidth}\raggedright
Alignment
\end{minipage} & \begin{minipage}[b]{\linewidth}\raggedright
Label
\end{minipage} \\
\midrule\noalign{}
\endhead
\bottomrule\noalign{}
\endlastfoot
1 & o.362 & 167.0992 & fires on names, citations, proper nouns and
formatting tokens \\
2 & o.490 & 164.5913 & line start and indentation tokens \\
3 & o.337 & 149.7576 & fires inside parentheses or mathematical
formulas \\
4 & o.895 & 132.7422 & variables, math symbols, and syntax in technical
text \\
5 & o.986 & 116.9061 & sentence/paragraph boundaries and transition
words \\
\end{longtable}

\paragraph{Head 2}\label{head-2}

\textbf{Read-aligned V subcomponents (top 5)}

\begin{longtable}[]{@{}
  >{\raggedright\arraybackslash}p{(\linewidth - 6\tabcolsep) * \real{0.2000}}
  >{\raggedright\arraybackslash}p{(\linewidth - 6\tabcolsep) * \real{0.2000}}
  >{\raggedright\arraybackslash}p{(\linewidth - 6\tabcolsep) * \real{0.3667}}
  >{\raggedright\arraybackslash}p{(\linewidth - 6\tabcolsep) * \real{0.2333}}@{}}
\toprule\noalign{}
\begin{minipage}[b]{\linewidth}\raggedright
Rank
\end{minipage} & \begin{minipage}[b]{\linewidth}\raggedright
Comp
\end{minipage} & \begin{minipage}[b]{\linewidth}\raggedright
Alignment
\end{minipage} & \begin{minipage}[b]{\linewidth}\raggedright
Label
\end{minipage} \\
\midrule\noalign{}
\endhead
\bottomrule\noalign{}
\endlastfoot
1 & v.984 & 83.9813 & fires on punctuation and symbols \\
2 & v.22 & 79.5923 & punctuation, syntax, and formatting tokens \\
3 & v.346 & 69.2600 & distinguishes function words (positive) and
content words (negative) \\
4 & v.1000 & 68.2968 & fires on punctuation, delimiters, and structural
boundaries \\
5 & v.531 & 66.0043 & opening parentheses, brackets, braces, and
quotes \\
\end{longtable}

\textbf{Write-aligned O subcomponents (top 5)}

\begin{longtable}[]{@{}
  >{\raggedright\arraybackslash}p{(\linewidth - 6\tabcolsep) * \real{0.2000}}
  >{\raggedright\arraybackslash}p{(\linewidth - 6\tabcolsep) * \real{0.2000}}
  >{\raggedright\arraybackslash}p{(\linewidth - 6\tabcolsep) * \real{0.3667}}
  >{\raggedright\arraybackslash}p{(\linewidth - 6\tabcolsep) * \real{0.2333}}@{}}
\toprule\noalign{}
\begin{minipage}[b]{\linewidth}\raggedright
Rank
\end{minipage} & \begin{minipage}[b]{\linewidth}\raggedright
Comp
\end{minipage} & \begin{minipage}[b]{\linewidth}\raggedright
Alignment
\end{minipage} & \begin{minipage}[b]{\linewidth}\raggedright
Label
\end{minipage} \\
\midrule\noalign{}
\endhead
\bottomrule\noalign{}
\endlastfoot
1 & o.923 & 432.9877 & first token of the sequence \\
2 & o.578 & 249.4896 & heterogeneous subcomponent / lack of clear
pattern \\
3 & o.180 & 227.2081 & diffuse firing on tokens within words/phrases \\
4 & o.866 & 208.1991 & predicts newlines and separators at line ends \\
5 & o.336 & 205.5378 & diverges between function words and complex
technical terms \\
\end{longtable}

\paragraph{Head 3}\label{head-3}

\textbf{Read-aligned V subcomponents (top 5)}

\begin{longtable}[]{@{}
  >{\raggedright\arraybackslash}p{(\linewidth - 6\tabcolsep) * \real{0.2000}}
  >{\raggedright\arraybackslash}p{(\linewidth - 6\tabcolsep) * \real{0.2000}}
  >{\raggedright\arraybackslash}p{(\linewidth - 6\tabcolsep) * \real{0.3667}}
  >{\raggedright\arraybackslash}p{(\linewidth - 6\tabcolsep) * \real{0.2333}}@{}}
\toprule\noalign{}
\begin{minipage}[b]{\linewidth}\raggedright
Rank
\end{minipage} & \begin{minipage}[b]{\linewidth}\raggedright
Comp
\end{minipage} & \begin{minipage}[b]{\linewidth}\raggedright
Alignment
\end{minipage} & \begin{minipage}[b]{\linewidth}\raggedright
Label
\end{minipage} \\
\midrule\noalign{}
\endhead
\bottomrule\noalign{}
\endlastfoot
1 & v.984 & 70.1569 & fires on punctuation and symbols \\
2 & v.745 & 57.7404 & formatting symbols, operators, and spatial
alignment \\
3 & v.946 & 53.8508 & distinguishes content words from function
words/symbols \\
4 & v.494 & 53.3743 & predicts line breaks or indentation in formatted
text \\
5 & v.346 & 52.9415 & distinguishes function words (positive) and
content words (negative) \\
\end{longtable}

\textbf{Write-aligned O subcomponents (top 5)}

\begin{longtable}[]{@{}
  >{\raggedright\arraybackslash}p{(\linewidth - 6\tabcolsep) * \real{0.2000}}
  >{\raggedright\arraybackslash}p{(\linewidth - 6\tabcolsep) * \real{0.2000}}
  >{\raggedright\arraybackslash}p{(\linewidth - 6\tabcolsep) * \real{0.3667}}
  >{\raggedright\arraybackslash}p{(\linewidth - 6\tabcolsep) * \real{0.2333}}@{}}
\toprule\noalign{}
\begin{minipage}[b]{\linewidth}\raggedright
Rank
\end{minipage} & \begin{minipage}[b]{\linewidth}\raggedright
Comp
\end{minipage} & \begin{minipage}[b]{\linewidth}\raggedright
Alignment
\end{minipage} & \begin{minipage}[b]{\linewidth}\raggedright
Label
\end{minipage} \\
\midrule\noalign{}
\endhead
\bottomrule\noalign{}
\endlastfoot
1 & o.311 & 333.3696 & fires universally on most tokens \\
2 & o.630 & 263.6867 & punctuation, symbols, and syntax in technical
text \\
3 & o.37 & 257.2931 & continuations of multi-token entities and compound
words \\
4 & o.300 & 251.2589 & code and structured text syntax/indentation \\
5 & o.180 & 177.4386 & diffuse firing on tokens within words/phrases \\
\end{longtable}

\paragraph{Head 4}\label{head-4}

\textbf{Read-aligned V subcomponents (top 5)}

\begin{longtable}[]{@{}
  >{\raggedright\arraybackslash}p{(\linewidth - 6\tabcolsep) * \real{0.2000}}
  >{\raggedright\arraybackslash}p{(\linewidth - 6\tabcolsep) * \real{0.2000}}
  >{\raggedright\arraybackslash}p{(\linewidth - 6\tabcolsep) * \real{0.3667}}
  >{\raggedright\arraybackslash}p{(\linewidth - 6\tabcolsep) * \real{0.2333}}@{}}
\toprule\noalign{}
\begin{minipage}[b]{\linewidth}\raggedright
Rank
\end{minipage} & \begin{minipage}[b]{\linewidth}\raggedright
Comp
\end{minipage} & \begin{minipage}[b]{\linewidth}\raggedright
Alignment
\end{minipage} & \begin{minipage}[b]{\linewidth}\raggedright
Label
\end{minipage} \\
\midrule\noalign{}
\endhead
\bottomrule\noalign{}
\endlastfoot
1 & v.346 & 146.1802 & distinguishes function words (positive) and
content words (negative) \\
2 & v.22 & 145.5668 & punctuation, syntax, and formatting tokens \\
3 & v.984 & 142.4020 & fires on punctuation and symbols \\
4 & v.745 & 129.9418 & formatting symbols, operators, and spatial
alignment \\
5 & v.1000 & 116.1570 & fires on punctuation, delimiters, and structural
boundaries \\
\end{longtable}

\textbf{Write-aligned O subcomponents (top 5)}

\begin{longtable}[]{@{}
  >{\raggedright\arraybackslash}p{(\linewidth - 6\tabcolsep) * \real{0.2000}}
  >{\raggedright\arraybackslash}p{(\linewidth - 6\tabcolsep) * \real{0.2000}}
  >{\raggedright\arraybackslash}p{(\linewidth - 6\tabcolsep) * \real{0.3667}}
  >{\raggedright\arraybackslash}p{(\linewidth - 6\tabcolsep) * \real{0.2333}}@{}}
\toprule\noalign{}
\begin{minipage}[b]{\linewidth}\raggedright
Rank
\end{minipage} & \begin{minipage}[b]{\linewidth}\raggedright
Comp
\end{minipage} & \begin{minipage}[b]{\linewidth}\raggedright
Alignment
\end{minipage} & \begin{minipage}[b]{\linewidth}\raggedright
Label
\end{minipage} \\
\midrule\noalign{}
\endhead
\bottomrule\noalign{}
\endlastfoot
1 & o.753 & 832.6620 & closing parentheses and brackets in code and
math \\
2 & o.630 & 591.0163 & punctuation, symbols, and syntax in technical
text \\
3 & o.411 & 538.0296 & code, markup, and technical formatting syntax \\
4 & o.860 & 522.9536 & structural and formatting markers vs content
words \\
5 & o.292 & 503.4616 & fires broadly on various tokens, promoting line
breaks and punctuation \\
\end{longtable}

\paragraph{Head 5}\label{head-5}

\textbf{Read-aligned V subcomponents (top 5)}

\begin{longtable}[]{@{}
  >{\raggedright\arraybackslash}p{(\linewidth - 6\tabcolsep) * \real{0.2000}}
  >{\raggedright\arraybackslash}p{(\linewidth - 6\tabcolsep) * \real{0.2000}}
  >{\raggedright\arraybackslash}p{(\linewidth - 6\tabcolsep) * \real{0.3667}}
  >{\raggedright\arraybackslash}p{(\linewidth - 6\tabcolsep) * \real{0.2333}}@{}}
\toprule\noalign{}
\begin{minipage}[b]{\linewidth}\raggedright
Rank
\end{minipage} & \begin{minipage}[b]{\linewidth}\raggedright
Comp
\end{minipage} & \begin{minipage}[b]{\linewidth}\raggedright
Alignment
\end{minipage} & \begin{minipage}[b]{\linewidth}\raggedright
Label
\end{minipage} \\
\midrule\noalign{}
\endhead
\bottomrule\noalign{}
\endlastfoot
1 & v.1000 & 57.2500 & fires on punctuation, delimiters, and structural
boundaries \\
2 & v.946 & 56.3107 & distinguishes content words from function
words/symbols \\
3 & v.346 & 53.8432 & distinguishes function words (positive) and
content words (negative) \\
4 & v.428 & 51.4916 & fragments of proper nouns, foreign, and technical
words \\
5 & v.984 & 50.2268 & fires on punctuation and symbols \\
\end{longtable}

\textbf{Write-aligned O subcomponents (top 5)}

\begin{longtable}[]{@{}
  >{\raggedright\arraybackslash}p{(\linewidth - 6\tabcolsep) * \real{0.2000}}
  >{\raggedright\arraybackslash}p{(\linewidth - 6\tabcolsep) * \real{0.2000}}
  >{\raggedright\arraybackslash}p{(\linewidth - 6\tabcolsep) * \real{0.3667}}
  >{\raggedright\arraybackslash}p{(\linewidth - 6\tabcolsep) * \real{0.2333}}@{}}
\toprule\noalign{}
\begin{minipage}[b]{\linewidth}\raggedright
Rank
\end{minipage} & \begin{minipage}[b]{\linewidth}\raggedright
Comp
\end{minipage} & \begin{minipage}[b]{\linewidth}\raggedright
Alignment
\end{minipage} & \begin{minipage}[b]{\linewidth}\raggedright
Label
\end{minipage} \\
\midrule\noalign{}
\endhead
\bottomrule\noalign{}
\endlastfoot
1 & o.923 & 227.1903 & first token of the sequence \\
2 & o.311 & 178.6259 & fires universally on most tokens \\
3 & o.630 & 160.5296 & punctuation, symbols, and syntax in technical
text \\
4 & o.180 & 156.3924 & diffuse firing on tokens within words/phrases \\
5 & o.300 & 144.0436 & code and structured text syntax/indentation \\
\end{longtable}

\subsection{Interaction graphs}\label{interaction-graphs}

\subsubsection{Gradient attributions}\label{gradient-attributions}

\label{app:gradient_attributions}

To understand how subcomponents interact with each other during the
forward pass, we compute gradient attributions between pairs of
subcomponents at adjacent layers in the computational graph. These
attributions form the edges of an interaction graph that visualizes the
flow of information through the decomposed model on a given prompt or
aggregated over the dataset.

Recall that each subcomponent \(c\) at weight matrix \(l\) has a
\emph{subcomponent activation}
\(a^l_{b,t,c} = (\vec{V}^l_c)^\top \vec{\varphi}^l_{b,t}\), where
\(\vec{\varphi}^l_{b,t}\) is the pre-weight hidden activation vector at
weight matrix \(l\) on batch element \(b\) at sequence position \(t\).
This is the projection of the input onto the right singular vector of
the subcomponent, and it determines how strongly the subcomponent
contributes to the layer's output.

For a source subcomponent \(c_1\) at layer \(l_1\) and a target
subcomponent \(c_2\) at layer \(l_2\) (where \(l_1\) feeds into \(l_2\)
in the computational graph), we define the \emph{gradient attribution}
on batch element \(b\) and at source sequence position \(t_1\) and
target sequence position \(t_2\) as:

\[\alpha(c_1 \to c_2; b, t_1, t_2) = \frac{\partial a^{l_2}_{b, t_2, c_2}}{\partial a^{l_1}_{b,t_1, c_1}} \cdot a^{l_1}_{b,t_1, c_1} \cdot g^{l_1}_{b,t_1,c_1}\]

where \(g^{l_1}_{b,t_1,c_1}\) is the causal importance of the source
subcomponent. The gradient \(\times\) activation product
\(\frac{\partial a^{l_2}_{b,t_2, c_2}}{\partial a^{l_1}_{b,t_1, c_1}} \cdot a^{l_1}_{b,t_1, c_1}\)
gives a first-order estimate of how much the source subcomponent's
activation contributes to the target subcomponent's activation.
Weighting by the causal importance \(g^{l_1}_{b,t_1,c_1}\) ensures that
subcomponents which are not causally important on a given datapoint
(i.e., those that can be ablated without affecting the output) do not
contribute to the attribution, even if they happen to have nonzero
activations and gradients.

For most adjacent layer pairs, source and target positions coincide
(\(t_1 = t_2\)), since MLP and attention projection layers operate
position-wise. However, for edges from key or value subcomponents to
attention output subcomponents within the same attention layer, the
source position \(t_1\) can be any position up to and including the
target position \(t_2\) (i.e., \(t_1 \leq t_2\), respecting the causal
attention mask). This reflects the fact that key and value activations
at earlier positions influence the attention output at later positions.

\textbf{Dataset-aggregated attributions.}

To obtain a summary of how subcomponents interact across the dataset, we
aggregate attributions over all datapoints and all valid position pairs:

\[A(c_1 \to c_2) = \sum^B_{b=1} \sum_{t_1, t_2} \frac{\partial
 a^{l_2}_{b,t_2,c_2}}{\partial a^{l_1}_{b,t_1,c_1}} \cdot a^{l_1}_{b,t_1,c_1} \cdot
 g^{l_1}_{b,t_1,c_1}\]

where the sum over \((t_1, t_2)\) ranges over \(t_1 = t_2\) for
position-wise layers and \(t_1 \leq t_2\) for key/value-to-output edges
in attention. In practice, we compute this sum over the training dataset
using a distributed pipeline across multiple GPUs. To make attributions
comparable across subcomponents with different activation scales and
different frequencies of causal importance, we normalize by the total
causal importance of the source and the root-mean-square activation of
the target:

\[\hat{A}(c_1 \to c_2) = \frac{A(c_1 \to c_2)}{\left(\sum_{b,t_1} g^{l_1}_{b,t_1,c_1}\right) \cdot
 \text{RMS}(a^{l_2}_{c_2})}\]

where
\(\text{RMS}(a^{l_2}_{c_2}) = \sqrt{\frac{1}{BT} \sum_{b,t} (a^{l_2}_{b,t,c_2})^2}\)
and \(BT\) is the total number of tokens processed. Dividing by the
source's cumulative causal importance puts the attribution on a
per-occurrence scale (analogous to averaging over only the datapoints
where the source is active), while dividing by the target's RMS
activation accounts for the target's overall magnitude. Together, these
normalizations allow meaningful comparison of attribution strengths
across edges in the graph.

We also compute an absolute-value variant
\(A_{\text{abs}}(c_1 \to c_2)\), which replaces the target activation
\(a^{l_2}_{c_2}\) with its absolute value \(|a^{l_2}_{c_2}|\) in the
backward pass. This variant captures the total magnitude of influence
irrespective of sign, and is useful for identifying strong interactions
where the signed attribution may cancel across datapoints.

\textbf{Prompt-level attributions.}

For analyzing individual prompts, we compute position-aware attributions
without aggregation. Given a prompt and a set of ``alive'' subcomponents
(those with nonzero causal importance at each position), we compute the
gradient attribution for each pair of alive source and target
subcomponents at each valid combination of source and target positions.
The resulting position-aware graph enables detailed analysis of how the
model processes a specific input.

The main changes are: separate \(t_s\) and \(t_t\) indices throughout,
an explicit paragraph explaining when they differ (K/V to O edges within
an attention layer), and the dataset sum now ranges over valid position
pairs rather than a single shared position.

\subsubsection{Pruning for specific behaviors: Post-hoc causal
importance
optimization}\label{pruning-for-specific-behaviors-post-hoc-causal-importance-optimization}

\label{app:posthoc_ci}

During VPD base training, the causal importance function \(\Gamma\) is
trained to predict which subcomponents are necessary to reconstruct the
target model's \emph{full output distribution across all sequence
positions}. However, when analyzing a specific behavior---such as the
model's prediction of a particular token at a particular position---many
causally important subcomponents will be irrelevant to that specific
behavior, even though they are necessary for reconstructing the full
output. To isolate only the subcomponents involved in a behavior of
interest, we optimize new causal importance values \emph{post hoc} on a
single prompt, using a reconstruction loss that targets only the
specific aspect of the output we wish to study.

\paragraph{Setup}\label{setup}

Given a trained VPD model and a prompt, we first run the model's trained
causal importance function to obtain the base causal importance values
\(g^l_{t,c}\) for all subcomponents on that prompt. We then identify the
set of \emph{alive} subcomponents for the prompt at each sequence
position \(t\): those for which \(g^l_{t,c} > 0\). Only alive
subcomponent causal importances are eligible for inclusion in the
post-hoc optimization, though masks for the other subcomponents (and the
\(\Delta\) components) can still be sampled stochastically to try to
ensure they remain ablatable.

We parameterize the new causal importances using pre-sigmoid parameters
\(\phi^l_{t,c}\), one per alive subcomponent at each position. The
causal importance values are obtained by passing these parameters
through the same lower-leaky (for sampling the masks used in the forward
pass) and upper leaky (for the
\(\mathcal{L}_{\text{importance-minimality}}\) and
\(\mathcal{L}_{\text{frequency-minimality}}\)) hard sigmoid functions
\(\sigma_{H,\text{lower}}\), \(\sigma_{H,\text{upper}}\) (see
\autoref{sec:vpd_ci_function}) used during base training. The parameters
\(\phi^l_{t,c}\) are initialized to the pre-sigmoid values produced by
the base causal importance function on this prompt, providing a warm
start. Non-alive subcomponents have their causal importance fixed at
zero throughout optimization.

\paragraph{Loss function}\label{loss-function}

The post-hoc optimization minimizes a combination of a reconstruction
loss and an importance minimality loss:

\[\mathcal{L}_{\text{post-hoc}} =\lambda_{\text{recon}} \cdot\mathcal{L}_{\text{recon}} + \lambda_{\text{min}} \cdot\mathcal{L}_{\text{importance-minimality}}\]

The reconstruction loss \(\mathcal{L}_{\text{recon}}\) is chosen to
target the specific behavior of interest. For example, to study how the
model predicts token \(y\) at position \(t^*\), we use a cross-entropy
loss at that position:

\[\mathcal{L}_{\text{recon}} = -\log p_{\text{masked}}(y \mid \vec{x}, t^*)\]

where \(p_{\text{masked}}\) denotes the output distribution of the model
with masks applied according to the post-hoc causal importances.
Alternatively, if we wish to reconstruct the model's full output
distribution at a specific position rather than targeting a particular
token, we can use a KL-divergence loss:

\[\mathcal{L}_{\text{recon}} = D_{\text{KL}}\!\left(p_{\text{target}}(\cdot \mid \vec{x}, t^*) \;\|\;
p_{\text{masked}}(\cdot \mid \vec{x}, t^*)\right)\]

The importance minimality loss
\(\mathcal{L}_{\text{importance-minimality}}\) has the same form as in
base training (\autoref{eq:minimal} and \autoref{eq:freq_minimality}),
applied to the post-hoc causal importances \(\tilde{g}^l_{t,c}\). This
loss encourages the optimization to find the sparsest set of
subcomponents that can still reconstruct the targeted behavior. The
coefficient \(\lambda_{\text{min}}\) controls the sparsity--fidelity
trade-off: larger values yield sparser graphs with fewer active
subcomponents, potentially at the cost of reconstruction quality.

\paragraph{Masking during
optimization}\label{masking-during-optimization}

As in base training, the post-hoc causal importances define masks on the
subcomponents via:

\[m^l_{t,c}(r) = \tilde{g}^l_{t,c} + (1 - \tilde{g}^l_{t,c}) r^l_{t,c}\]

where \(r^l_{t,c} \in [0,1]\). On each optimization step, we sample
masks by drawing \(r^l_{t,c}\), either stochastically uniformly or
adversarially, and compute the reconstruction loss under those masks.
This is done to try to ensure that the optimization satisfies the same
mechanistic faithfulness criterion as base training: subcomponents
marked as unimportant must be ablatable in any combination without
affecting the targeted output.\footnote{Or at least in almost any
  combination, see \autoref{sec:vpd_recon_motivation} for discussion.}

\[\mathcal{L}_{\text{post-hoc}} = \lambda_{\text{recon}} \cdot \mathcal{L}_{\text{recon}}
+\lambda_{\text{min}} \cdot \mathcal{L}_{\text{importance-minimality}}
+ \lambda_{\text{recon}} \cdot \mathcal{L}_{\text{adversarial-recon}}\]

where \(\mathcal{L}_{\text{adversarial-recon}}\) is computed similarly
to \autoref{eq:adv_recon}, but using the post-hoc causal importances and
the targeted reconstruction loss. There is also one additional
constraint imposed on the adversarial sampler compared to VPD base
training: Only alive subcomponents on the prompt have their masks
adversarially optimized, other subcomponents have their masks drawn
stochastically. This is because we want to prevent the adversary from
finetuning on data dependent noise inside the many inactive
subcomponents of the model, see \autoref{sec:vpd_methods-adv}. In base
training, this is accomplished by forcing the adversary to use the same
\(r^l_c\) for many data points. For post-hoc optimization we cannot do
this, because we only have a single prompt available. But the causal
importance function has already pre-filtered the subcomponents to
exclude those that were not involved in computing the prompt at all, so
we attempt to sidestep this issue by restricting the adversary to
subcomponents that were alive on the original prompt.

\paragraph{Optimization procedure}\label{optimization-procedure}

We optimize the pre-sigmoid parameters \(\phi^l_{t,c}\) using AdamW with
a cosine learning rate schedule and brief linear warmup. The model
weights and subcomponent parameters (\(U\), \(V\)) are frozen
throughout; only the post-hoc causal importance parameters are updated.
The optimization typically converges within a few hundred steps, since
it starts from a good initialization and optimizes over a single prompt
rather than a dataset. The result is a set of refined causal importance
values \(\tilde{g}^l_{t,c}\) that are sparser than the base values: many
subcomponents that were causally important for the full output are
driven to zero importance when only a specific behavior is targeted. The
surviving subcomponents---those with \(\tilde{g}^l_{t,c} > 0\)---form
the nodes of the interaction graph for that behavior, and gradient
attributions (\autoref{app:gradient_attributions}) are then computed
between them.

\subsection{Nonlinear parameter subcomponent
interactions}\label{nonlinear-parameter-subcomponent-interactions}

\label{app:interactions-gis-vs-coact}

In our case studies, we traced the relationships between subcomponent
activations in particular computations using attributions. However, this
is not a complete account of how the model computes its outputs.
Attributions only measure how strongly one subcomponent activation
influences another; they do not describe the actual functional
relationship between them. To fully reverse engineer neural networks
with VPD, we will need some account of how downstream subcomponent
activations are actually computed from upstream ones.

For some matrices (such as MLP Up, query, key, and value projection
matrices), this should not be difficult. The connections to their
preceding subcomponent activations are linear (apart from the RMS
norms), so they can be understood almost entirely as linear combinations
of preceding subcomponent activations. The RMS norms are non-linear, but
not very functionally expressive, so their role in the computation can
hopefully be understood without too much trouble.

For MLP Down projection and attention output matrices, however,
functionally rich nonlinearities in the computational graph separate
them from preceding subcomponent activations: Neurons in the case of MLP
Down projections, and attention heads in the case of attention output
matrices. For MLP Down projection subcomponents in particular, every
subcomponent activation is a linear combination of many MLP neuron
activations, each of which is potentially a nonlinear function of all
MLP Up matrix subcomponent activations.

One might therefore worry that the nonlinear interactions between MLP Up
matrix subcomponent activations could be inherently very complicated. We
cannot exclude this possibility at present, but there are some
theoretical and empirical reasons to think that these interactions may
be much simpler than the raw number of nonlinearities might suggest.

\subsubsection{Theoretical argument}\label{theoretical-argument}

To the extent that the network implements different circuits in the same
MLP---such as a lookup for which city the Eiffel Tower is in and a
modular addition algorithm for the months of the year---it is actively
incentivised to avoid nonlinear interactions between them. Otherwise,
the circuits would interfere with each other, potentially producing
wrong results. So, to the extent that two subcomponents parametrize two
unrelated circuits, they should not interact much. From our
interpretations of the subcomponents in
\autoref{sec:param-comps-interpretable}, it appears that many of them
are quite specialized to very different contexts, and thus presumably
would not interact substantially. There are components consisting of
clusters of related subcomponents, such as those for bracket closing
from our analysis in \autoref{sec:case-studies-bracket}, and these
presumably could interact quite a bit. But smaller blocks of mutual
interaction would still be much easier to analyze than a single block of
all subcomponents interacting with all other subcomponents. There also
appear to be some subcomponents, such as those that are
nearly-always-active, that would presumably interact nonlinearly with
almost everything else. But characterizing these interactions for a
reasonably small number of subcomponents still seems quite feasible.

\subsubsection{Preliminary empirical
investigation}\label{preliminary-empirical-investigation}

We can approximately measure the interaction strength between MLP Up
projection matrix subcomponents at neurons. Specifically, we compute
interaction matrices \(I_{c,c'}\) that crudely measure two things:

\begin{enumerate}
\def\labelenumi{\arabic{enumi}.}
\tightlist
\item
  \textbf{Weight overlap}: How strongly different subcomponents
  \(c, c'\) connect to the same neurons with sizeable weights, and
\item
  \textbf{Activation overlap}: How often they are causally important
  with large activations at the same batch and sequence index:
\end{enumerate}

\begin{equation}
I^l_{c,c'}
  :=
    \frac{
      \left(
        \sum_{i}
        \vert U^l_{i,c}\vert
        \vert U^l_{i,c'}\vert
      \right)
    }{
      \left(
        \sum_i \vert U^l_{i,c}\vert^2
      \right)
    }
    \frac{
      \left(
        \sum_{b,t}
        \vert g^l_{b,t,c} a^l_{b,t,c}\vert
        \vert g^l_{b,t,c'} a^l_{b,t,c'}\vert
      \right)
    }{
      \left(
        \sum_{b,t}\vert g^l_{b,t,c} a^l_{b,t,c}\vert^2
      \right)
    }
\end{equation}

Intuitively, \(I^l_{c,c'}\) measures how often subcomponents \(c\) and
\(c'\) make a large contribution to the preactivations of the same MLP
neurons \(i\) at the same batch and sequence indices \(b, t\). If the
contribution of subcomponent \(c\) to the preactivation of neuron \(i\)
is much larger in magnitude than the corresponding contribution of
subcomponent \(c'\)
(i.e.~\(U^l_{i,c} g^l_{b,t,c} a^l_{b,t,c} \gg U^l_{i,c'} g^l_{b,t,c'} a^l_{b,t,c'}\)),
then \(c'\) will be mostly irrelevant for determining the nonlinear
response of neuron \(i\) on that data point.

The matrix is normalised such that all diagonal entries equal \(1.0\),
and each row can be read as estimating the interaction strength between
subcomponent \(c\) and other subcomponents \(c'\), relative to the
self-interaction of \(c\). If an off-diagonal entry \(I_{c,c'}\) is much
smaller than \(1.0\), this indicates that subcomponent \(c'\) does not
substantially interact with subcomponent \(c\) at the neurons.
\footnote{Note that \(I\) is not symmetric: \(I_{c,c'}\) and
  \(I_{c',c}\) can differ. This is intentional. If subcomponent \(c\)
  influences a neuron's preactivation much more strongly than
  subcomponent \(c'\), the computational pathway of the latter is likely
  heavily influenced by the former, but not vice versa.}

We expect some off-diagonal entries to be large. For example, for
subcomponents that form part of the same component. But generally
speaking, the fewer large off-diagonal entries there are, the easier it
should be to describe the computation in an MLP Up terms of components
without considering many inter-component interactions.

\begin{figure}[htbp]
\centering
\includegraphics[width=0.85\textwidth,height=0.8\textheight,keepaspectratio]{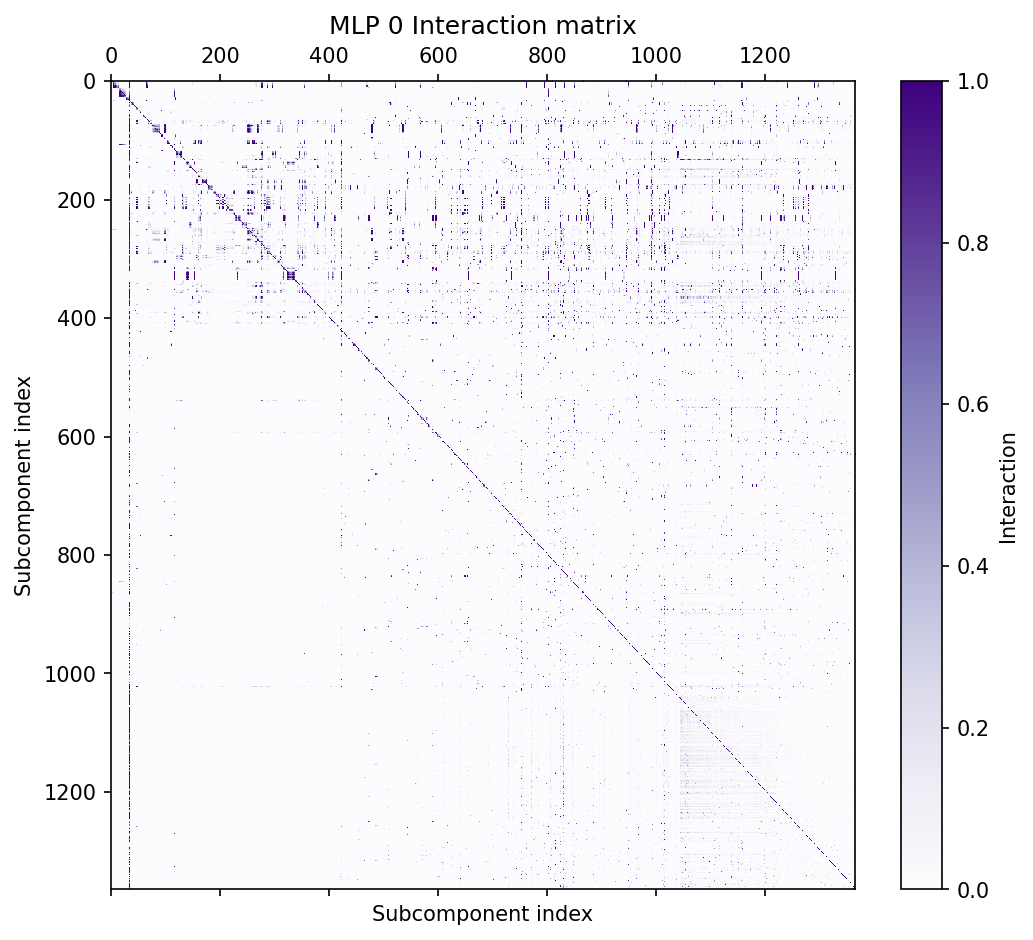}
\caption[]{Entries of the interaction matrix \(I_{c,c'}\) for the layer \(0\) MLP Up projection matrix subcomponents. Entries greater than \(1.0\) are clamped to \(1\). Indices are sorted by the components that each subcomponent belong to. Entries \(\geq 1.0\) in a row indicate that the nonlinear interaction between subcomponents \(c'\) and \(c\) is large compared to the self-interaction of \(c\) on the diagonal. Many interactions are either block-diagonal, indicating they take place inside higher-rank components, or arranged along vertical and horizontal lines, indicating they are caused by a relatively small number of highly interacting components.}\label{fig:I_h}
\end{figure}

\begin{figure}[htbp]
\centering
\includegraphics[width=0.85\textwidth,height=0.8\textheight,keepaspectratio]{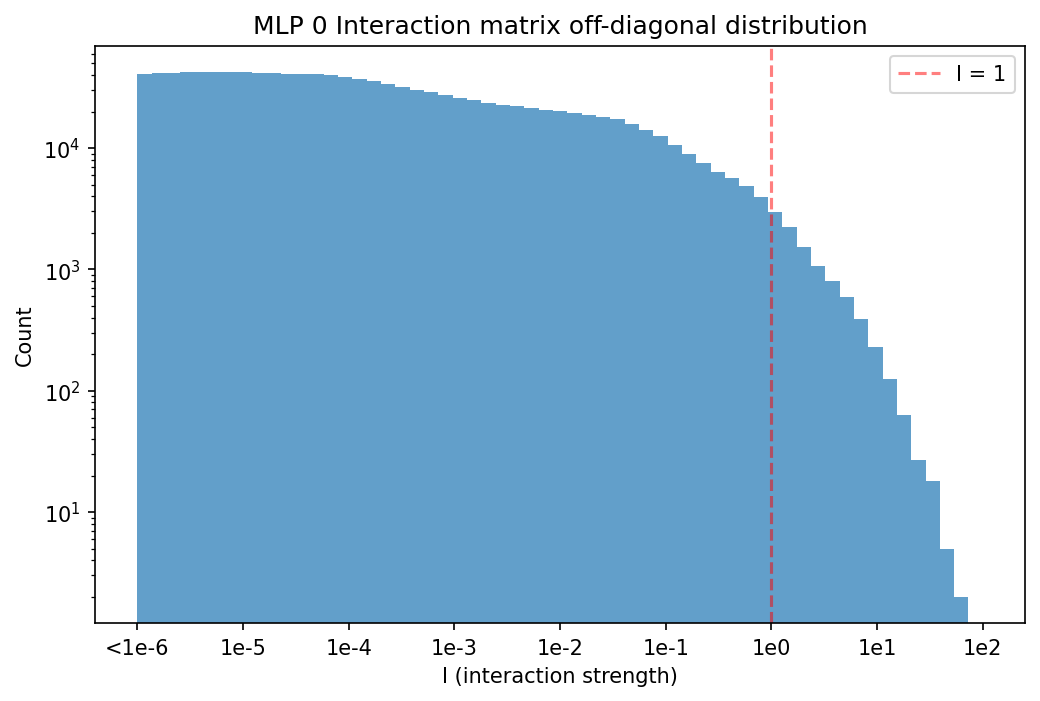}
\caption[]{Histogram of matrix entries of the interaction matrix \(I_{c,c'}\) for the layer \(0\) up projection subcomponents. Most entries are much smaller than the self-interaction \(1.0\).}\label{fig:I_dist}
\end{figure}

\autoref{fig:I_h} shows the \(I\) matrix entries for the layer 0 MLP Up
matrix, with the subcomponent indices sorted by the higher-rank
components they belong to. \autoref{fig:I_dist} shows a histograms of
the \(I\) entries. We can see that while there are certainly quite a few
large off-diagonal entries, many of them represent interactions within
components (the diagonal blocks), or interactions of a small set of
highly interacting subcomponents with all others (the vertical and
horizontal stripes). Plots for the interaction matrices of other layers
can be found in the next subsection. We stress that this is a very
preliminary investigation and the matrices \(I_{c,c'}\) are a crude and
imprecise measure of nonlinear interactivity in many ways.

For example, they do not quantify how much particular nonlinear
interactions between subcomponents actually influence downstream
observables like the model output.

Ultimately, what will matter in practice is whether we can use parameter
components to interpret nonlinear interactions well enough to reverse
engineer the algorithms neural networks have learned to implement. While
we may have some reasons for optimism on this question, we cannot
provide a real answer to it yet.

\subsubsection{Interaction matrix plots for all
MLPs}\label{interaction-matrix-plots-for-all-mlps}

\label{app:non_linear_plots}

Here, we show raw heatmaps and histograms for the interaction matrices
\(I_{c,c'}\) quantifying nonlinear interactions between MLP Up
projection matrix subcomponents at neurons for the other three mlp
layers. For the heatmaps, indices are sorted by the components that
subcomponents belong to. Entries \(\geq 1.0\) in a row indicate that the
nonlinear interaction between subcomponents \(c'\) and \(c\) is large
compared to the self-interaction of \(c\) on the diagonal. Many
interactions are either block-diagonal, indicating they take place
inside higher-rank components, or arranged along vertical and horizontal
lines, indicating they are caused by a relatively small number of highly
interacting components. The layer 1 mlp has particularly many
interactions, which may be an additional indicator that the VPD
decomposition of this transformer layer is somewhat pathological.

\begin{figure}[htbp]
\centering
\includegraphics[width=0.85\textwidth,height=0.8\textheight,keepaspectratio]{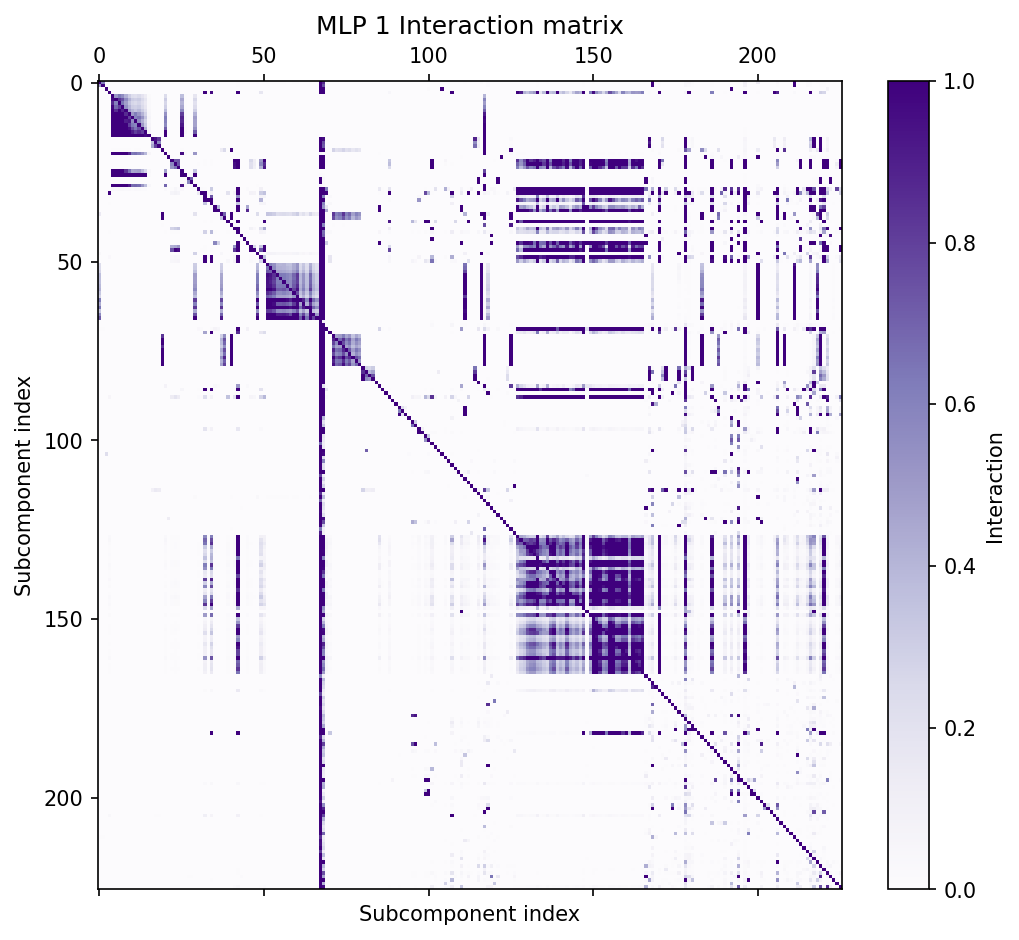}
\caption[]{Entries of the interaction matrix \(I_{c,c'}\) for the layer 1 MLP Up projection subcomponents. Indices are sorted by the components subcomponents belong to.}\label{fig:I_h_1_mlp_c_fc}
\end{figure}

\begin{figure}[htbp]
\centering
\includegraphics[width=0.85\textwidth,height=0.8\textheight,keepaspectratio]{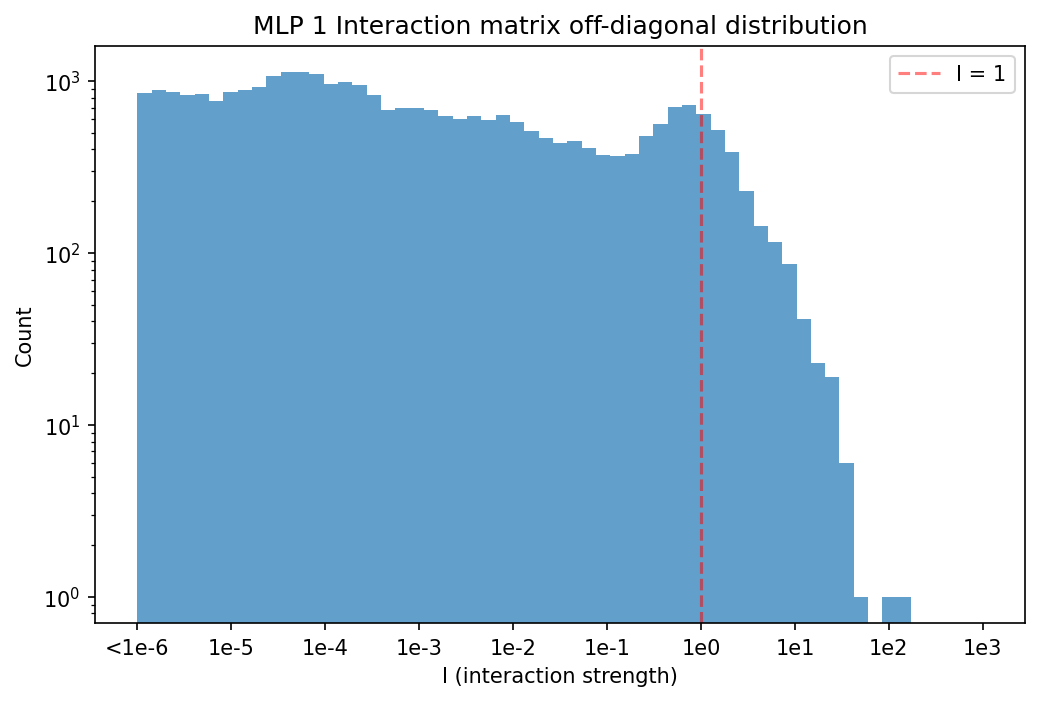}
\caption[]{Histogram of the entries of the interaction matrix \(I_{c,c'}\) for the layer 1 MLP Up projection subcomponents.}\label{fig:I_dist_h_1_mlp_c_fc}
\end{figure}

\begin{figure}[htbp]
\centering
\includegraphics[width=0.85\textwidth,height=0.8\textheight,keepaspectratio]{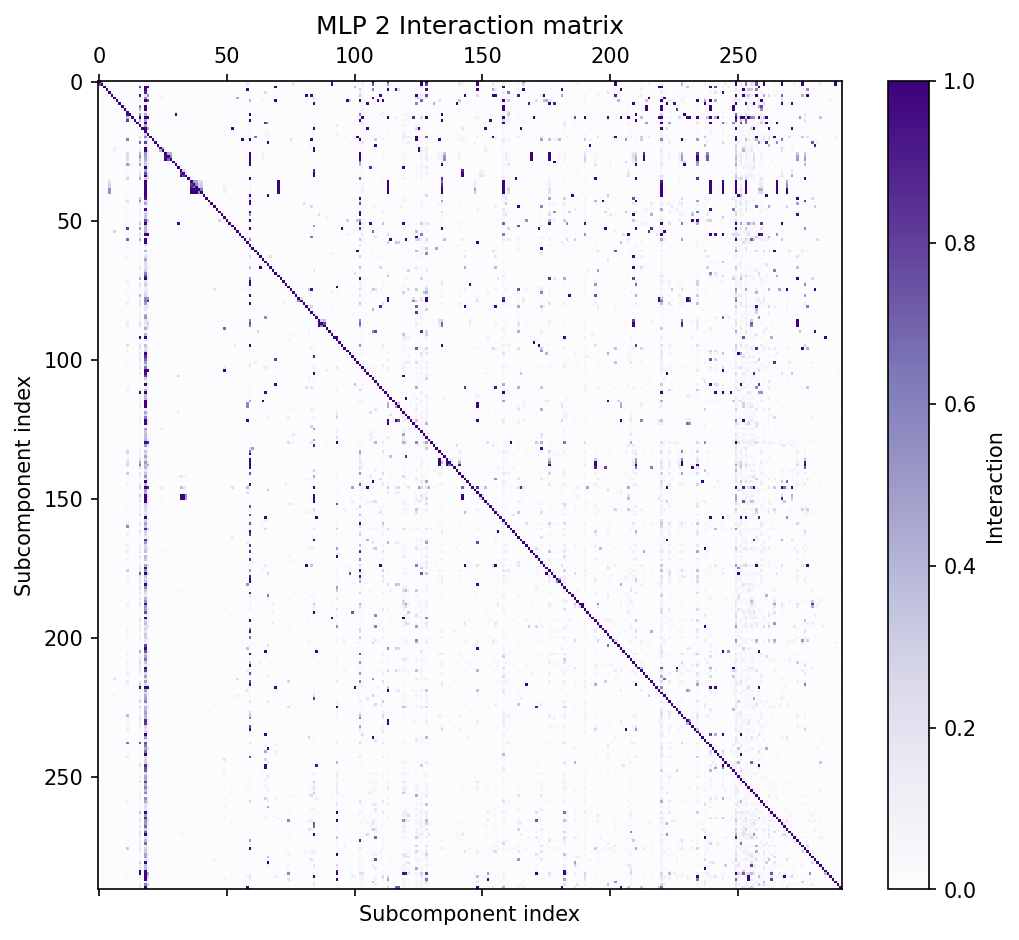}
\caption[]{Entries of the interaction matrix \(I_{c,c'}\) for the layer 2 MLP Up projection subcomponents. Indices are sorted by the components subcomponents belong to.}\label{fig:I_h_2_mlp_c_fc}
\end{figure}

\begin{figure}[htbp]
\centering
\includegraphics[width=0.85\textwidth,height=0.8\textheight,keepaspectratio]{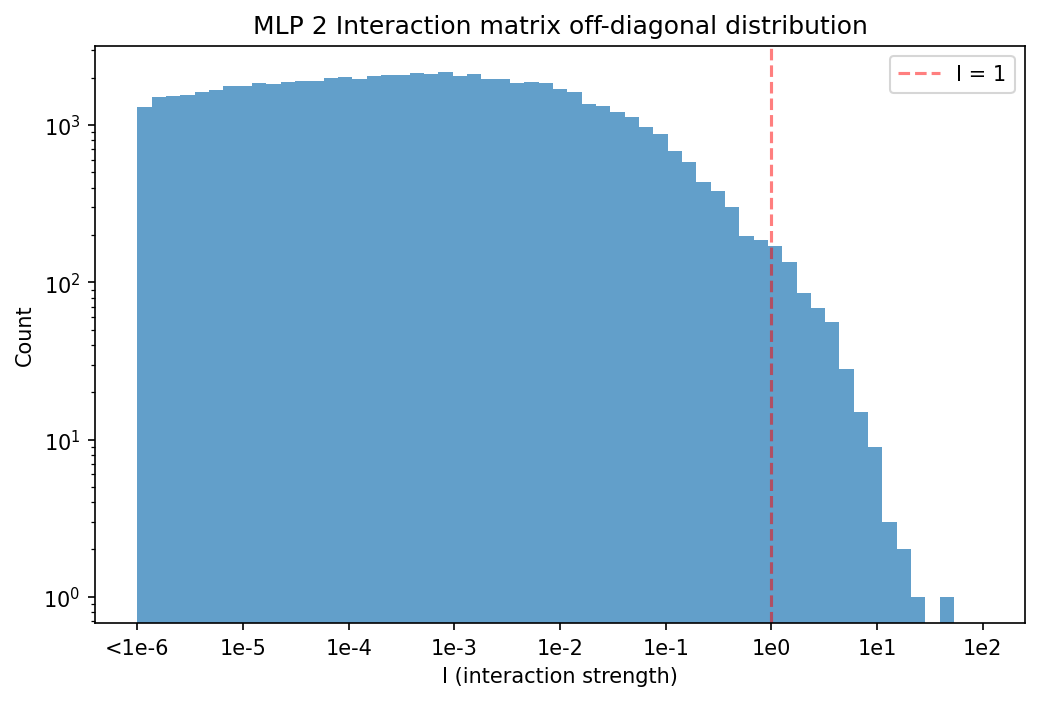}
\caption[]{Histogram of the entries of the interaction matrix \(I_{c,c'}\) for the layer 2 MLP Up projection subcomponents.}\label{fig:I_dist_h_2_mlp_c_fc}
\end{figure}

\begin{figure}[htbp]
\centering
\includegraphics[width=0.85\textwidth,height=0.8\textheight,keepaspectratio]{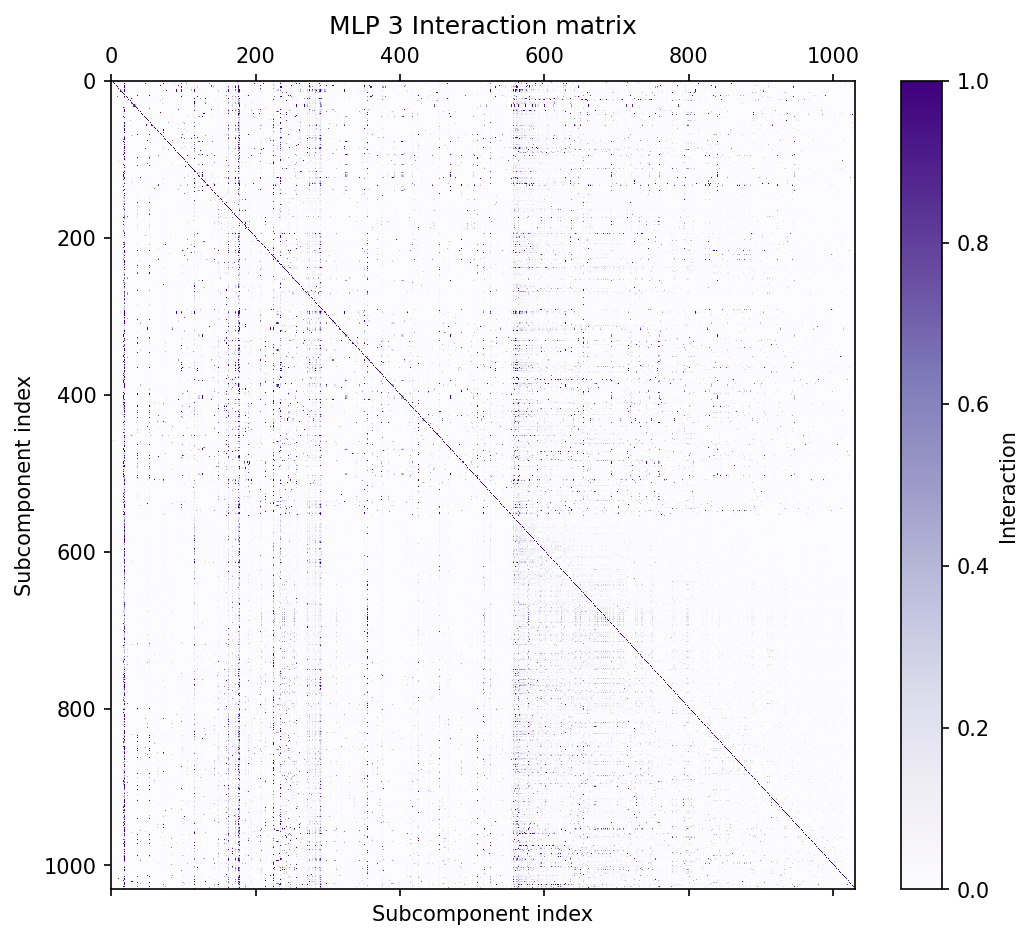}
\caption[]{Entries of the interaction matrix \(I_{c,c'}\) for the layer 3 MLP Up projection subcomponents. Indices are sorted by the components subcomponents belong to.}\label{fig:I_h_3_mlp_c_fc}
\end{figure}

\begin{figure}[htbp]
\centering
\includegraphics[width=0.85\textwidth,height=0.8\textheight,keepaspectratio]{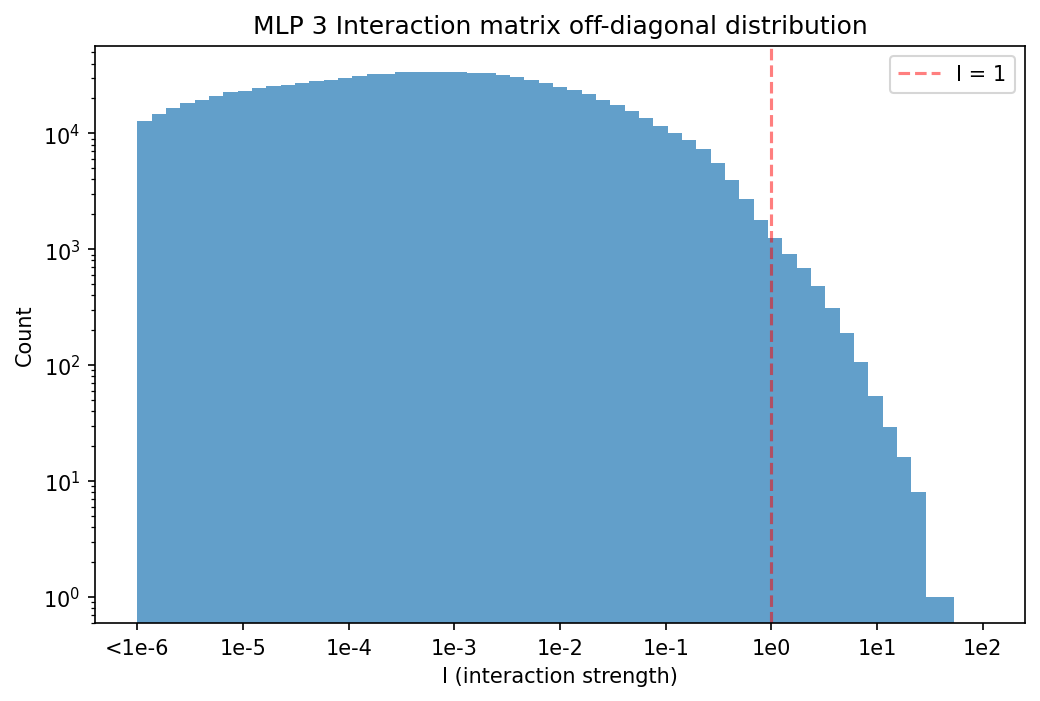}
\caption[]{Histogram of the entries of the interaction matrix \(I_{c,c'}\) for the layer 3 MLP Up projection subcomponents.}\label{fig:I_dist_h_3_mlp_c_fc}
\end{figure}
\bibliographystyle{plainnat}
\bibliography{refs}
\end{document}